\documentclass[runningheads, envcountsame, a4paper]{llncs}
\usepackage{graphicx}
\usepackage{microtype}
\usepackage{graphicx}
\usepackage{subfigure}
\usepackage{booktabs} 
\usepackage{algorithm,algorithmic}
\usepackage[hyphens]{url}
\usepackage[inline]{enumitem}
\usepackage[misc]{ifsym}

\usepackage{amsmath}
\usepackage{amssymb,amsfonts,dsfont,mathrsfs,bm,bbm}
\usepackage{xcolor}
\DeclareMathOperator*{\Argmax}{argmax}
\DeclareMathOperator{\prob}{Pr}
\DeclareMathOperator{\Expect}{\mathbb{E}}

\begin{document}
\title{Bayesian Crowdsourcing with Constraints
	\thanks{Work in this paper was supported by NSF grant 1901134.}}
\titlerunning{Bayesian Crowdsourcing with Constraints}

%
\author{Panagiotis A. Traganitis {\Letter}\and Georgios B. Giannakis}
\authorrunning{P. A. Traganitis and G. B. Giannakis}
%
\institute{Electrical \& Computer Engineering Department, and Digital Technology Center,\\ University of Minnesota, Minneapolis MN 55414, USA\\
\email{$\{$traga003,georgios$\}$@umn.edu}.
}
\toctitle{Bayesian Crowdsourcing with Constraints}
\tocauthor{Panagiotis~A.~Traganitis and Georgios~B.~Giannakis}
\maketitle              
\begin{abstract}
Crowdsourcing has emerged as a powerful paradigm for efficiently labeling large datasets and performing various learning tasks, by leveraging crowds of human annotators. When additional information is available about the data, constrained or semi-supervised crowdsourcing approaches that enhance the aggregation of labels from human annotators are well motivated. This work deals with constrained crowdsourced classification with instance-level constraints, that capture relationships between pairs of data. A Bayesian algorithm based on variational inference is developed, and its quantifiably improved performance, compared to unsupervised crowdsourcing, is analytically and empirically validated on several crowdsourcing datasets.

\end{abstract}
\section{Introduction}
{C}{rowdsourcing}, as the name suggests, harnesses crowds of human annotators, using services such as Amazon's Mechanical Turk~\cite{MTurk}, to perform various learning tasks such as labeling, image tagging and natural language annotations, among others~\cite{crowdsourcing}. Even though crowdsourcing can be efficient and relatively inexpensive, inference of true labels from the noisy responses provided by multiple annotators of unknown expertise can be challenging, especially in the typical \emph{unsupervised} scenario, where no ground-truth data is available.

It is thus practical to look for side information that can be beneficial to the crowdsourcing task, either from experts or from physical constraints associated with the task. For example, queries with known answers may be injected in crowdsourcing tasks in Amazon's Mechanical Turk, in order to assist with the evaluation of annotator reliability. Methods that leverage such information fall under the constrained or semi-supervised learning paradigm~\cite{ssl}.
Seeking improved performance in the crowdsourcing task, we focus on constrained crowdsourcing, and investigate instance-level or pairwise constraints, such as must- and cannot-link constraints, which show up as side information in unsupervised tasks, such as clustering~\cite{constrained_clustering}. Compared 
to label constraints, instance-level ones provide `weaker information,' as they describe relationships between pairs of data, instead of anchoring the label of a datum. Semi-supervised learning with few label constraints is typically employed when acquiring ground-truth labels is time consuming or expensive, e.g. annotation of biomedical images from a medical professional. Instance-level constraints on the other hand are used when domain knowledge is easier to encode between pairs of data, e.g. road lane identification from GPS data.

To accommodate such available information, the present work capitalizes on recent advances in Bayesian inference. Relative to deterministic methods, the Bayesian approach allows for seamless integration of prior information for the crowdsourcing task, such as annotator performance from previous tasks, and at the same time enables uncertainty quantification of the fused label estimates and parameters of interest. Our major contributions can be summarized as follows: 
\begin{enumerate*}
    \item[(i)] We develop a Bayesian algorithm for crowdsourcing with pairwise constraints;
    \item[(ii)] We derive novel error bounds for the unsupervised variational Bayes crowdsourcing algorithm, in Thm. \ref{thm:VBEM}. These error bounds are extended for the proposed constrained algorithm in Thm. \ref{thm:VBEM_instance}; and 
    \item[(iii)] Guided by the aforementioned theoretical analysis, we provide a constraint selection scheme.
\end{enumerate*}
In addition to the error bounds, the performance of the proposed algorithm is evaluated with extensive numerical tests. Those corroborate that there are classification performance gains to be harnessed, even when using weaker side information, such as the aforementioned must- and cannot-link constraints. 

\noindent\textbf{Notation:} Unless otherwise noted, lowercase bold fonts, $\bm{x}$, denote
vectors, uppercase ones, $\mathbf{X}$, represent matrices, and calligraphic uppercase, $\mathcal{X}$, stand for sets. The $(i,j)$th entry of matrix $\mathbf{X}$ is denoted by
$[\mathbf{X}]_{ij}$. $\prob$ or $p$ denotes probability, or the probability mass function; $\sim$ denotes  ``distributed as,'' $|\mathcal{X}|$ is the cardinality of set $\mathcal{X}$, $\Expect[\cdot]$ denotes expectation, and $\mathds{1}(\mathcal{A})$ is the indicator function for the event $\mathcal{A}$, that takes value $1$, when $\mathcal{A}$ occurs, and is $0$ otherwise.

\section{Problem formulation and preliminaries}
\label{sec:problem_statement}
Consider a dataset consisting of $N$ data $\{x_n\}_{n=1}^{N}$, with each datum belonging to one of $K$ classes with corresponding labels $\{y_n\}_{n=1}^{N}$; that is, $y_n = k$ if $x_n$ belongs to class $k$. Consider now $M$ annotators that observe $\{x_n\}_{n=1}^{N}$, and provide estimates of labels. Let $\check{y}_n^{(m)}\in\{1,\ldots,K\}$ be the label estimate of $x_n$ assigned by the $m$-th annotator. If an annotator has not provided a response for datum $n$ we set $\check{y}_n^{(m)} = 0.$ Let $\check{\mathbf{Y}}$ be the $M\times N$ matrix of annotator responses with entries $[\check{\mathbf{Y}}]_{mn} = \check{y}_n^{(m)}$, and $\mathbf{y} := [y_1,\ldots,y_N]^{\top}$ the $N\times 1$ vector of ground truth labels. 
The task of \emph{crowdsourced classification} is: Given only the annotator responses in $\check{\mathbf{Y}}$, the goal is find the ground-truth label estimates $\{\hat{y}_n\}_{n=1}^N$. 

Per datum $x_n$, the true label $y_n$ is assumed drawn from a categorical distribution with parameters $\bm{\pi} := [\pi_1,\ldots,\pi_K]^\top$, where $\pi_k := \prob(y_n = k)$. Further, consider that each learner has a fixed probability of deciding that a datum belongs to class $k'$, when presented with a datum of class $k$; thus, annotator behavior is presumed invariant across the dataset.The performance of an annotator $m$ is then characterized by the so-called \emph{confusion} matrix $\mathbf{\Gamma}^{(m)}$, whose $(k,k')$-th entry is {$[\mathbf{\Gamma}^{(m)}]_{k,k'} := \gamma_{k,k'}^{(m)} = \prob\left(\check{y}_n^{(m)} = k' | y_n = k\right).$}
The $K\times K$ confusion matrix showcases the statistical behavior of an annotator, as each row provides the annotator's probability of deciding the correct class, when presented with a datum from each class. Collect all annotator confusion matrices in $\mathbf{\Gamma}:=[\mathbf{\Gamma}^{(1)},\ldots,\mathbf{\Gamma}^{(m)}].$
Responses of different annotators per datum $n$ are presumed conditionally independent, given the ground-truth label $y_n$; that is, $p\left(\check{y}_{n}^{(1)} = k_1,\ldots,\check{y}_{n}^{(M)}=k_M | y_n = k\right) 
= \prod_{m=1}^{M} p\left(\check{y}_{n}^{(m)} = k_m | y_n = k\right). $
The latter is a standard assumption that is commonly employed in crowdsourcing works~\cite{dawid1979maximum,jaffe2015estimating,zhang2014spectral}. Finally, most annotators are assumed to be better than random.

\subsection{Prior works}
Arguably the simplest approach to fusing crowdsourced labels is majority voting, where the estimated label of a datum is the one most annotators agree upon. This presumes that all annotators are equally ``reliable,'' which may be unrealistic. Aiming at high-performance label fusion, several  approaches estimate annotator parameters, meaning the confusion matrices as well as class priors. A popular approach is joint maximum likelihood (ML) estimation of the unknown labels $\mathbf{y}$, and the aforementioned {confusion matrices} using the expectation-maximization (EM) algorithm~\cite{dawid1979maximum}. As EM guarantees convergence to a locally optimal solution, alternative estimation methods have been recently advocated. Spectral methods invoke second- and third-order moments of annotator responses to infer the unknown annotator parameters~\cite{jaffe2015estimating,zhang2014spectral,traganitis2018,kargas_xiao}.  

A Bayesian treatment of crowdsourced learning, termed \emph{Bayesian Classifier Combination} was introduced in~\cite{BCC_Kim}. This approach used Gibbs sampling to estimate the parameters of interest, while~\cite{BCC_VB} introduced a variational Bayes EM (VBEM) method for the same task. Other Bayesian approaches infer communities of annotators, enhancing the quality of aggregated labels~\cite{cbcc,bayesian_nonparam}. For sequential or networked data, \cite{Rodrigues2014,pmid_hmm_em,traganitis_ensemble_dep_dsw,traganitis2019_networked_sequential} advocated variational inference and EM-based alternatives. All aforementioned approaches utilize only $\check{\mathbf{Y}}$. When features $\{x_n\}_{n=1}^{N}$ are also available, parametric models~\cite{raykar}, {approaches based on Gaussian Processes~\cite{vgp,musicgenre_senpoldata}, or deep learning \cite{aggnet,deep_rodrigues,kdd_crowdSSL_deep,aaai_crowd_SSL_deep} can be employed to classify the data, and simultaneously learn a classifier}. 

Current constrained or semi-supervised approaches to crowdsourcing extend the EM algorithm of Dawid and Skene~\cite{tang}, by including a few ground-truth labels. The model in \cite{yan_semi} includes a graph between datapoints to enable label prediction for data that have not been annotated, while \cite{m3v_active} puts forth a max-margin majority vote method when label constraints are available. When features $\{x_n \}_{n=1}^{N}$ are available, \cite{kajino2012learning} relies on a parametric model to learn a binary classifier from crowd responses and expert labels.

The present work develops a Bayesian \emph{constrained} approach to crowdsourced classification, by adapting the popular variational inference framework \cite{variational}. In addition, it provides novel performance analysis for both Bayesian unsupervised and semi-supervised approaches. The proposed method does not require features $\{x_n \}_{n=1}^{N}$, but relies only on annotator labels -- a first attempt at incorporating instance-level constraints for the crowdsourced classification task.

\section{Variational inference for crowdsourcing}
\label{sec:VB}
Before presenting our {constrained} approach, this section will recap a Bayesian treatment of the crowdsourcing problem. First, variational Bayes is presented in Sec.~\ref{ssec:VI}, followed by an inference algorithm for crowdsourcing~\cite{BCC_VB} in Sec.~\ref{ssec:IBCC_VBEM}. A novel performance analysis for the VBEM algorithm is also presented in Sec.~\ref{ssec:IBCC_VBEM}.

\subsection{Variational Bayes}
\label{ssec:VI}
Consider a set of observed data collected in $\mathbf{X}$, and a set of latent variables and parameters in $\mathbf{Z}$ that depend on $\mathbf{X}$. Variational Bayes seeks $\mathbf{Z}$ that maximize the marginal of 
$\mathbf{X}$, by treating both latent variables and parameters as random~\cite{variational}. EM in contrast treats $\mathbf{Z}$ as deterministic and provides point estimates. The log-marginal of $\mathbf{X}$ can be written as
\begin{align}
	\ln p(\mathbf{X}) & = \int q(\mathbf{Z})\ln\frac{p(\mathbf{X},\mathbf{Z})}{q(\mathbf{Z})}d\mathbf{Z} - \int q(\mathbf{Z})\ln\frac{p(\mathbf{Z}|\mathbf{X})}{q(\mathbf{Z})}d\mathbf{Z} 
	= L(q) + {\rm KL}(q||p) \label{eq:log_obs}
\end{align}
where $L(q): = \int q(\mathbf{Z})\ln\frac{p(\mathbf{X},\mathbf{Z})}{q(\mathbf{Z})}d\mathbf{Z}$,
and ${\rm KL}(q||p) = -\int q(\mathbf{Z})\ln\frac{p(\mathbf{Z}|\mathbf{X})}{q(\mathbf{Z})}d\mathbf{Z}$ denotes the Kullback-Leibler divergence between pdfs $q$ and $p$~\cite{cover2012elements}. This expression is maximized for 
$q(\mathbf{Z}) = p(\mathbf{Z}|\mathbf{X})$, however when $p(\mathbf{Z}|\mathbf{X})$ is an intractable pdf, one may seek distributions $q$ from a prescribed tractable family $\mathcal{Q}$, such that ${\rm KL}(q||p)$ is minimized. One such family is the family of factorized distributions, which also go by the name of mean field distributions. Under the mean field paradigm, the variational distribution $q(\mathbf{Z})$ is decomposed into a product of single variable factors, $q(\mathbf{Z}) = \prod_i q(\mathbf{Z}_i)$, with $q(\mathbf{Z}_i)$ denoting the variational distribution corresponding to the variable $\mathbf{Z}_i.$

It can be shown that the optimal updates for each factor are given by~\cite{variational}
\begin{equation}
	\ln q^*(\mathbf{Z}_i) = \Expect_{-\mathbf{Z}_i}\left[\ln p(\mathbf{X},\mathbf{Z}) \right] + c
\end{equation}
where $c$ is an appropriate constant, and the ${-\mathbf{Z}_i}$ subscript denotes that the expectation is taken w.r.t. the terms in $q$ that do not involve $\mathbf{Z}_i$, that is $\prod_{j\neq i}q(\mathbf{Z}_j)$. These optimal factors can be estimated iteratively, and such a procedure is guaranteed to converge at least to a local maximum of \eqref{eq:log_obs}.

\subsection{Variational EM for crowdsourcing}
\label{ssec:IBCC_VBEM}
Next, we will outline how variational inference can be used to derive an iterative algorithm for crowdsourced classification~\cite{BCC_VB}. The Bayesian treatment of the crowdsourcing problem dictates the use of prior distributions on the parameters of interest, namely  $\bm{\pi}$, and $\mathbf{\Gamma}$. The probabilities $\bm{\pi}$ are assigned a Dirichlet distribution prior with parameters $\bm{\alpha}_{0}:=[\alpha_{0,1},\ldots,\alpha_{0,K}]^\top$, that is $\bm{\pi}\sim{\rm Dir}(\bm{\pi};\bm{\alpha}_{0})$, whereas for an annotator $m$, the columns $\{\bm{\gamma}_k^{(m)} \}_{k=1}^{K}$ of its confusion matrix are considered independent, and $\bm{\gamma}_k^{(m)}$ is assigned  a Dirichlet distribution prior with parameters $\bm{\beta}_{0,k}^{(m)} := [\beta_{0,k,1}^{(m)},\ldots,\beta_{0,k,K}^{(m)}]^\top$, respectively. These priors on the parameters of interest are especially useful when only few data have been annotated, and can also capture annotator behavior from previous crowdsourcing tasks. The joint distribution of $\mathbf{y},\check{\mathbf{Y}},\bm{\pi}$ and $\mathbf{\Gamma}$ is 
\begin{align}
	& p(\mathbf{y},\bm{\pi},\check{\mathbf{Y}},\mathbf{\Gamma};\bm{\alpha}_{0},\mathbf{B}_0)  \label{eq:IBCC_joint_prob} 
	 = \prod_{n=1}^{N}\pi_{y_n}\prod_{m=1}^{M}\prod_{k'=1}^{K}(\gamma_{y_n,k'}^{(m)})^{\delta_{n,k'}^{(m)}}p(\bm{\pi};\bm{\alpha}_{0})p(\mathbf{\Gamma};\mathbf{B}_0) 
\end{align}
with $\mathbf{B}_0$ collecting all prior parameters $\bm{\beta}_{0,k}^{(m)}$ for $k=1,\ldots,K$ and $m=1,\ldots,M$, and $\delta_{n,k}^{(m)}:=\mathds{1}(\check{y}_n^{(m)} = k)$. The parametrization on $\bm{\alpha}_{0},\mathbf{B}_0$ will be henceforth implicit for brevity.

With the goal of estimating the unknown variables and parameters of interest, we will use the approach outlined in the previous subsection, to approximate $p(\mathbf{y},\bm{\pi},\mathbf{\Gamma}|\check{\mathbf{Y}})$ using a variational distribution $q(\mathbf{y},\bm{\pi},\mathbf{\Gamma})$. Under the mean-field class, this variational distribution factors across the unknown variables as $q(\mathbf{y},\bm{\pi},\mathbf{\Gamma}) = q(\mathbf{y})q(\bm{\pi})q(\mathbf{\Gamma}).$ 
Since the data are assumed i.i.d. [cf. Sec.~\ref{sec:problem_statement}], $q(\mathbf{y})$ is further decomposed into $q(\mathbf{y}) = \prod_{n=1}^{N}q(y_n).$ In addition, since annotators are assumed independent and the columns of their confusion matrices are also independent, we have 
$q(\mathbf{\Gamma}) = \prod_{m=1}^{M}\prod_{k=1}^{K}q(\bm{\gamma}_{k}^{(m)})$. The variational Bayes EM algorithm is an iterative algorithm, with each iteration consisting of two steps: the variational E-step, where the latent variables $\mathbf{y}$ are estimated; and the variational M-step, where the distribution of the parameters of interest $\bm{\pi},\mathbf{\Gamma}$ are estimated.

At iteration $t+1$, the variational distribution for the unknown label $y_n$ is given by
\begin{equation}
	\ln q_{t+1}(y_n) = \Expect_{-y_n}\left[\ln p(\mathbf{y},\bm{\pi}, \check{\mathbf{Y}},\mathbf{\Gamma})  \right] + c. \label{eq:q_upd_y_n}
\end{equation}
The subscript of $q$ denotes the iteration index. The expectation in \eqref{eq:q_upd_y_n} is taken w.r.t. the terms in $q_{t}$ that do not involve $y_n$, that is $\prod_{n'\neq n}q_{t}(y_{n'})q_{t}(\bm{\pi})q_{t}(\mathbf{\Gamma})$. Upon expanding $p(\bm{y},\bm{\pi},\check{\mathbf{Y}},\mathbf{\Gamma})$, \eqref{eq:q_upd_y_n} becomes
\begin{align}
	& \ln q_{t+1}(y_n)  = \Expect_{\bm{\pi}}\left[\ln\pi_{y_n}\right]  + \Expect_{\mathbf{\Gamma}}\left[\sum_{m=1}^{M}\sum_{k'=1}^{K}\delta_{n,k'}^{(m)}\ln\gamma_{y_n,k'}^{(m)} \right] + c \label{eq:q_upd_y_n_2}.
\end{align}
Accordingly, the update for the class priors in $\bm{\pi}$ is 
\begin{equation}
	\ln q_{t+1}(\bm{\pi}) = \Expect_{-\bm{\pi}}\left[\ln p(\mathbf{y},\bm{\pi}, \check{\mathbf{Y}},\mathbf{\Gamma})  \right] + c \label{eq:q_upd_pi}
\end{equation}
where the expectation is taken w.r.t. $q_{t+1}(\mathbf{y})q_t(\mathbf{\Gamma})$. 
{Based on \eqref{eq:q_upd_pi}, it can be shown \cite{BCC_VB} that}
\begin{equation}
	q_{t+1}(\bm{\pi}) \propto \text{Dir}(\bm{\pi} ; \bm{\alpha}_{t+1})
\end{equation}
where $\bm{\alpha}_{t+1} := [\alpha_{t+1,1},\ldots,\alpha_{t+1,K}]^\top$ with $\alpha_{t+1,k} = N_{t+1,k} + \alpha_{0,k}$, $N_{t+1,k} :=\sum_{n=1}^{N}q_{t+1}(y_n = k) $.
{As a direct consequence of this, the term involving $\bm{\pi}$ in \eqref{eq:q_upd_y_n_2} is given by $\Expect_{\bm{\pi}}\left[\ln\pi_{k}\right] = \psi\left(\alpha_{t,k}\right) - \psi\left(\bar{\alpha}_{t}\right),$} 
with $\psi$ denoting the digamma function, and $\bar{\alpha}_{t} := \sum_{k=1}^{K}\alpha_{t,k}$.

For the $k$-th column of $\mathbf{\Gamma}^{(m)}$ the update takes the form
\begin{align}
	& \ln q_{t+1}(\bm{\gamma}_{k}^{(m)}) = \Expect_{-\bm{\gamma}_{k}^{(m)}}\left[\ln p(\mathbf{y},\bm{\pi}, \check{\mathbf{Y}},\mathbf{\Gamma}) \right] + c  \label{eq:q_upd_gamma_col}
\end{align}
Using steps similar to the update of $q(\bm{\pi})$, one can show that
\begin{equation}
	q_{t+1}(\bm{\gamma}_{k}^{(m)}) \propto \text{Dir}(\bm{\gamma}_{k}^{(m)} ; \bm{\beta}_{t+1,k}^{(m)})
\end{equation}
with $\bm{\beta}_{t+1,k}^{(m)} := [\beta_{t+1,k,1}^{(m)},\ldots,\beta_{t+1,k,1}^{(m)}]$, $\beta_{t+1,k,k'}^{(m)} := N_{t+1,k,k'}^{(m)} + \beta_{0,k,k'}^{(m)}$ and  $N_{t+1,k,k'}^{(m)}:= \sum_{n=1}^{N}q_{t+1}(y_n = k)\delta_{n,k'}^{(m)}$.
Consequently at iteration $t+1$, and upon defining $\bar{\beta}_{t+1,k}^{(m)}:=\sum_{\ell}^{K}\beta_{t+1,k,\ell}^{(m)}$, we have $\Expect_\mathbf{\Gamma}[\ln\gamma_{k,k'}^{(m)}] = \psi\left(\beta_{t+1,k,k'}^{(m)}\right) -
	\psi\left(\bar{\beta}_{t+1,k}^{(m)} \right).$
 Given initial values for $\Expect_{\bm{\pi}}\left[\ln\pi_{y_n}\right]$ and  $\Expect_{\mathbf{\Gamma}}\left[\sum_{m=1}^{M}\ln\gamma_{y_n,\check{y}_n^{(m)}}^{(m)} \right]$ per variational E-step, first the variational distribution for each datum $q(y_n)$ is computed using \eqref{eq:q_upd_y_n}. Using the recently computed $q(\mathbf{y})$ at the variational M-step, the variational distribution for class priors $q(\bm{\pi})$ is updated via \eqref{eq:q_upd_pi} and the variational distributions for each column of the confusion matrices $q(\bm{\gamma}_{k}^{(m)})$ are updated via \eqref{eq:q_upd_gamma_col}.  
The E- and M-steps are repeated until convergence. Finally, the fused labels $\hat{\mathbf{y}}$ are recovered at the last iteration $T$ as
\begin{equation}
	\hat{y}_n = \Argmax_{k} \: q_{T}(y_n = k).
	\label{eq:final_labels}
\end{equation}
The overall computational complexity of this algorithm is $\mathcal{O}(NKMT)$. The updates of this VBEM algorithm are very similar to the corresponding updates of the EM algorithm for crowdsourcing~\cite{dawid1979maximum}.

\subsubsection{Performance analysis}
\label{sssec:IBCC_VBEM_thm}
Let $\bm{\pi}^*$ comprise the optimal class priors, $\mathbf{\Gamma}^*$ the optimal confusion matrices, and  $\mu_m := \prob(\check{y}_n^{(m)} \neq 0)$ the probability an annotator will provide a response.
The following theorem establishes that when VBEM is properly initialized, the estimation errors for $\bm{\pi},\mathbf{\Gamma}$ and the labels $\mathbf{y}$ are bounded. Such proper initializations can be achieved, for instance, using the spectral approaches of \cite{zhang2014spectral,traganitis2018,kargas_xiao}.
\begin{theorem}
	\label{thm:VBEM}
	Suppose that $\gamma_{k,k'}^{*(m)}\geq \rho_\gamma$ for all $k,k',m$, $\pi_k^{*}\geq\rho_\pi$ for all $k$, and that VBEM is initialized so that $|\Expect[\pi_k] - \pi_k^*|\leq\varepsilon_{\pi,0}$ for all $k$ and 
	$|\Expect[\gamma_{k,k'}^{(m)}] - \gamma_{k,k'}^{*(m)}|\leq\varepsilon_{\gamma,0}$ for all $k,k',m$. It then holds w.p. at least $1-\nu$ that for iterations $t=1,\ldots,T$, update \eqref{eq:q_upd_y_n} yields
	\begin{align}
		&\max_{k} |q_t(y_n = k) - \mathds{1}(y_n = k)| \leq \varepsilon_{q,t}:= K\exp(-U) \notag\\
		& U := D+ f_\pi(\varepsilon_{\pi,t-1}) + Mf_\gamma (\varepsilon_{\gamma,t-1})\notag,
	\end{align}
	where $D$ is a quantity related to properties of the dataset and the annotators, while $f_\pi,f_\gamma$ are decreasing functions.
	Consequently for $k,k'=1,\ldots,K$, $m=1,\ldots,M$ updates \eqref{eq:q_upd_pi} and \eqref{eq:q_upd_gamma_col} yield respectively
	\begin{align}
		& |\Expect[\pi_k] - \pi_k^*|\leq \epsilon_{\pi,k,t} := \frac{N(\varepsilon_{q,t} + g_\pi(\nu)) + \alpha_{0,k} + \rho_\pi\bar{\alpha}_0}{N + \bar{\alpha}_0}  \\
		& |\Expect[\gamma_{k,k'}^{(m)}] - \gamma_{k,k'}^{*(m)}|\leq \epsilon_{\gamma,k,k',t}^{(m)} :=  \frac{2N(g_\gamma(\nu) + \varepsilon_{q,t}) + \beta_{0,k,k'}^{(m)} + \bar{\beta}_{0,k}^{(m)} }{N\mu_m\pi_k^* - N\frac{g_\gamma(\nu)}{\gamma_{k,k'}^{*(m)}} - N\varepsilon_{q,t} + \bar{\beta}_{k}^{(m)}} 
	\end{align}
	with $g_\pi(\nu)$ and $g_\gamma(\nu)$ being decreasing functions of $\nu$ and $\varepsilon_{\pi,t}:=\max_{k}\epsilon_{\pi,k,t}, \varepsilon_{\gamma,t}:= \max_{k,k',m} \epsilon_{\gamma,k,k',t}^{(m)}$ .
\end{theorem}
Detailed theorem proofs are deferred to Appendix~A of the supplementary material. Theorem \ref{thm:VBEM} shows that lowering the upper bound on label errors $\varepsilon_{q}$, will reduce the estimation error upper bounds of $\{\gamma_{k,k'}^{(m)}\}$ and $\bm{\pi}$. This in turn can further reduce $\varepsilon_{q,t}$, as it is proportional to $\varepsilon_\pi$, and $\varepsilon_\gamma$. With a VBEM algorithm for crowdsourcing and its performance analysis at hand, the ensuing section will introduce our {constrained} variational Bayes algorithm that can incorporate additional information to enhance label aggregation.

\section{Constrained crowdsourcing}
\label{sec:semi-supervised}
This section deals with a {constrained (or semi-supervised)} variational Bayes approach to crowdsourcing. {Here, additional information to the crowdsourcing task is available in the form of pairwise constraints, that indicate relationships between pairs of data.} Throughout this section, $N_C$ will denote the number of available constraints that are collected in the set $\mathcal{C}$. 

First, we note that when label constraints are available, the aforementioned VBEM algorithm can readily handle them, in a manner similar to \cite{tang}. Let $\mathcal{C}$ denote the set of indices for data with label constraints $\{ y_n\}_{n\in\mathcal{C}}$ available. These constraints can then be incorporated by fixing the values of $\{q(y_n)\}_{n\in\mathcal{C}}$ to $1$ for all iterations. The variational distributions for data $n\not\in\mathcal{C}$ are updated according to \eqref{eq:q_upd_y_n_2}, while confusion matrices and prior probabilities according to \eqref{eq:q_upd_gamma_col} and \eqref{eq:q_upd_pi}, respectively. 

Next, we consider the case of instance-level constraints, which are the main focus of this work. Such information may be easier to obtain than the label constraints of the previous subsection, as pairwise constraints  encapsulate relationships between pairs of data and not ``hard'' label information. The pairwise constraints considered here take the form of must-link and cannot-link relationships, and are collected respectively in the sets $\mathcal{C}_{\rm ML}$ and $\mathcal{C}_{\rm CL}$, and $\mathcal{C} = \mathcal{C}_{\rm ML}\cup\mathcal{C}_{\rm CL}$, $N_{\rm ML} = |\mathcal{C}_{\rm ML}|, N_{\rm CL} = |\mathcal{C}_{\rm CL}|$. All constraints consist of tuples $(i,j)\in\{1,\ldots,N\}\times\{1,\ldots,N\}$.  A must-link constraint $(i,j), i\neq j$ indicates that two data points, $\bm{x}_i$ and $\bm{x}_j$  must belong to the same class, i.e. $y_i = y_j$, whereas a cannot-link constraint $(i',j')$  indicates that two points $\bm{x}_{i'}$ and $\bm{x}_{j'}$ are not in the same class; $y_{i'} \neq y_{j'}$. Note that instance level constraints naturally describe a graph $\mathcal{G}$, whose $N$ nodes correspond to the data $\{x_n\}_{n=1}^{N}$, and whose (weighted) edges are the available constraints. This realization is the cornerstone of the proposed algorithm, suggesting that instance-level constraints can be incorporated in the crowdsourcing task by means of a probabilistic graphical model.

Specifically, we will encode constraints in the marginal pmf of the unknown labels $p(\mathbf{y})$ using a Markov Random Field (MRF), which implies that for all $n=1,\ldots,N$, the local Markov property holds, that is $ p(y_n|\mathbf{y}_{-n}) = p(y_n | \mathbf{y}_{\mathcal{C}_n})$,
where $\mathbf{y}_{-n}$ denotes a vector containing all labels except $y_n$, while $\mathbf{y}_{\mathcal{C}_n}$ a vector containing labels for $y_{n'}$, where $n'\in\mathcal{C}_n,$ and  $\mathcal{C}_n$ is a set containing indices such that $n': (n,n')\in\mathcal{C}$. By the Hammersley-Clifford theorem~\cite{HammersleyClifford}, the marginal pmf of the unknown labels is given by  
\begin{equation}
	p(\mathbf{y}) = \frac{1}{Z}\exp\left(\sum_{n=1}^{N}\ln\pi_{y_n} + \eta \underset{n'\in\mathcal{C}_n}{\sum} V(y_n,y_{n'}) \right)
	\label{eq:MRF_p_y}
\end{equation}
where $Z$ is the (typically intractable) normalization constant, and $\eta > 0$ is a tunable parameter that specifies how much weight we assign to the constraints. Here, we define the $V$ function as $V(y_n,y_{n'}) = w_{n,n'}\mathds{1}(y_n = y_{n'})$, where 
\begin{equation}
    w_{n,n'} = \mathds{1}\left((n,n')\in\mathcal{C}_{\rm ML}\right) - \mathds{1}\left((n,n')\in\mathcal{C}_{\rm CL}\right)
\end{equation}
The weights $w_{n,n'}$ can also take other real values, indicating the confidence one has per  constraint $(n,n')$, but here we will focus on ${-1,0,1}$ values. This $V(y_n,y_{n'})$ term promotes the same label for data with must-link constraints, whereas it penalizes similar labels for data with cannot-link constraints. Note that more sophisticated choices for $V$ can also be used. Nevertheless, this particular choice leads to a simple algorithm with quantifiable performance, as will be seen in the ensuing sections. Since the number of constraints is typically small, for the majority of data the term $\sum V(y_n,y_{n'})$ will be $0$. Thus, the $\ln\pi$ term in the exponent of \eqref{eq:MRF_p_y} acts similarly to the prior probabilities of Sec.~\ref{ssec:IBCC_VBEM}. Again, adopting a Dirichlet prior ${\rm Dir} (\bm{\pi};\bm{\alpha}_{0}),$ and $\bm{\gamma}_k^{(m)}\sim {\rm Dir}(\bm{\gamma}_k^{(m)}; \bm{\beta}_{0,k}^{(m)})$ for all $m,k$, 
and using mean-field VB as before, the variational update for the $n$-th label becomes
\begin{align}
	&\ln q_{t+1}(y_n = k)  = \Expect_{\bm{\pi}}\left[\ln\pi_{k}\right] + c \label{eq:q_upd_MRF_y}
	\\ & + 
	\Expect_{\mathbf{\Gamma}}\left[\sum_{m=1}^{M}\sum_{k'=1}^{K}{\delta_{n,k'}^{(m)}}\ln\gamma_{k,k'}^{(m)} \right] + \eta\sum_{n'\in\mathcal{C}_{n}}w_{n,n'}q_{t}(y_{n'} = k) . \notag
\end{align}
where we have used $\Expect_{y_{n'}}[\mathds{1}(y_{n'} = k)] = q_{t}(y_{n'} = k).$
The variational update for label $y_n$ is similar to the one in \eqref{eq:q_upd_y_n_2}, with the addition of the $\eta\sum_{n'\in\mathcal{C}_{n}}w_{n,n'}q_{t}(y_{n'} = k)$ term that captures the labels of the data that are related (through the instance constraints) to the $n$-th datum.
Updating label distributions via \eqref{eq:q_upd_MRF_y}, updates for $\bm{\pi}$ and $\mathbf{\Gamma}$ remain identical to those in VBEM [cf. Sec.~\ref{ssec:IBCC_VBEM}]. The instance-level semi-supervised crowdsourcing algorithm is outlined in Alg.~\ref{alg:VBEM_instance}. As with its plain vanilla counterpart of Sec.~\ref{ssec:IBCC_VBEM}, Alg.~\ref{alg:VBEM_instance} maintains the asymptotic complexity of $\mathcal{O}(NMKT)$. 
\begin{algorithm}[tb]
\caption{Crowdsourcing with instance-level constraints}\label{alg:VBEM_instance}
\begin{algorithmic}
\STATE {\bfseries Input:} Annotator responses $\check{\mathbf{Y}}$, initial $\bm{\pi}$, $\mathbf{\Gamma}$, $\{q_0(y_n = k\}_{n,k=1}^{N,K}$, constraints $\mathcal{C}$, parameters $\bm{\alpha}_0$, $\mathbf{B}_0$, $\eta$
\STATE {\bfseries Output:} Estimates $\Expect[\bm{\pi}]$, $\Expect[\mathbf{\Gamma}]$, $\hat{\mathbf{y}}$.
\WHILE{not converged}
\STATE Update $q_{t+1}(y_n)$ via \eqref{eq:q_upd_MRF_y}
\STATE Update $q_{t+1}(\bm{\pi})$ using $q_{t+1}(y_n)$, \eqref{eq:q_upd_pi}
\STATE Update $q_{t+1}(\bm{\gamma}_{k}^{(m)})$ $\forall m,k$, using $q_{t+1}(y_n)$,  \eqref{eq:q_upd_gamma_col}.
\STATE $t\leftarrow t + 1$
\ENDWHILE
\STATE Estimate fused data labels using \eqref{eq:final_labels}.
\end{algorithmic}
\end{algorithm}

\subsection{Performance analysis}
Let $\tilde{C}$ with cardinality $|\tilde{C}|=\tilde{N}_C$ be a set comprising indices of data that take part in at least one constraint, and let $\tilde{C}^{c}$ with $|\tilde{C}^{c}| = \bar{N}_C$ denote its complement. The next theorem quantifies the performance of Alg.~\ref{alg:VBEM_instance}.
\begin{theorem}
	\label{thm:VBEM_instance}
	Consider the same setup as in Thm.~\ref{thm:VBEM}, instance level constraints collected in $\mathcal{C}$, and the initalization of Alg.~\ref{alg:VBEM_instance} satisfying
	$\max_{k} |q_{0}(y_n = k) - \mathds{1}(y_n = k)| \leq \varepsilon_{q,0}$ for all $k,n$.
	Then the following hold w.p. at least $1-\nu$: For iterations $t=1,\ldots,T$, update \eqref{eq:q_upd_MRF_y} yields for $n\in\tilde{\mathcal{C}}$
	\begin{align}
		&\max_{k} |q_t(y_n = k) - \mathds{1}(y_n = k)| \leq \tilde{\varepsilon}_{q,t}\label{eq:thm3_qupd1}\\
		& \tilde{\varepsilon}_{q,t} := \max_{n} K\exp\left(-U  - \eta W_n\right) = \max_{n} \varepsilon_{q,t}\exp(-\eta W_n)\notag \\
		& W_n:= N_{\rm ML,n}(1 - 2\varepsilon_{q,t-1}) - 2N_{\rm CL,n}\varepsilon_{q,t-1} + N_{\rm CL,n,min} \notag
	\end{align}
	with $N_{\rm ML,n}, N_{\rm CL,n}$ denoting the number of must- and cannot-link constraints for datum $n$ respectively, and $N_{\rm CL,n,min}:=\min_{k} N_{\rm CL,n,k}$, where $N_{\rm CL,n,k}$ is the number of cannot-link constraints of datum $n$, that belong to class $k$.
	For data without constraints, that is $n\in\tilde{\mathcal{C}}^{c}$
	\begin{align*}
		&\max_{k} |q_t(y_n = k) - \mathds{1}(y_n = k)| \leq \varepsilon_{q,t} = K\exp(-U),
	\end{align*}
	with $U$ as defined in Thm.~\ref{thm:VBEM}.
	For $k,k'=1,\ldots,K$, $m=1,\ldots,M$ updates \eqref{eq:q_upd_pi} and \eqref{eq:q_upd_gamma_col} yield respectively {\small
	\begin{align}
		& |\Expect[\pi_k] - \pi_k^*|\leq \epsilon_{\pi,k,t} := \label{eq:thm3_piupd} \frac{\tilde{N}_C\tilde{\varepsilon}_{q,t} + \bar{N}_C\varepsilon_{q,t} + Ng_\pi(\nu) + \alpha_{0,k} + \rho_\pi\bar{\alpha}_0}{N + \bar{\alpha}_0}  \\
		& |\Expect[\gamma_{k,k'}^{(m)}] - \gamma_{k,k'}^{*(m)}|\leq \epsilon_{\gamma,k,k',t}^{(m)} := \label{eq:thm3_gammaupd} \frac{2Ng_\gamma(\nu) + 2\tilde{N}_C\tilde{\varepsilon}_{q,t} + 2\bar{N}_C\varepsilon_{q,t} + \beta_{0,k,k'}^{(m)} + \bar{\beta}_{0,k}^{(m)} }{N\mu_m\pi_k^* - N\frac{g_\gamma(\nu)}{\gamma_{k,k'}^{*(m)}} - \tilde{N}_C\tilde{\varepsilon}_{q,t} - \bar{N}_C\varepsilon_{q,t} + \bar{\beta}_{k}^{(m)}}  
	\end{align}}
	and $\varepsilon_{\pi,t}:=\max_{k}\epsilon_{\pi,k,t}, \varepsilon_{\gamma,t}:= \max_{k,k',m} \epsilon_{\gamma,k,k',t}^{(m)}$.
\end{theorem}
Comparing Thm.~\ref{thm:VBEM_instance} to Thm.~\ref{thm:VBEM}, the error bounds for the labels involved in constraints introduced by Alg.~\ref{alg:VBEM_instance} will be smaller than the corresponding bounds by VBEM, as long as $\max_n W_n > 0$. This in turn reduces the error bounds for the parameters of interest, thus justifying the improved performance of Alg.~\ref{alg:VBEM_instance} when provided with good initialization. Such an initialization can be achieved by using spectral methods, or, by utilizing the output of the aforementioned unconstrained VBEM algorithm.

\subsection{Choosing $\eta.$}
\label{ssec:choosing_params}
Proper choice of $\eta$ is critical for the performance of the proposed algorithm, as it represents the weight assigned to the available constraints. Here, using the number of violated constraints $N_{\rm V} := \sum_{(n,n')\in\mathcal{C}_{\rm ML}} \mathds{1}(y_n \neq y_{n'}) + \sum_{(n,n')\in\mathcal{C}_{\rm CL}} \mathds{1}(y_n = y_{n'})$ as a proxy for performance, grid search can be used to select $\eta$ from a (ideally small) set of possible values $\mathcal{H}.$

\subsection{Selecting instance-level constraints}
\label{sssec:instance_constraint_select}
In some cases, acquiring pairwise constraints, can be costly and time consuming. This motivates judicious selection of data we would prefer to query constraints for. Here, we outline an approach for selecting constraints using only the annotator responses $\check{\mathbf{Y}}$. In addition to providing error bounds for VBEM under instance level constraints, Thm. \ref{thm:VBEM_instance} reveals how to select the $N_C$ constraints in $\mathcal{C}$. Equations \eqref{eq:thm3_piupd} and \eqref{eq:thm3_gammaupd} show that in order to minimize the error bounds of the priors and confusion matrices, the data with the largest label errors $|q(y_n = k) - \mathds{1}(y_n = k)|$ should be included in $\mathcal{C}.$ The smaller error bounds of \eqref{eq:thm3_piupd} and \eqref{eq:thm3_gammaupd} then result in smaller error bounds for the labels, thus improving the overall classification performance. Let $\tilde{\mathcal{C}}_u$ denote the set of data for which we wish to reduce the label error. In order to minimize the error $\tilde{\varepsilon}_q$ per element of $\tilde{\mathcal{C}}_u$, the term $W_n$ in the exponent of \eqref{eq:thm3_qupd1} should be maximized. To this end, data in $\tilde{\mathcal{C}}_u$ should be connected with instance-level constraints to data that exhibit small errors $\varepsilon_q$. Data connected to each $n\in\tilde{\mathcal{C}}_u$, are collected in the set $\tilde{\mathcal{C}}_c(n).$

Since the true label errors are not available, one has to resort to surrogates for them. One class of such surrogates, typically used in active learning \cite{active_survey2013}, are the so-called uncertainty measures. These quantities estimate how uncertain the classifier (in this case the crowd) is about each datum. Intuitively, data for which the crowd is highly uncertain are more likely to be misclassified. In our setup, such uncertainty measures can be obtained using the results provided by VBEM, and specifically the label posteriors $\{q_T(y_n)\}_{n=1}^{N}$. As an alternative, approximate posteriors can be found without using the VBEM algorithm, by taking a histogram of annotator responses per datum. Using these posteriors, one can measure the uncertainty of the crowd for each datum. Here, we opted for the so-called best-versus-second-best uncertainty measure, which per datum $n$ is given by
\begin{equation}
    H(\hat{y}_n) = \max_{k}q_T(y_n = k) - \max_{k'\neq k}q_T(y_n = k').
    \label{eq:uncertainty_measure}
\end{equation}
This quantity shows how close the two largest posteriors for each datum are; larger values imply that the crowd is highly certain for a datum, whereas smaller ones indicate uncertainty. 

Using the results of VBEM and the uncertainty measure in \eqref{eq:uncertainty_measure}, $\left \lfloor{N_C/K}\right \rfloor$  are randomly selected without replacement, to be included in $\tilde{\mathcal{C}}_u$, with probabilities $\lambda_n \propto 1-H(\hat{y}_n)$.
For each $n\in\tilde{\mathcal{C}}_u$, $K$ data are randomly selected, with probabilities $\lambda_n^{\dagger} \propto H(\hat{y}_n)$, and are included in $\tilde{\mathcal{C}}_{c}(n)$. Using this procedure, data in $\tilde{\mathcal{C}}_{c}(n), n\in\tilde{\mathcal{C}}_u$ are likely the ones that the crowd is certain of.  Finally, the links $\{(n,n'),  n\in\tilde{\mathcal{C}}_u, n'\in\tilde{\mathcal{C}}_{c}(n)\}$ are queried and included in the constraint set $\mathcal{C}$. Note here that each uncertain data point is connected to $K$ certain ones, to increase the likelihood that its label error will decrease.  
\begin{remark}
        The principles outlined in this subsection can be leveraged to develop active semi-supervised crowdsourcing algorithms. We defer such approaches to future work.
\end{remark}


\section{Numerical tests}
\label{sec:numerical_tests}
Performance of the proposed semi-supervised approach is evaluated here on several popular crowdsourcing datasets. Variational Bayes with instance-level constraints (abbreviated as \emph{VB - ILC}) [cf. Sec.~\ref{sec:semi-supervised}, Alg.~\ref{alg:VBEM_instance}] is compared to majority voting (abbreviated as \emph{MV}), the variational Bayes method of Sec.~\ref{ssec:IBCC_VBEM} that does not include side-information (abbreviated as \emph{VB}), and the EM algorithm of~\cite{dawid1979maximum} (abbreviated as \emph{DS}) that also does not utilize any side information. Here, simple baselines are chosen to showcase the importance of including constraints in the crowdsourcing task. { To further show the effect of pairwise constraints \emph{VB - ILC} is compared to Variational Bayes using label constraints (abbreviated as \emph{VB - LC}), as outlined at the beginning of Sec. \ref{sec:semi-supervised}. }
\begin{table}[tb]
	\centering
    \caption{Dataset properties}
	\label{tab:datasets}
	\begin{tabular}{|c | c | c | c | c|}
		\hline
		Dataset & $N$ & $M$ & $K$ & $\tilde{\delta}$  \\ \hline\hline
		RTE & $800$ & $164$ & $2$ & $48.78$ \\ \hline
		Sentence Polarity & $5,000$ & $203$ & $2$ & $136.68$ \\ \hline 
		Dog & $807$ & $109$ & $5$ & $74.03$  \\ \hline
		Web & $2,665$ & $177$ & $5$ & $87.94$ \\ \hline
	\end{tabular}
\end{table}

\begin{figure}[tb]
	\centering
	    \begin{minipage}{0.45\textwidth}
        \centering
        \includegraphics[width=0.9\textwidth]{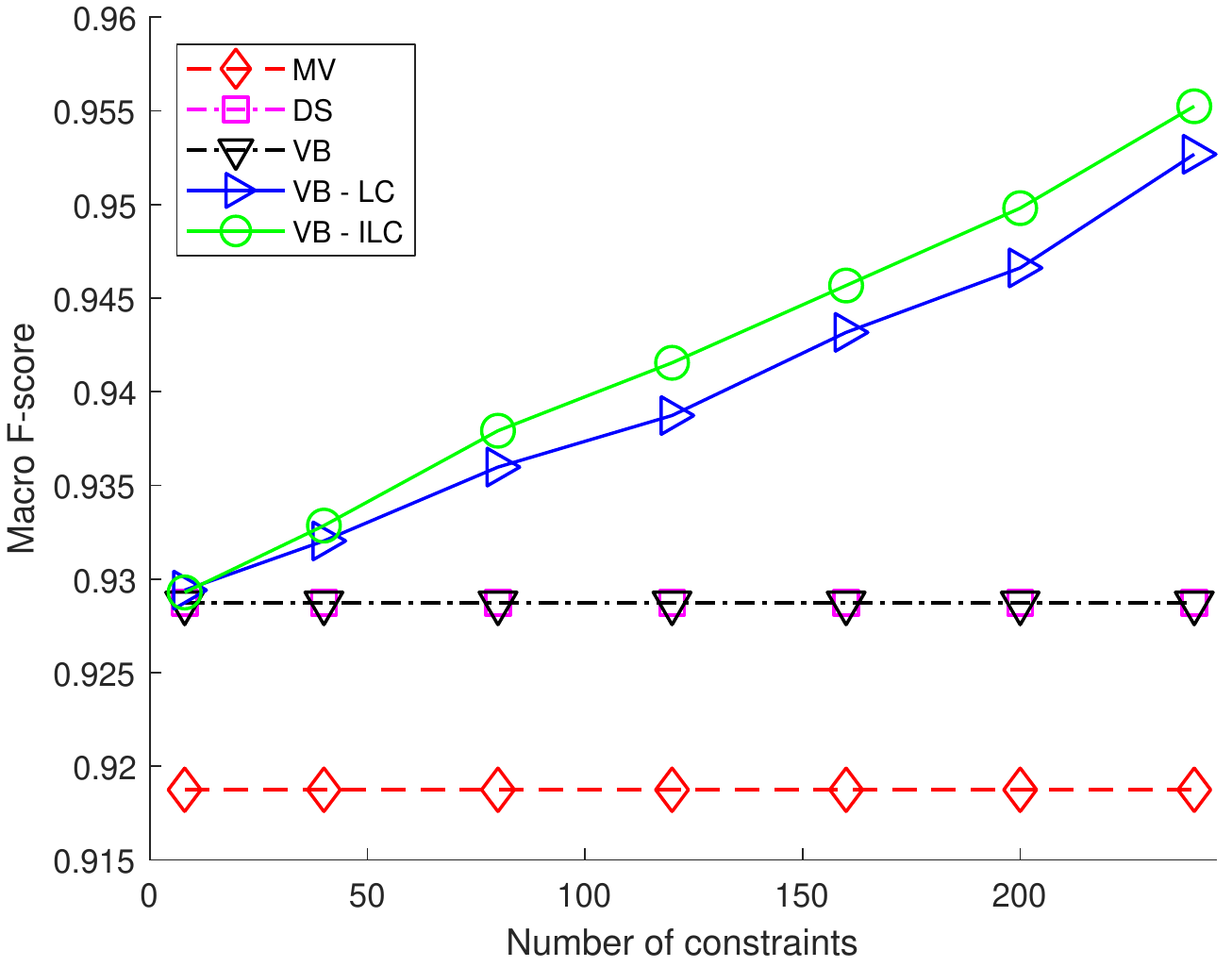} 
        \caption{Macro F-score for the RTE dataset, with the same number of constraints for \emph{VB-LC} and \emph{VB - ILC}} 	\label{fig:rte_macro}
    \end{minipage}\hfill
    \begin{minipage}{0.45\textwidth}
        \centering
        \includegraphics[width=0.9\textwidth]{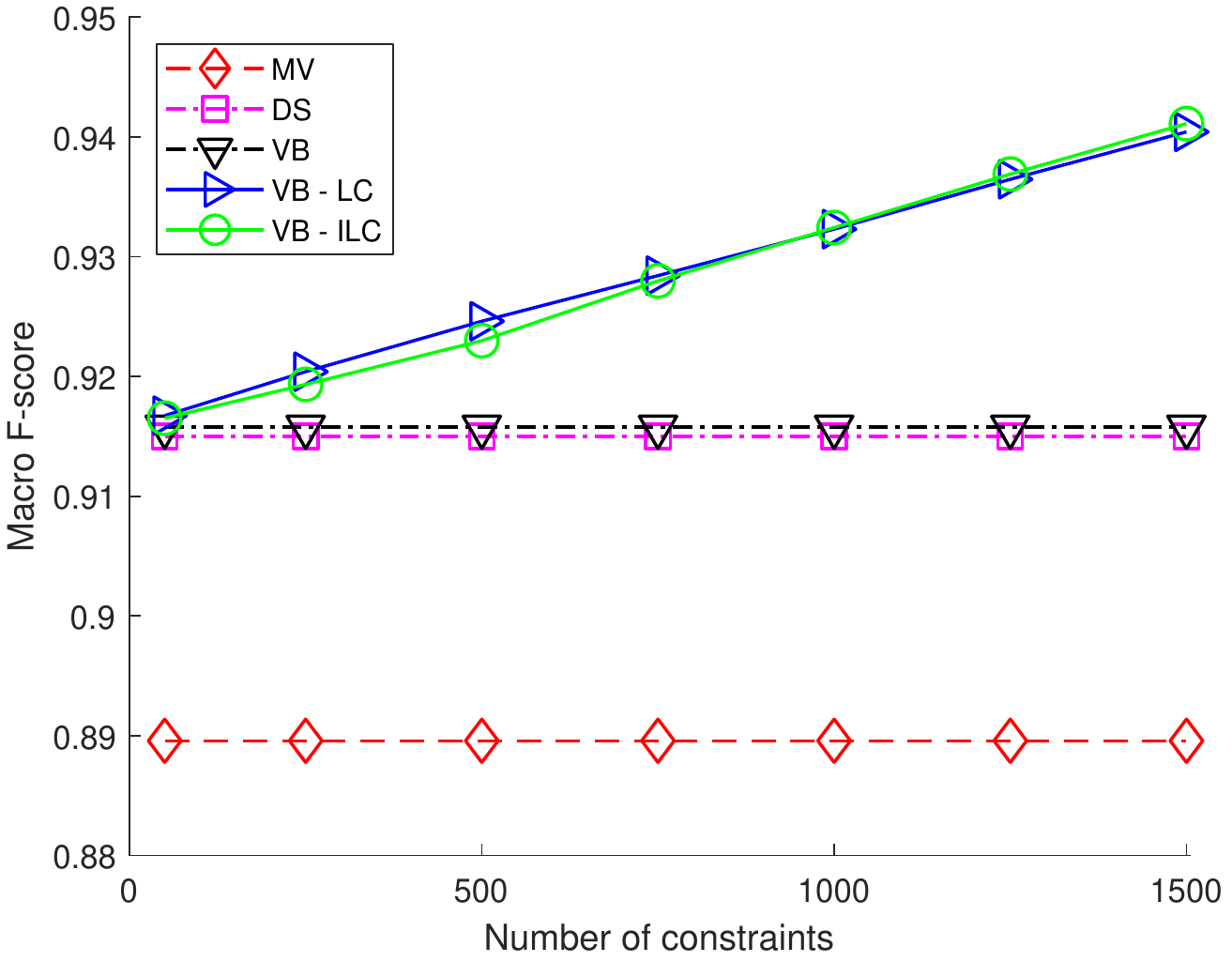} 
        \caption{Macro F-score for the Sentence Polarity dataset, with the same number of constraints for \emph{VB-LC} and \emph{VB - ILC}}
	\label{fig:senpol_macro}
    \end{minipage}
\end{figure}

As figure of merit, the macro-averaged F-score \cite{powers2011evaluation} is adopted to indicate the per-class performance of an algorithm. The datasets considered here are the  RTE \cite{cheapnfast}, Sentence Polarity \cite{musicgenre_senpoldata},  Dog \cite{imagenet}, and Web \cite{minimax_crowd}; see Table~\ref{tab:datasets} for their properties, { where $\tilde{\delta}$ denotes the average number of responses per annotator.}
For the datasets presented here, results indicating the micro-averaged F-score, alongside {dataset descriptions and} results with $6$ additional datasets are included in Appendix B of the supplementary material. 
MATLAB \cite{MATLAB:2019} was used throughout, and all results represent the average over $20$ Monte Carlo runs. In all experiments \emph{VB} and \emph{DS} are initialized using majority voting, while \emph{VB - LC} and \emph{VB - ILC} are initialized using the results of \emph{VB}, since, as shown in Thm.~\ref{thm:VBEM_instance} \emph{VB-ILC} requires good initialization. For the RTE, Dog, and Sentence Polarity datasets, the prior parameters are set as $\bm{\alpha_0} = \bm{1}$, where $\bm{1}$ denotes the all-ones vector, and $\bm{\beta}_{0,k}^{(m)}$ is a vector with $K$ at its $k$-th entry and ones everywhere else, for all $k,m$. For the Web dataset, all priors are set to uniform as this provides the best performance for the VB based algorithms. The values of $\eta$ for Alg.~\ref{alg:VBEM_instance} are chosen as described in Sec.~\ref{ssec:choosing_params} from the set $\mathcal{H}=\{0.01, 0.05, 0.1, 0.2, 0.5, 1, 2, 5, 10, 20, 100, 500\}.$
When instance-level constraints in $\mathcal{C}$ are provided to \emph{VB-ILC}, constraints that can be logically derived from the ones in $\mathcal{C}$ are also included. For example, if $(i,j)\in\mathcal{C}_{\rm ML}$ and $(j,k)\in\mathcal{C}_{\rm ML}$, a new must-link constraint $(i,k)$ will be added. Similarly, if $(i,j)\in\mathcal{C}_{\rm ML}$ and $(j,k)\in\mathcal{C}_{\rm CL}$, then a cannot-link constraint $(i,k)$ will be added.
\begin{figure}[tb]
	\centering
	    \begin{minipage}{0.45\textwidth}
        \centering
        \includegraphics[width=0.9\textwidth]{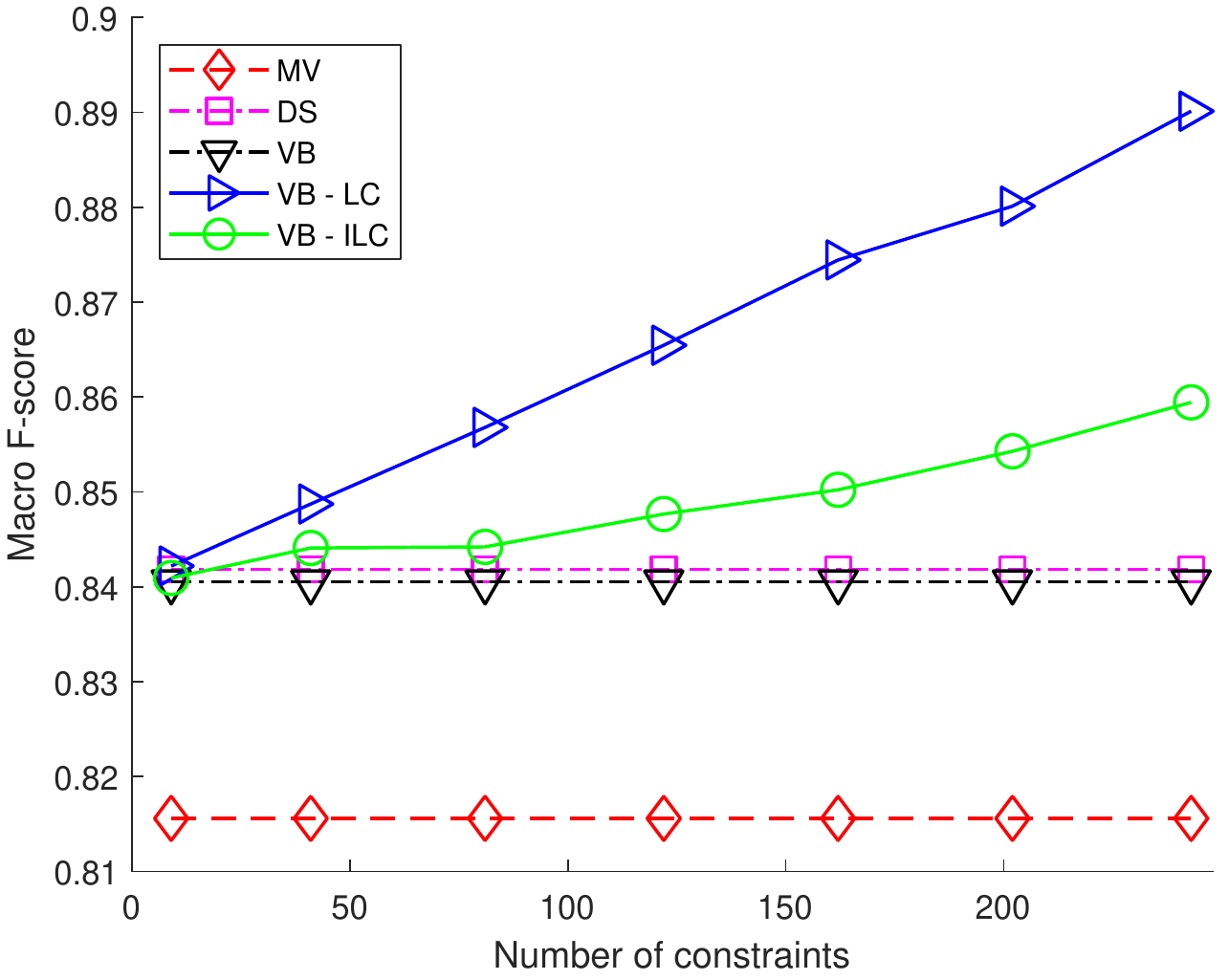} 
        \caption{Macro F-score for the Dog dataset, with the same number of constraints for \emph{VB-LC} and \emph{VB - ILC}}
	\label{fig:dog_macro}
    \end{minipage}\hfill
    \begin{minipage}{0.45\textwidth}
        \centering
        \includegraphics[width=0.9\textwidth]{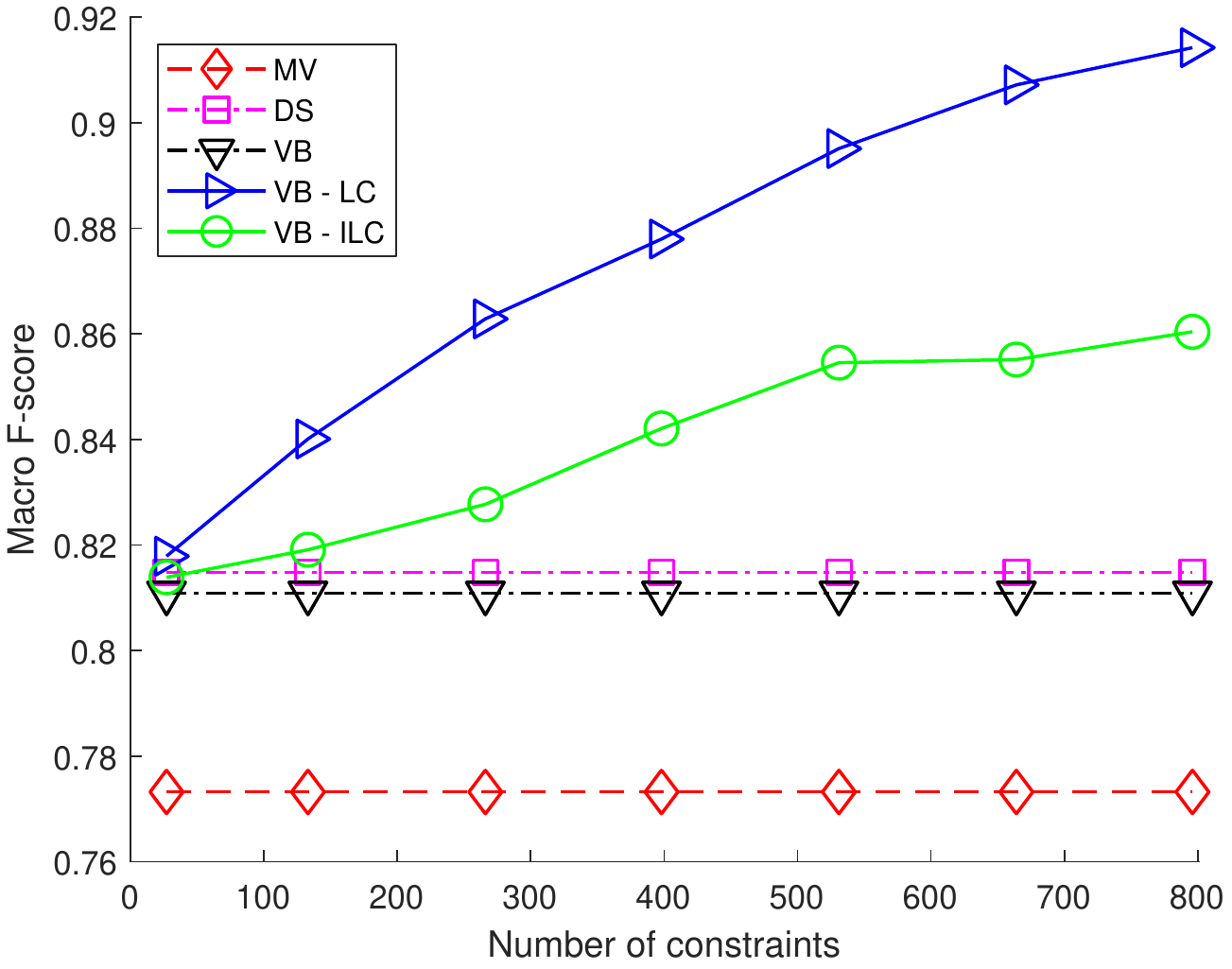} 
        \caption{Macro F-score for the Web dataset, with the same number of constraints for \emph{VB-LC} and \emph{VB - ILC}}
	\label{fig:web_macro}
    \end{minipage}
\end{figure}

\begin{figure}[tb]
	\centering
	    \begin{minipage}{0.45\textwidth}
        \centering
        \includegraphics[width=0.9\textwidth]{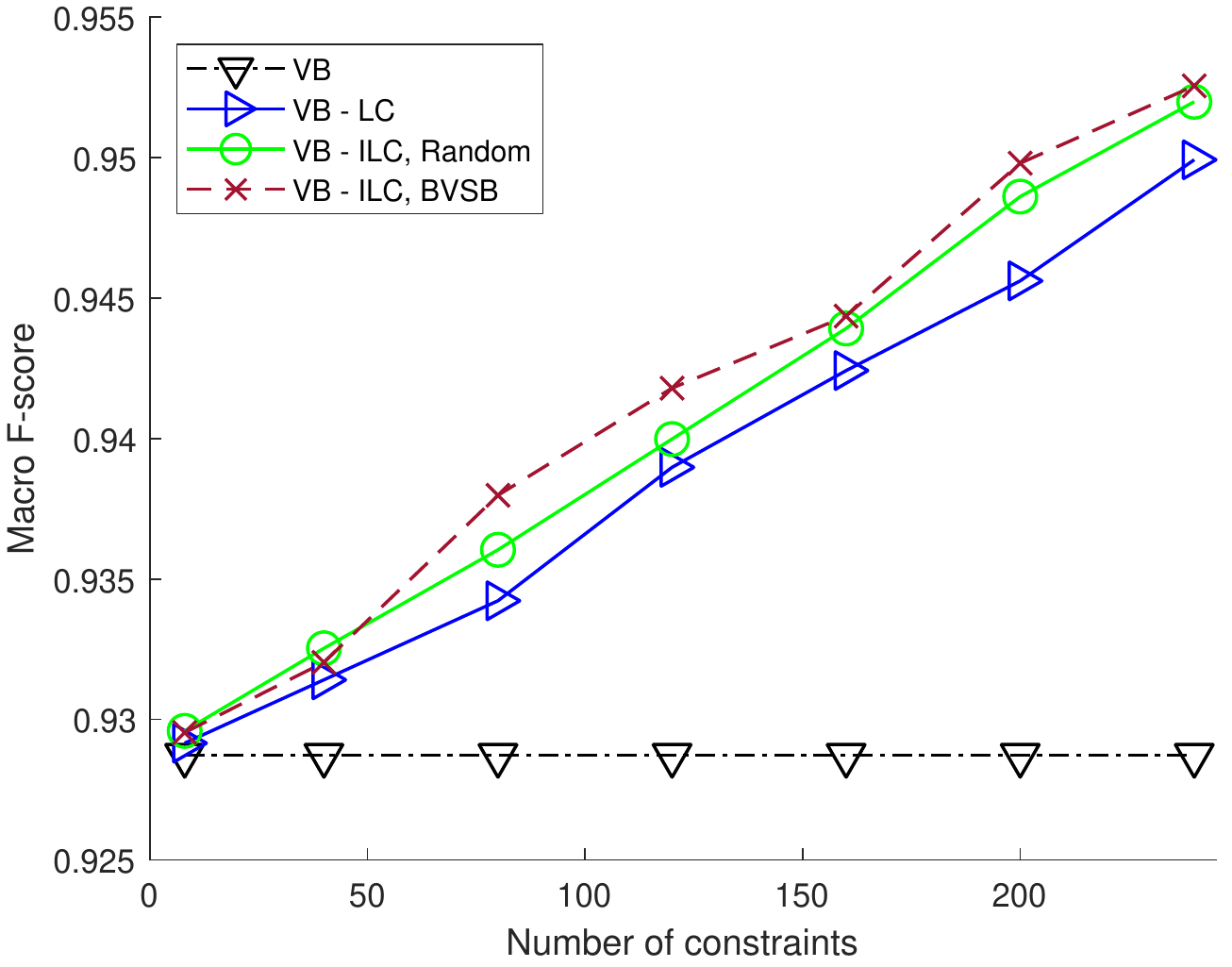} 
	\caption{Macro F-score for the RTE dataset, with random and uncertainty based constraint selection for  \emph{VB-ILC}}
	\label{fig:rte_macro_exp2}
    \end{minipage}\hfill
    \begin{minipage}{0.45\textwidth}
        \centering
	        \includegraphics[width=0.9\textwidth]{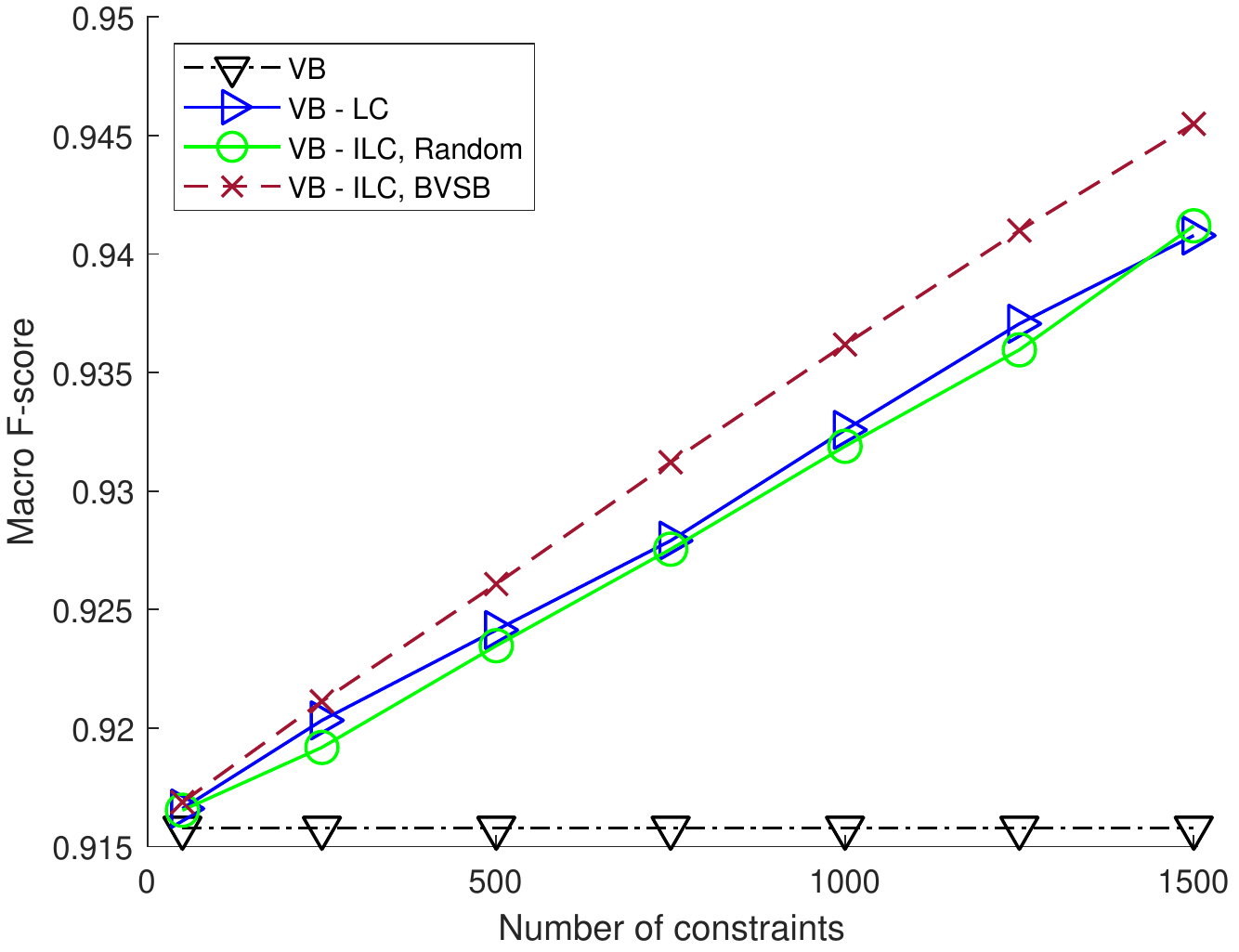} 
	\caption{Macro F-score for the Sentence polarity dataset, with random and uncertainty based constraint selection for \emph{VB-ILC}}
	\label{fig:senpol_macro_exp2}
    \end{minipage}
\end{figure}

\begin{figure}[tb]
	\centering
	    \begin{minipage}{0.45\textwidth}
        \centering
        \includegraphics[width=0.9\textwidth]{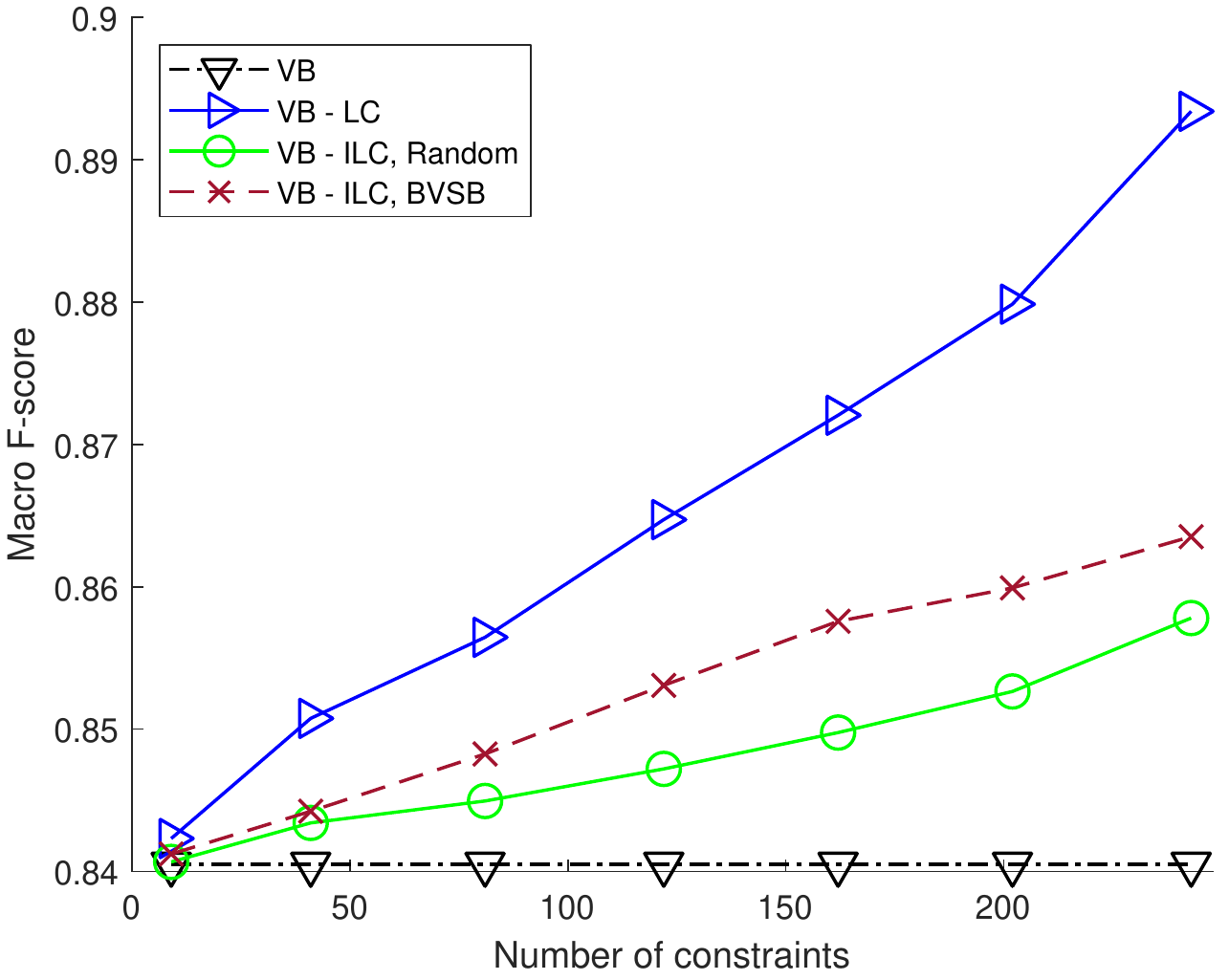} 
	\caption{Macro F-score for the Dog dataset, with random and uncertainty based constraint selection for \emph{VB-ILC}}
	\label{fig:dog_macro_exp2}
    \end{minipage}\hfill
    \begin{minipage}{0.45\textwidth}
        \centering
        \includegraphics[width=0.9\textwidth]{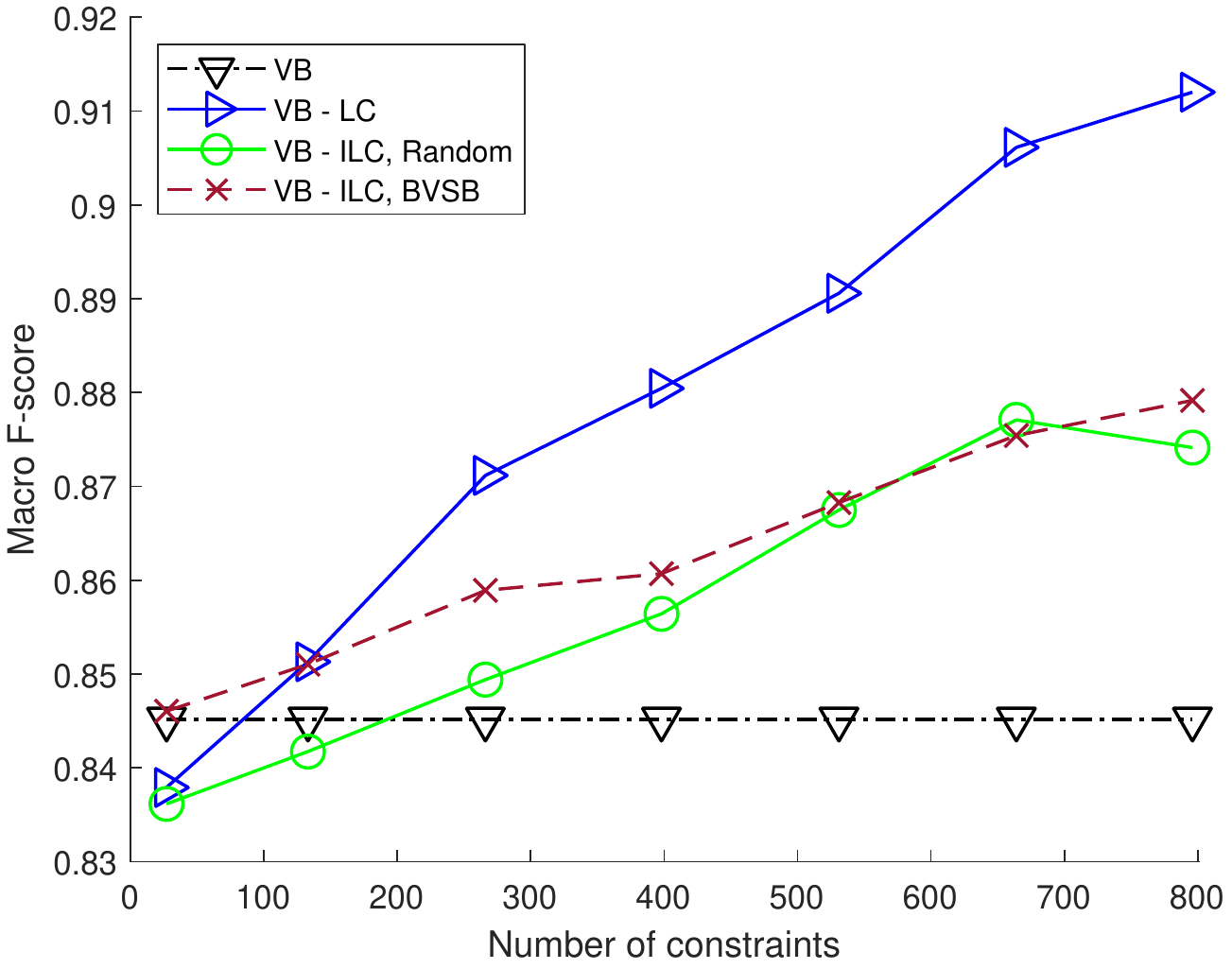} 
	\caption{Macro F-score for the Web dataset, with random and uncertainty based constraint selection for \emph{VB-ILC}}
	\label{fig:web_macro_exp2}
    \end{minipage}
\end{figure}

\begin{figure}[tb]
	\centering
	    \begin{minipage}{0.45\textwidth}
        \centering
        \includegraphics[width=0.9\textwidth]{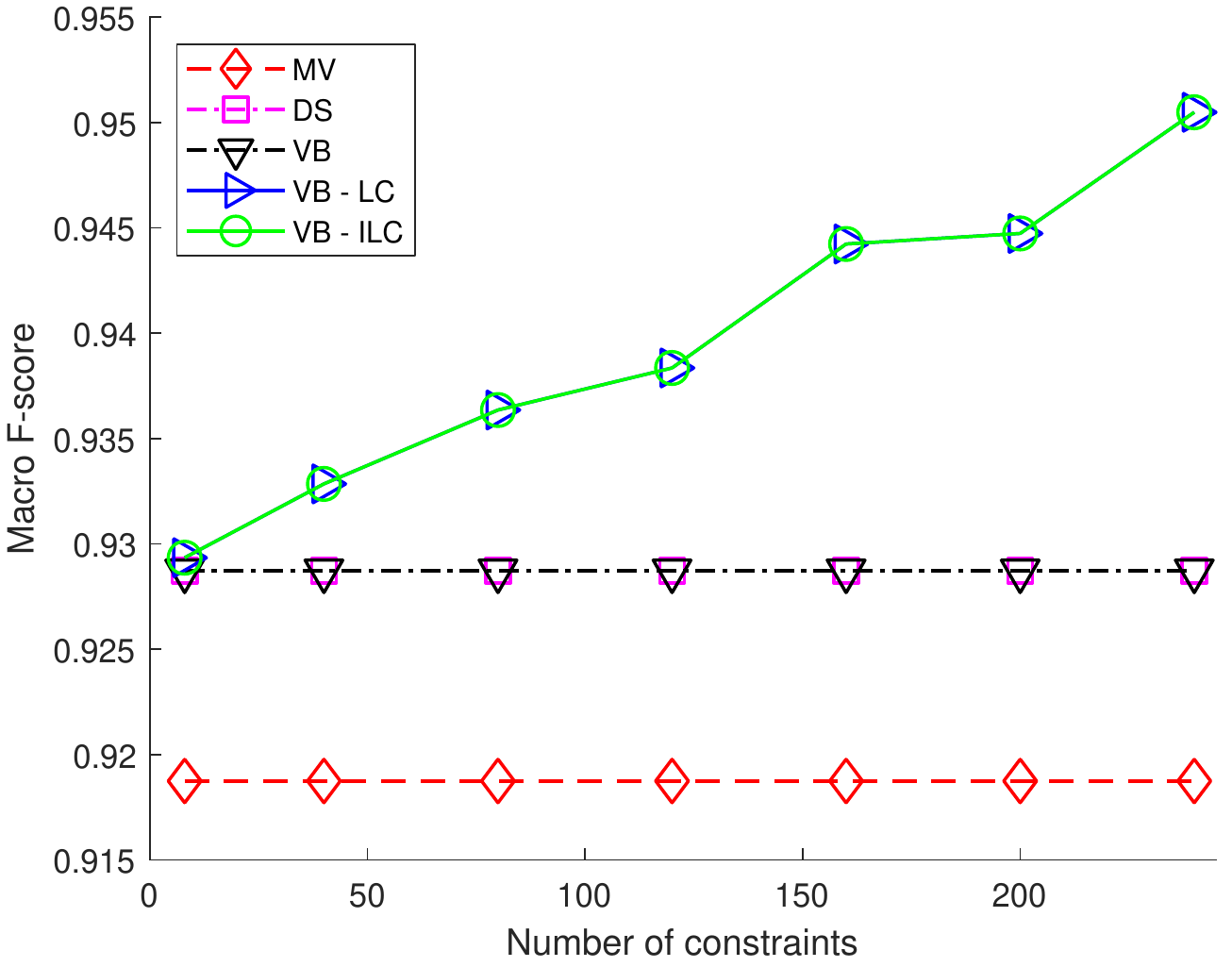} 
	\caption{Macro F-score for the RTE dataset, with label derived constraints for \emph{VB-ILC}}
	\label{fig:rte_macro_exp3}
    \end{minipage}\hfill
    \begin{minipage}{0.45\textwidth}
        \centering
        \includegraphics[width=0.9\textwidth]{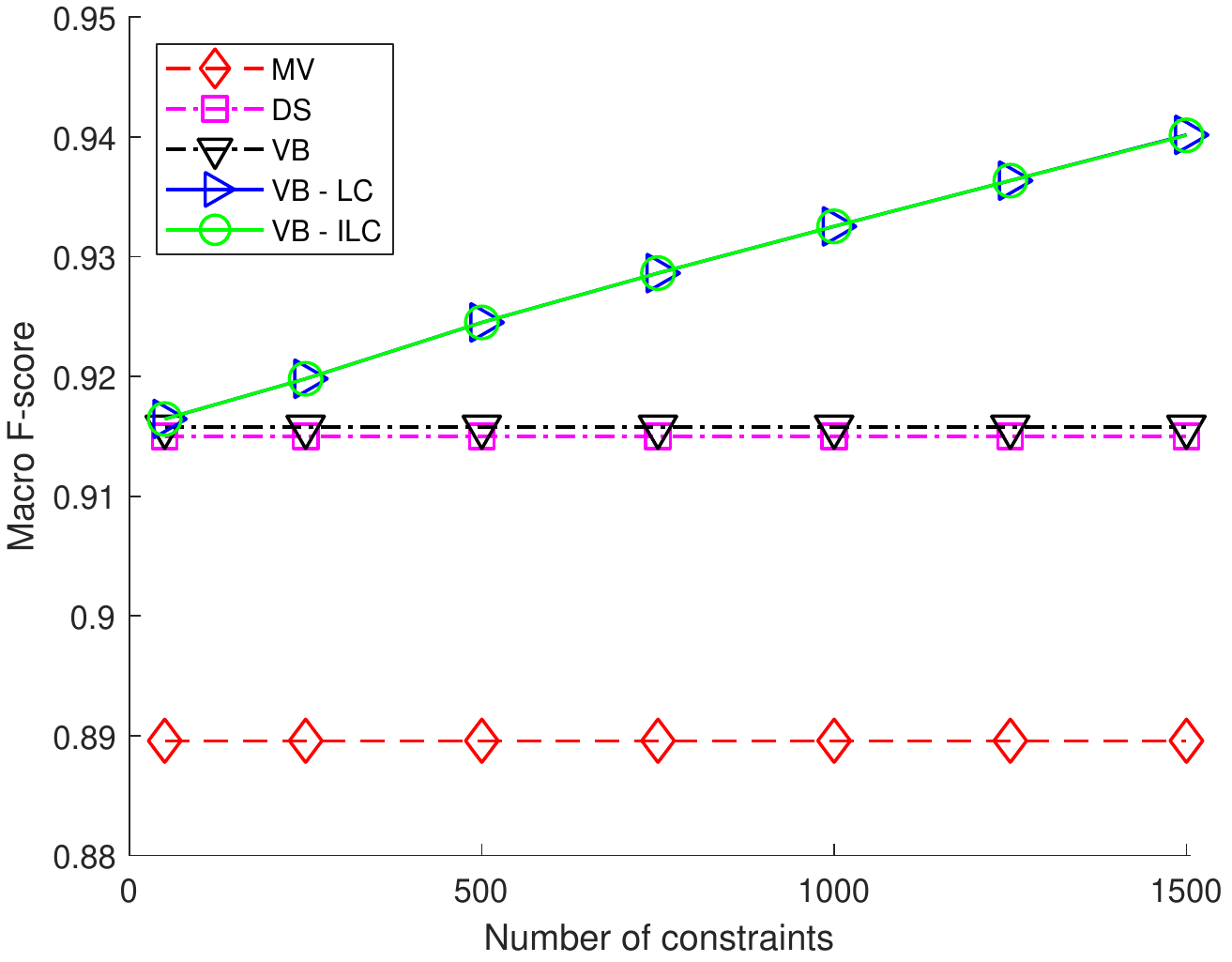} 
	\caption{Macro F-score for the Sentence Polarity dataset, with label derived constraints for \emph{VB-ILC}}
	\label{fig:senpol_macro_exp3}
    \end{minipage}
\end{figure}

\begin{figure}[tb]
	\centering
	    \begin{minipage}{0.45\textwidth}
        \centering
        \includegraphics[width=0.9\textwidth]{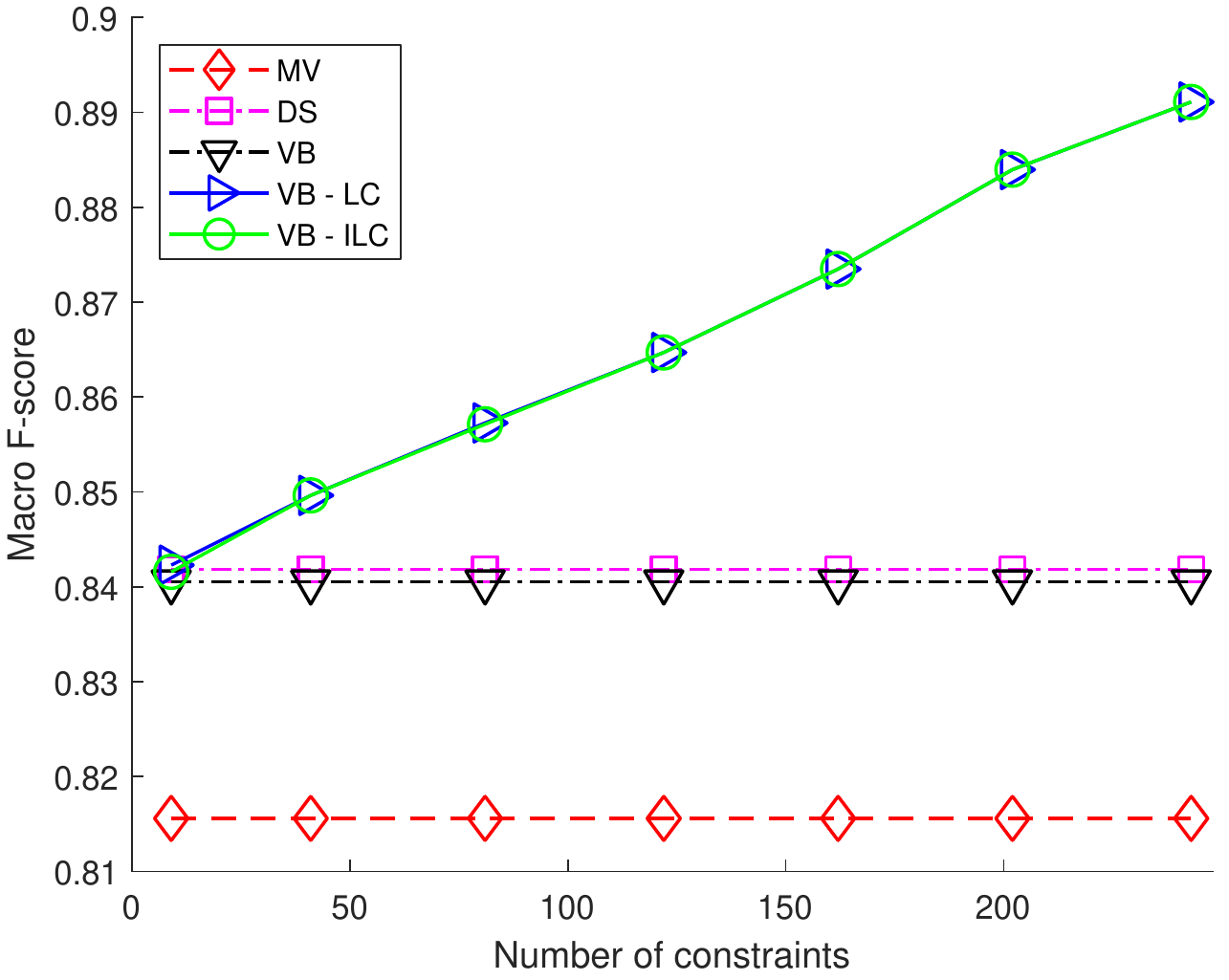} 
	\caption{Macro F-score for the Dog dataset, with label derived constraints for \emph{VB-ILC}}
	\label{fig:dog_macro_exp3}
    \end{minipage}\hfill
    \begin{minipage}{0.45\textwidth}
        \centering
        \includegraphics[width=0.9\textwidth]{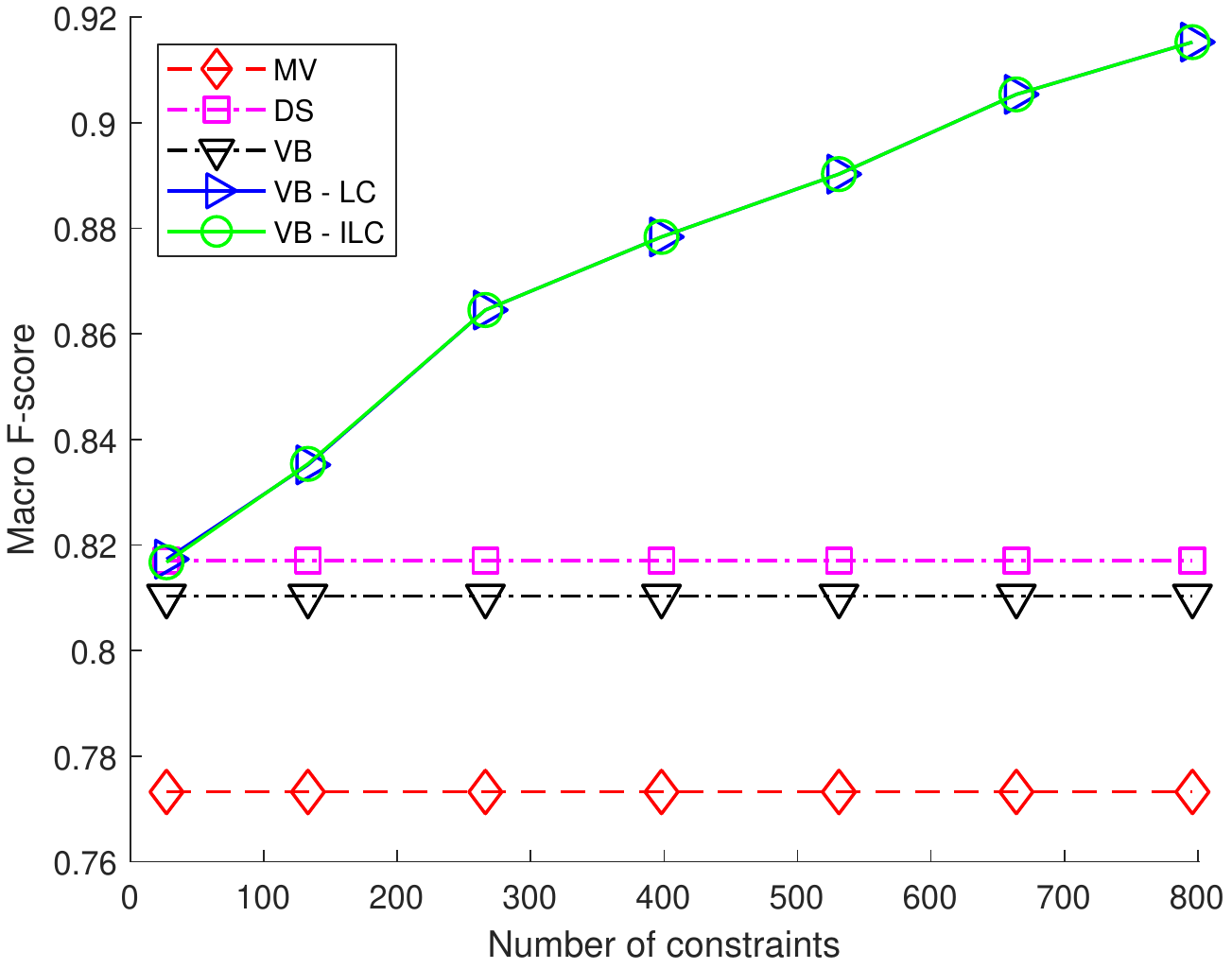} 
	\caption{Macro F-score for the Web dataset, with label derived constraints for \emph{VB-ILC}}
	\label{fig:web_macro_exp3}
    \end{minipage}
\end{figure}

{Figs. \ref{fig:rte_macro}-\ref{fig:web_macro} show classification performance when \emph{VB-LC} and \emph{VB-ILC} are provided with $N_C$ randomly selected constraints, that is $N_C$ label constraints for \emph{VB-LC} and $N_C$ instance-level constraints for \emph{VB-ILC}. \emph{VB - LC} and \emph{VB - ILC} exhibit improved performance as the number of constraints increases. As expected, \emph{VB - LC} outperforms \emph{VB - ILC}, when $K>2$, since label information is stronger than instance-level information. Interestingly, for $K=2$, \emph{VB-ILC} exhibits comparable performance to \emph{VB-LC} in the Sentence polarity dataset, and outperforms \emph{VB-LC} in the RTE dataset. Nevertheless, \emph{VB-ILC} outperforms its unsupervised counterparts as $N_C$ increases, corroborating that even relatively weak constraints are useful. The performance of the constraint selection scheme of Sec. \ref{sssec:instance_constraint_select} is evaluated in Figs.~\ref{fig:rte_macro_exp2}-\ref{fig:web_macro_exp2}. The uncertainty sampling based variant of  \emph{VB - ILC} is denoted as \emph{VB - ILC, BVSB}. Uncertainty sampling for selecting label constraints is clearly beneficial, as it provides performance gains compared to randomly selecting constraints in all considered datasets.  Figs.~\ref{fig:rte_macro_exp3}-\ref{fig:web_macro_exp3} show the effect of providing \emph{VB-ILC} with constraints derived from $N_C$ label constraints. When provided with $N_C$ label constraints from $K$ classes, and with $z_k$ denoting the proportion of constraints from class $k$, the resulting number of must-link constraints is $N_{\rm ML} = \sum_{k=1}^{K} {N_C z_k \choose 2}$, since for each class every two points must be connected. The number of  cannot-link constraints is $N_{\rm CL} = \sum_{k=1}^{K}\sum_{k'=1}^{K-1} N_C^2 z_k z_{k'+1}$, as for each class every point must be connected to all points belonging to other classes. In this case, \emph{VB - ILC} almost matches the performance of \emph{VB - LC}. This indicates that when provided with an adequate number of instance-level constraints, \emph{VB - ILC} performs as well as if label constraints had been provided.}

\section{Conclusions}
\label{sec:conclusions}
This paper investigated constrained crowdsourcing with pairwise constraints, that encode relationships between data. 
The performance of the proposed algorithm was analytically and empirically evaluated on popular real crowdsourcing datasets.
Future research will involve distributed and online implementations of the proposed algorithm, and other types of constraints alongside semi-supervised crowdsourcing with dependent annotators, and dependent data.


\end{document}


\title{Supplementary material for Bayesian Crowdsourcing with Constraints}


\maketitle

\titleformat{\section}{\normalfont\Large\bfseries}{\appendixname~\thesection.}{1em}{}

\begin{appendices}
\section{Technical proofs}
To prove Theorems 1 and 2, we will follow a similar procedure to the one used in \cite{zhang2014spectral} for deriving error bounds for the Expectation maximization algorithm. The iteration subscript is omitted in this Appendix for brevity. Before proceeding with the proofs of the theorems, a few supporting lemmata are provided.
\subsection{Supporting lemmata}
The following lemmata will be used to assist in the proofs of the propositions and theorems. For the remainder of this supplementary material define an annotators probability of providing a response as $\mu_m := \prob(\check{y}_n^{(m)} \neq 0)$, and $D_\gamma := \min_{k\neq k'}\frac{1}{M}\sum_{m=1}^{M}\mu_m {\rm KL}(\gamma_{k}^{*(m)},\gamma_{k'}^{*(m)})$, with $\rm KL$ denoting the Kullback-Leibler divergence~\cite{cover2012elements}. 
Lemma~\ref{lemma:concentration} provides a probabilistic characterization of the dataset as well as annotator responses. Lemmata~\ref{lemma:lnpi} and \ref{lemma:lngamma} establish bounds related to the variational updates of the parameters $\bm{\pi}, \mathbf{\Gamma}$.
\begin{lemma}
\label{lemma:concentration}
Consider the events
\begin{align}
    &\mathcal{E}_1: \sum_{m=1}^{M}\sum_{k'=1}^{K}\delta_{n,k'}^{(m)}\ln\frac{\gamma_{y_n,k'}^{*(m)}}{\gamma_{k,k'}^{*(m)}} \geq MD_\gamma/2 \quad \text{ for } k\neq y_n, n=1,\ldots,N \\
    &\mathcal{E}_2: \left| \sum_{n=1}^{N}\mathbbm{1}(y_n = k)\delta_{n,k'}^{(m)} - N\mu_m\pi_k^*\gamma_{k,k'}^{*(m)}   \right| \leq N t_{k,k'}^{(m)} \quad \text{ for } k,k'=1,\ldots,K, m=1,\ldots,M \\
    &\mathcal{E}_3: \left| \sum_{n=1}^{N}\mathbbm{1}(y_n = k)\mathbbm{1}(\check{y}_{n}^{(m)} \neq 0) - N\mu_m\pi_k^*\right| \leq N\frac{t_{k,k'}^{(m)}}{\gamma_{k,k'}^{*(m)}} \quad \text{ for } k,k'=1,\ldots,K, m=1,\ldots,M \\
    &\mathcal{E}_4: \left|\sum_{n=1}^{N}\mathbbm{1}(y_n = k)  -N\pi_k^* \right|
    \leq N r_k \quad \text{ for } k=1,\ldots,K, 
\end{align}
where $t_{k,k'}^{(m)} \leq \mu_m\pi_k^*\gamma_{k,k'}^{*(m)},$ and $r_k\leq \pi_k^*.$
Then $\prob(\mathcal{E}_1\cap\mathcal{E}_2\cap\mathcal{E}_3\cap\mathcal{E}_4) \geq 1-\nu$, with
\begin{equation}
    \nu :=  KN\exp\left(-\frac{MD_\gamma}{33\ln\rho_\gamma}\right) + \sum_{m=1}^{M}\sum_{k=1}^{K}4\exp\left(-\frac{N (t_{k,k'}^{(m)})^2}{3\pi_k^*\mu_m\gamma_{k,k'}^{*(m)}}\right)+ \sum_{k=1}^{K}2\exp\left(-\frac{N r_k^2}{3\pi_k^*}\right) \label{eq:nu}
\end{equation}
\end{lemma}
\begin{proof}
In \cite[Lemma 10]{zhang2014spectral} the following concentration bounds for events $\mathcal{E}_1,\mathcal{E}_2$ and $\mathcal{E}_3$ are established
\begin{align}
    & \prob(\mathcal{E}_1) \geq 1 - KN\exp\left(-\frac{MD_\gamma}{33\ln\rho_\gamma}\right) \label{eq:p_event_1}\\
    & \prob(\mathcal{E}_2) \geq 1 -  2\sum_{m=1}^{M}\sum_{k=1}^{K}\exp\left(-\frac{N (t_{k,k'}^{(m)})^2}{3\pi_k^*\mu_m\gamma_{k,k'}^{*(m)}} \right) \label{eq:p_event_2}\\
    & \prob(\mathcal{E}_3) \geq 1 - 2\sum_{m=1}^{M}\sum_{k=1}^{K}\exp\left(-\frac{N (t_{k,k'}^{(m)})^2}{3\pi_k^*\mu_m\gamma_{k,k'}^{*(m)}} \right). \label{eq:p_event_3}
\end{align}
For event $\mathcal{E}_4$, use of the Chernoff bound~\cite{concentration} yields, for $r_k > 0$,
\begin{align}
    \prob\left(\left|\sum_{n=1}^{N}\mathbbm{1}(y_n = k) -N\pi_k^* \right| \geq N r_k\right) \leq 2\exp\left(-\frac{Nr_k^2}{3\pi_k^*} \right).
\end{align}
Then 
\begin{equation}
    \label{eq:p_event_4}
    1 - \prob(\mathcal{E}_4) \geq \sum_{k=1}^{K}2\exp\left(-\frac{N r_k^2}{3\pi_k^*} \right).
\end{equation}
Combining \eqref{eq:p_event_1},\eqref{eq:p_event_2},\eqref{eq:p_event_3} and \eqref{eq:p_event_4} for all $k,k',m$ one recovers
\begin{align}
    & \prob(\mathcal{E}_1\cap\mathcal{E}_2\cap\mathcal{E}_3\cap\mathcal{E}_4)\geq 1 - \sum_{i=1}^{4}(1-\prob(\mathcal{E}_i)) \notag\\ & \geq 1 - KN\exp\left(-\frac{MD_\gamma}{33\ln\rho_\gamma}\right) -  \sum_{m=1}^{M}\sum_{k=1}^{K}4\exp\left(-\frac{N (t_{k,k'}^{(m)})^2}{3\pi_k^*\mu_m\gamma_{k,k'}^{*(m)}}\right) - \sum_{k=1}^{K}2\exp\left(-\frac{Nr_k^2}{3\pi_k^*}\right).
\end{align}
Finally, note that values (or bounds) for $t_{k',k}^{(m)}$ and $r_k$ can be derived in a similar manner to \cite{zhang2014spectral}.
\end{proof}

\begin{lemma}
\label{lemma:lnpi}
Suppose that $\pi_k^*\geq\rho_\pi>0$ and $|\Expect[\pi_k] - \pi_k^*|\leq\varepsilon_\pi$, for $k=1,\ldots,K,$ and some $\varepsilon_{\pi} >0.$ Then 
\begin{equation}
    \Expect\left[\ln\frac{\pi_k}{\pi_k^*} \right] \geq f_\pi(\varepsilon_\pi)  \quad k=1,\ldots,K
\end{equation}
with $f_\pi(\varepsilon_\pi)$ being a decreasing function of $\varepsilon_\pi$.
\end{lemma}
\begin{proof}
Since $q(\bm{\pi})$ is a Dirichlet distribution parametrized by $\bm{\alpha}=[\alpha_1,\ldots,\alpha_K]$, it holds that
\begin{equation}
    \Expect\left[\ln\pi_k\right] = \psi(\alpha_k) - \psi(\bar{\alpha}).
    \label{eq:pi_psi}
\end{equation}
For $x>1/2$ the digamma function lies in the interval $\psi(x)\in\left(\ln(x-1/2),\ln x\right).$ Using this fact, \eqref{eq:pi_psi} can be bounded as
\begin{equation}
    \Expect\left[\ln\pi_k\right] = \psi(\alpha_k) - \psi(\bar{\alpha}) \geq 
    \ln(\alpha_k - 1/2) - \ln(\bar{\alpha}) = \ln\left(\frac{\alpha_k}{\bar{\alpha}} - \frac{1}{2\bar{\alpha}} \right) = 
    \ln\left(\Expect[\pi_k] - \frac{1}{2(N + \bar{\alpha}_0)}\right)
    \label{eq:Elnpi_k}
\end{equation}
where we have assumed that $\alpha_{0,k}\geq 1/2$ and the last equality is due to $\Expect[\pi_k] = \alpha_k/\bar{\alpha}$ and $\bar{\alpha} = \sum_{k'}\alpha_{k'} = \sum_{k'}(N_{k'} + \alpha_{0,k'}) = N + \bar{\alpha}_0.$ Then
\begin{equation}
    \Expect\left[\ln\frac{{\pi}_k}{{\pi}_k^*} \right] = \Expect[\ln\pi_k] - \ln\pi_k^*\geq \ln\left(\frac{\Expect[\pi_k]}{\pi_k^*} - \frac{1}{2\pi_k^*(N+\bar{\alpha}_0)}\right).
\end{equation}
From the assumption of the lemma, we have that 
\begin{equation}
    |\Expect[\pi_k] - \pi_k^*|\leq\varepsilon_\pi \Rightarrow 1 - \frac{\varepsilon_\pi}{\rho_\pi}\leq\frac{\Expect[\pi_k]}{\pi_k^*}\leq 1 + \frac{\varepsilon_\pi}{\rho_\pi}
    \label{eq:pi_abs_ineq}
\end{equation}
therefore 
\begin{equation}
    \Expect\left[\ln\frac{{\pi}_k}{{\pi}_k^*} \right] \geq
    \ln\left(\frac{\rho_\pi - \varepsilon_\pi}{\rho_\pi} - \frac{1}{2\rho_\pi(N+\bar{\alpha}_0)}\right) :=f_\pi(\varepsilon_\pi)
\end{equation}
which completes the proof.
\end{proof}

\begin{lemma}
\label{lemma:lngamma}
Suppose that $\gamma_{k,k'}^{*(m)}\geq\rho_\gamma>0$ and $|\Expect[\gamma_{k,k'}^{(m)}] - \gamma_{k,k'}^{*(m)}|\leq\varepsilon_\gamma$, for $k,k'=1,\ldots,K, m=1,\ldots,M,$ and some $\varepsilon_\gamma>0$. Then 
\begin{equation}
    \Expect\left[\ln\frac{\gamma_{k,k'}^{(m)}}{\gamma_{k,k'}^{*(m)}} \right] \geq f_\gamma(\varepsilon_\gamma)  \quad k,k'=1,\ldots,K \quad m=1,\ldots,M
\end{equation}
with $f_\gamma(\varepsilon_\gamma)$ being a decreasing function of $\varepsilon_\gamma$.
\end{lemma}
\begin{proof}
Using similar arguments as in Lemma~\ref{lemma:lnpi}, one arrives at
\begin{equation}
    \Expect\left[ \ln\frac{\gamma_{k,k'}^{(m)}}{\gamma_{k,k'}^{*(m)}} \right]\geq
    \ln\left(\frac{\rho_\gamma - \varepsilon_\gamma}{\rho_\gamma} - \frac{1}{2\rho_\gamma\bar{\beta}_{0}^{(m)}}   \right):=f_\gamma(\varepsilon_\gamma).
\end{equation}
\end{proof}

\subsection{Proof of Theorem 1}
In order to prove Thm.~1, Lemma~\ref{lemma:q_y} establishes an error bound for the variational updates of labels $q(y_n)$, whereas Lemmata~\ref{lemma:pi_err} and~\ref{lemma:gamma_err} establish error bounds for the updates of the parameters $\bm{\pi}$ and $\mathbf{\Gamma}.$ 
For the next lemmata we define $D_\pi:= \min_{n,k\neq y_n} \ln\frac{\pi_{y_n}^*}{\pi_k^*}$, and we assume that the event $\mathcal{E}_1\cap\mathcal{E}_2\cap\mathcal{E}_3\cap\mathcal{E}_4$ from Lemma~\ref{lemma:concentration} holds.  
\begin{lemma}
\label{lemma:q_y}
Suppose that $\mathcal{E}_1$ holds [cf. Lemma~\ref{lemma:concentration}] and that, $\gamma_{k,k'}^{*(m)}\geq \rho_\gamma>0$, $|\Expect[\gamma_{k,k'}^{(m)}] - \gamma_{k,k'}^{*(m)}|\leq\varepsilon_\gamma$ for all $k,k',m$ and some $\varepsilon_\gamma >0$, and  $\pi_k^{*}\geq\rho_\pi>0$, $|\Expect[\pi_k] - \pi_k^*|\leq\varepsilon_\pi$  for all $k$ and some $\varepsilon_\pi>0$. Then under the variational update for $q(y_n)$ the following holds for $n=1,\ldots,N$
\begin{align}
    &\max_{k} |q(y_n = k) - \mathbbm{1}(y_n = k)| \leq
    K\exp(-U) \\
    & \notag U:= D_\pi+2f_\pi(\varepsilon_\pi) + M\left(D_\gamma/2 + 2f_\gamma (\varepsilon_\gamma)\right)
\end{align}
\end{lemma}
\begin{proof}
For a single datum $n$, let $k = y_n$ be the true label of datum $n$. Consider an index $\ell\neq k$ and define the following quantity 
\begin{align}
    A_\ell & := \Expect[\ln\frac{\pi_{k}}{\pi_\ell}] + \sum_{m=1}^{M}\sum_{k'=1}^{K}\delta_{n,k'}^{(m)}\Expect\left[\ln\frac{\gamma_{k,k'}^{(m)}}{\gamma_{\ell,k'}^{(m)}}\right] \\ & = \ln\frac{\pi_{k}^{*}}{\pi_\ell^{*}} + \Expect\left[\ln\frac{\pi_{k}}{\pi_{k}^{*}}\right] - \Expect\left[\ln\frac{\pi_\ell}{\pi_\ell^*}\right]
    + \sum_{m=1}^{M}\sum_{k'=1}^{K}\delta_{n,k'}^{(m)}\ln\frac{\gamma_{k,k'}^{*(m)}}{\gamma_{\ell,k'}^{*(m)}} + \sum_{m=1}^{M}\sum_{k'=1}^{K}\delta_{n,k'}^{(m)}\left[\Expect\left[\ln\frac{\gamma_{k,k'}^{(m)}}{\gamma_{k,k'}^{*(m)}}\right] - \Expect\left[\ln\frac{\gamma_{\ell,k'}^{(m)}}{\gamma_{\ell,k'}^{*(m)}}\right] \right].
\end{align}
From the results of lemmata \ref{lemma:lnpi}, \ref{lemma:concentration} (event $\mathcal{E}_1$) and \ref{lemma:lngamma} $A_\ell$ can be lower-bounded as
\begin{equation}
    A_\ell \geq D_\pi + 2f_{\pi}(\varepsilon_\pi) + M(D_\gamma/2 + 2f_\gamma(\varepsilon_\gamma)). \label{eq:A_k_lower_bound}
\end{equation}
Now note that the variational update for $q(y_n = \ell)$ is
\begin{equation}
    q(y_n = \ell) = \frac{1}{Z_q}\exp\left(\Expect[\ln\pi_\ell] + \sum_m\sum_{k'}\delta_{n,k'}^{(m)}\Expect[\ln\gamma_{\ell,k'}^{(m)}]\right)
    \label{eq:q_upd}
\end{equation}
where $Z_q$ is the appropriate normalization constant, therefore 
\begin{align}
    A_\ell = \Expect[\ln\pi_{k}] + \sum_m\sum_{k'}\delta_{n,k'}^{(m)}\Expect[\ln\gamma_{k,k'}^{(m)}] - \Expect[\ln\pi_\ell] - \sum_m\sum_{k'}\delta_{n,k'}^{(m)}\Expect[\ln\gamma_{\ell,k'}^{(m)}] = u(k) - \ln q(y_n = \ell) - \ln Z_q, \label{eq:A_k_diff}
\end{align}
where the last equality is due to \eqref{eq:q_upd} and by defining $u(k):=\Expect[\ln\pi_{k}] + \sum_m\sum_{k'}\delta_{n,k'}^{(m)}\Expect[\ln\gamma_{k,k'}^{(m)}]$. Rearranging \eqref{eq:A_k_diff} it is straightforward to show that
\begin{equation}
    \ln q(y_n = \ell) = u(k) - \ln Z_q - A_\ell \Rightarrow q(y_n = \ell) = \frac{\exp(u(k))}{Z_q}\exp(-A_\ell) \leq \exp(-A_\ell), \label{eq:q_A_k_ineq}
\end{equation}
where the last inequality is due to $\exp(u(k))/{Z_q}\leq 1.$ Combining \eqref{eq:q_A_k_ineq} with \eqref{eq:A_k_lower_bound} yields
\begin{equation}
    q(y_n = \ell) \leq \exp\left(-D_\pi - 2f_{\pi}(\varepsilon_\pi) - M(D_\gamma/2 + 2f_\gamma(\varepsilon_\gamma) \right).
\end{equation}
Finally, 
\begin{equation}
    q(y_n = k) = 1 - \sum_{\ell\neq k}q(y_n = \ell) \geq 1 - K\exp\left(-D_\pi - 2f_{\pi}(\varepsilon_\pi) - M(D_\gamma/2 + 2f_\gamma(\varepsilon_\gamma) \right) 
\end{equation}
completes the proof. 
\end{proof}

\begin{lemma}
\label{lemma:pi_err}
Suppose that $\mathcal{E}_4$ holds [cf. Lemma~\ref{lemma:concentration}] and for $n=1,\ldots,N$ $\max_k |q(y_n = k) - \mathbbm{1}(y_n = k)|\leq \varepsilon_q$ holds, for some $\varepsilon_q > 0$, and also $\pi_k^*\geq\rho_\pi>0$ for $k=1,\ldots,K$. Then
\begin{equation}
    |\Expect[\pi_k] - \pi_k^*|\leq \frac{N(\varepsilon_q + g_\pi(\nu)) + \alpha_{0,k} + \rho_\pi\bar{\alpha}_0}{N + \bar{\alpha}_0} \quad k=1,\ldots,K
\end{equation}
where $g_\pi$ is a decreasing function of $\nu$.
\end{lemma}
\begin{proof}
Since $\bm{\pi}\sim\textrm{Dir}(\bm{\pi};\bm{\alpha}),$ the expected value of $\pi_k$ is given by $\Expect[\pi_k] = \alpha_k/\bar{\alpha}$. Now consider
\begin{align}
    & |\alpha_k - N\pi_k^*|  = |N_k + \alpha_{0,k} - N\pi_k^*| = |\sum_{n=1}^{N}q(y_n = k) - \sum_{n=1}^{N}\mathbbm{1}(y_n=k) + \alpha_{0,k} + \sum_{n=1}^{N}\mathbbm{1}(y_n = k) - N\pi_k^*| \notag\\
    & \leq |\sum_{n=1}^{N}\left(q(y_n = k) - \mathbbm{1}(y_n = k)\right)| + |\sum_{n=1}^{N}\mathbbm{1}(y_n = k) - N\pi_k^*| + \alpha_{0,k} \leq N\varepsilon_q + Nr_k + \alpha_{0,k} = N\varepsilon_q + Ng_\pi(\nu) + \alpha_{0,k} \label{eq:alpha_k_diff}
\end{align}
where the inequality is due to the assumption and Lemma~\ref{lemma:concentration}, and the last equality is derived from expressing $r_k$ as a function of $\nu$ in \eqref{eq:nu}.
Next, since $\bar{\alpha} = \sum_{k'}\alpha_{k'} = \sum_{k'}(N_{k'} + \alpha_{0,k'}) = N + \bar{\alpha}_0$ we have 
\begin{align}
    & |\Expect[\pi_k] - \pi_k^*| = \left|\frac{\alpha_k}{\bar{\alpha}} - \pi_k^*\right| = \left|\frac{\alpha_k - \pi_k^*\bar{\alpha}}{\bar{\alpha}}\right| = \left|\frac{\alpha_k - N\pi_k^* - \pi_k^*\bar{\alpha}_0}{\bar{\alpha}}\right| \notag\\ & \leq \left|\frac{\alpha_k - N\pi_k^* - \rho_\pi\bar{\alpha}_0}{\bar{\alpha}}\right|\leq \frac{|\alpha_k  - N\pi_k^*| + |\rho_\pi\bar{\alpha_0}|}{N + \bar{\alpha}_0} \label{eq:pi_upd_diff}
\end{align}
Combining \eqref{eq:alpha_k_diff} with \eqref{eq:pi_upd_diff} completes the proof.
\end{proof}

\begin{lemma}
\label{lemma:gamma_err}
Suppose that $\mathcal{E}_2\cap\mathcal{E}_3$ holds [cf. Lemma~\ref{lemma:concentration}] and for $n=1,\ldots,N$ $\max_k |q(y_n = k) - \mathbbm{1}(y_n = k)|\leq \varepsilon_q$ holds, for some $\varepsilon_q > 0$. Then
\begin{equation}
    |\Expect[\gamma_{k,k'}^{(m)}] - \gamma_{k,k'}^{*(m)}|\leq \frac{2Ng_\gamma(\nu) + 2N\varepsilon_q + \beta_{0,k,k'}^{(m)} + \bar{\beta}_{0,k}^{(m)} }{N\mu_m\pi_k^* - N\frac{g_\gamma(\nu)}{\gamma_{k,k'}^{*(m)}} - N\varepsilon_q + \bar{\beta}_{k}^{(m)}}  \quad k ,k'=1,\ldots,K \quad m=1,\ldots,M
\end{equation}
where $g_\gamma$ is a decreasing function of $\nu$. 
\end{lemma}
\begin{proof}
Since $q(\bm{\gamma}_k^{(m)}\sim\textrm{Dir}(\bm{\gamma}_k^{(m)};\bm{\beta}_k^{(m)})$ for $k=1,\ldots,K$ and $m=1,\ldots,M$, the expected value of the $k'$-th entry of $\bm{\gamma}_k^{(m)}$, $\gamma_{k,k'}^{(m)}$ is given by $\Expect[\gamma_{k,k'}^{(m)}] = \beta_{k,k'}^{(m)} / \bar{\beta}_k^{(m)}$. Consider 
\begin{align}
    &\left|\beta_{k,k'}^{(m)} - N\pi_k^*\gamma_{k,k'}^{*(m)}\mu_m \right| = 
    \left|\sum_{n=1}^{N}q(y_n = k)\delta_{n,k'}^{(m)} - \sum_{n=1}^{N}\mathbbm{1}(y_n = k)\delta_{n,k'}^{(m)} + \beta_{0,k,k'}^{(m)} +  \sum_{n=1}^{N}\mathbbm{1}(y_n = k)\delta_{n,k'}^{(m)} - N\pi_k^*\gamma_{k,k'}^{*(m)} \mu_m \right| \notag\\
    & \leq \left| \sum_{n=1}^{N}\mathbbm{1}(y_n = k)\delta_{n,k'}^{(m)} - N\pi_k^*\gamma_{k,k'}^{*(m)}\mu_m\right| + \left|\sum_{n=1}^{N}\left(q(y_n = k) - \mathbbm{1}(y_n = k)\right)\delta_{n,k'}^{(m)} \right| + \beta_{0,k,k'}^{(m)} \notag \\
    & \leq N t_{k,k'}^{(m)} + N\varepsilon_q + \beta_{0,k,k'}^{(m)} = Ng_\gamma(\nu) + N\varepsilon_q + \beta_{0,k,k'}^{(m)}  \label{eq:gamma_upd_numerator}
\end{align}
where the last inequality is due to Lemma~\ref{lemma:concentration} and the assumption, and the last equality arises by expressing $t_{k,k'}^{(m)}$ as a function of $\nu$. Similarly, due to Lemma~\ref{lemma:concentration} we have,
\begin{align}
    & \left|\bar{\beta}_{k}^{(m)} - N\pi_k^*\mu_m \right| \notag\\ & = \left|\sum_{n=1}^{N} q(y_n = k)\mathbbm{1}(\check{y}_n^{(m}) \neq 0) - \sum_{n=1}^{N}\mathbbm{1}(y_n = k)\mathbbm{1}(\check{y}_n^{(m)} \neq 0) + \bar{\beta}_{0,k}^{(m)} +  \sum_{n=1}^{N}\mathbbm{1}(y_n = k)\mathbbm{1}(\check{y}_n^{(m)} \neq 0) - N\pi_k^*\mu_m +  \right| \notag \\
    & \leq \left|\sum_{n=1}^{N}\left(q(y_n = k) - \mathbbm{1}(y_n = k)\right)\mathbbm{1}(\check{y}_n^{(m)}\neq 0) \right| + \left|\sum_{n=1}^{N}\mathbbm{1}(y_n = k)\mathbbm{1}(\check{y}_n^{(m)} \neq 0) - N\pi_k^*\mu_m \right| + \bar{\beta}_{0,k}^{(m)} \notag \\ & \leq N\varepsilon_q + N\frac{g_\gamma(\nu)}{\gamma_{k,k'}^{*(m)}} + \bar{\beta}_{0,k}^{(m)}. \label{eq:gamma_upd_denominator}
\end{align}
Then, combining \eqref{eq:gamma_upd_numerator} and \eqref{eq:gamma_upd_denominator} one can show
\begin{align}
    & \left|\Expect[\gamma_{k,k'}^{(m)}] - \gamma_{k,k'}^{*(m)} \right| = 
    \left| \frac{\beta_{k,k'}^{(m)} - N\pi_k^*\gamma_{k,k'}^{*(m)}\mu_m + N\pi_k^*\gamma_{k,k'}^{*(m)}\mu_m}{\bar{\beta}_{k}^{(m)} - N\pi_k^*\mu_m +  N\pi_k^*\mu_m} - \gamma_{k,k'}^{*(m)}\right| \notag\\ & = 
    \left| \frac{\beta_{k,k'}^{(m)} - N\pi_k^*\gamma_{k,k'}^{*(m)}\mu_m - \gamma_{k,k'}^{*(m)}\left(\bar{\beta}_k^{(m)} - N\pi_k^*\mu_m \right) }{\bar{\beta}_{k}^{(m)} - N\pi_k^*\mu_m +  N\pi_k^*\mu_m} \right| 
    \leq \frac{2Ng_\gamma(\nu) + 2N\varepsilon_q + \beta_{0,k,k'}^{(m)} + \bar{\beta}_{0,k}^{(m)} }{N\mu_m\pi_k^* - N\frac{g_\gamma(\nu)}{\gamma_{k,k'}^{*(m)}} - N\varepsilon_q + \bar{\beta}_{k}^{(m)}} 
\end{align}
\end{proof}

Assuming that event $\mathcal{E}_1\cap\mathcal{E}_2\cap\mathcal{E}_3\cap\mathcal{E}_4$, defined in Lemma~\ref{lemma:concentration} holds, combining lemmata~\ref{lemma:q_y}, \ref{lemma:pi_err} and, \ref{lemma:gamma_err}, for every iteration and defining $D:=D_\pi + MD_\gamma/2$ yields the statement of the theorem.

\subsection{Proof of Theorem 2}
In the case of crowdsourcing with pairwise constraints, Alg.~1 differs from VBEM in the update of the variational distribution for the unknown labels $q(\mathbf{y}).$ At iteration $t$ the update for $q(\mathbf{y})$ uses $q_{t-1}(\mathbf{y}).$ Here, the proof of Theorem 2 follows the proof of Theorem 1 and Lemma~\ref{lemma:q_y_pw} takes the place of lemma~\ref{lemma:q_y}. Here, $\tilde{\mathcal{C}}$ is the set containing indices of data involved in at least one constraint.
\begin{lemma}
\label{lemma:q_y_pw}
Suppose that $\mathcal{E}_1$ holds [cf. Lemma~\ref{lemma:concentration}] and that, $\gamma_{k,k'}^{*(m)}\geq \rho_\gamma>0$, $|\Expect[\gamma_{k,k'}^{(m)}] - \gamma_{k,k'}^{*(m)}|\leq\varepsilon_\gamma$ for all $k,k',m$, and some $\varepsilon_\gamma >0$, and  $\pi_k^{*}\geq\rho_\pi>0$, $|\Expect[\pi_k] - \pi_k^*|\leq\varepsilon_\pi$  for all , and some $\varepsilon_\pi > 0$. Also suppose that from the previous iteration of Alg. 1 $\max_k |q(y_n = k) - \mathbbm{1}(y_n = k)|\leq\varepsilon_q$ holds for some $\varepsilon_q>0$. Then under the variational update for $q(y_n)$ the following holds for $n\in\tilde{\mathcal{C}}$
\begin{align}
    &\max_{k} |q(y_n = k) - \mathbbm{1}(y_n = k)| \leq
    K\exp(-U-\eta W) \\
    & \notag U:= D_\pi+2f_\pi(\varepsilon_\pi) + M\left(D_\gamma/2 + 2f_\gamma (\varepsilon_\gamma)\right) \\
    & \notag W_n:= \left( N_{\rm ML,n}(1 - 2\varepsilon_q) - 2N_{\rm CL,n}\varepsilon_q + N_{\rm CL,n,min} \right)
\end{align}
and for $n\in\tilde{\mathcal{C}}^{c}$
\begin{align}
    &\max_{k} |q(y_n = k) - \mathbbm{1}(y_n = k)| \leq
    K\exp(-U).
\end{align}
\end{lemma}
\begin{proof}
For data that are not involved in constraints the results of Lemma~\ref{lemma:q_y} hold. 
Let $k = y_n$ be the true label of datum $n\in\tilde{\mathcal{C}}$. 
For  $\ell\neq k$ and with $A_\ell$ defined as in lemma~\ref{lemma:q_y} consider the quantity
\begin{align}
    & \tilde{A}_\ell := A_\ell + \eta\sum_{(n,n')\in\mathcal{C}}w_{n,n'}q(y_{n'} = k) - \eta\sum_{(n,n')\in\mathcal{C}}w_{n,n'}q(y_{n'} = \ell) \\ 
    & = A_\ell + \eta\sum_{(n,n')\in\mathcal{C}_{\rm ML}}\mathbbm{1}(y_{n'} = k)
    + \eta\sum_{(n,n')\in\mathcal{C}_{\rm ML}}\left(q(y_{n'} = k) -\mathbbm{1}(y_{n'} = k)\right) \notag\\ & - \eta\sum_{(n,n')\in\mathcal{C}_{\rm CL}}\mathbbm{1}(y_{n'} = k)
     - \eta\sum_{(n,n')\in\mathcal{C}_{\rm CL}}\left(q(y_{n'} = k) -\mathbbm{1}(y_{n'} = k)\right) \notag\\ & -\eta\sum_{(n,n')\in\mathcal{C}_{\rm ML}}\mathbbm{1}(y_{n'} = \ell) 
    - \eta\sum_{(n,n')\in\mathcal{C}_{\rm ML}}\left(\mathbbm{1}(y_{n'} = \ell) - q(y_{n'} = \ell)\right) \notag \\
    & + \eta\sum_{(n,n')\in\mathcal{C}_{\rm CL}}\mathbbm{1}(y_{n'} = \ell) + \eta\sum_{(n,n')\in\mathcal{C}_{\rm CL}}\left(q(y_{n'} = \ell) - \mathbbm{1}(y_{n'} = \ell) \right).\notag
\end{align}
Now, note that $\sum_{(n,n')\in\mathcal{C}_{\rm ML}}\mathbbm{1}(y_{n'} = k) = N_{\rm ML,n}$ is the number of must-link constraints for datum $n$, $\sum_{(n,n')\in\mathcal{C}_{\rm ML}}\mathbbm{1}(y_{n'} = \ell) = 0$,  $\sum_{(n,n')\in\mathcal{C}_{\rm CL}}\mathbbm{1}(y_{n'} = k) = 0$, and $\sum_{(n,n')\in\mathcal{C}_{\rm CL}}\mathbbm{1}(y_{n'} = \ell) = N_{\rm CL,n,\ell}$, where $N_{\rm CL,n,k}$ denotes the number of cannot-link constraints of $n$ that belong to class $k$. In addition, from the assumption of the lemma, it holds that  whereas $q(y_{n'} = \ell)\leq\varepsilon_q$, and $q(y_{n'} = k) \leq 1$. Then $\tilde{A}_\ell$ can be lower-bounded as
\begin{align}
    \tilde{A}_\ell & \geq A_\ell + \eta N_{\rm ML,n}(1-\varepsilon_q) - \eta N_{\rm CL,n}\varepsilon_q - \eta N_{\rm ML,n}\varepsilon_q + \eta N_{\rm CL,n,\ell} - \eta N_{\rm CL,n}\varepsilon_q \notag  \\
    & = A_\ell+ \eta \left(N_{\rm ML,n}(1 - 2\varepsilon_q) - 2N_{\rm CL,n}\varepsilon_q + N_{\rm CL,n,\ell}\right).
\end{align}
Repeating the steps of lemma~\ref{lemma:q_y} yields
\begin{align}
    q(y_n = \ell) \leq \exp\left(-A_\ell - \eta\left( N_{\rm ML,n}(1 - 2\varepsilon_q) - 2N_{\rm CL,n}\varepsilon_q + N_{\rm CL,n,\ell} \right) \right)
\end{align}
and
\begin{align}
    q(y_n = k) \geq 1 - K\exp\left(-D_\pi - 2f_{\pi}(\varepsilon_\pi) - M(D_\gamma(\nu) + 2f_\gamma(\varepsilon_\gamma)) - \eta\left( N_{\rm ML,n}(1 - 2\varepsilon_q) - N_{\rm CL,n}(1 + \varepsilon_q) + N_{\rm CL,n,min} \right)\right) 
\end{align}
with $N_{\rm CL,n,min} = \min_{\ell}N_{\rm CL,n,\ell}.$
\end{proof}

Next, since the error bounds are different for data involved in constraints Lemmata \ref{lemma:pi_err_instance} and \ref{lemma:gamma_err_instance} take this into account to derive error bounds for the updates of the parameters of interest.
Letting $|\tilde{C}| = {N}_{\tilde{C}}$ denote the number of data having at least one constraint, and $\bar{N}_C = N - \tilde{N}_C$, the updates for $\bm{\pi}$ and $\mathbf{\Gamma}$ yield 
\begin{lemma}
\label{lemma:pi_err_instance}
Suppose that for $n\in{\mathcal{C}^{c}}$ $\max_k |q(y_n = k) - \mathbbm{1}(y_n = k)|\leq \varepsilon_q$ holds, for some $\varepsilon_q > 0$, for $n\in\mathcal{C}$ $\max_k |q(y_n = k) - \mathbbm{1}(y_n = k)|\leq\tilde{\varepsilon}_q$ holds, for some $\tilde{\varepsilon}_q > 0$ and also $\pi_k^*\geq\rho_\pi>0$ for $k=1,\ldots,K$ Then
\begin{equation}
    |\Expect[\pi_k] - \pi_k^*|\leq \frac{\tilde{N}_C\tilde{\varepsilon}_{q} + \bar{N}_C\varepsilon_q + Ng_\pi(\nu) + \alpha_{0,k} + \rho_\pi\bar{\alpha}_0}{N + \bar{\alpha}_0} \quad k=1,\ldots,K
\end{equation}
\end{lemma}
\begin{proof}
Repeating the procedure of Lemma~\ref{lemma:pi_err} we have
\begin{align}
    & |\alpha_k - N\pi_k^*|  = |\sum_{n=1}^{N}q(y_n = k) - \sum_{n=1}^{N}\mathbbm{1}(y_n=k) + \alpha_{0,k} + \sum_{n=1}^{N}\mathbbm{1}(y_n = k) - N\pi_k^*| \notag\\
    & \leq |\sum_{n\in\tilde{C}}\left(q(y_n = k) - \mathbbm{1}(y_n = k)\right)|  + |\sum_{n\in\tilde{C}^{c}}\left(q(y_n = k) - \mathbbm{1}(y_n = k)\right)| + |\sum_{n=1}^{N}\mathbbm{1}(y_n = k) - N\pi_k^*| + \alpha_{0,k} \notag \\
    &\leq {N}_{\tilde{C}}\tilde{\varepsilon}_q + \bar{N}_C \varepsilon_q + g_\pi(\nu) + \alpha_{0,k} \label{eq:alpha_k_diff_inst}
\end{align}
 Combining  \eqref{eq:alpha_k_diff_inst} with \eqref{eq:pi_upd_diff} completes the proof.
\end{proof}

\begin{lemma}
\label{lemma:gamma_err_instance}
Suppose that for $n\in{\mathcal{C}}^{c}$, $\max_k |q(y_n = k) - \mathbbm{1}(y_n = k)|\leq \varepsilon_q$ holds, for some $\varepsilon_q > 0$, and  for $n\in\tilde{\mathcal{C}}$ $\max_k |q(y_n = k) - \mathbbm{1}(y_n = k)|\leq\tilde{\varepsilon}_q$ holds, for some $\tilde{\varepsilon}_q>0$. Then
\begin{equation}
    |\Expect[\gamma_{k,k'}^{(m)}] - \gamma_{k,k'}^{*(m)}|\leq \frac{2Ng_\gamma(\nu) + 2\tilde{N}_C\tilde{\varepsilon}_q + 2\bar{N}_C\varepsilon_q + \beta_{0,k,k'}^{(m)} + \bar{\beta}_{0,k}^{(m)} }{N\mu_m\pi_k^* - N\frac{g_\gamma(\nu)}{\gamma_{k,k'}^{*(m)}} - \tilde{N}_C\tilde{\varepsilon}_q - \bar{N}_C\varepsilon_q + \bar{\beta}_{k}^{(m)}}  \quad k ,k'=1,\ldots,K \quad m=1,\ldots,M
\end{equation}
where $g_\gamma$ is a decreasing function of $\nu$.
\end{lemma}
\begin{proof}
Repeating the steps of lemma~\ref{lemma:gamma_err} one shows
\begin{align}
    &\left|\beta_{k,k'}^{(m)} - N\pi_k^*\gamma_{k,k'}^{*(m)}\mu_m \right| 
     \leq \left| \sum_{n=1}^{N}\mathbbm{1}(y_n = k)\delta_{n,k'}^{(m)} - N\pi_k^*\gamma_{k,k'}^{*(m)}\mu_m\right| + \left|\sum_{n=1}^{N}\left(q(y_n = k) - \mathbbm{1}(y_n = k)\right)\delta_{n,k'}^{(m)} \right| + \beta_{0,k,k'}^{(m)} \notag \\
    & \leq N g_\gamma(\nu) + \tilde{N}_C\tilde{\varepsilon_q} + \bar{N}_C\varepsilon_q + \beta_{0,k,k'}^{(m)} \label{eq:gamma_upd_numerator_inst}
\end{align}
where the last inequality is due to Lemma~\ref{lemma:concentration} and the assumption. Invoking lemma~\ref{lemma:concentration} again we have
\begin{align}
    & \left|\bar{\beta}_{k}^{(m)} - N\pi_k^*\mu_m \right| 
    \leq \left|\sum_{n=1}^{N}\left(q(y_n = k) - \mathbbm{1}(y_n = k)\right)\mathbbm{1}(\check{y}_n^{(m)}\neq 0) \right| + \left|\sum_{n=1}^{N}\mathbbm{1}(y_n = k)\mathbbm{1}(\check{y}_n^{(m)} \neq 0) - N\pi_k^*\mu_m \right| + \bar{\beta}_{0,k}^{(m)} \notag \\ & \leq \tilde{N}_C\tilde{\varepsilon_q} + \bar{N}_C\varepsilon_q + N\frac{g_\gamma(\nu)}{\gamma_{k,k'}^{*(m)}} + \bar{\beta}_{0,k}^{(m)}. \label{eq:gamma_upd_denominator_inst}
\end{align}
Combining \eqref{eq:gamma_upd_numerator_inst} and \eqref{eq:gamma_upd_denominator_inst} yields
\begin{align}
    & \left|\Expect[\gamma_{k,k'}^{(m)}] - \gamma_{k,k'}^{*(m)} \right|
    \leq \frac{2Ng_\gamma(\nu) + 2\tilde{N}_C\tilde{\varepsilon}_q + 2\bar{N}_C\varepsilon_q + \beta_{0,k,k'}^{(m)} + \bar{\beta}_{0,k}^{(m)} }{N\mu_m\pi_k^* - N\frac{g_\gamma(\nu)}{\gamma_{k,k'}^{*(m)}} - \tilde{N}_C\tilde{\varepsilon}_q - \bar{N}_C\varepsilon_q + \bar{\beta}_{k}^{(m)}} 
\end{align}
completing the proof.
\end{proof}
Assuming that event $\mathcal{E}_1\cap\mathcal{E}_2\cap\mathcal{E}_3\cap\mathcal{E}_4$, defined in Lemma~\ref{lemma:concentration} holds, combining lemmata~\ref{lemma:q_y_pw}, \ref{lemma:pi_err_instance} and, \ref{lemma:gamma_err_instance}, for every iteration and defining $D:=D_\pi + MD_\gamma/2$ yields the statement of the theorem. 

\clearpage
\section{Additional numerical tests}
This appendix includes additional numerical tests for the proposed semi-supervised algorithms, that corroborate the results presented in the main paper. The datasets used are the Bluebird \cite{multidim_wisdom}, Dog \cite{imagenet}, Web \cite{minimax_crowd}, ZenCrowd India and US \cite{ZenCrowd}, Music Genre and Sentence polarity \cite{musicgenre_senpoldata}, RTE \cite{cheapnfast}, TREC \cite{Lease11-trec} and Product \cite{crowder} datasets. Dataset properties are listed in Tab.~\ref{tab:datasets}.

For the Bluebird dataset, annotators from Amazon's Mechanical Turk~\cite{MTurk} were tasked with classifying $N=108$ images of birds into $K=2$ classes: ``Indigo Bunting'' or ``Blue Grosbeak''. The Dog dataset consists of $807$ images of dogs which are classified into $5$ breeds by $109$ annotators. For the Web dataset, annotators $177$  were tasked with classifying $2,665$ website results into $5$ classes denoting relevance to a corresponding search query. For the ZenCrowd India and US datasets, annotators were asked to classify whether an identifier was relevant or irrelevant to a particular entity. The total number of identifier entity pairs was $2,040$.
The Music genre dataset contains $700$ song samples (each of duration $30$secs), belonging into $10$ music categories, annotated by $44$  annotators. The sentence polarity dataset contains $5,000$ sentences from movie reviews, classified into $2$ categories (positive or negative), annotated by $203$ annotators.
For the RTE dataset $164$ annotators from Amazon's Mechanical Turk were provided with $800$ pairs of sentences and asked to classify for each pair whether the second sentence can be inferred from the first. The TREC dataset consists of $19,033$ documents whose relevance or not to a query is assessed by $762$ annotators. Note that, for the TREC dataset, the ground-truth is known only for $2,275$ data and as such all metrics are evaluated on that subset of $2,275$ data. The number of available constraints that can be generated for our tests are limited by the amount of available ground-truth labels.  The Product dataset contains $8,315$ pairs of descriptions of products. In this dataset $176$ annotators are tasked with classifying whether the pairs of descriptions correspond to the same product or not.

\begin{table}[tb]
\centering
 \begin{tabular}{|c || c | c | c | c | c | c | c | c | c | c||} 
 \hline
 Dataset & Bluebird & Dog & Web & Music Genre & ZenCrowd India & ZenCrowd US & RTE & Sentence Polarity & TREC & Product\\ [0.5ex] 
 \hline\hline
 $N$ & $108$ & $807$ & $2,665$ & $700$ & $2,040$ & $2,040$ & $800$ & $5,000$ & $19,033$ & $8,315$ \\ 
 \hline
 $M$ & $39$ & $109$ & $177$ & $44$ & $25$ & $74$ & $164$ & $203$ & $762$ & $176$  \\
 \hline
 $K$ & $2$ & $5$ & $5$ & $10$ & $2$ & $2$ & $2$ & $2$ & $2$ & $2$ \\
 \hline
 $\tilde{\delta}$ & $108$ & $74.03$ & $87.94$ & $66.93$ & $418.2$ & $105.16$ & $48.78$ &  $136.68$ & $116$ & $141.73$ \\  [1ex] 
 \hline
\end{tabular}
\caption{Datasets and their properties}
\label{tab:datasets}
\end{table}

\subsection{Same number of randomly sampled constraints}
\begin{figure}
    \centering
    \begin{minipage}{0.32\textwidth}
        \centering
        \begin{subfigure}[b]{0.9\textwidth}
         \centering
         \includegraphics[width=\textwidth]{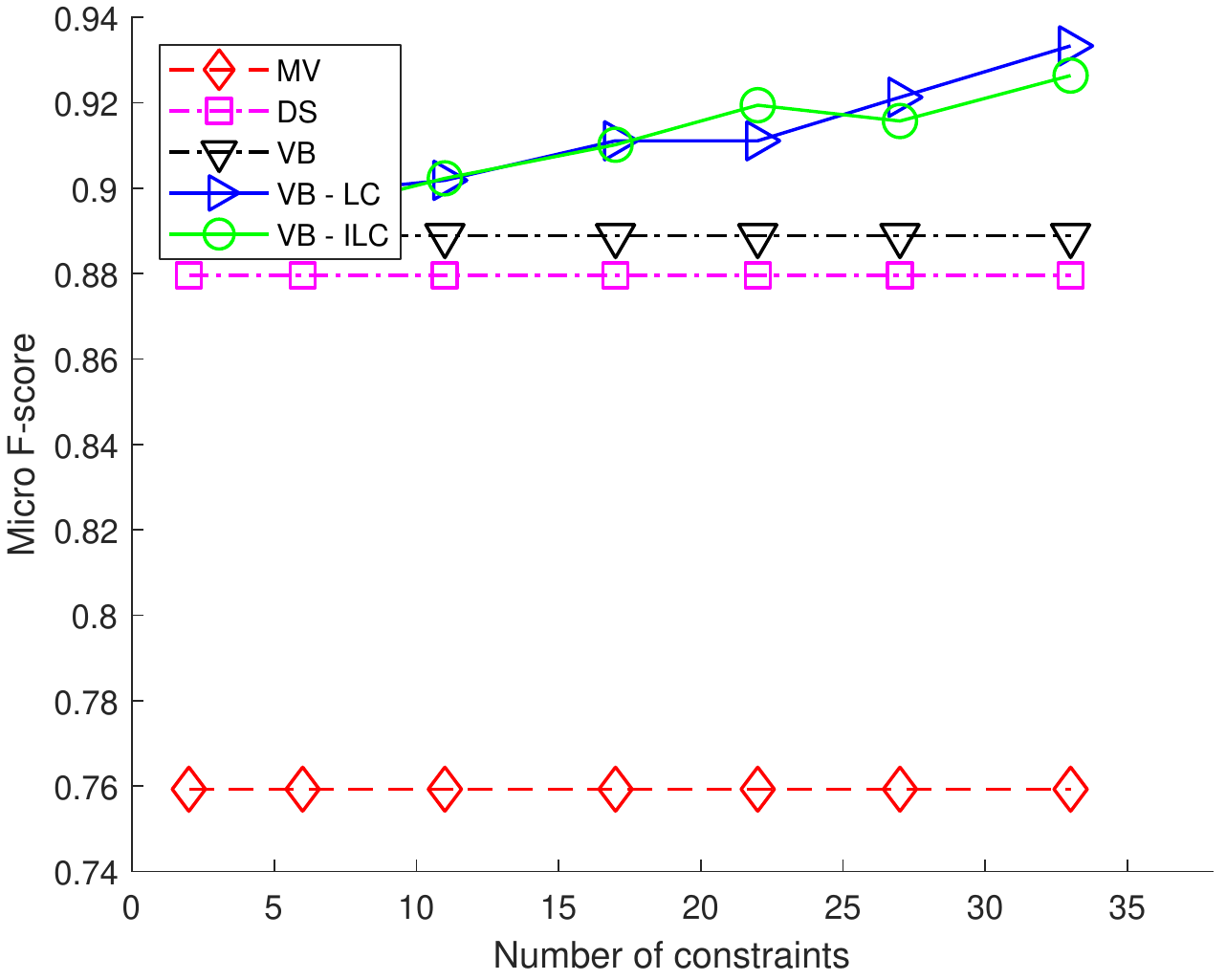}
         \caption{Micro F-score}
         \label{fig:bluebird_micro}
     \end{subfigure}\\
     \begin{subfigure}[b]{0.9\textwidth}
         \centering
         \includegraphics[width=\textwidth]{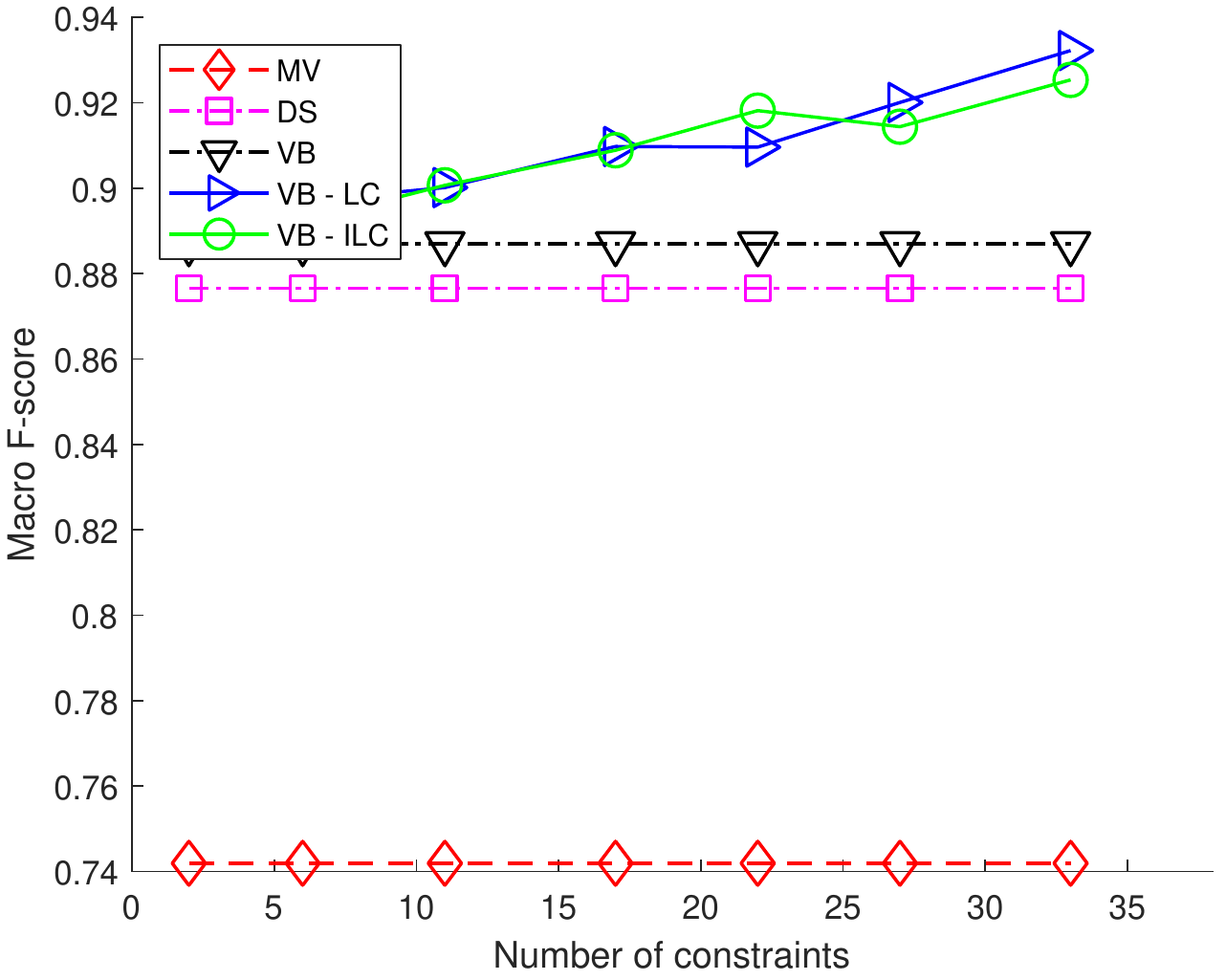}
         \caption{Macro F-score}
         \label{fig:bluebird_macro}
     \end{subfigure}
        \caption{Results for the Bluebird~\cite{multidim_wisdom} dataset, with randomly sampled constraints.}
        \label{fig:bluebird_res}
    \end{minipage}\hfill
    \begin{minipage}{0.32\textwidth}
        \centering
     \begin{subfigure}[b]{0.9\textwidth}
         \centering
         \includegraphics[width=\textwidth]{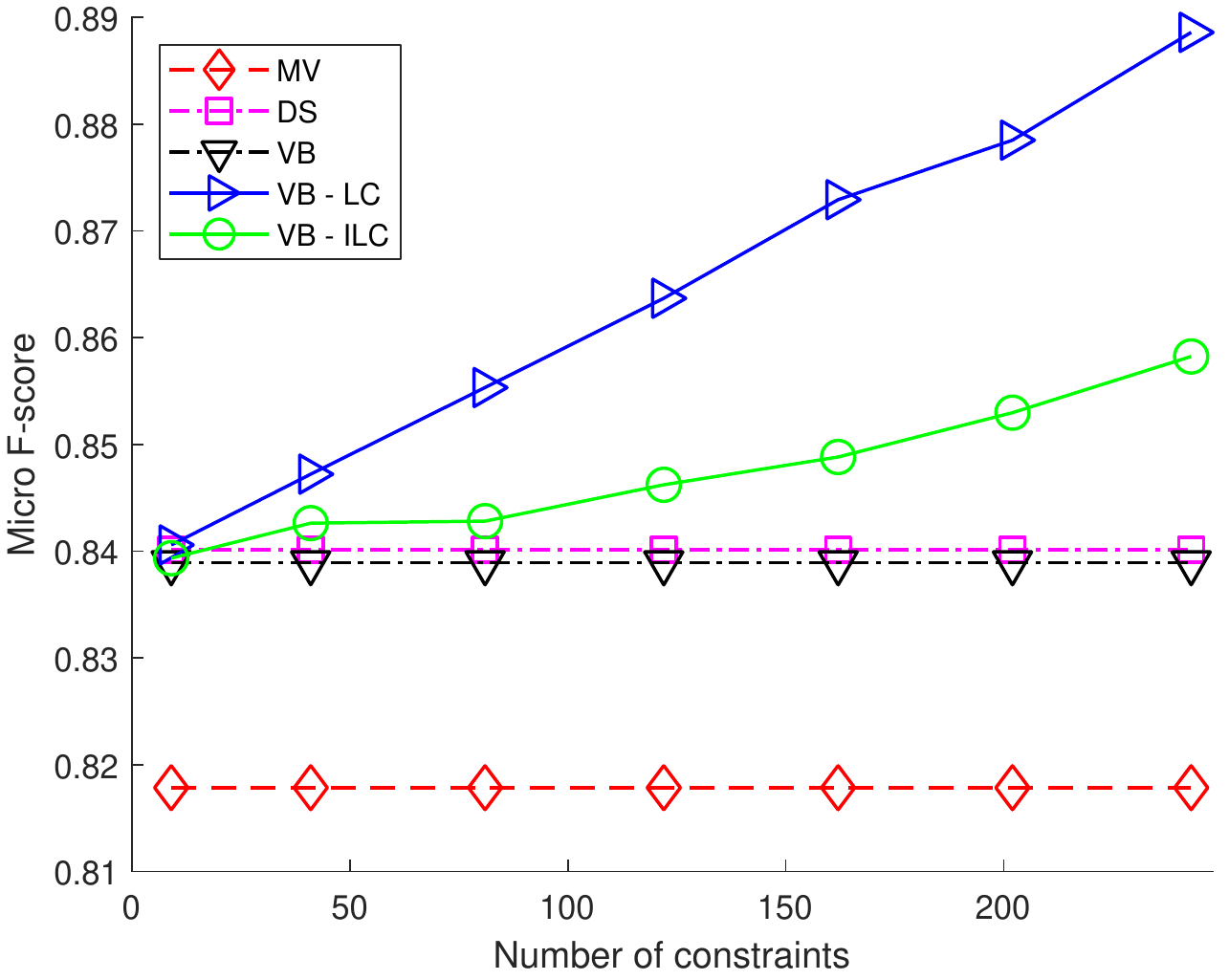}
         \caption{Micro F-score}
         \label{fig:dog_micro}
     \end{subfigure}
     \begin{subfigure}[b]{0.9\textwidth}
         \centering
         \includegraphics[width=\textwidth]{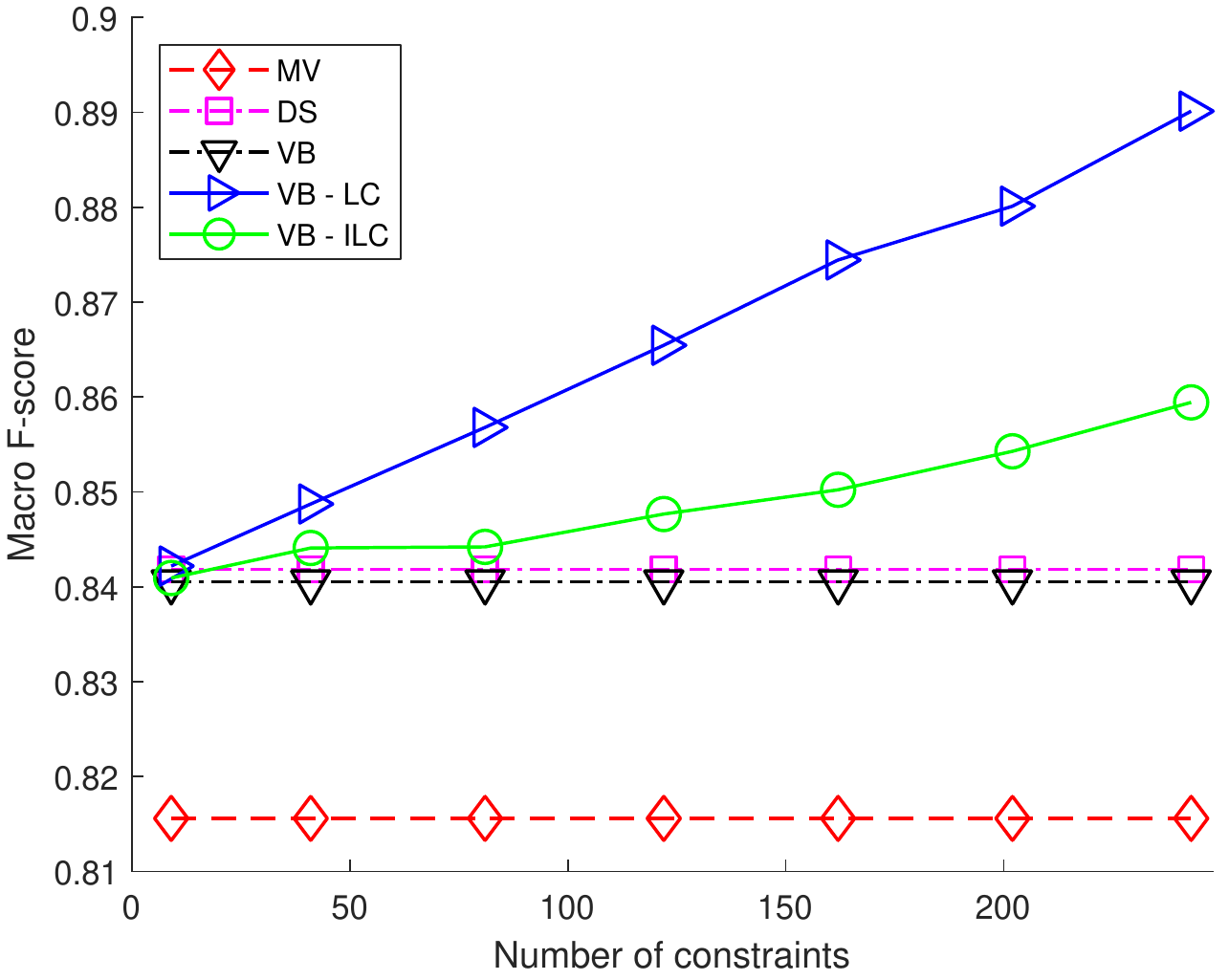}
         \caption{Macro F-score}
         \label{fig:dog_macro}
     \end{subfigure}
        \caption{Results for the Dog~\cite{imagenet} dataset, with randomly sampled constraints.}
        \label{fig:dog_res}
    \end{minipage}
    \hfill
    \begin{minipage}{0.32\textwidth}
        \centering
        \begin{subfigure}[b]{0.9\textwidth}
         \centering
         \includegraphics[width=\textwidth]{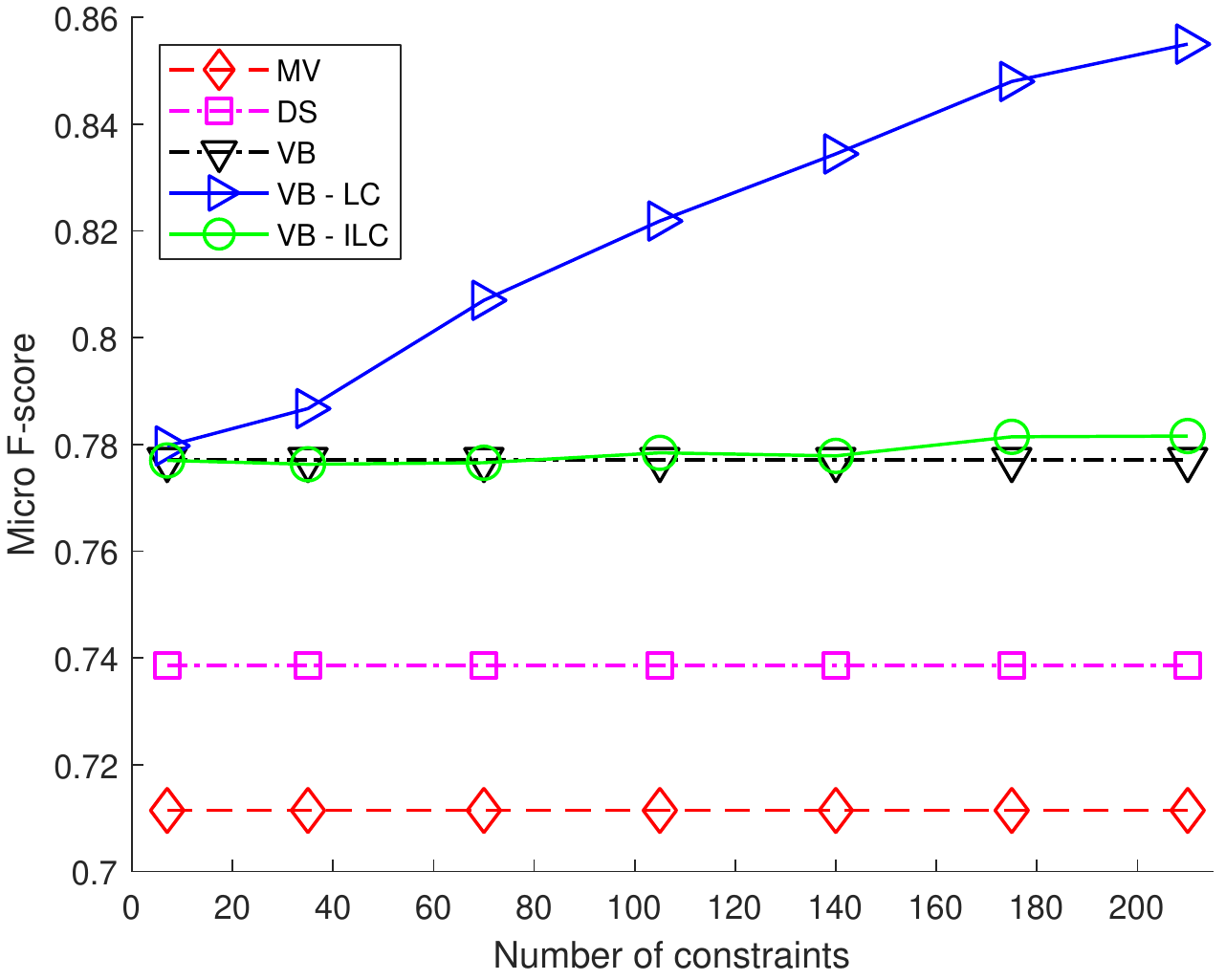}
         \caption{Micro F-score}
         \label{fig:music_micro}
     \end{subfigure}
     \begin{subfigure}[b]{0.9\textwidth}
         \centering
         \includegraphics[width=\textwidth]{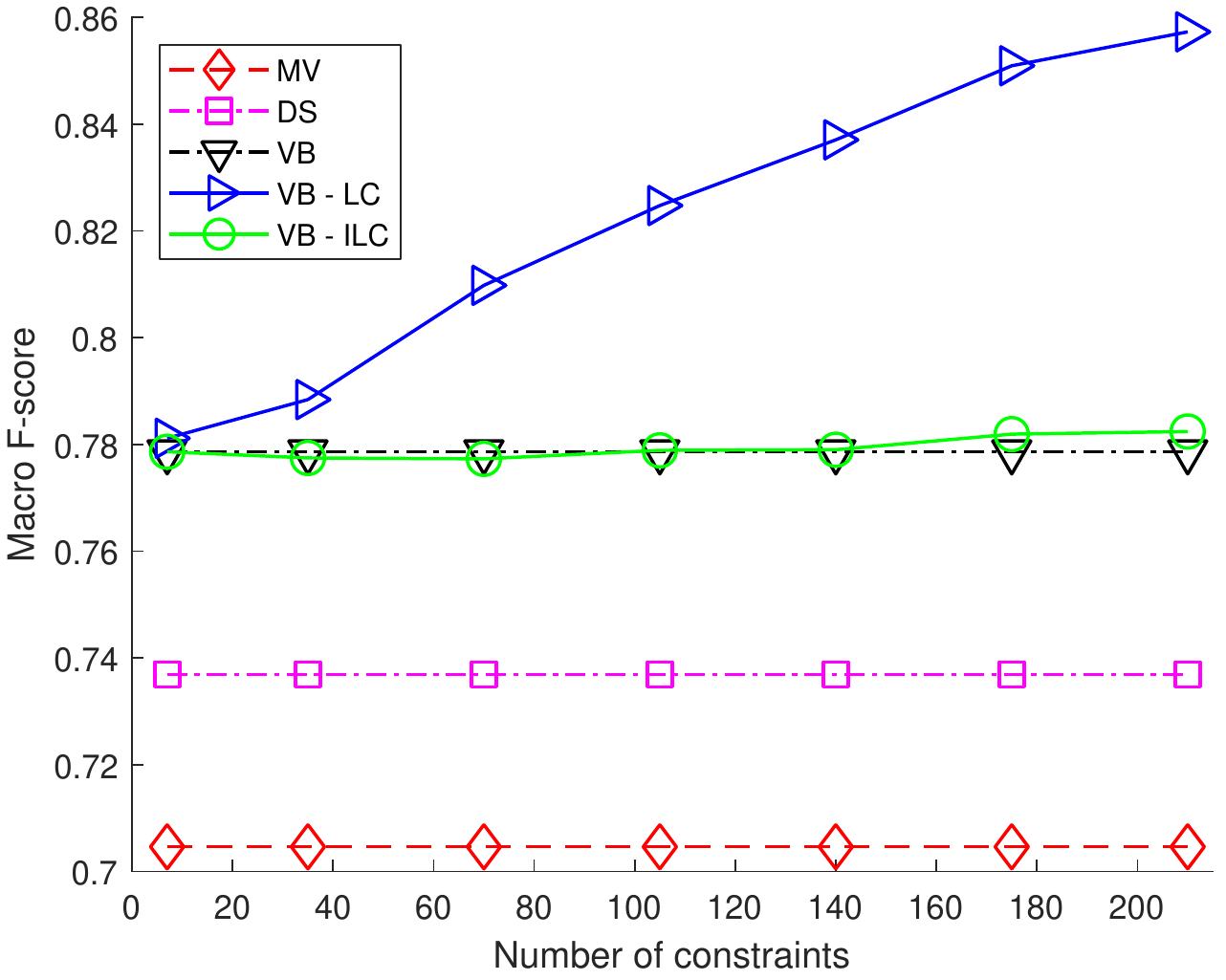}
         \caption{Macro F-score}
         \label{fig:music_macro}
     \end{subfigure}
        \caption{Results for the Music Genre~\cite{musicgenre_senpoldata} dataset, with randomly sampled constraints.}
        \label{fig:music_res}
    \end{minipage}
\end{figure}

\begin{figure}
    \centering
    \begin{minipage}{0.32\textwidth}
        \centering
        \begin{subfigure}[b]{0.9\textwidth}
         \centering
         \includegraphics[width=\textwidth]{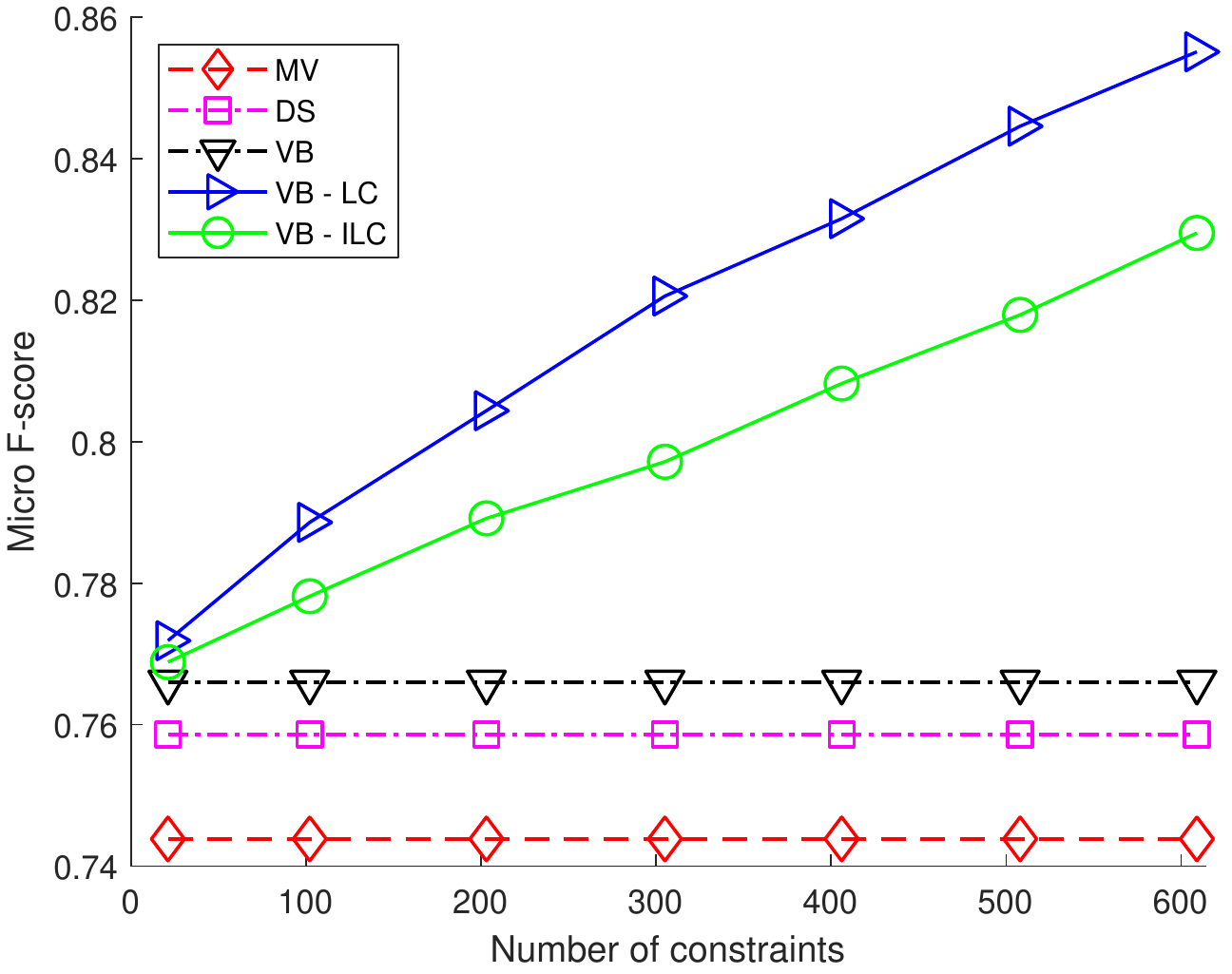}
         \caption{Micro F-score}
         \label{fig:ZenCrowd_in_micro}
     \end{subfigure}
     \begin{subfigure}[b]{0.9\textwidth}
         \centering
         \includegraphics[width=\textwidth]{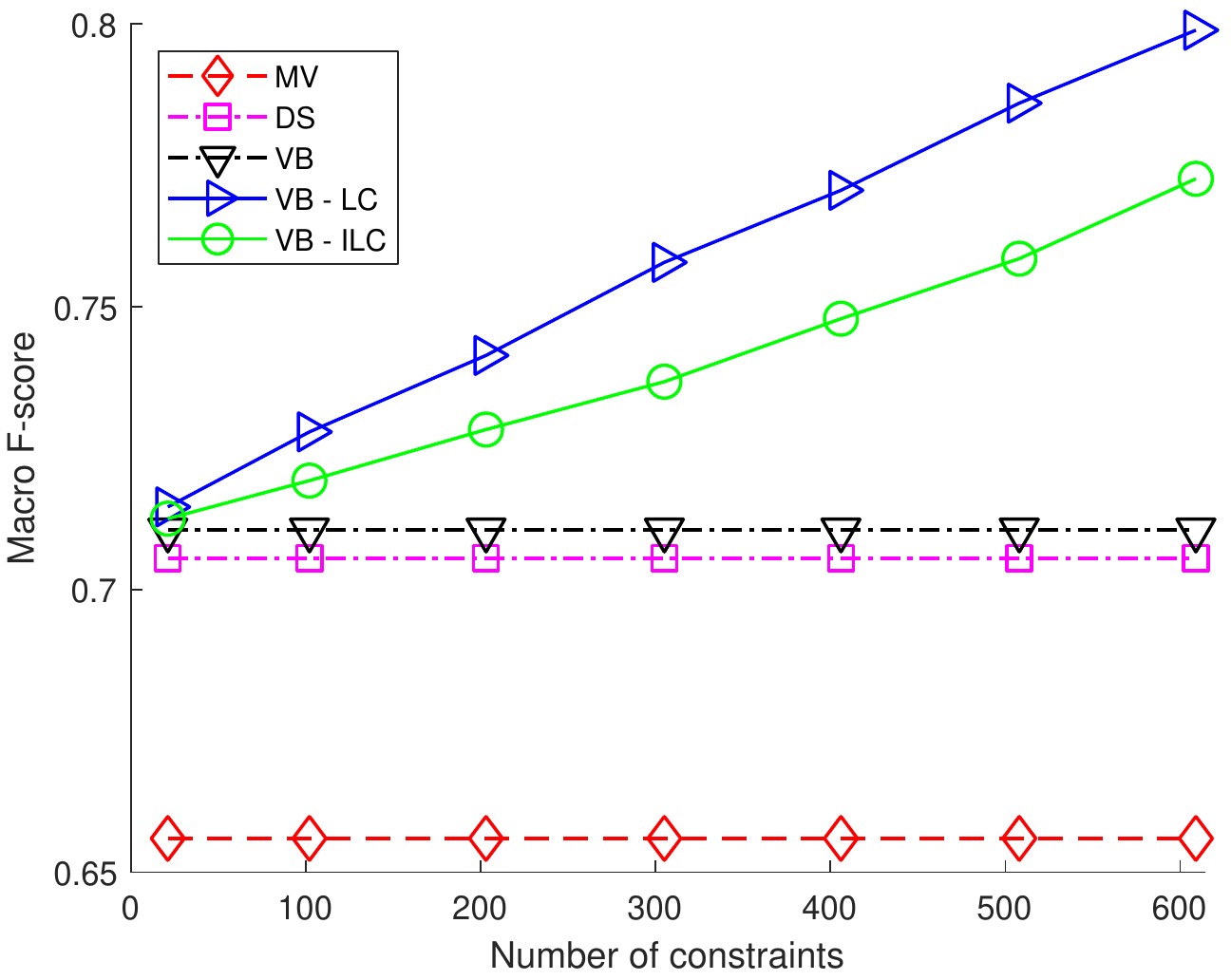}
         \caption{Macro F-score}
         \label{fig:ZenCrowd_in_macro}
     \end{subfigure}
        \caption{Results for the ZenCrowd India~\cite{ZenCrowd} dataset, with randomly sampled constraints.}
        \label{fig:ZenCrowd_in_res}
    \end{minipage}\hfill
    \begin{minipage}{0.32\textwidth}
        \centering
        \begin{subfigure}[b]{0.9\textwidth}
         \centering
         \includegraphics[width=\textwidth]{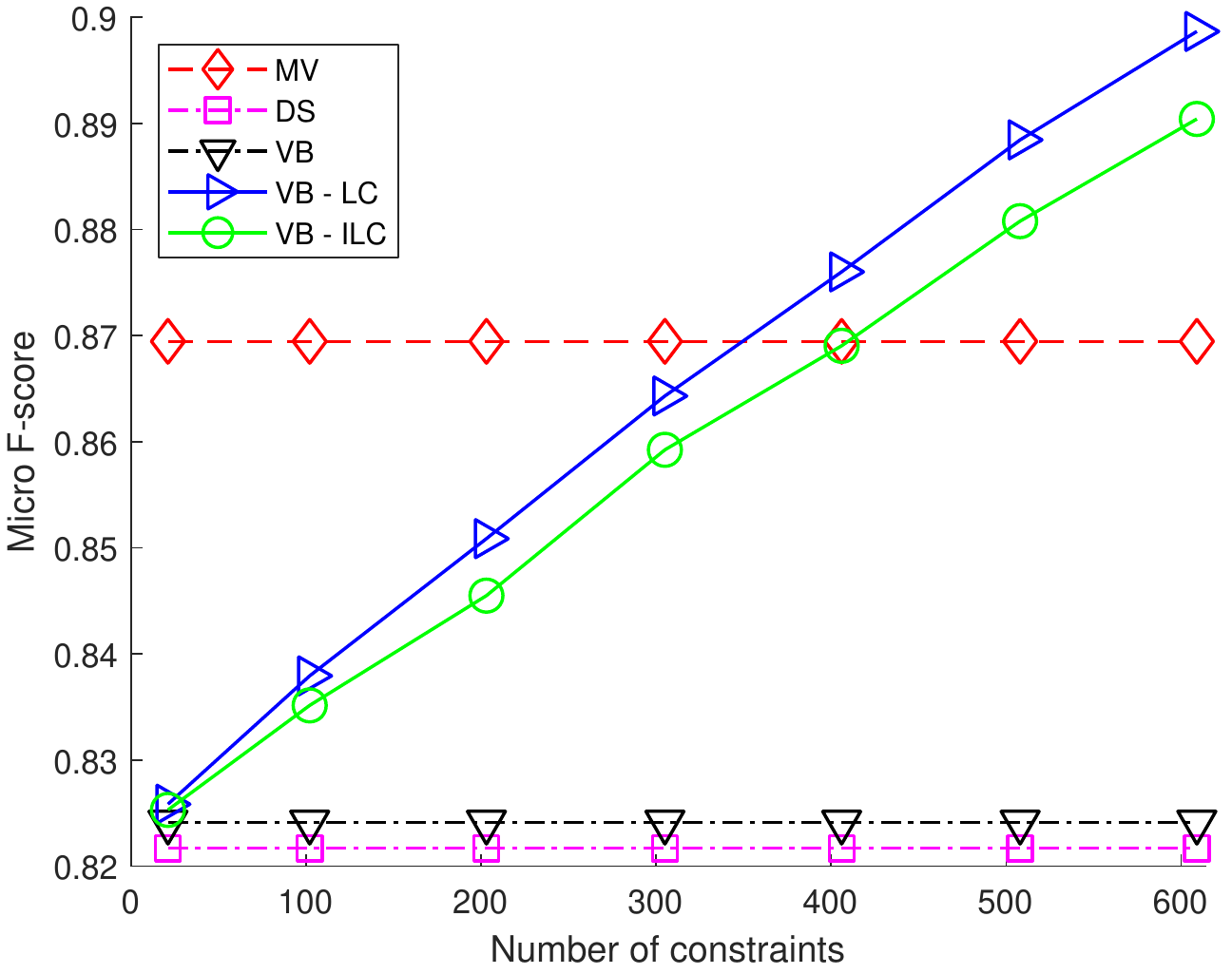}
         \caption{Micro F-score}
         \label{fig:ZenCrowd_us_micro}
     \end{subfigure}
     \begin{subfigure}[b]{0.9\textwidth}
         \centering
         \includegraphics[width=\textwidth]{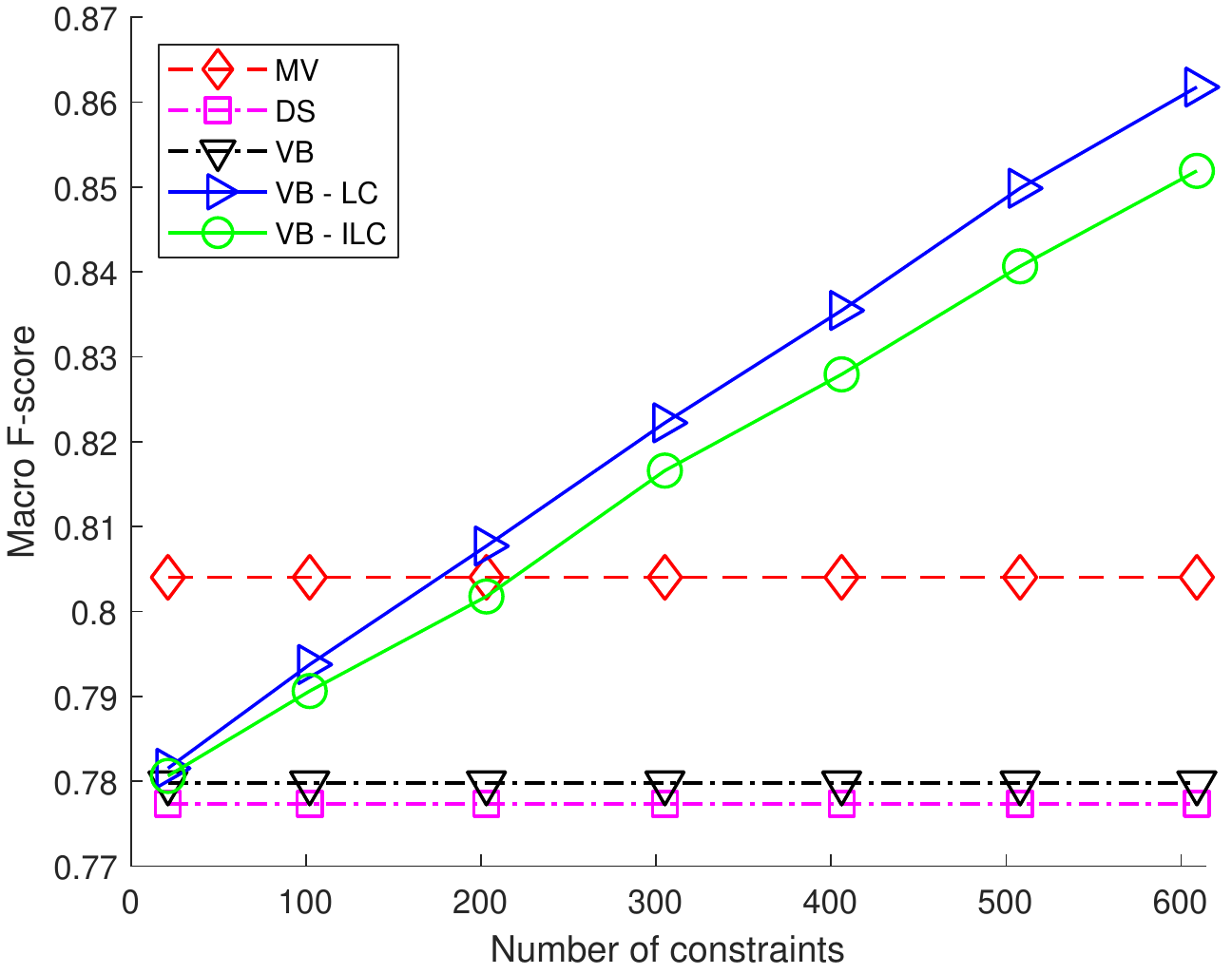}
         \caption{Macro F-score}
         \label{fig:ZenCrowd_us_macro}
     \end{subfigure}
        \caption{Results for the ZenCrowd US~\cite{ZenCrowd} dataset, with randomly sampled constraints.}
        \label{fig:ZenCrowd_us_res}
    \end{minipage}
    \hfill
    \begin{minipage}{0.32\textwidth}
        \centering
        \begin{subfigure}[b]{0.9\textwidth}
         \centering
         \includegraphics[width=\textwidth]{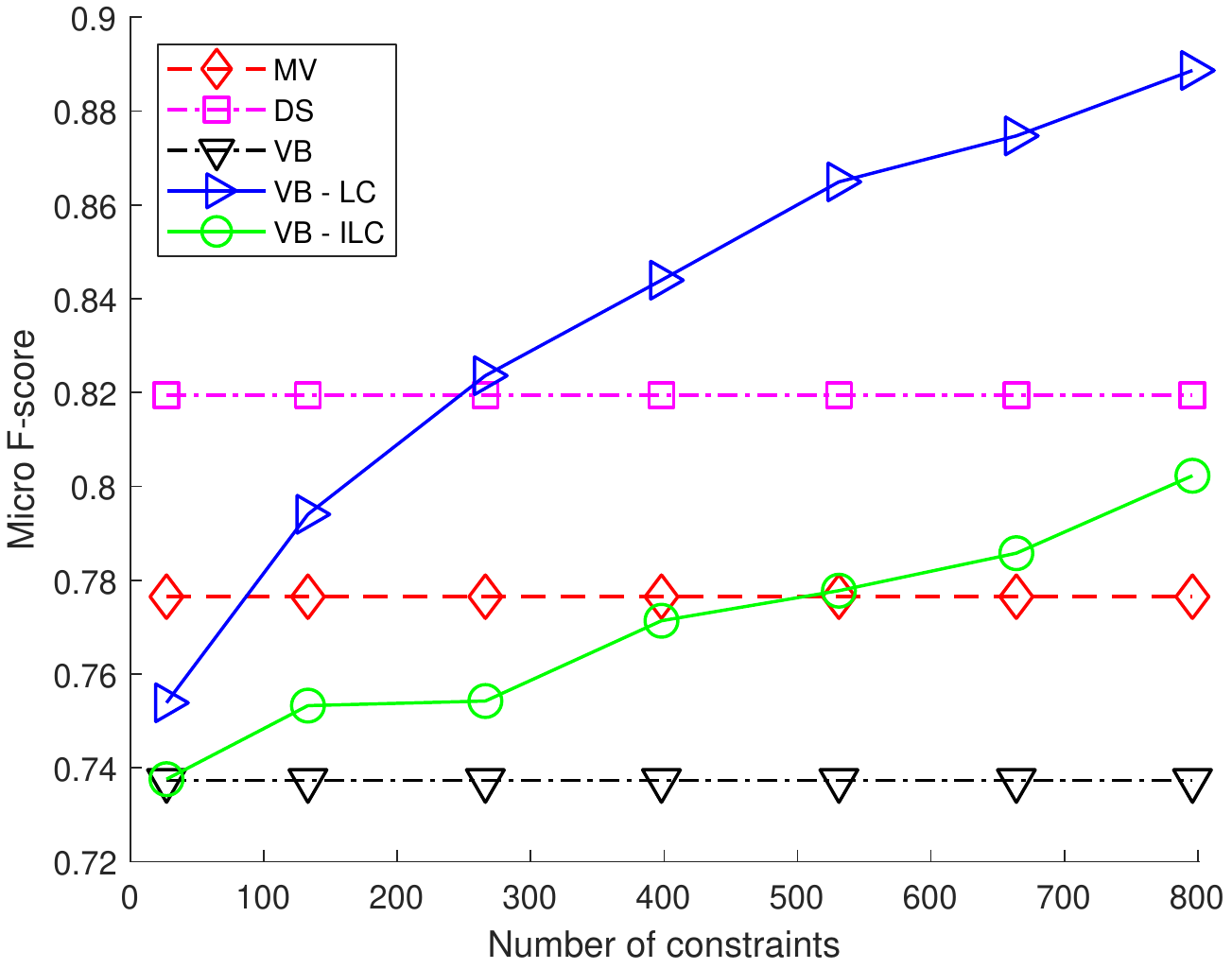}
         \caption{Micro F-score}
         \label{fig:web_micro}
     \end{subfigure}
     \begin{subfigure}[b]{0.9\textwidth}
         \centering
         \includegraphics[width=\textwidth]{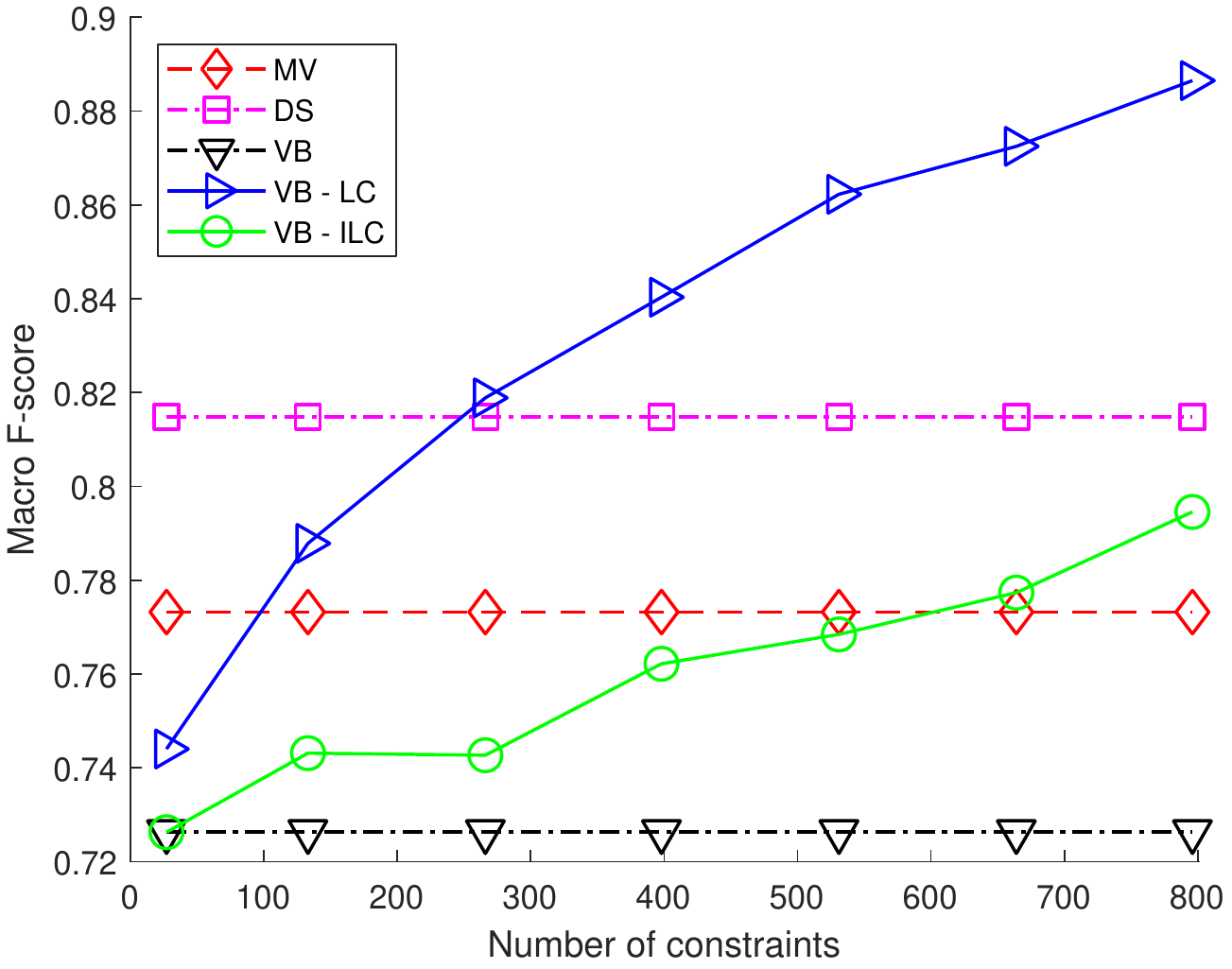}
         \caption{Macro F-score}
         \label{fig:web_macro}
     \end{subfigure}
        \caption{Results for the Web~\cite{minimax_crowd} dataset, with randomly sampled constraints.}
        \label{fig:web_res}
    \end{minipage}
\end{figure}

\begin{figure}
    \centering
    \begin{minipage}{0.32\textwidth}
        \centering
        \begin{subfigure}[b]{0.9\textwidth}
         \centering
         \includegraphics[width=\textwidth]{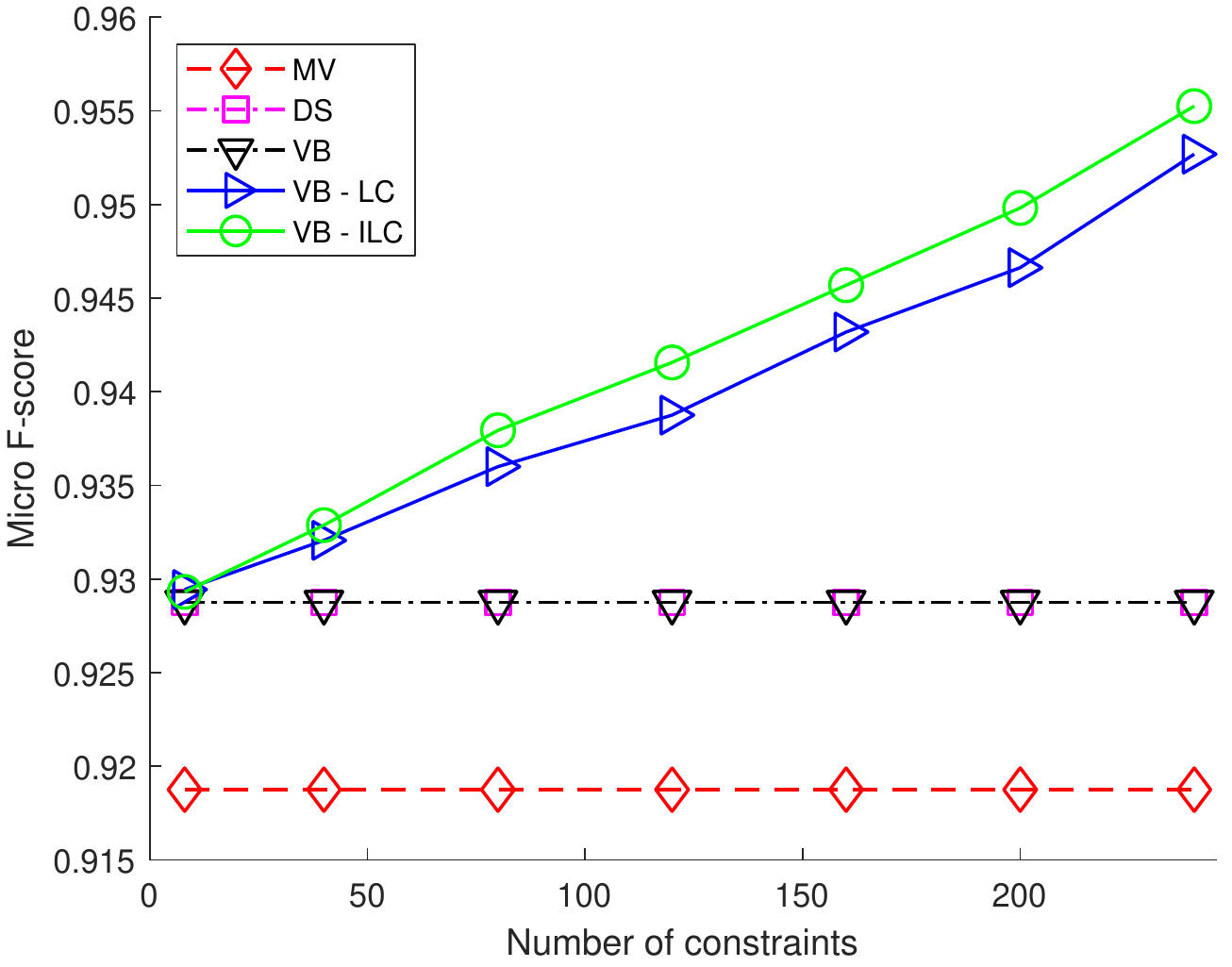}
         \caption{Micro F-score}
         \label{fig:rte_micro}
     \end{subfigure}
     \begin{subfigure}[b]{0.9\textwidth}
         \centering
         \includegraphics[width=\textwidth]{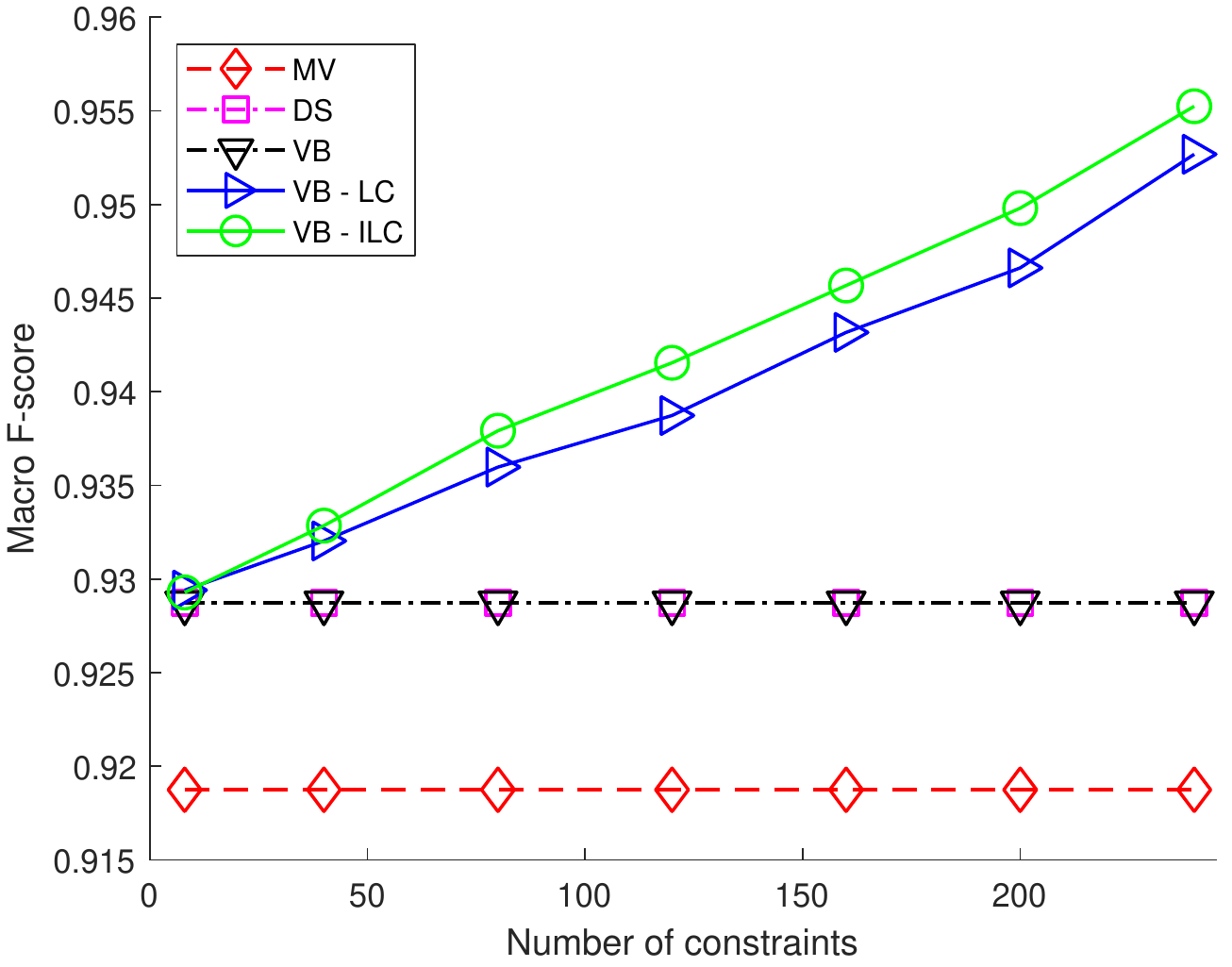}
         \caption{Macro F-score}
         \label{fig:rte_macro}
     \end{subfigure}
        \caption{Results for the RTE~\cite{cheapnfast} dataset, with randomly sampled constraints.}
        \label{fig:rte_res}
    \end{minipage}\hfill
    \begin{minipage}{0.32\textwidth}
        \centering
        \begin{subfigure}[b]{0.9\textwidth}
         \centering
         \includegraphics[width=\textwidth]{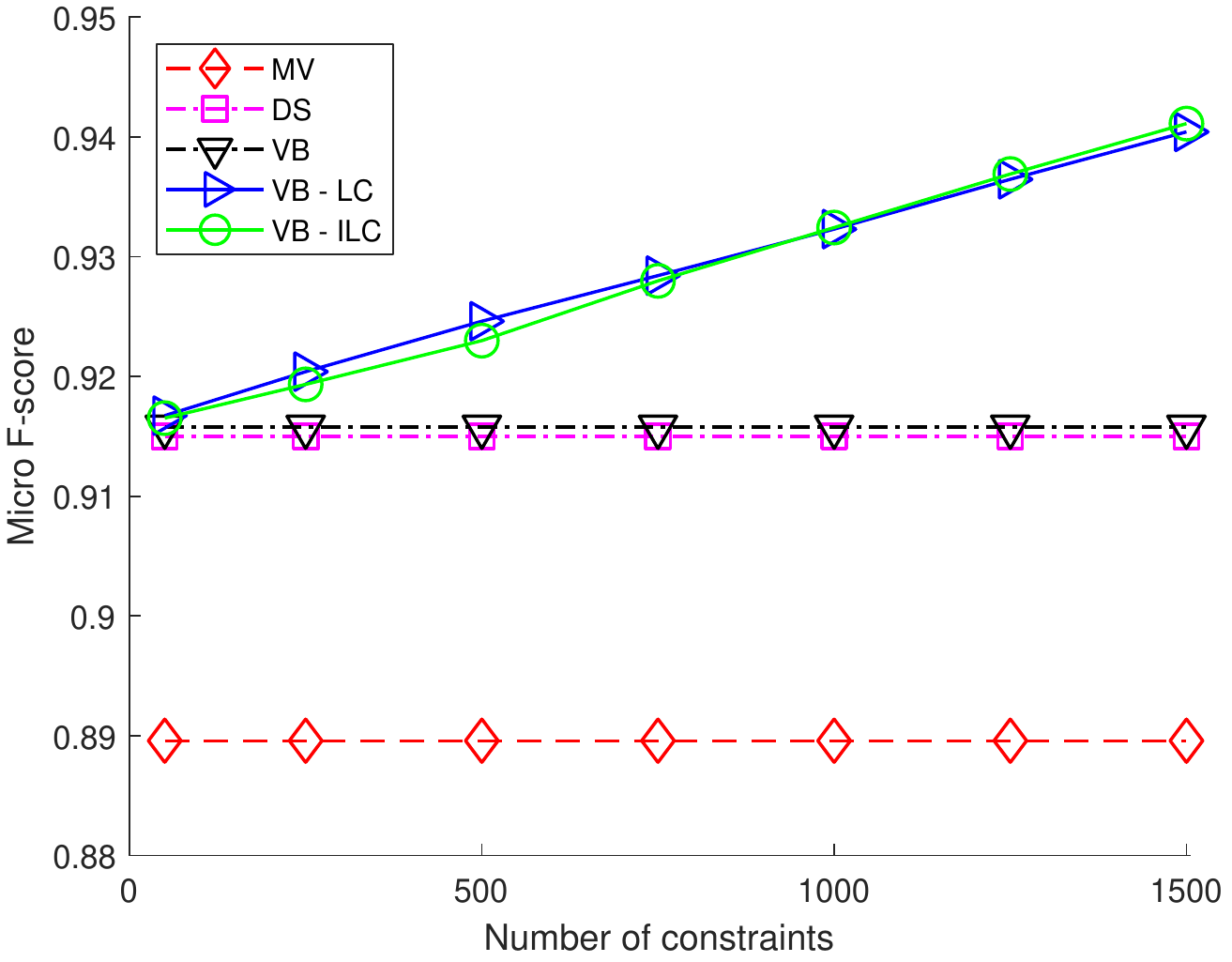}
         \caption{Micro F-score}
         \label{fig:sentencepolarity_micro}
     \end{subfigure}
     \begin{subfigure}[b]{0.9\textwidth}
         \centering
         \includegraphics[width=\textwidth]{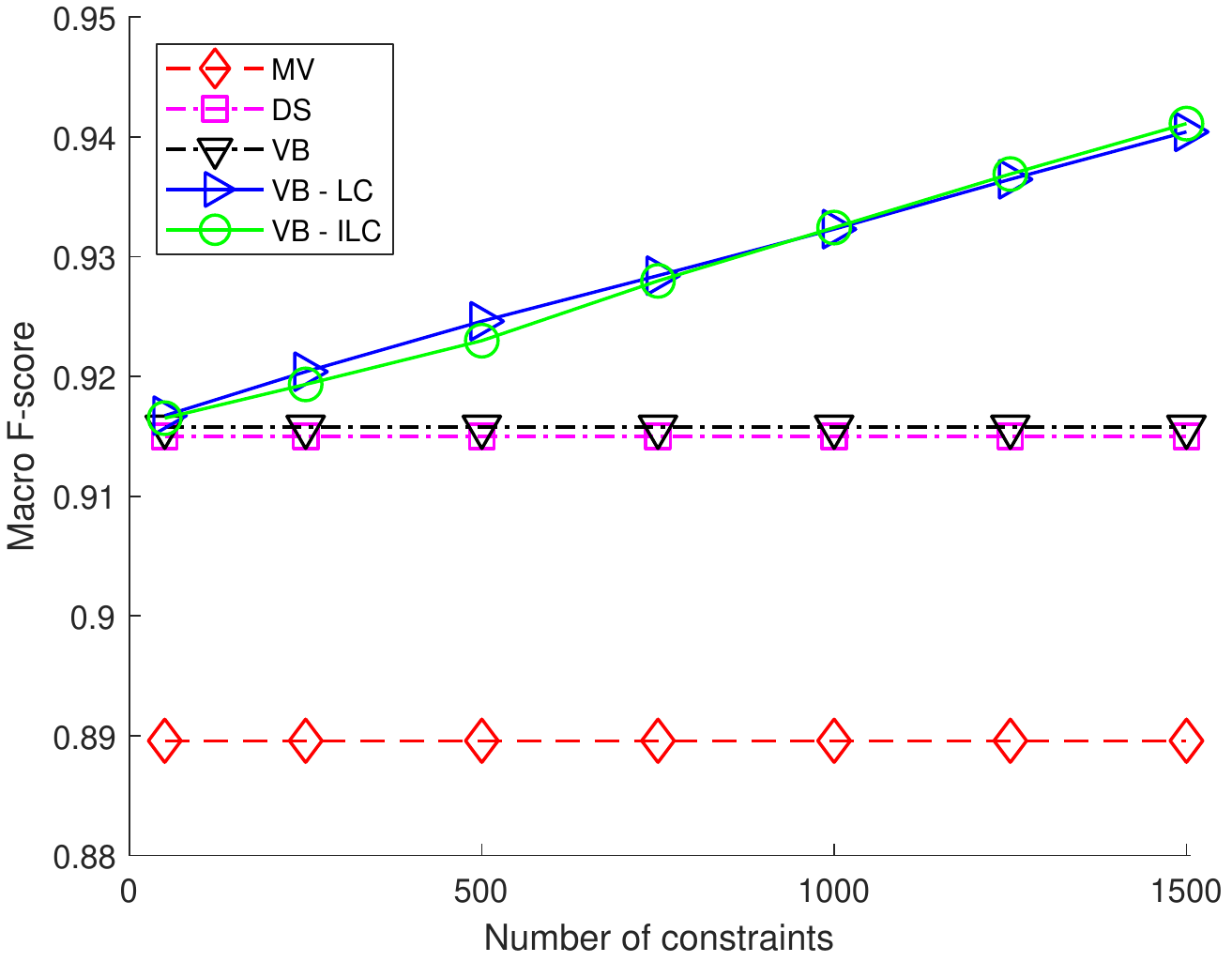}
         \caption{Macro F-score}
         \label{fig:sentencepolarity_macro}
     \end{subfigure}
        \caption{Results for the Sentence Polarity~\cite{musicgenre_senpoldata} dataset, with randomly sampled constraints.}
        \label{fig:sentencepolarity_res}
    \end{minipage}
    \hfill
    \begin{minipage}{0.32\textwidth}
        \centering
        \begin{subfigure}[b]{0.9\textwidth}
         \centering
         \includegraphics[width=\textwidth]{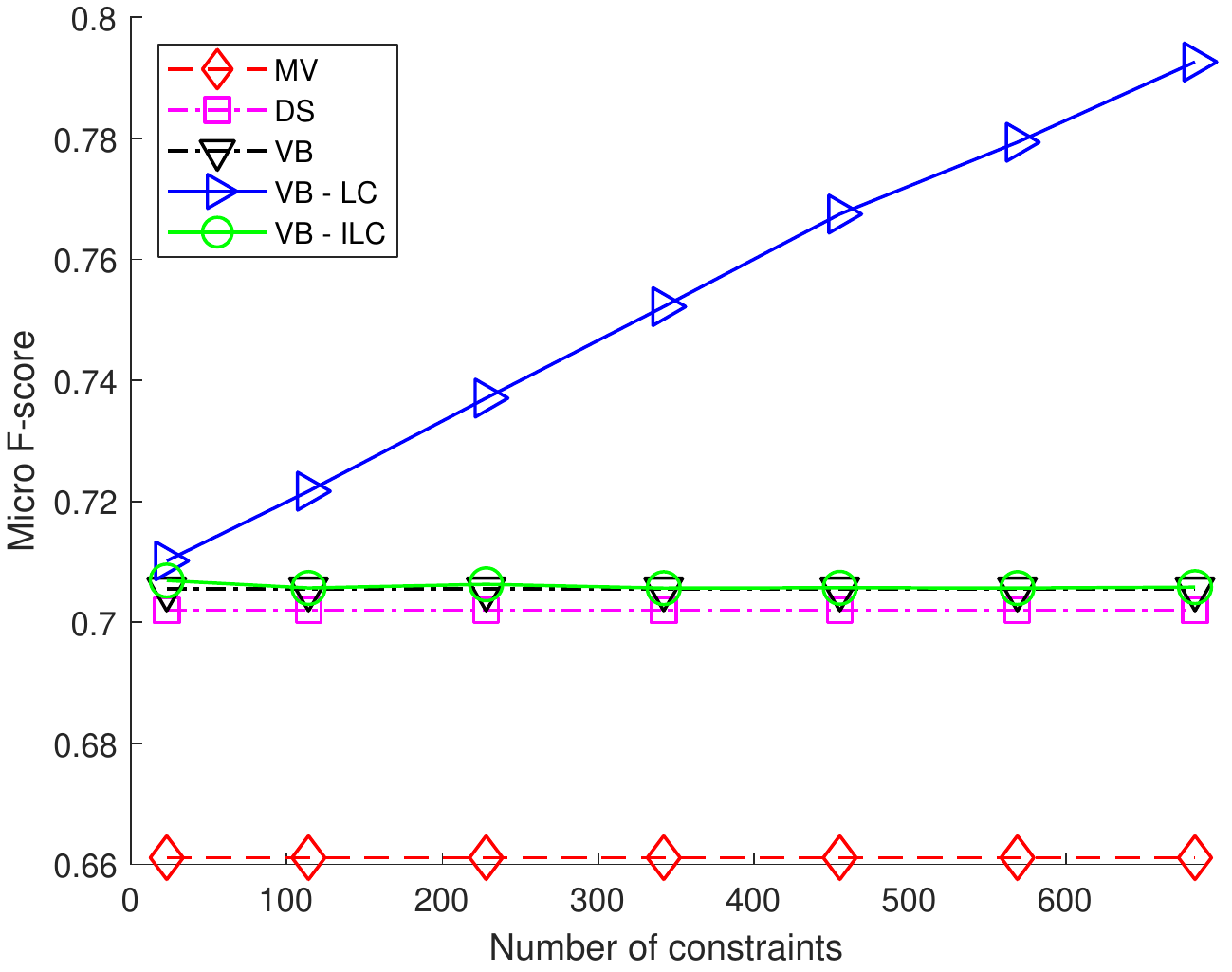}
         \caption{Micro F-score}
         \label{fig:trec_micro}
     \end{subfigure}
     \begin{subfigure}[b]{0.9\textwidth}
         \centering
         \includegraphics[width=\textwidth]{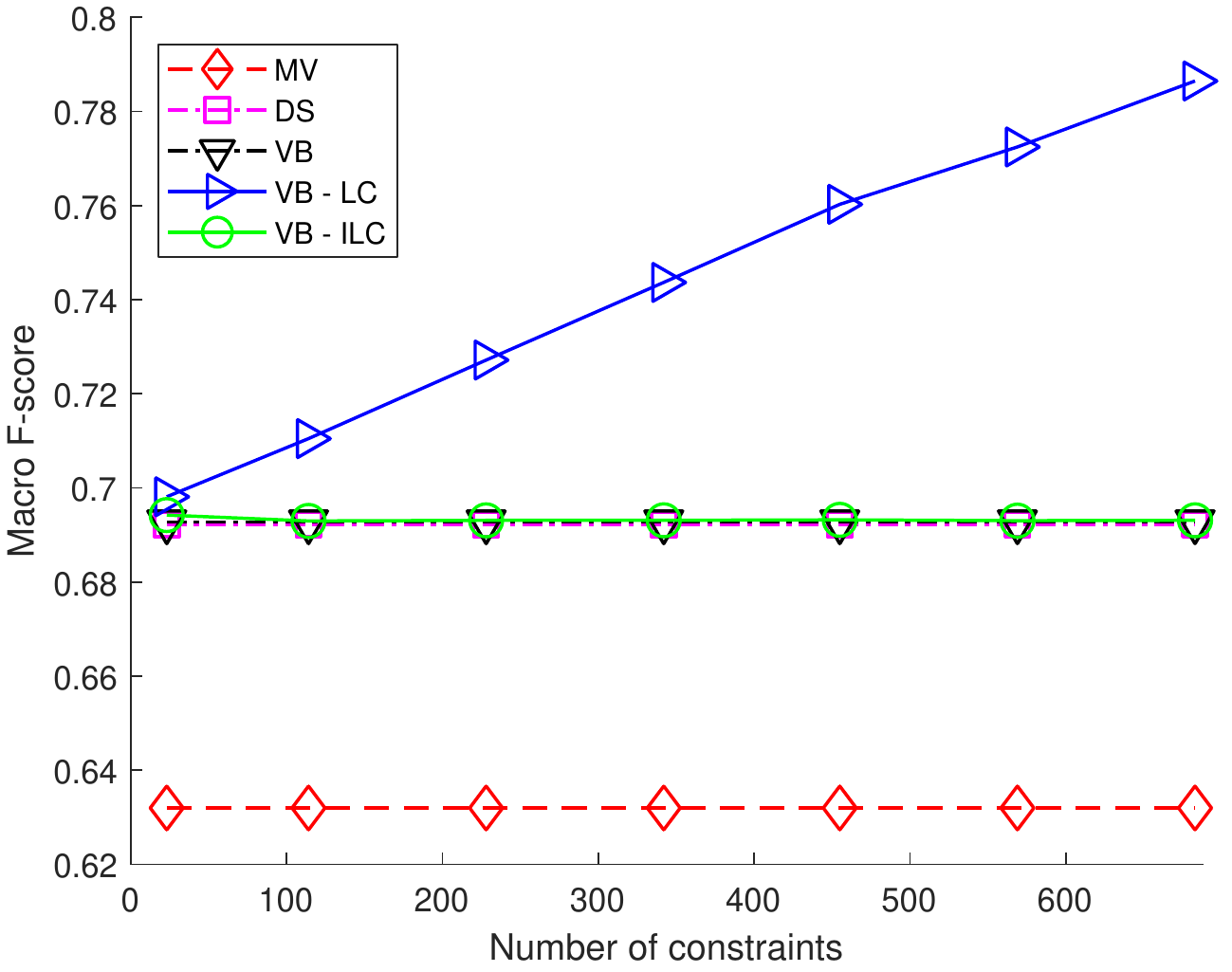}
         \caption{Macro F-score}
         \label{fig:trec_macro}
     \end{subfigure}
        \caption{Results for the TREC~\cite{Lease11-trec} dataset, with randomly sampled constraints.}
        \label{fig:trec_res}
    \end{minipage}
\end{figure}

\begin{figure}[tb]
     \centering
     \begin{subfigure}[b]{0.3\textwidth}
         \centering
         \includegraphics[width=\textwidth]{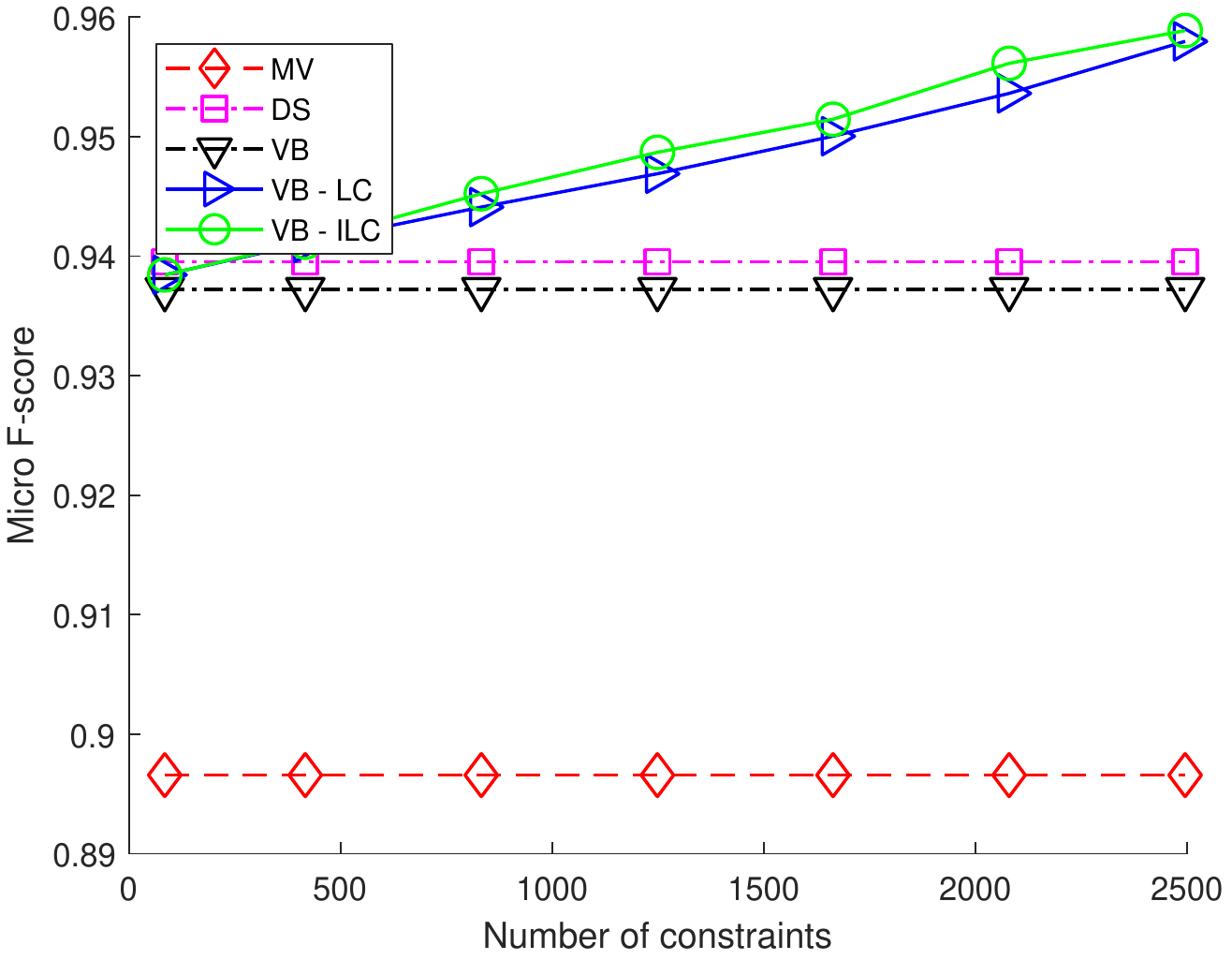}
         \caption{Micro F-score}
         \label{fig:prod_micro}
     \end{subfigure}\\
     \begin{subfigure}[b]{0.3\textwidth}
         \centering
         \includegraphics[width=\textwidth]{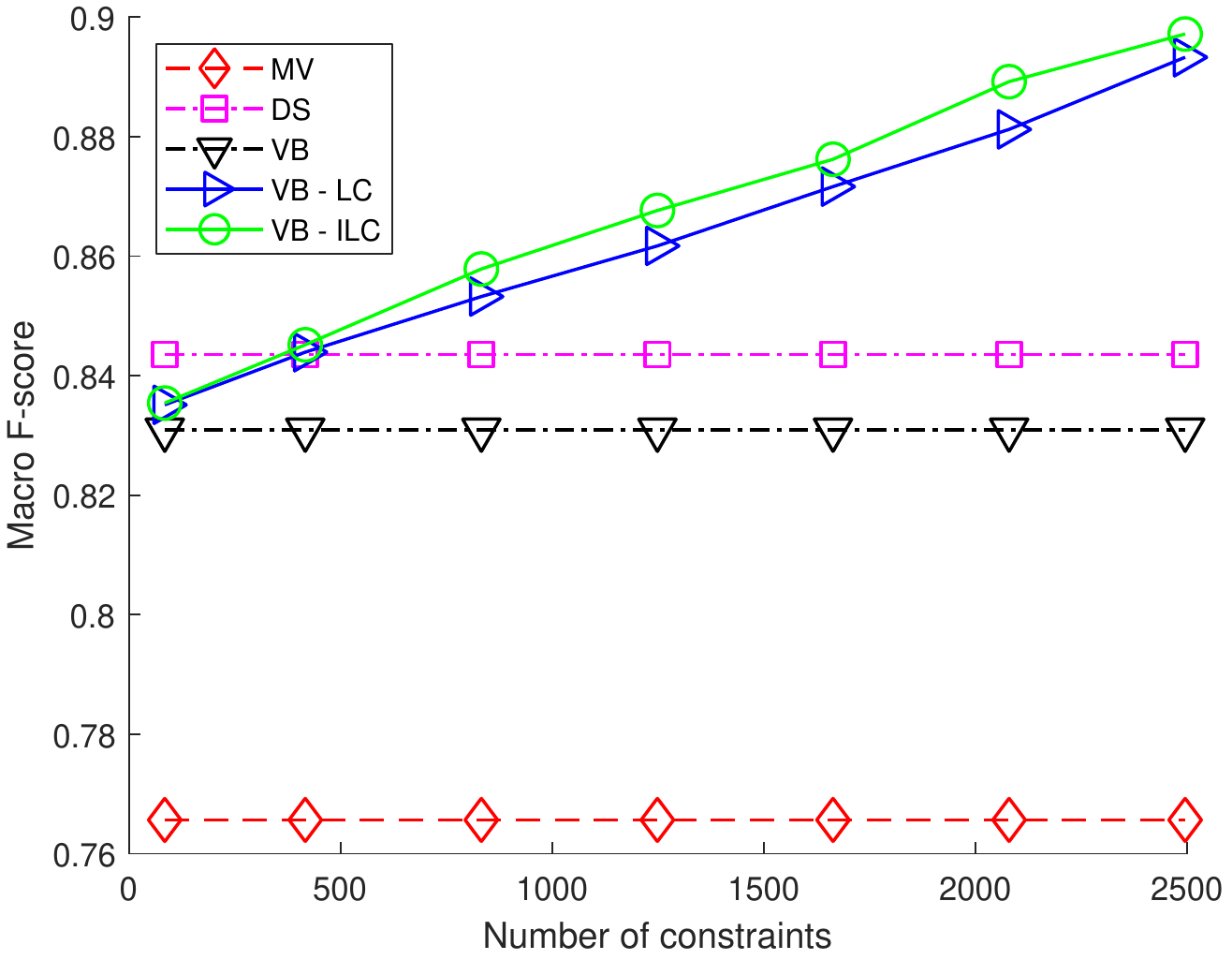}
         \caption{Macro F-score}
         \label{fig:prod_macro}
     \end{subfigure}
        \caption{Results for the Product~\cite{crowder} dataset, with randomly sampled constraints.}
        \label{fig:prod_res}
\end{figure}

\begin{figure}[tb]
\centering
\begin{subfigure}[b]{0.24\textwidth}
\centering
\includegraphics[width=\textwidth]{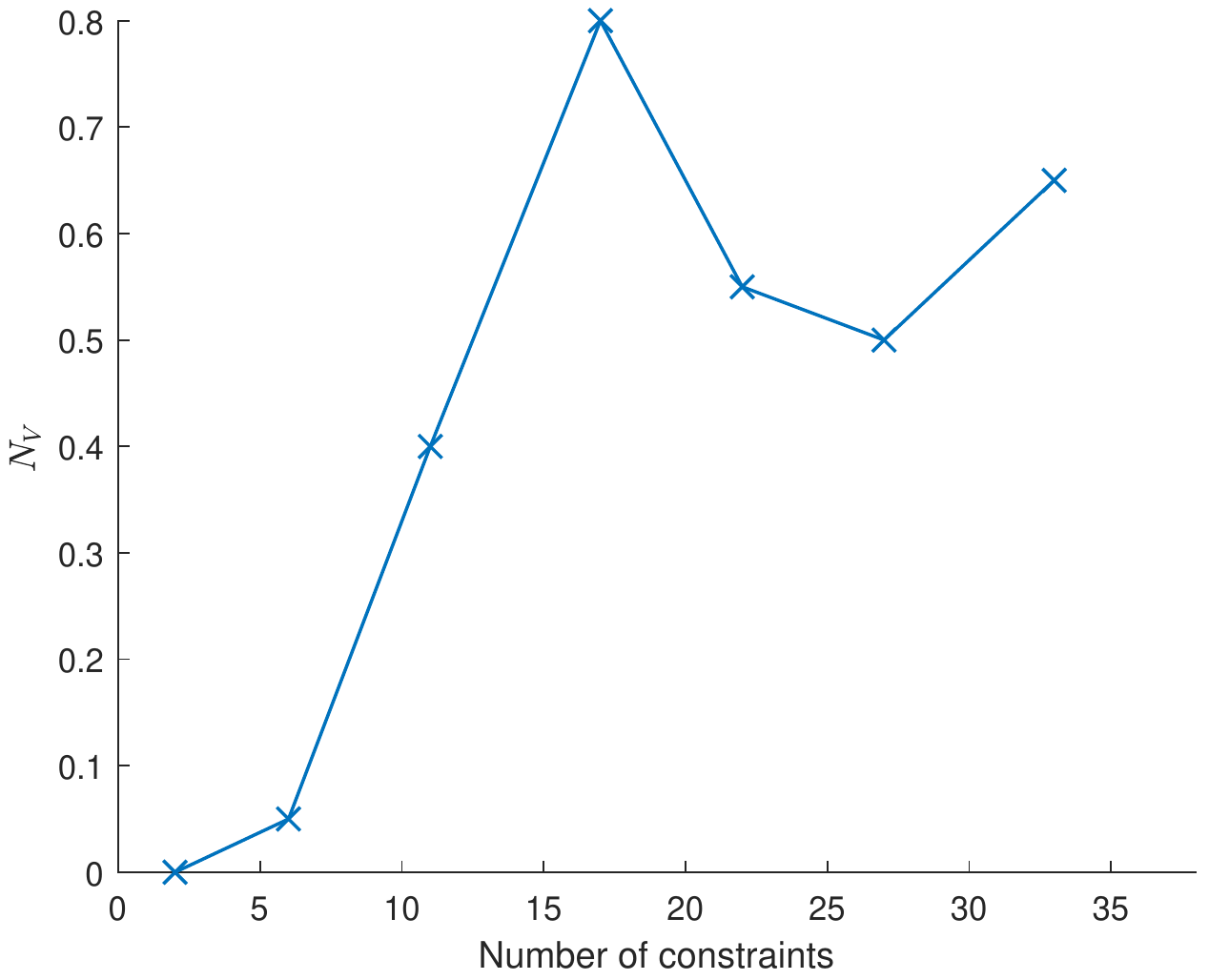}
\caption{Bluebird}
\label{fig:exp1:num_violated:bluebird}
\end{subfigure}
\begin{subfigure}[b]{0.24\textwidth}
\centering
\includegraphics[width=\textwidth]{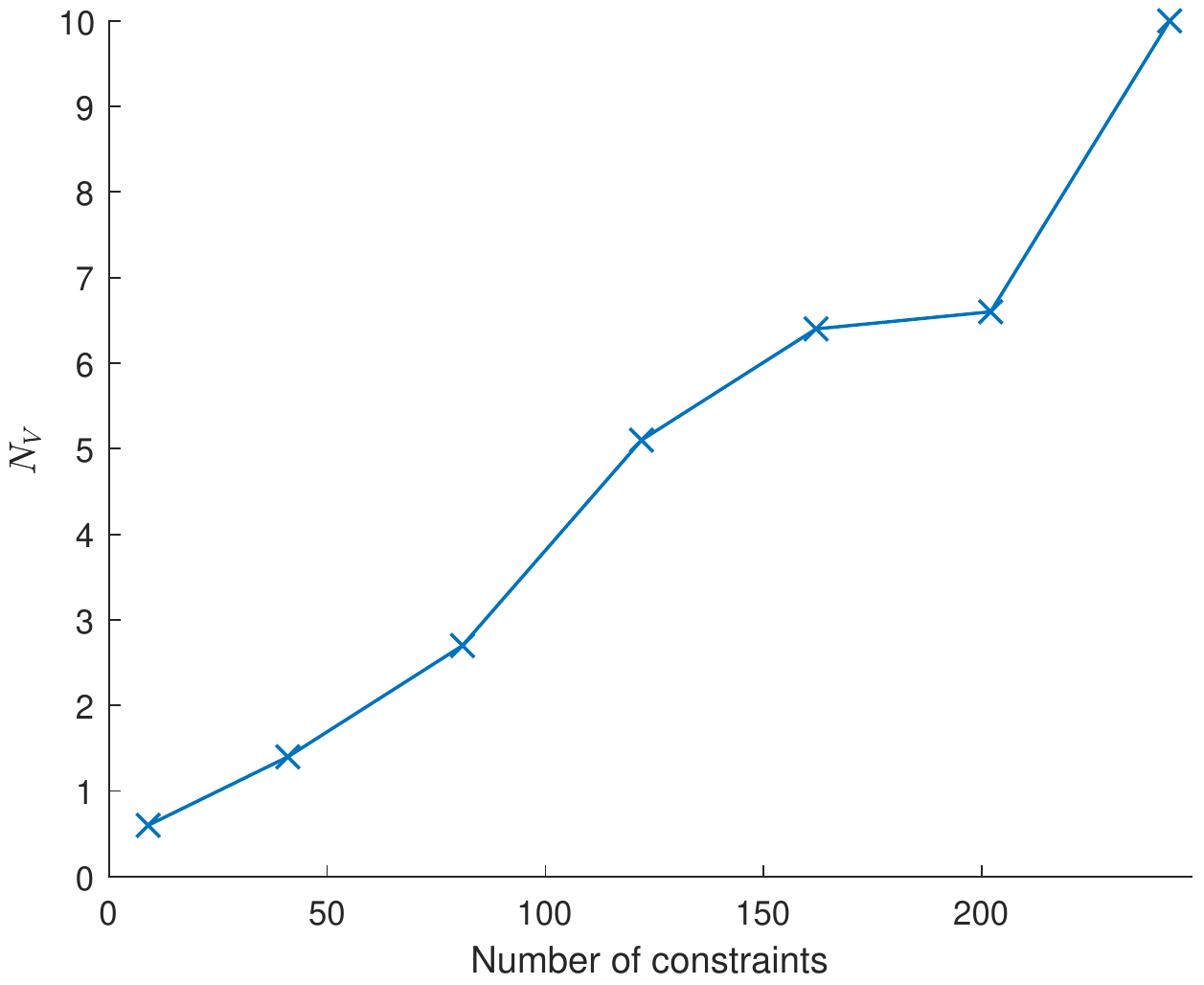}
\caption{Dog}
\label{fig:exp1:num_violated:dog}
\end{subfigure}
\begin{subfigure}[b]{0.24\textwidth}
\centering
\includegraphics[width=\textwidth]{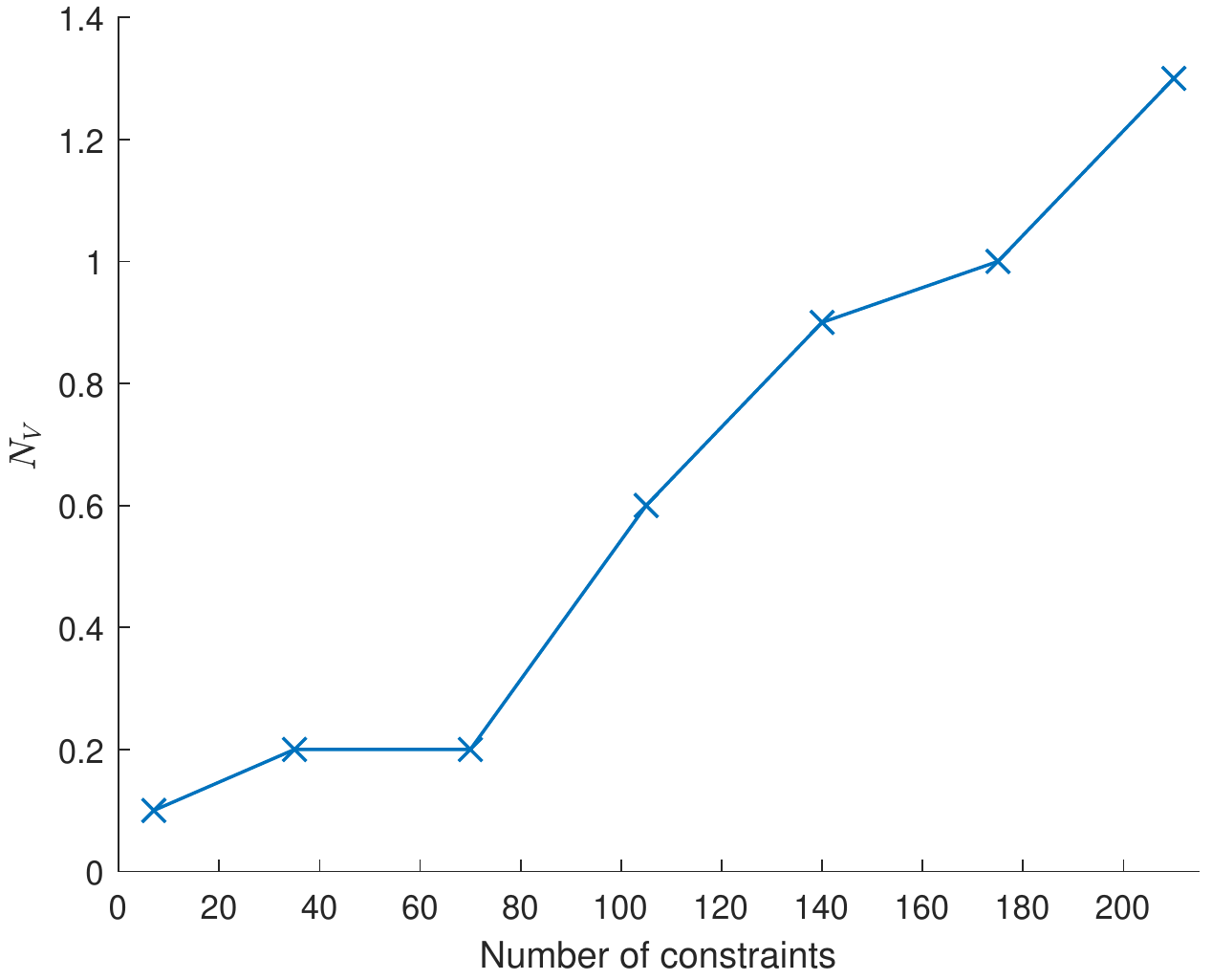}
\caption{Music Genre}
\label{fig:exp1:num_violated:music}
\end{subfigure}
\begin{subfigure}[b]{0.24\textwidth}
\centering
\includegraphics[width=\textwidth]{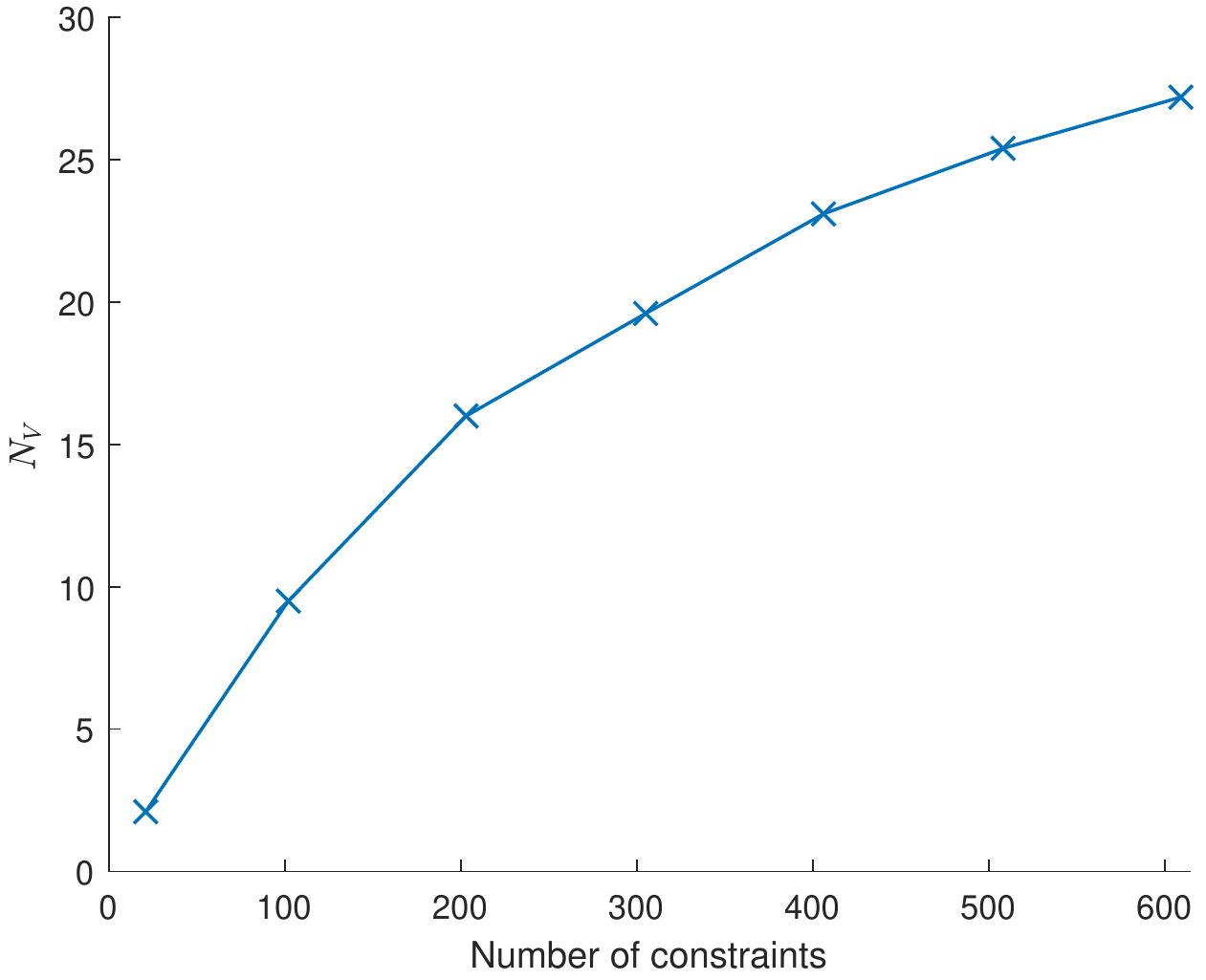}
\caption{ZenCrowd India}
\label{fig:exp1:num_violated:ZenCrowd_in}
\end{subfigure}
\\
\begin{subfigure}[b]{0.24\textwidth}
\centering
\includegraphics[width=\textwidth]{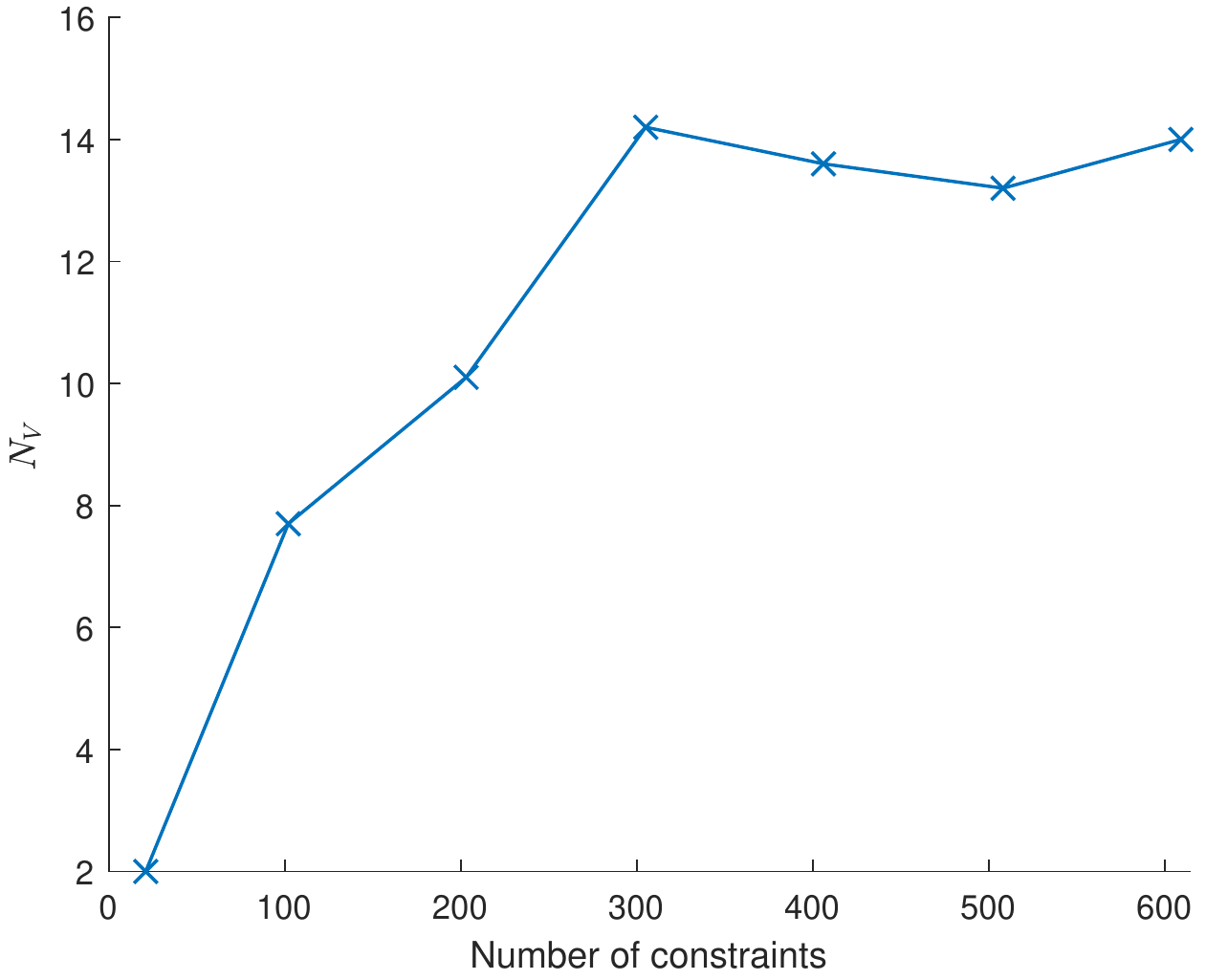}
\caption{ZenCrowd US}
\label{fig:exp1:num_violated:ZenCrowd_US}
\end{subfigure}
\begin{subfigure}[b]{0.24\textwidth}
\centering
\includegraphics[width=\textwidth]{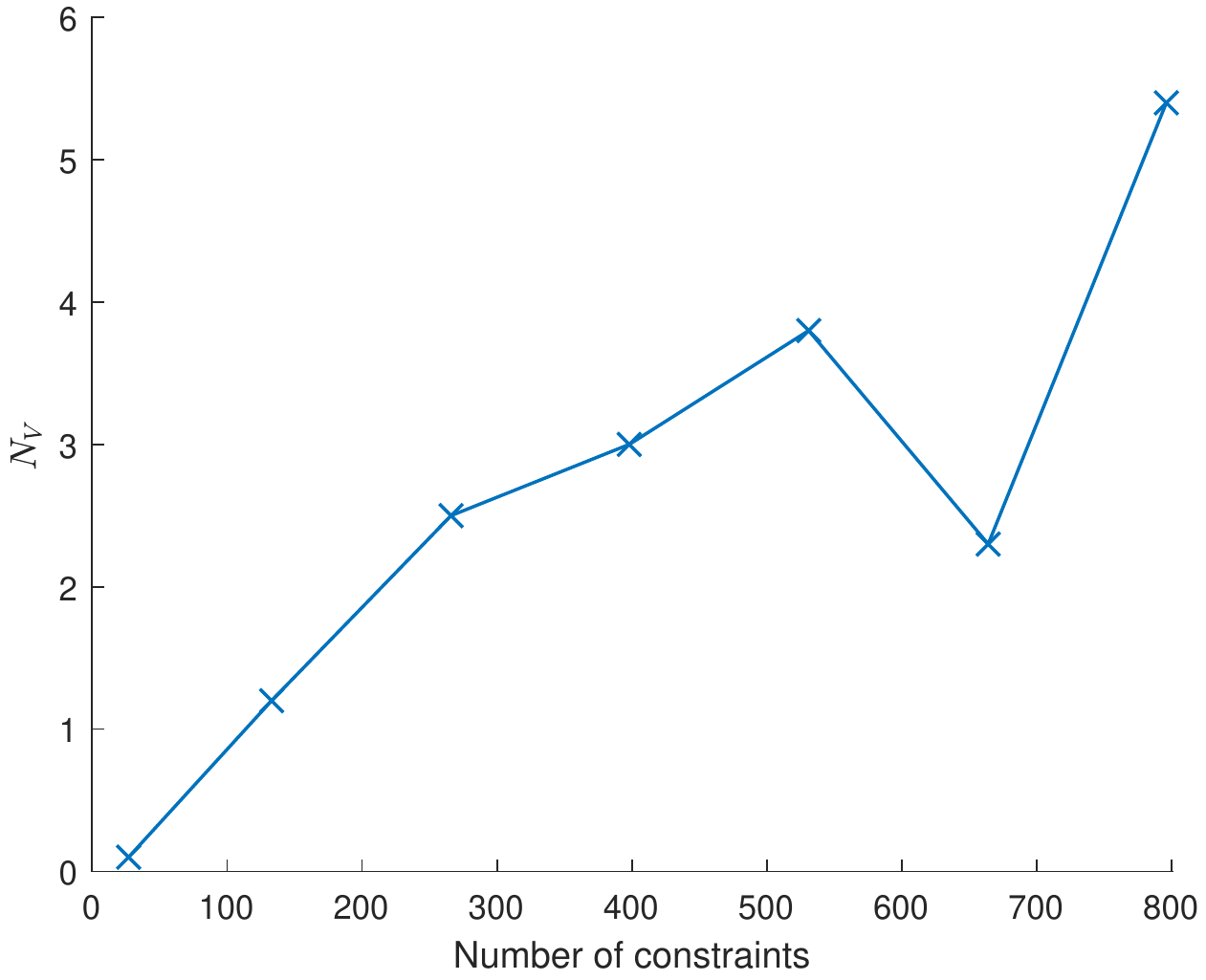}
\caption{Web}
\label{fig:exp1:num_violated:web}
\end{subfigure}
\begin{subfigure}[b]{0.24\textwidth}
\centering
\includegraphics[width=\textwidth]{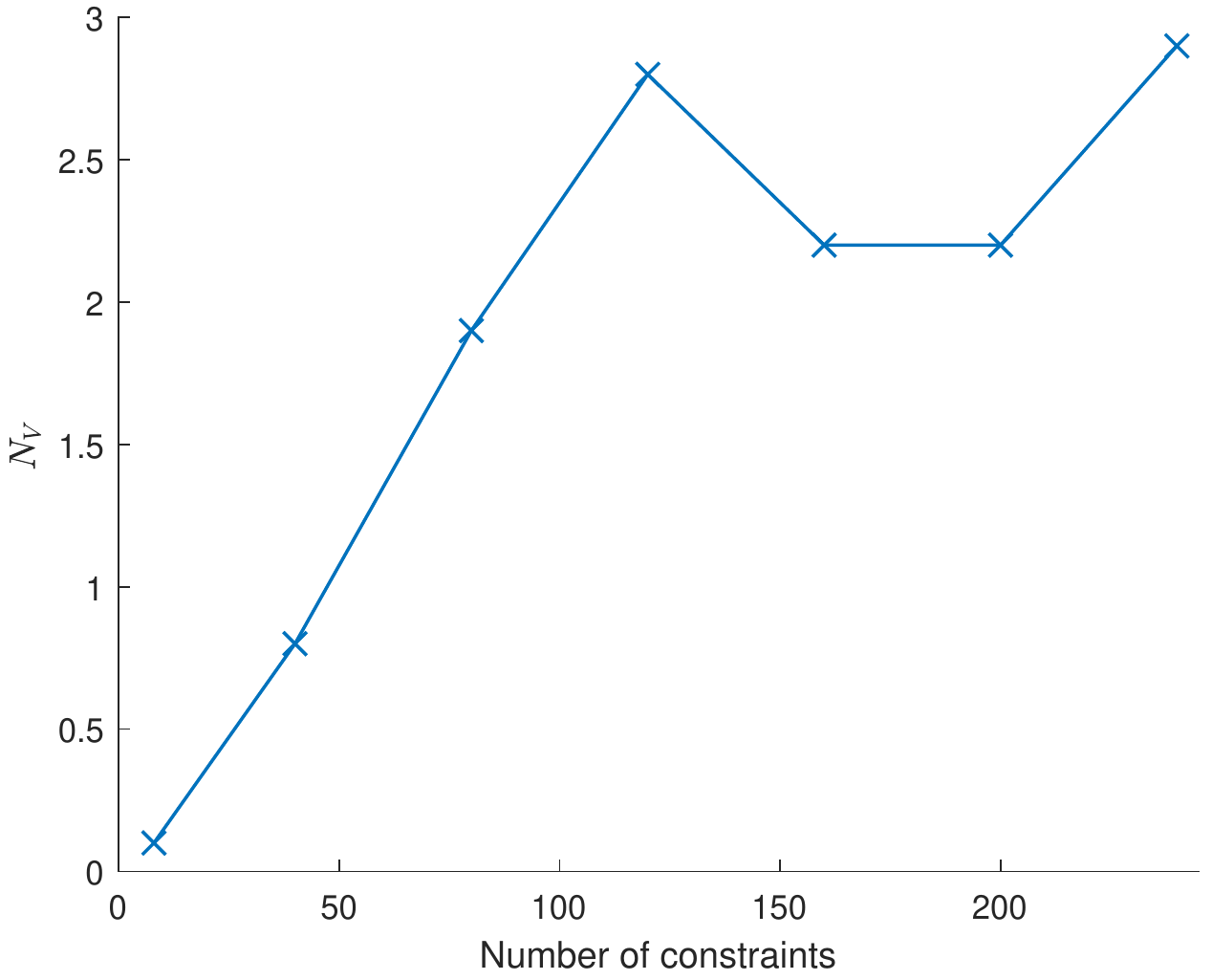}
\caption{RTE}
\label{fig:exp1:num_violated:rte}
\end{subfigure}
\begin{subfigure}[b]{0.24\textwidth}
\centering
\includegraphics[width=\textwidth]{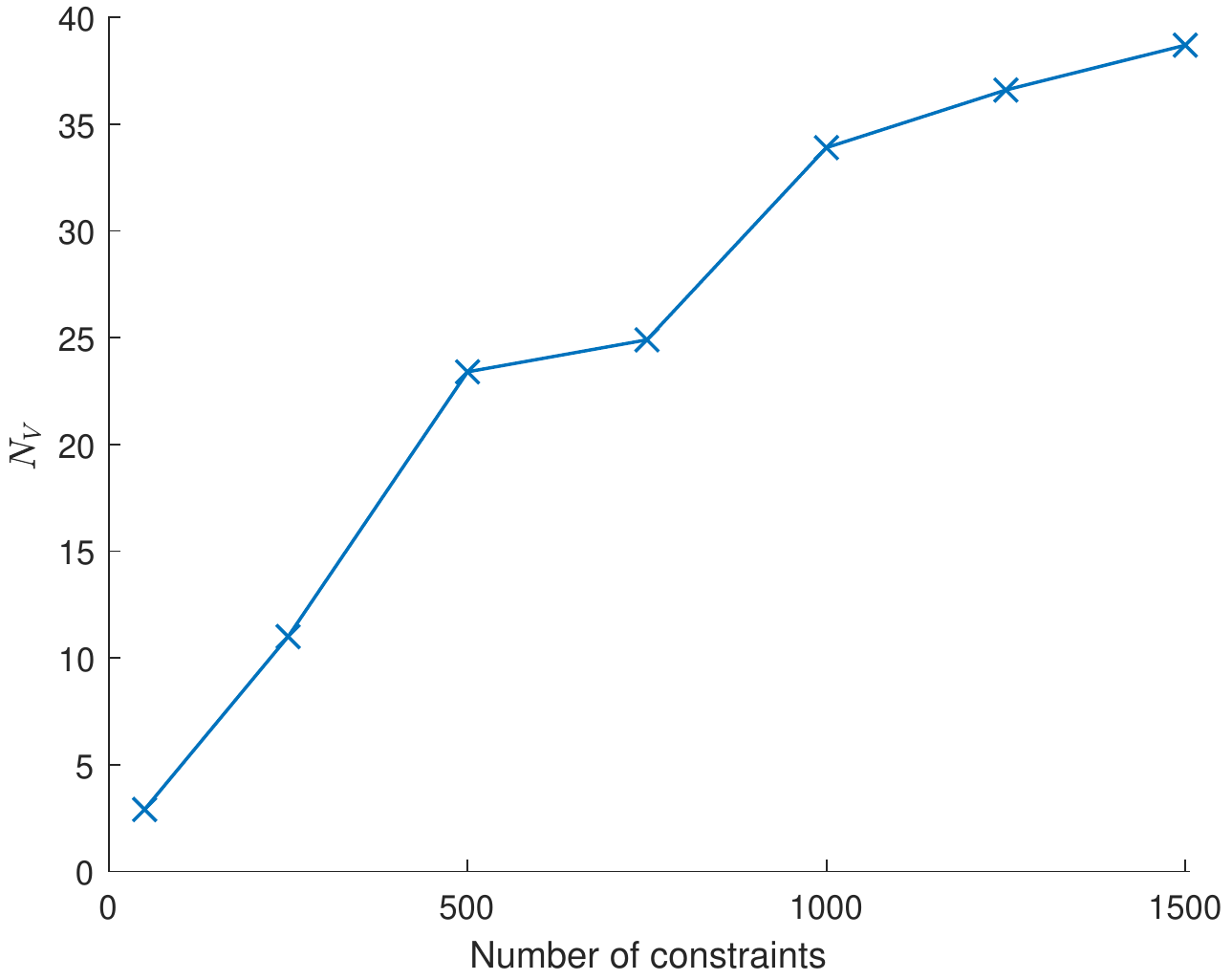}
\caption{Sentence Polarity}
\label{fig:exp1:num_violated:sen_polarity}
\end{subfigure}
\\
\begin{subfigure}[b]{0.24\textwidth}
\centering
\includegraphics[width=\textwidth]{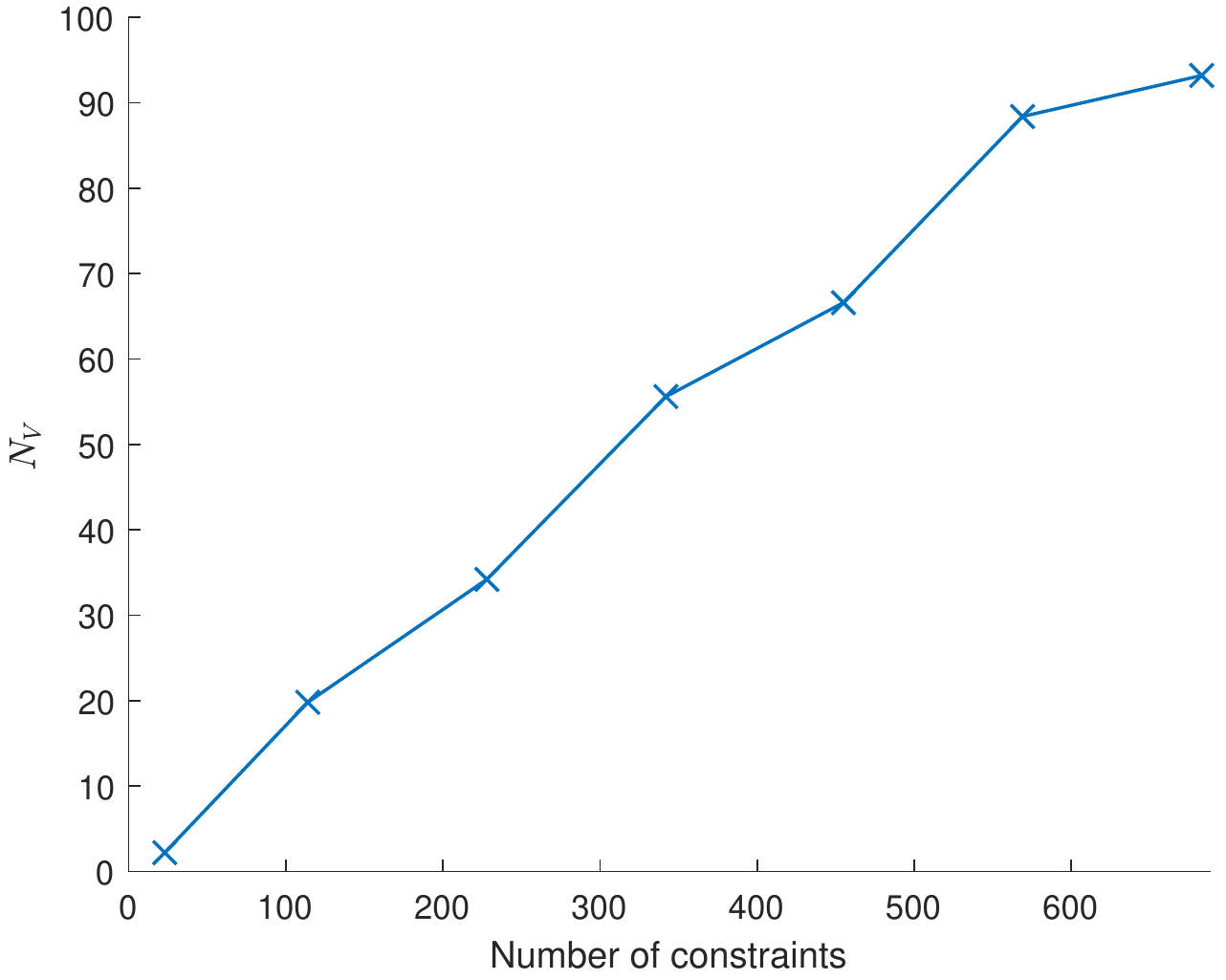}
\caption{TREC}
\label{fig:exp1:num_violated:trec}
\end{subfigure}
\begin{subfigure}[b]{0.24\textwidth}
\centering
\includegraphics[width=\textwidth]{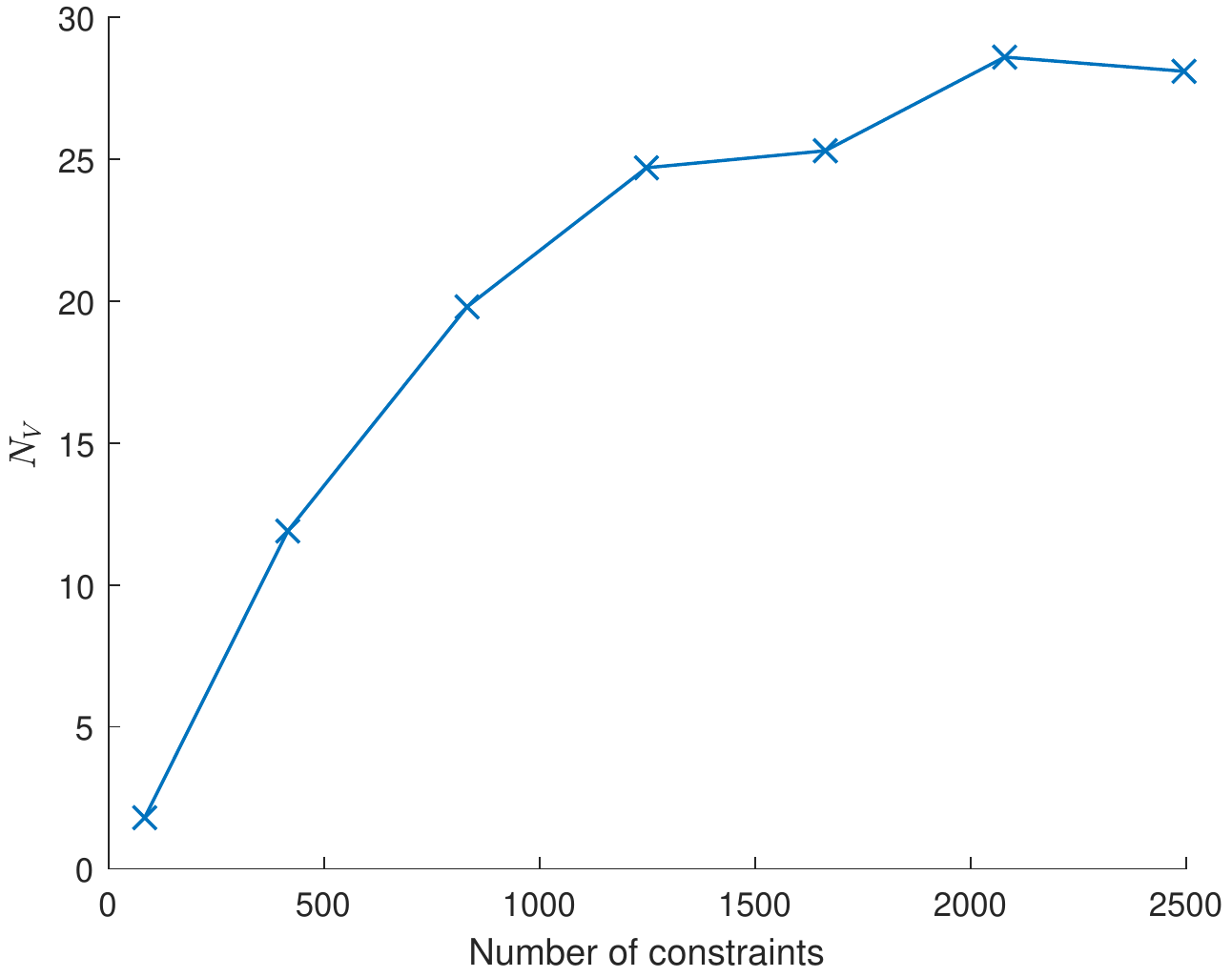}
\caption{Product}
\label{fig:exp1:num_violated:prod}
\end{subfigure}
\caption{Number of violated constraints for \emph{VB - ILC} per dataset, with randomly selected constraints.}
\label{fig:exp1:num_violated}
\end{figure}

For the first set of numerical tests,  \emph{VB - LC} and \emph{VB - ILC} are provided with the same amount of constraints, that is $N_C$ label constraints for \emph{VB - LC} and $N_C$ instance-level constraints for \emph{VB - ILC}. In this scenario, \emph{VB - LC} is expected to outperform \emph{VB - ILC}, since instance-level constraints provide ``weaker'' information than label constraints. Figs.~\ref{fig:bluebird_res}-\ref{fig:prod_res} show the results w.r.t. Micro and Macro-averaged F-scores for the datasets considered, as the number of constraints increases. As expected in most datasets \emph{VB - LC} outperforms \emph{VB - ILC}, however both perform better than their unsupervised counterparts, that do not take into account label or instance level information. For the ZenCrowd US dataset, in Figs.~\ref{fig:ZenCrowd_us_res}, \emph{VB - LC} and \emph{VB - ILC} require more constraints to outperform majority voting and the EM algorithm of Dawid and Skene~\cite{dawid1979maximum} respectively. This is probably due to the worse performance of VB compared to \emph{MV} and \emph{DS} in this datasets, when provided with the uniform priors mentioned in the main paper. Interestingly, in the RTE and Product datasets (Fig.~\ref{fig:prod_res}) \emph{VB - ILC} outperforms \emph{VB - LC}. Fig.~\ref{fig:exp1:num_violated} shows the number of violated constraints $N_V$ for \emph{VB - ILC} for each dataset, as the number of randomly selected constraints increases. For all datasets, as the number of constraints increases, so does $N_V$, however $N_V$ remains at relatively low levels, except for the TREC dataset, where $N_V$ is a significant fraction of the constraints. This can potentially explain the marginal improvements of \emph{VB - ILC} over \emph{VB} in this dataset.

\subsection{Constraints selected via uncertainty sampling}
\begin{figure}
    \centering
    \begin{minipage}{0.32\textwidth}
        \centering
        \begin{subfigure}[b]{0.9\textwidth}
         \centering
         \includegraphics[width=\textwidth]{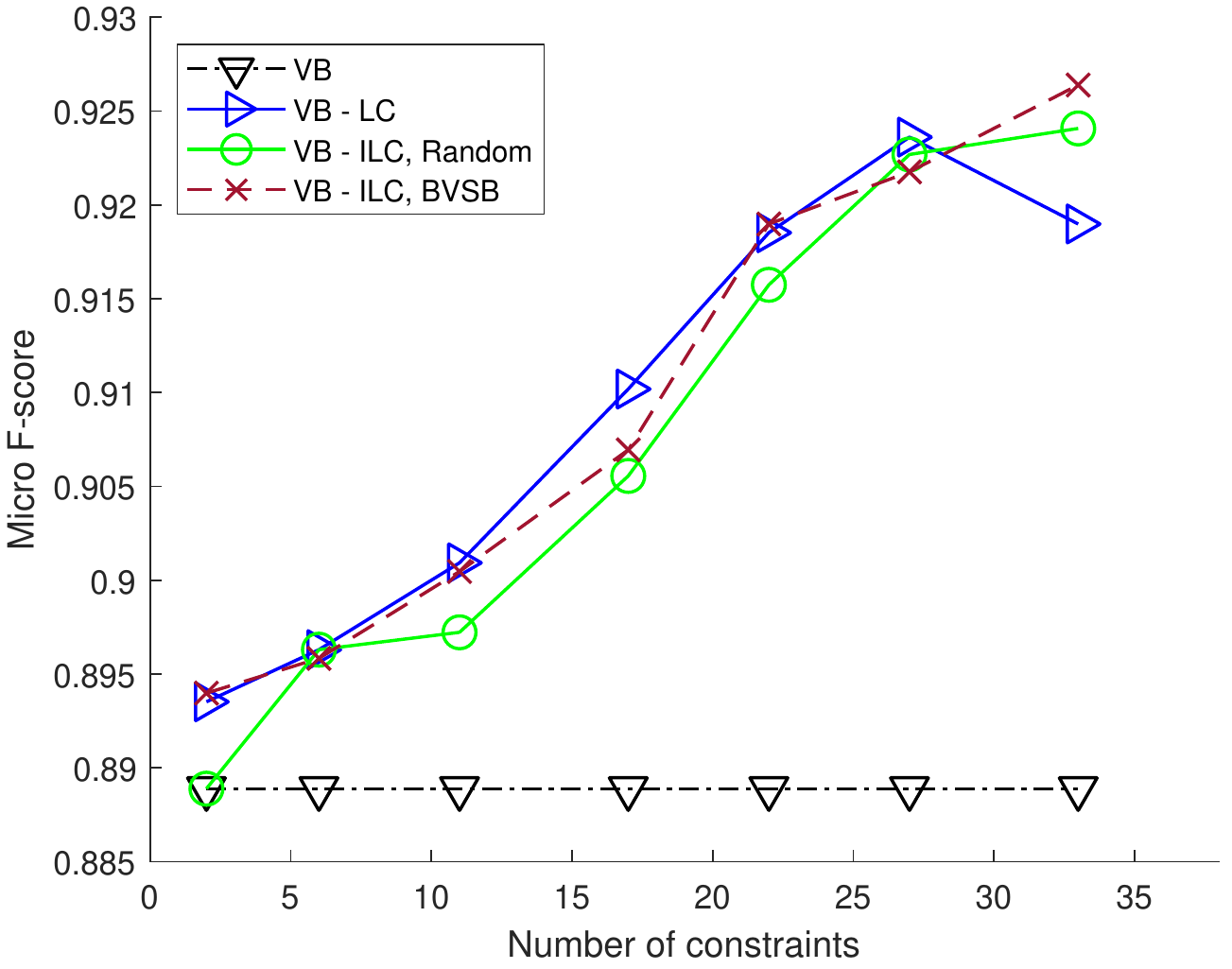}
         \caption{Micro F-score}
         \label{fig:bluebird_micro_exp2}
     \end{subfigure}\\
     \begin{subfigure}[b]{0.9\textwidth}
         \centering
         \includegraphics[width=\textwidth]{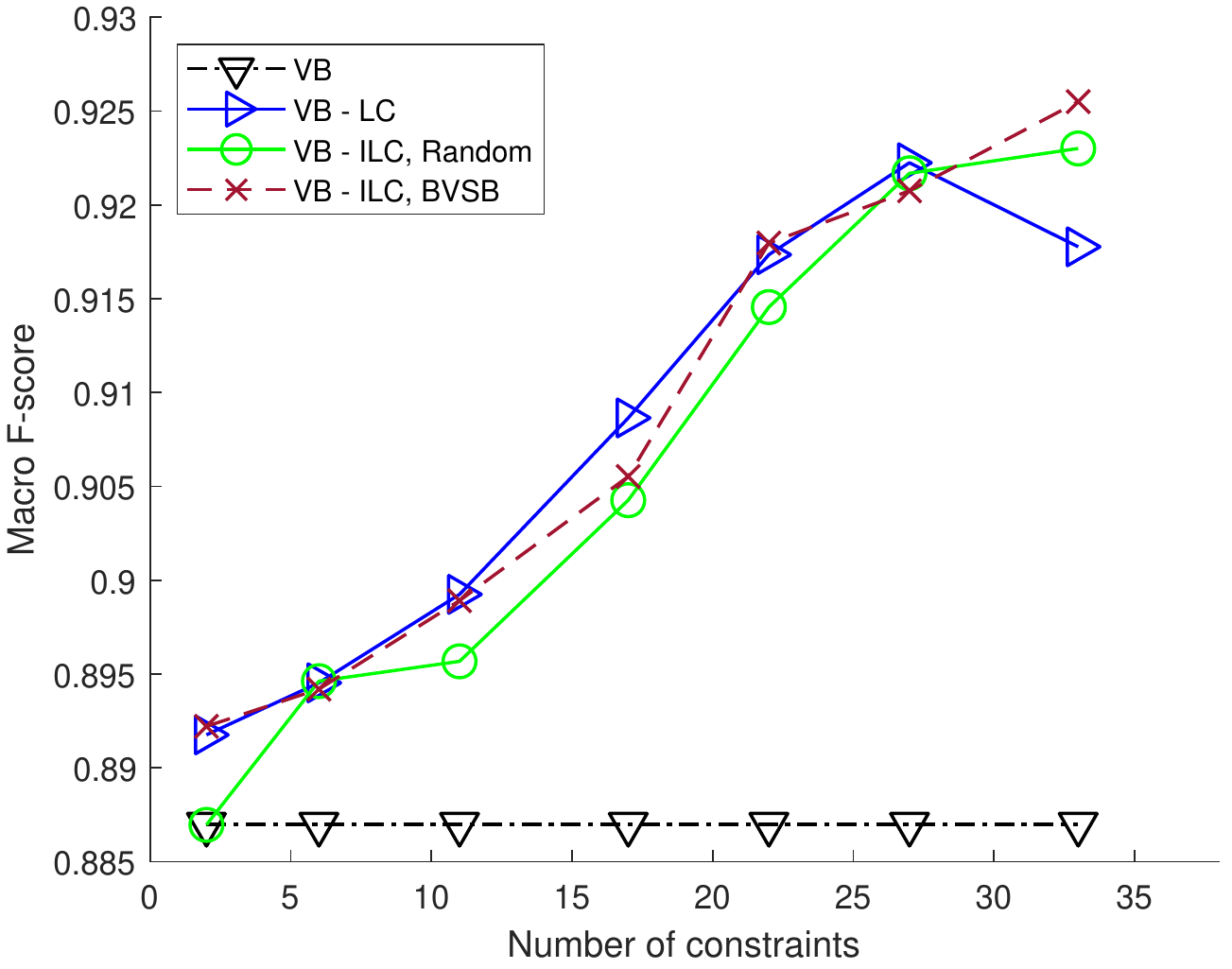}
         \caption{Macro F-score}
         \label{fig:bluebird_macro_exp2}
     \end{subfigure}
        \caption{Results for the Bluebird~\cite{multidim_wisdom} dataset, with constraints selected via uncertainty sampling.}
        \label{fig:bluebird_res_exp2}
    \end{minipage}\hfill
    \begin{minipage}{0.32\textwidth}
        \centering
     \begin{subfigure}[b]{0.9\textwidth}
         \centering
         \includegraphics[width=\textwidth]{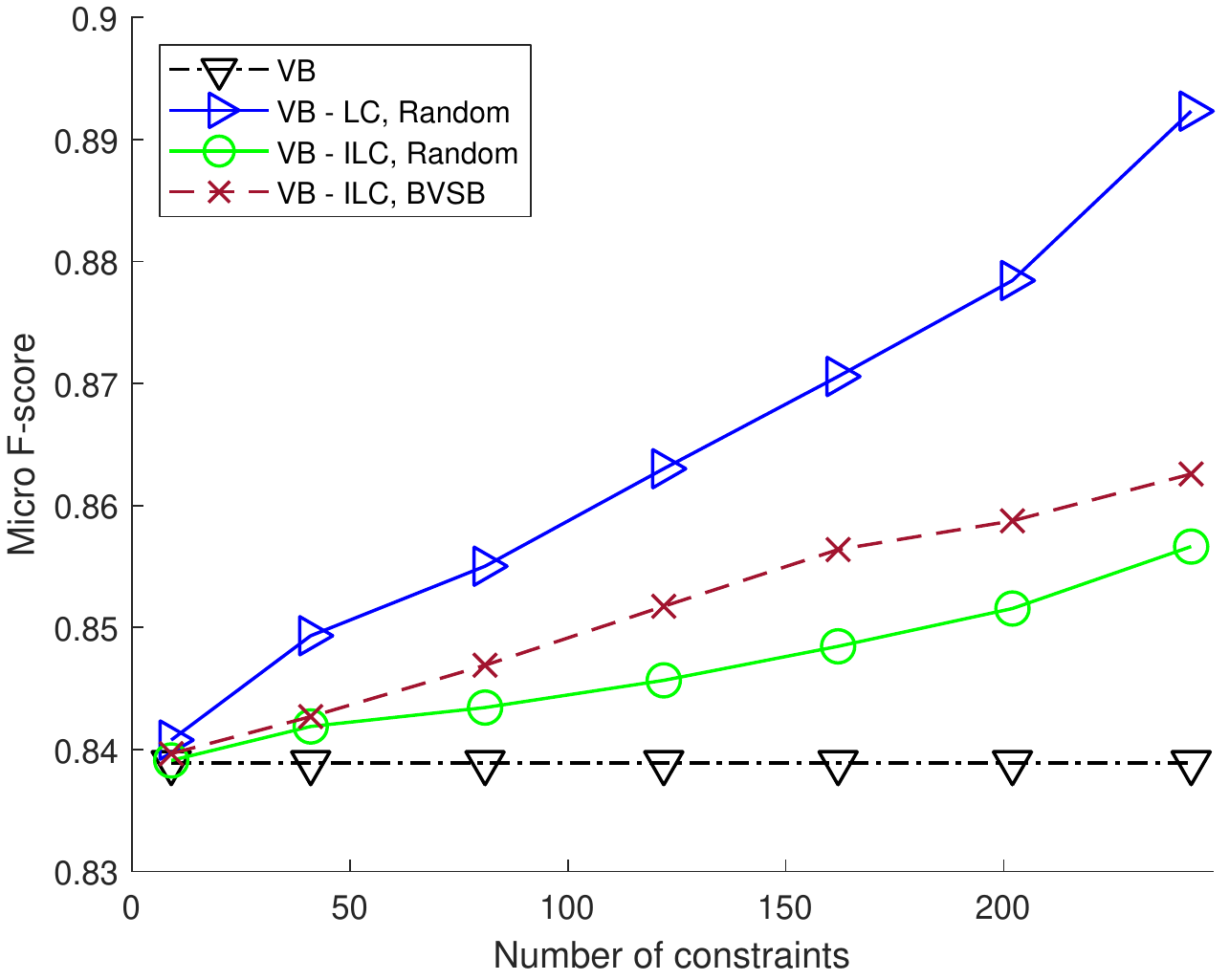}
         \caption{Micro F-score}
         \label{fig:dog_micro_exp2}
     \end{subfigure}
     \begin{subfigure}[b]{0.9\textwidth}
         \centering
         \includegraphics[width=\textwidth]{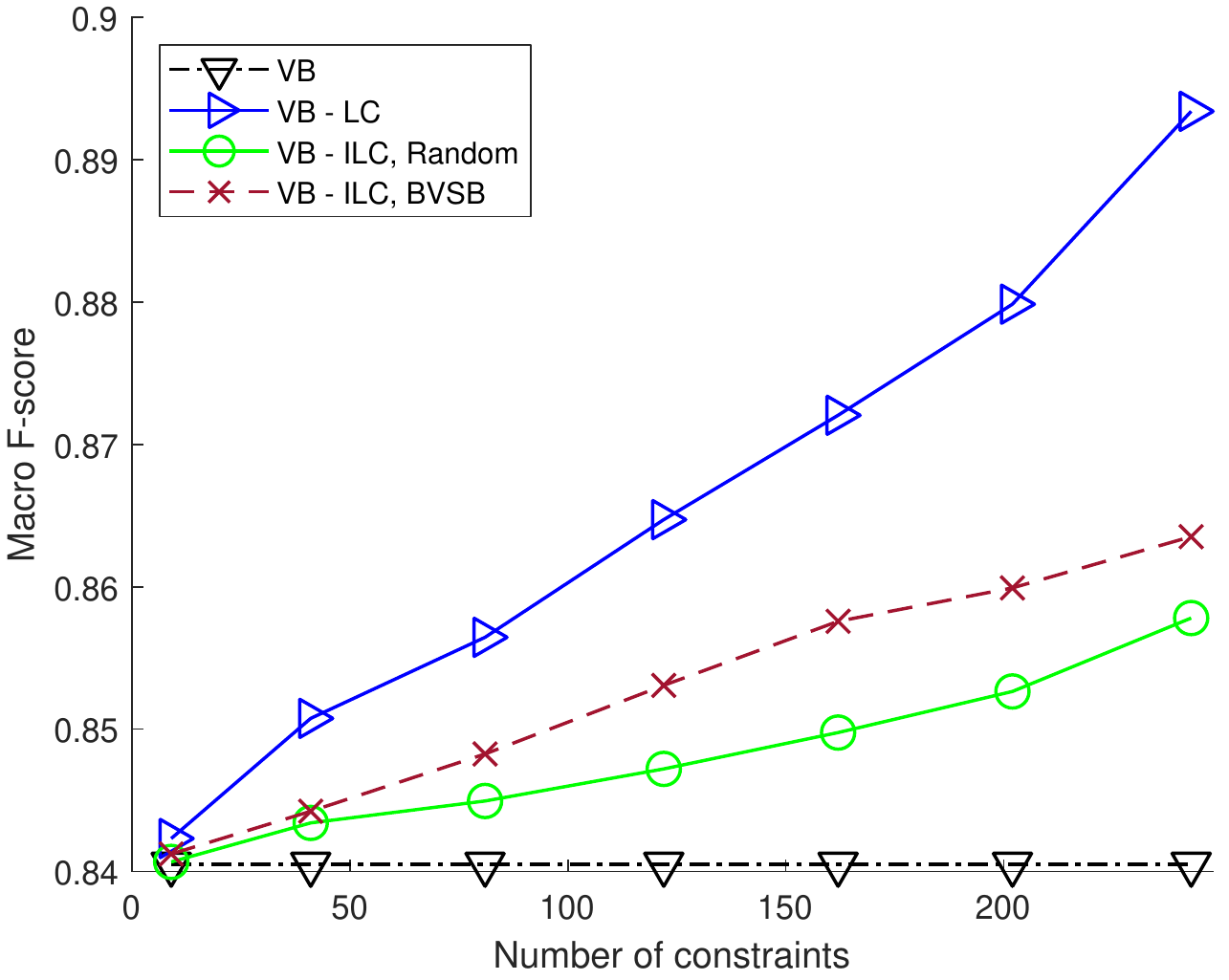}
         \caption{Macro F-score}
         \label{fig:dog_macro_exp2}
     \end{subfigure}
        \caption{Results for the Dog~\cite{imagenet} dataset, with constraints selected via uncertainty sampling.}
        \label{fig:dog_res_exp2}
    \end{minipage}
    \hfill
    \begin{minipage}{0.32\textwidth}
        \centering
        \begin{subfigure}[b]{0.9\textwidth}
         \centering
         \includegraphics[width=\textwidth]{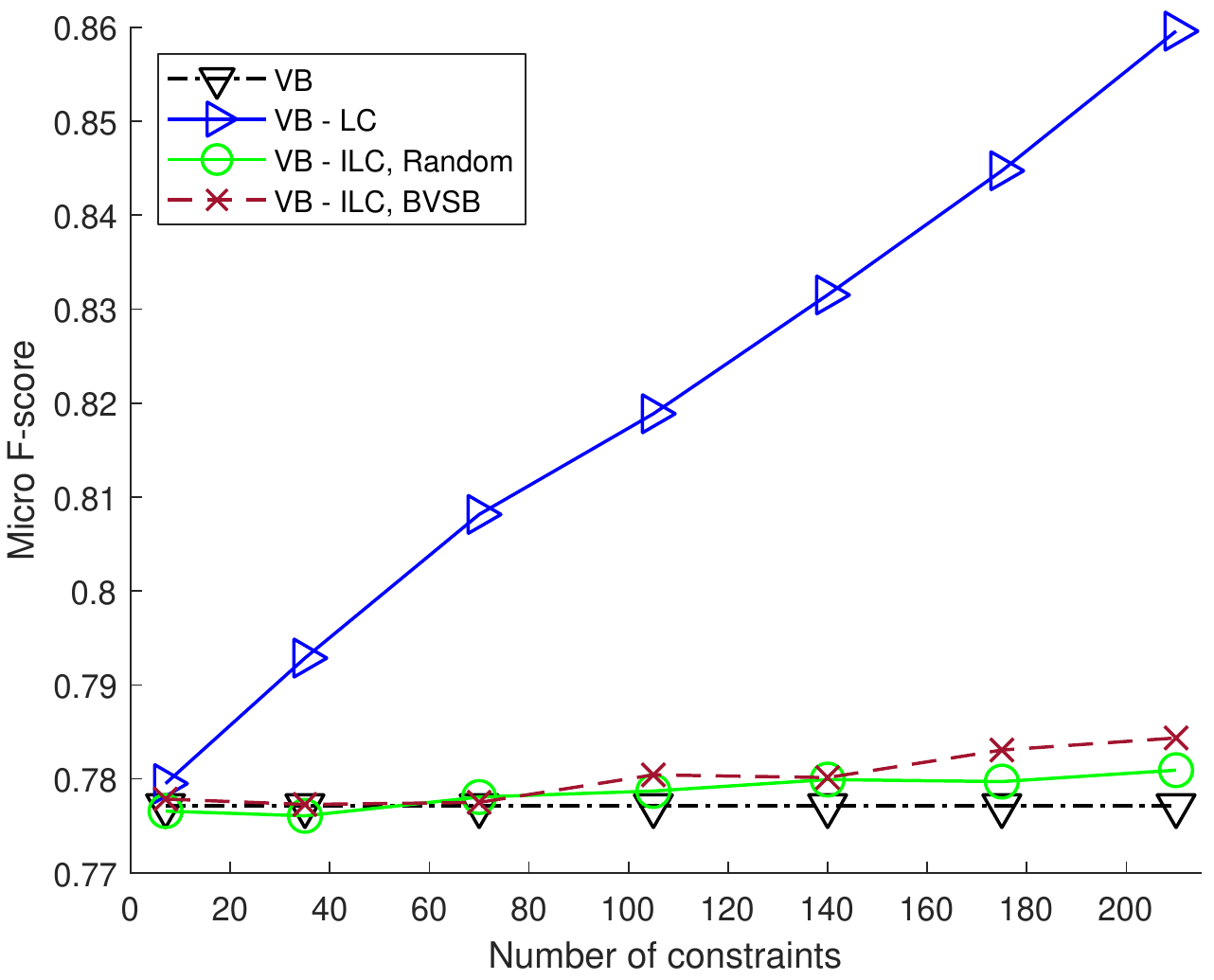}
         \caption{Micro F-score}
         \label{fig:music_micro_exp2}
     \end{subfigure}
     \begin{subfigure}[b]{0.9\textwidth}
         \centering
         \includegraphics[width=\textwidth]{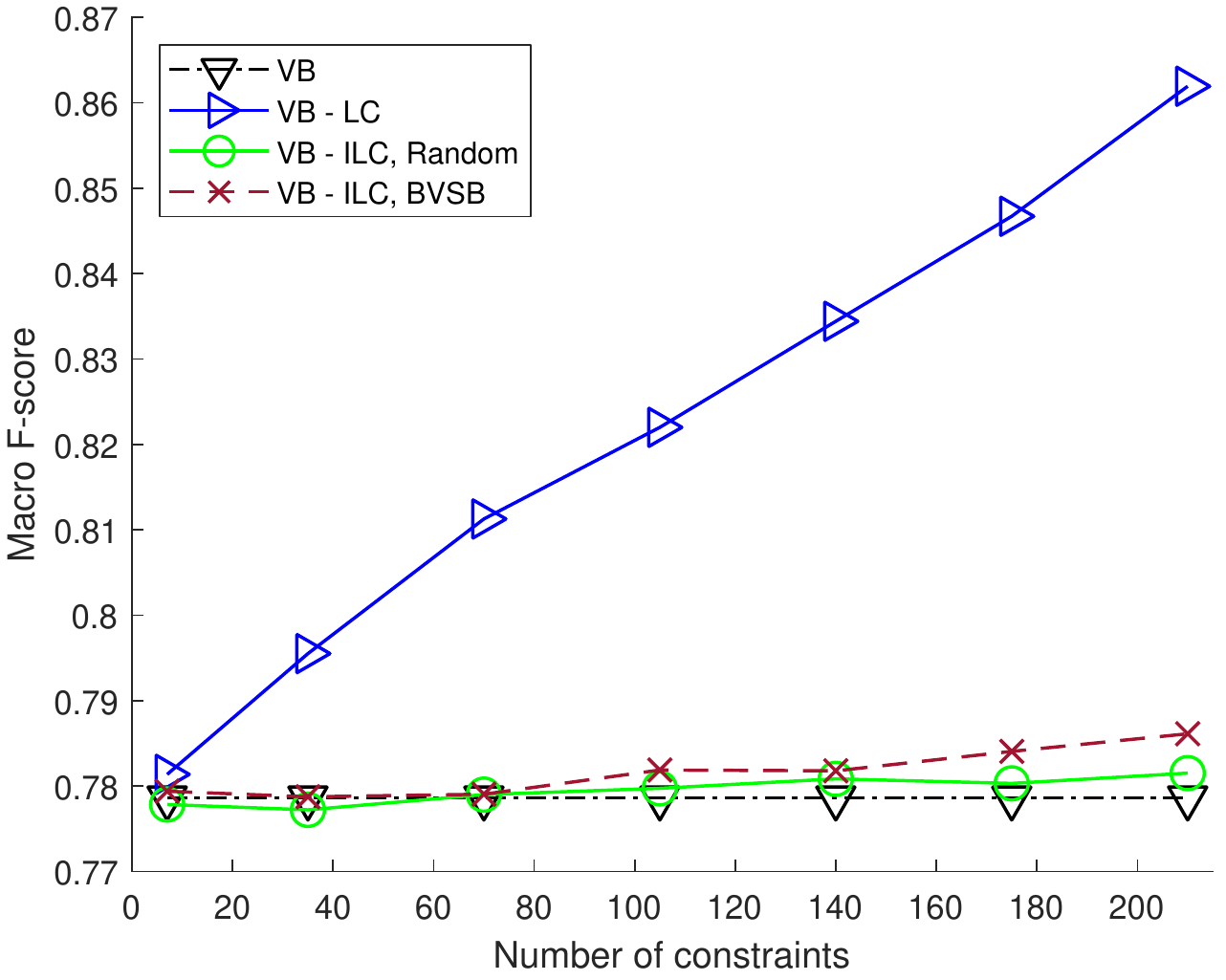}
         \caption{Macro F-score}
         \label{fig:music_macro_exp2}
     \end{subfigure}
        \caption{Results for the Music Genre~\cite{musicgenre_senpoldata} dataset, with constraints selected via uncertainty sampling.}
        \label{fig:music_res_exp2}
    \end{minipage}
\end{figure}

\begin{figure}
    \centering
    \begin{minipage}{0.32\textwidth}
        \centering
        \begin{subfigure}[b]{0.9\textwidth}
         \centering
         \includegraphics[width=\textwidth]{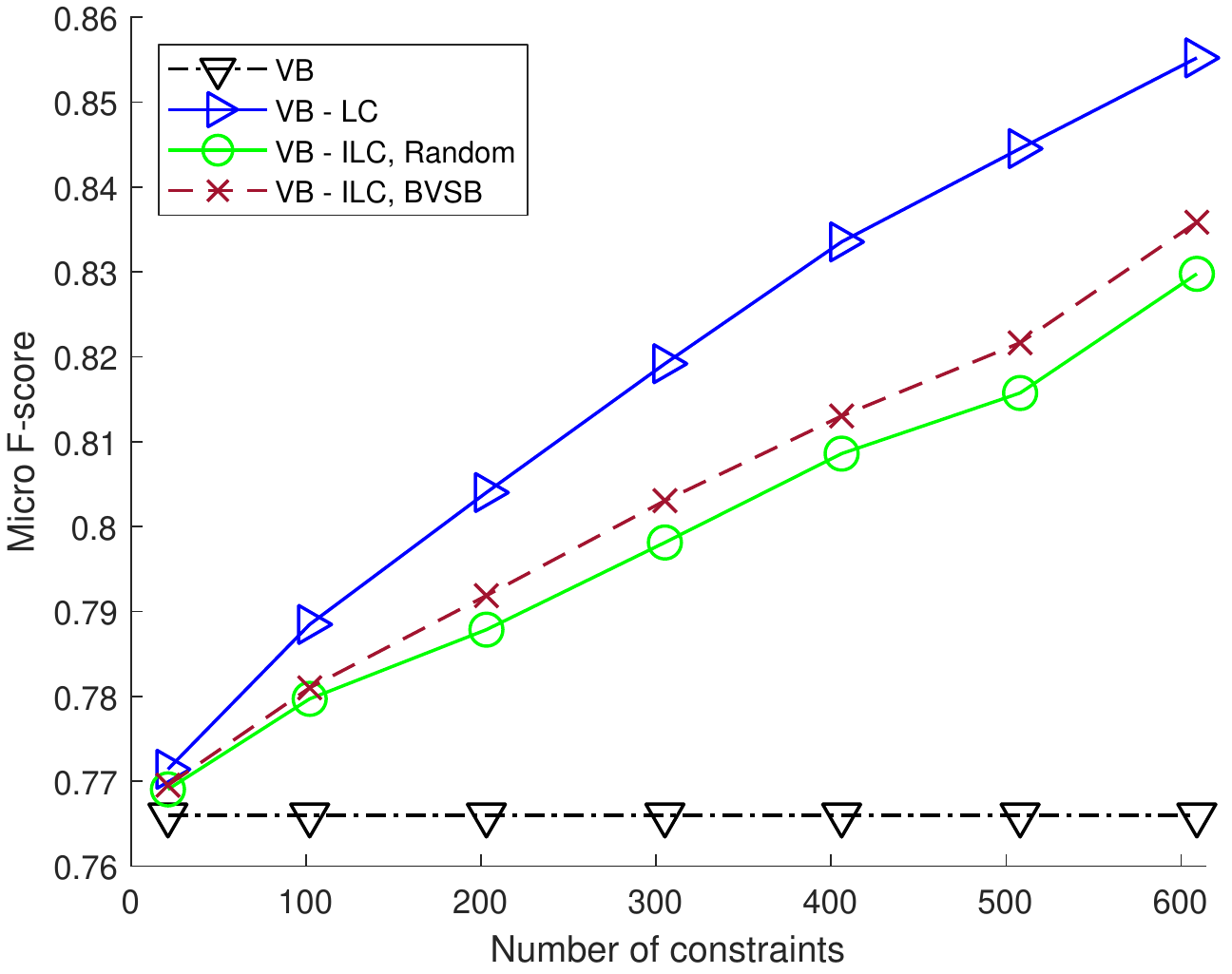}
         \caption{Micro F-score}
         \label{fig:ZenCrowd_in_micro_exp2}
     \end{subfigure}
     \begin{subfigure}[b]{0.9\textwidth}
         \centering
         \includegraphics[width=\textwidth]{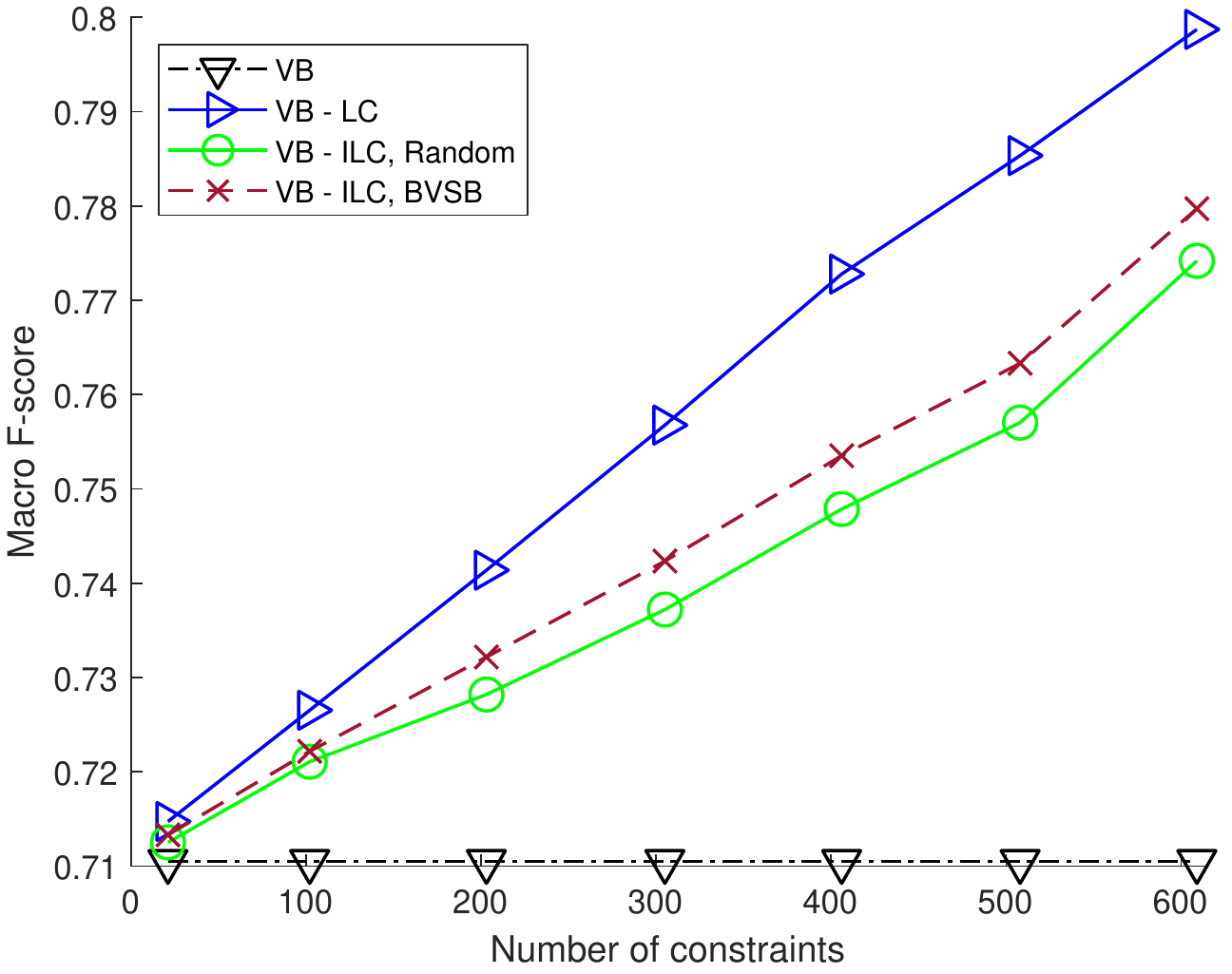}
         \caption{Macro F-score}
         \label{fig:ZenCrowd_in_macro_exp2}
     \end{subfigure}
        \caption{Results for the ZenCrowd India~\cite{ZenCrowd} dataset, with constraints selected via uncertainty sampling.}
        \label{fig:ZenCrowd_in_res_exp2}
    \end{minipage}\hfill
    \begin{minipage}{0.32\textwidth}
        \centering
        \begin{subfigure}[b]{0.9\textwidth}
         \centering
         \includegraphics[width=\textwidth]{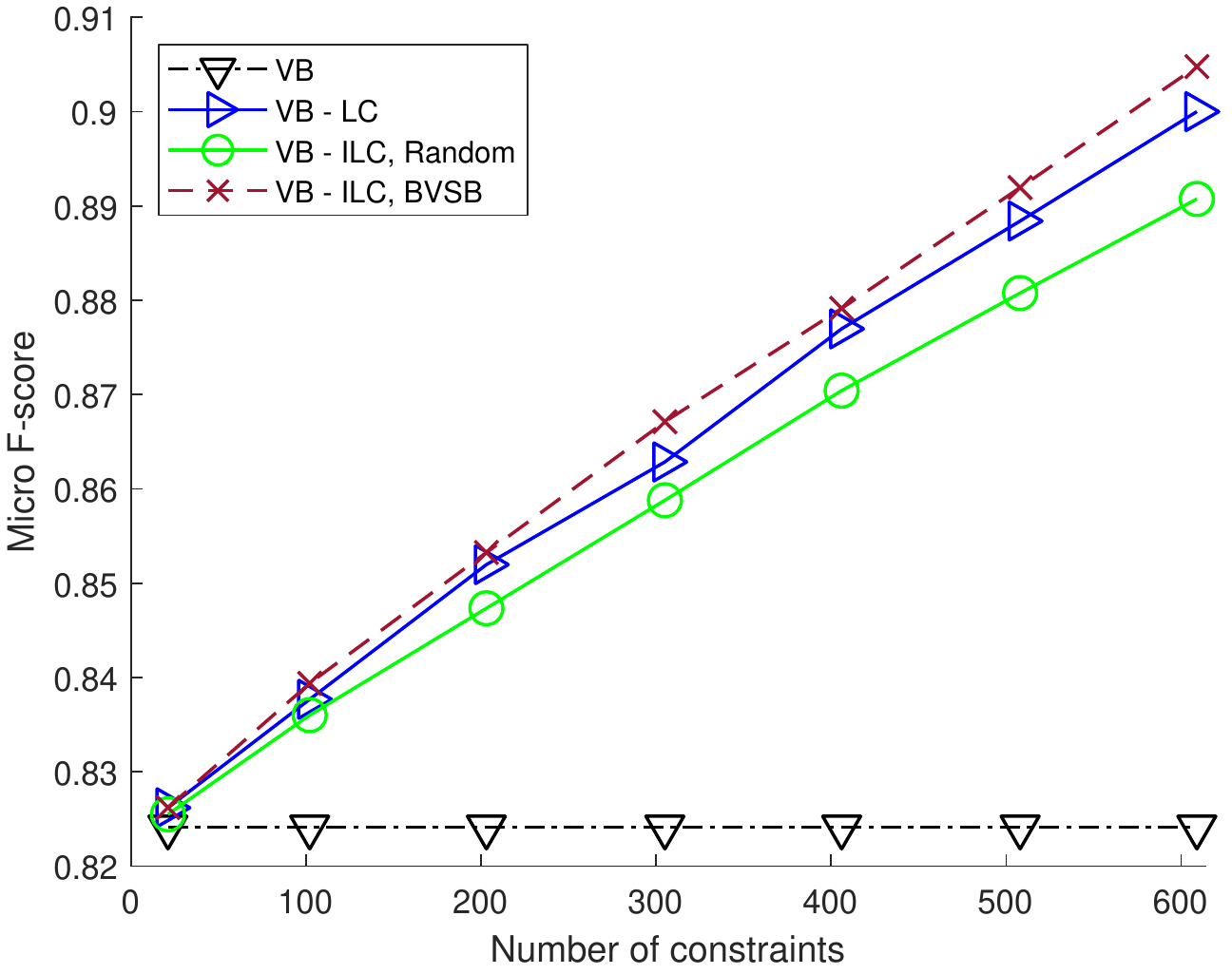}
         \caption{Micro F-score}
         \label{fig:ZenCrowd_us_micro_exp2}
     \end{subfigure}
     \begin{subfigure}[b]{0.9\textwidth}
         \centering
         \includegraphics[width=\textwidth]{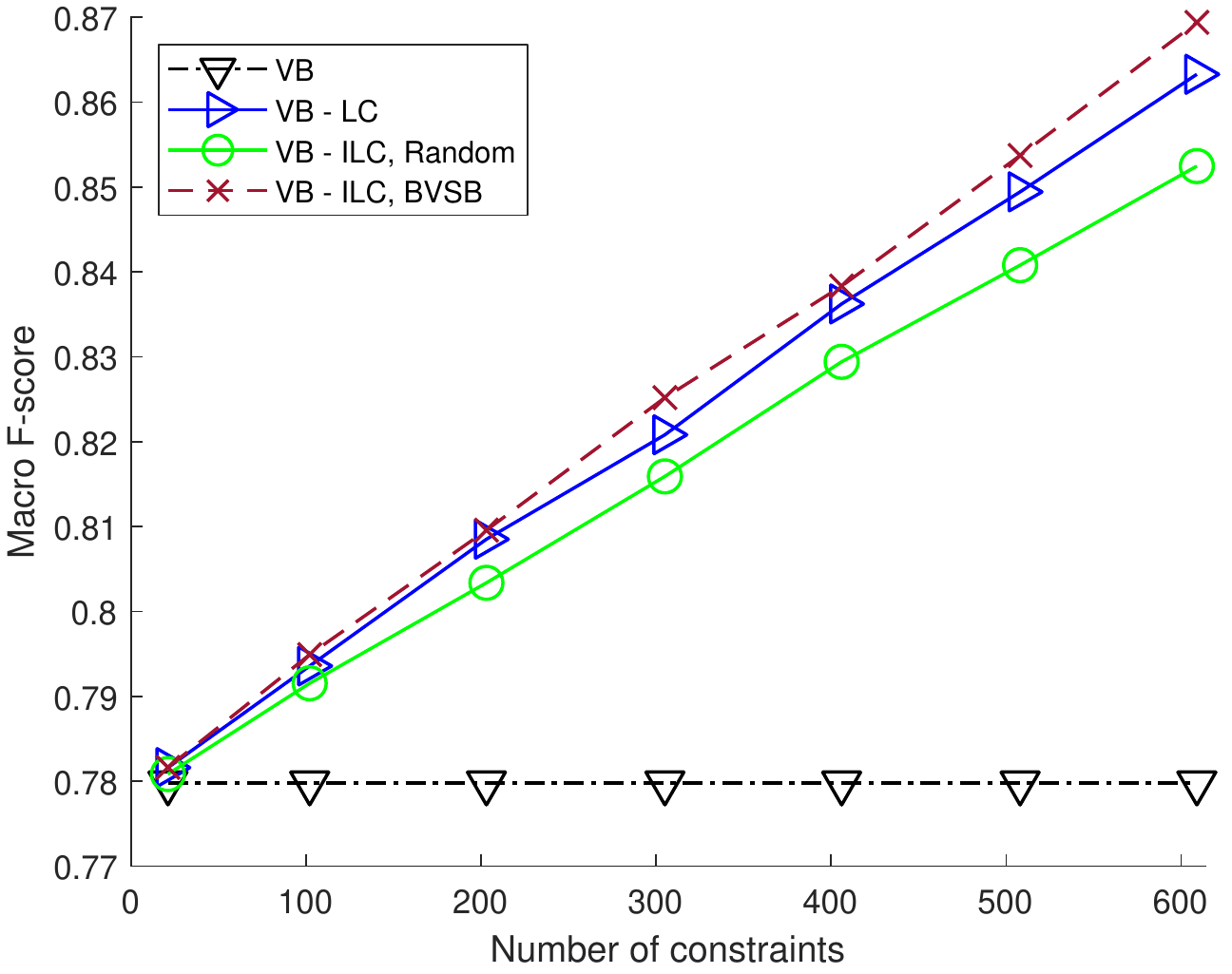}
         \caption{Macro F-score}
         \label{fig:ZenCrowd_us_macro_exp2}
     \end{subfigure}
        \caption{Results for the ZenCrowd US~\cite{ZenCrowd} dataset, with constraints selected via uncertainty sampling.}
        \label{fig:ZenCrowd_us_res_exp2}
    \end{minipage}
    \hfill
    \begin{minipage}{0.32\textwidth}
        \centering
        \begin{subfigure}[b]{0.9\textwidth}
         \centering
         \includegraphics[width=\textwidth]{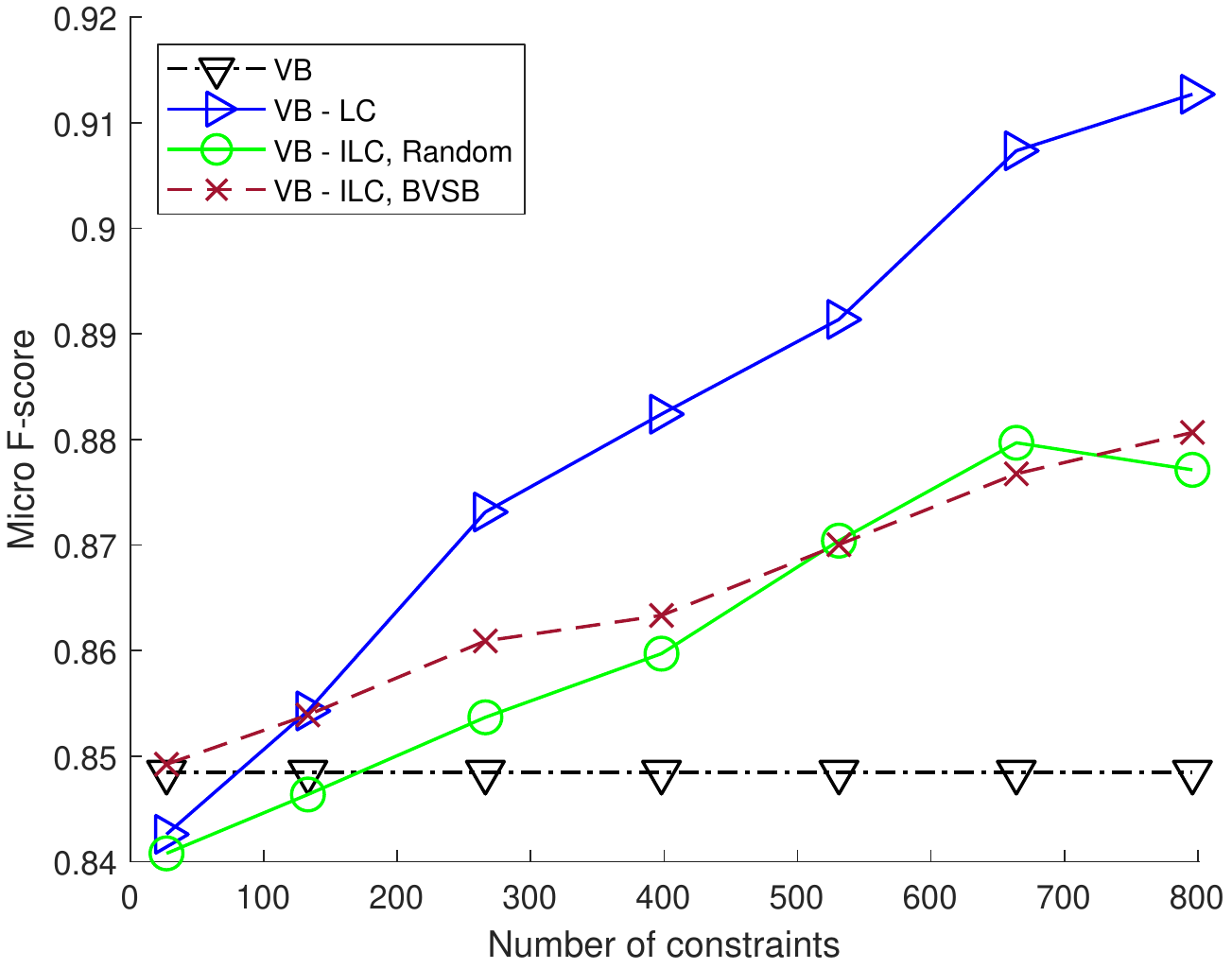}
         \caption{Micro F-score}
         \label{fig:web_micro_exp2}
     \end{subfigure}
     \begin{subfigure}[b]{0.9\textwidth}
         \centering
         \includegraphics[width=\textwidth]{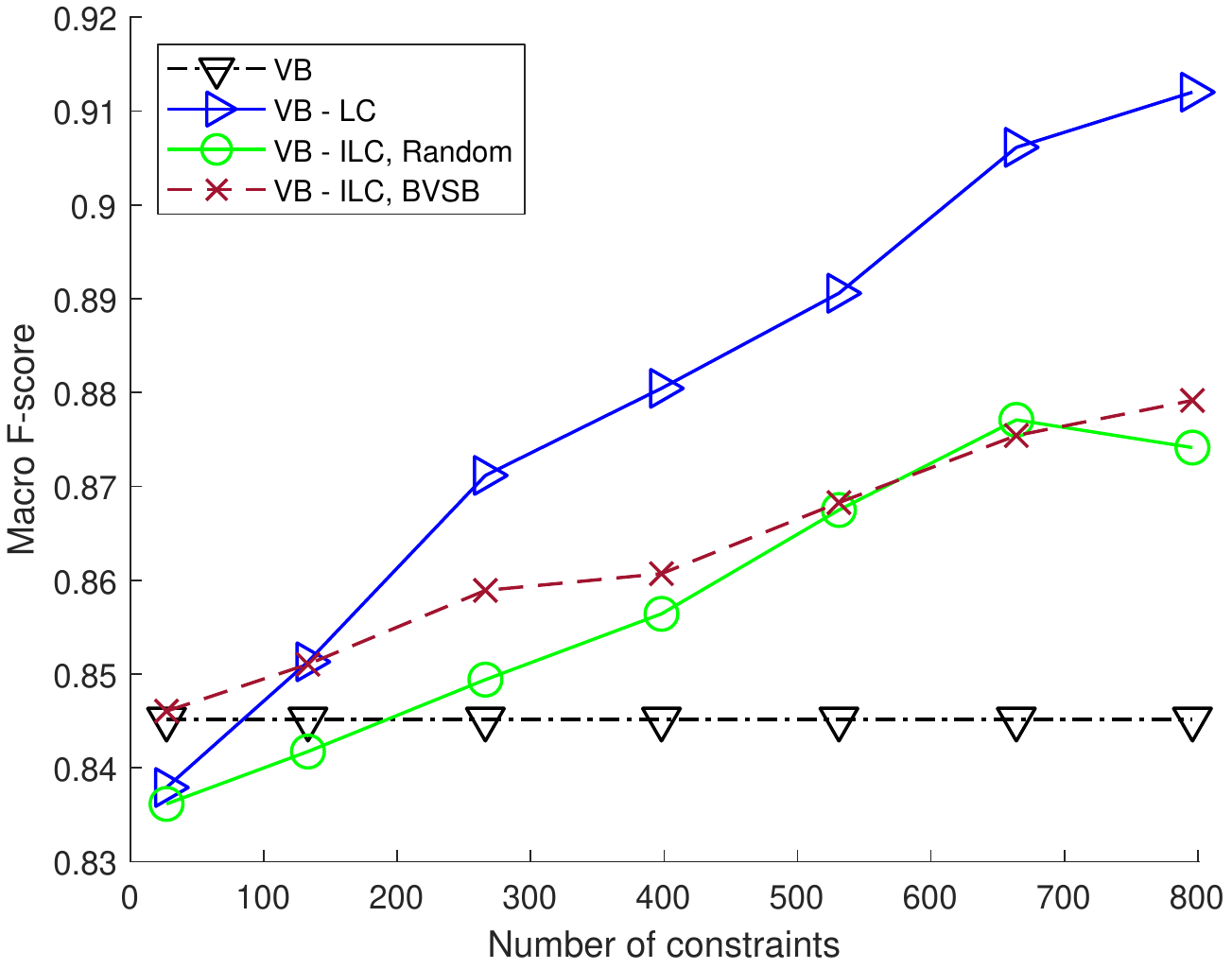}
         \caption{Macro F-score}
         \label{fig:web_macro_exp2}
     \end{subfigure}
        \caption{Results for the Web~\cite{minimax_crowd} dataset, with constraints selected via uncertainty sampling.}
        \label{fig:web_res_exp2}
    \end{minipage}
\end{figure}

\begin{figure}
    \centering
    \begin{minipage}{0.32\textwidth}
        \centering
        \begin{subfigure}[b]{0.9\textwidth}
         \centering
         \includegraphics[width=\textwidth]{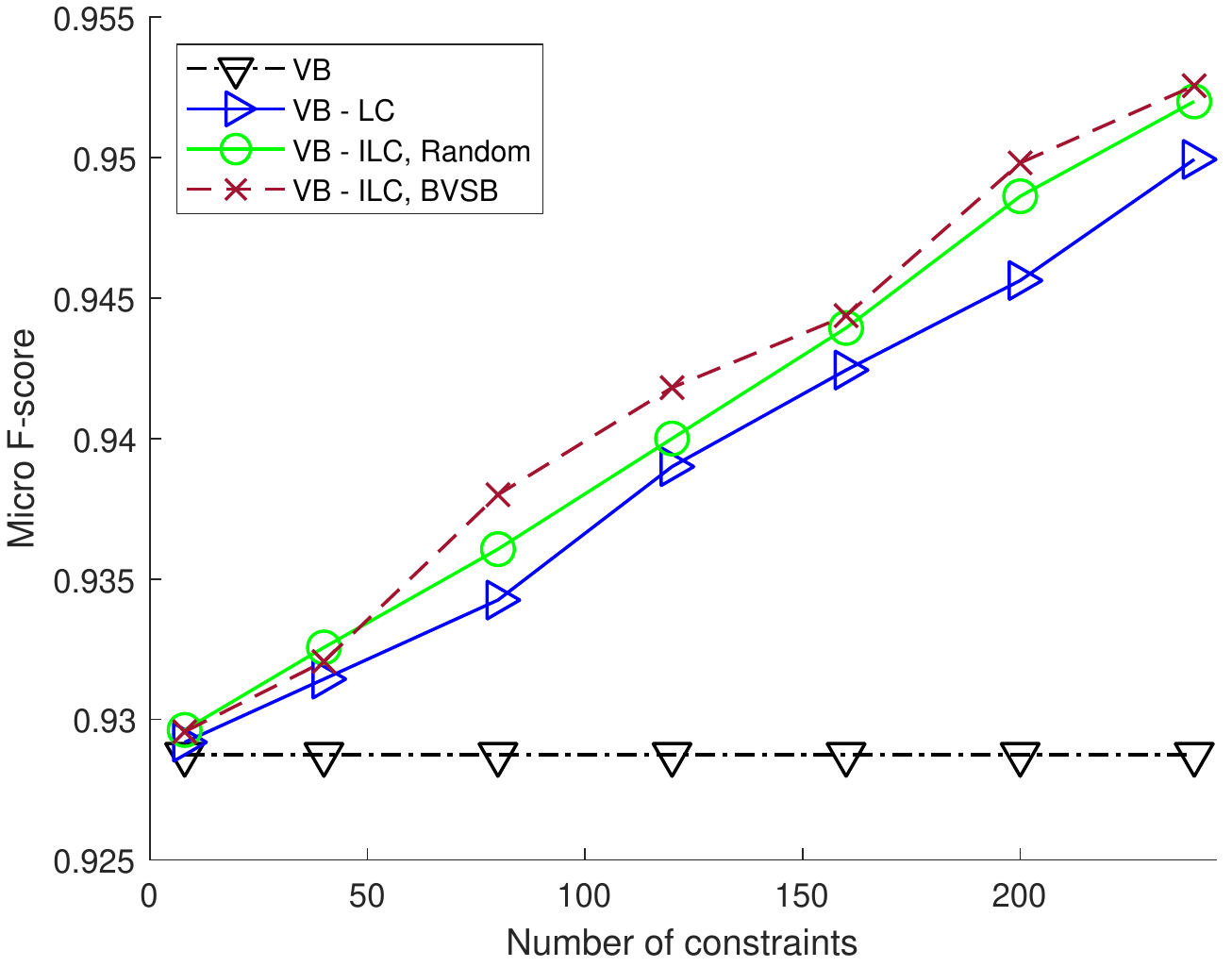}
         \caption{Micro F-score}
         \label{fig:rte_micro_exp2}
     \end{subfigure}
     \begin{subfigure}[b]{0.9\textwidth}
         \centering
         \includegraphics[width=\textwidth]{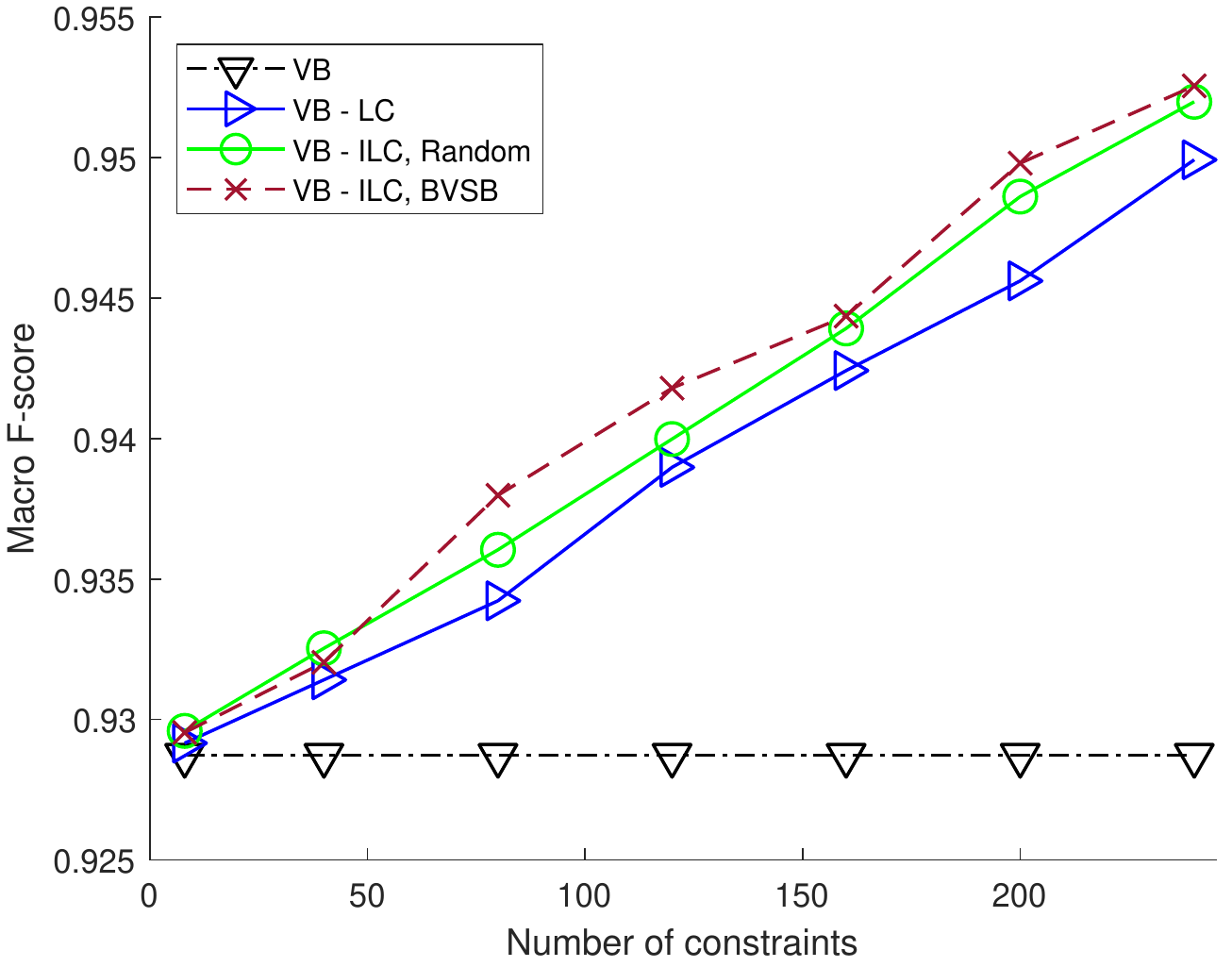}
         \caption{Macro F-score}
         \label{fig:rte_macro_exp2}
     \end{subfigure}
        \caption{Results for the RTE~\cite{cheapnfast} dataset, with constraints selected via uncertainty sampling.}
        \label{fig:rte_res_exp2}
    \end{minipage}\hfill
    \begin{minipage}{0.32\textwidth}
        \centering
        \begin{subfigure}[b]{0.9\textwidth}
         \centering
         \includegraphics[width=\textwidth]{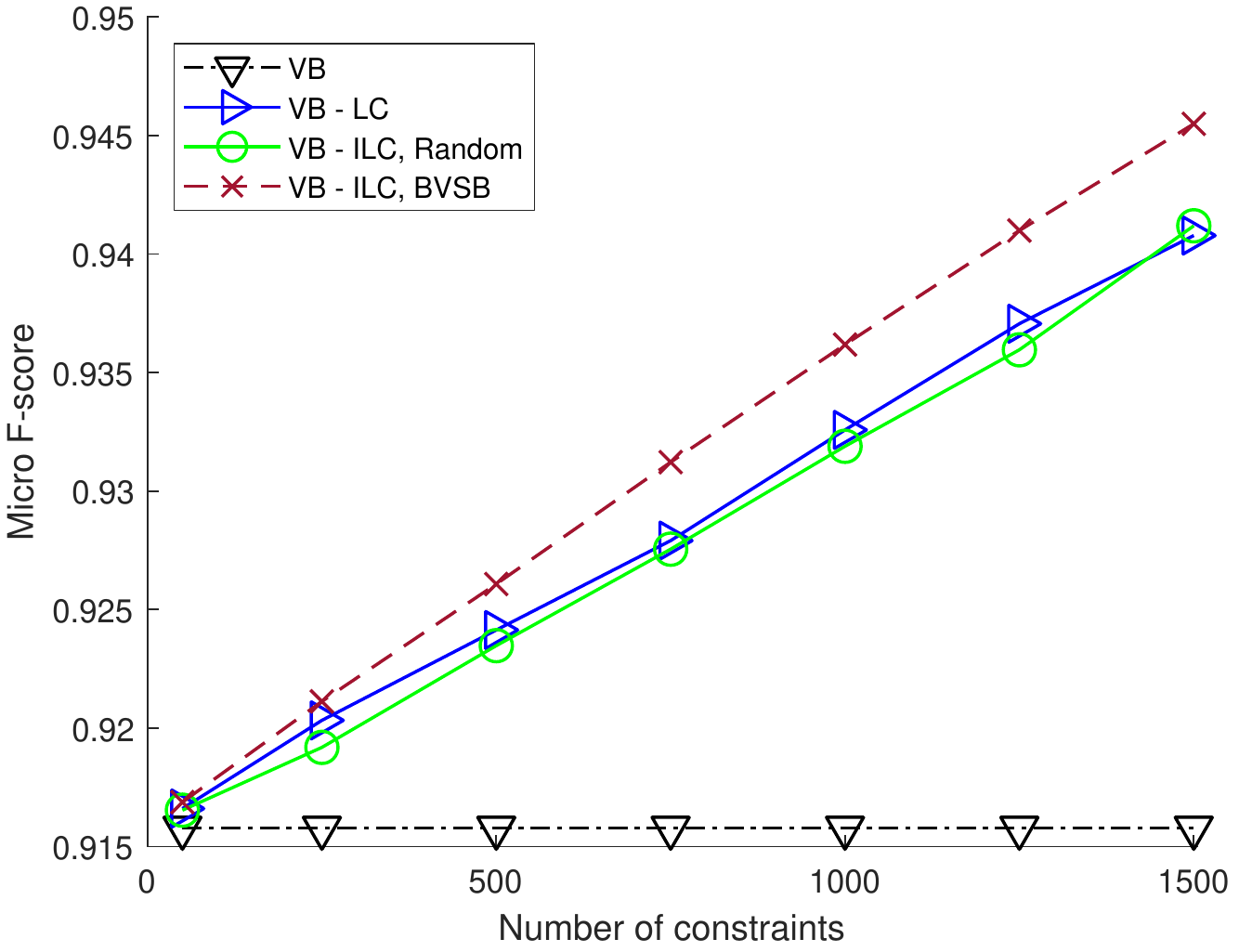}
         \caption{Micro F-score}
         \label{fig:sentencepolarity_micro_exp2}
     \end{subfigure}
     \begin{subfigure}[b]{0.9\textwidth}
         \centering
         \includegraphics[width=\textwidth]{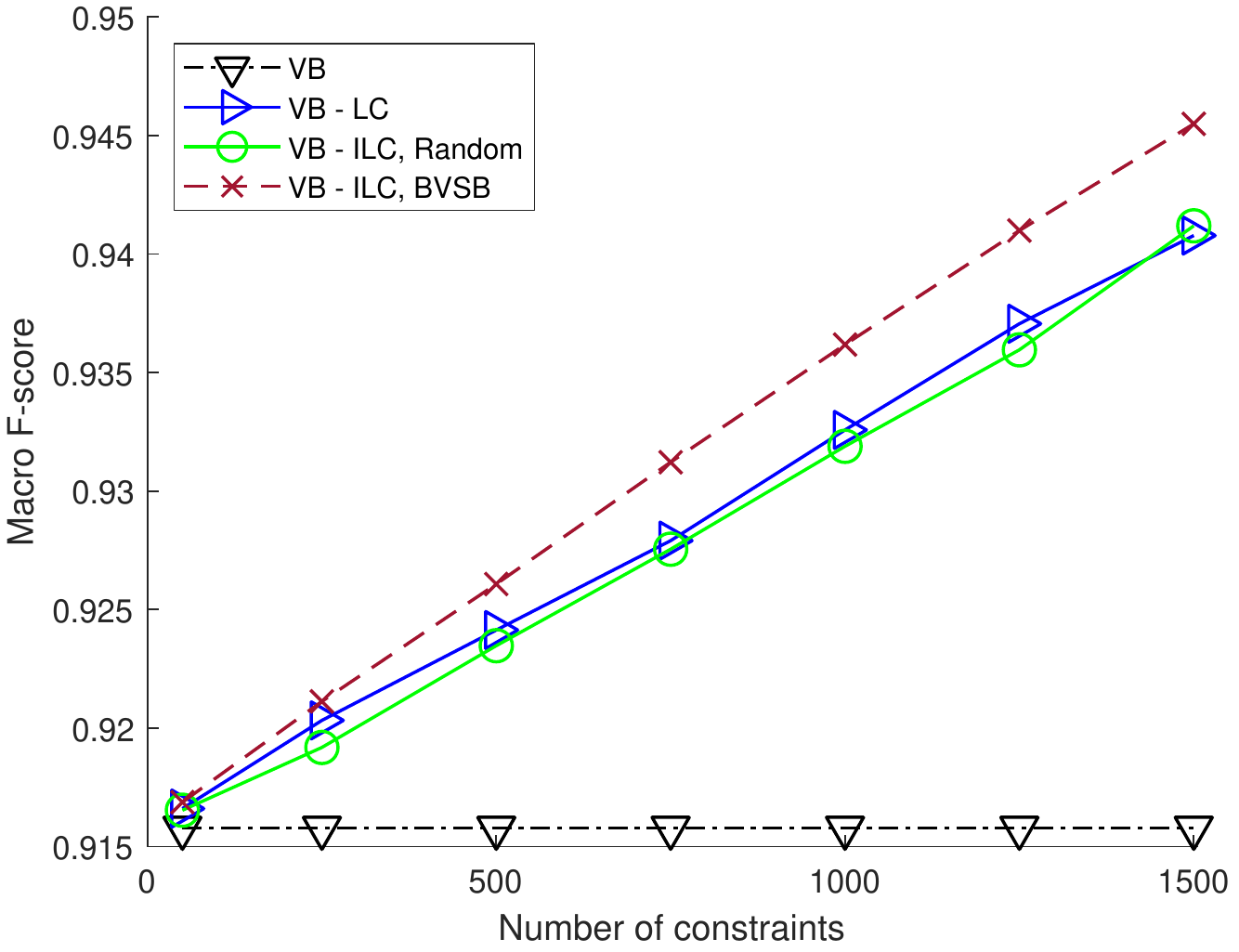}
         \caption{Macro F-score}
         \label{fig:sentencepolarity_macro_exp2}
     \end{subfigure}
        \caption{Results for the Sentence Polarity~\cite{musicgenre_senpoldata} dataset, with constraints selected via uncertainty sampling.}
        \label{fig:sentencepolarity_res_exp2}
    \end{minipage}
    \hfill
    \begin{minipage}{0.32\textwidth}
        \centering
        \begin{subfigure}[b]{0.9\textwidth}
         \centering
         \includegraphics[width=\textwidth]{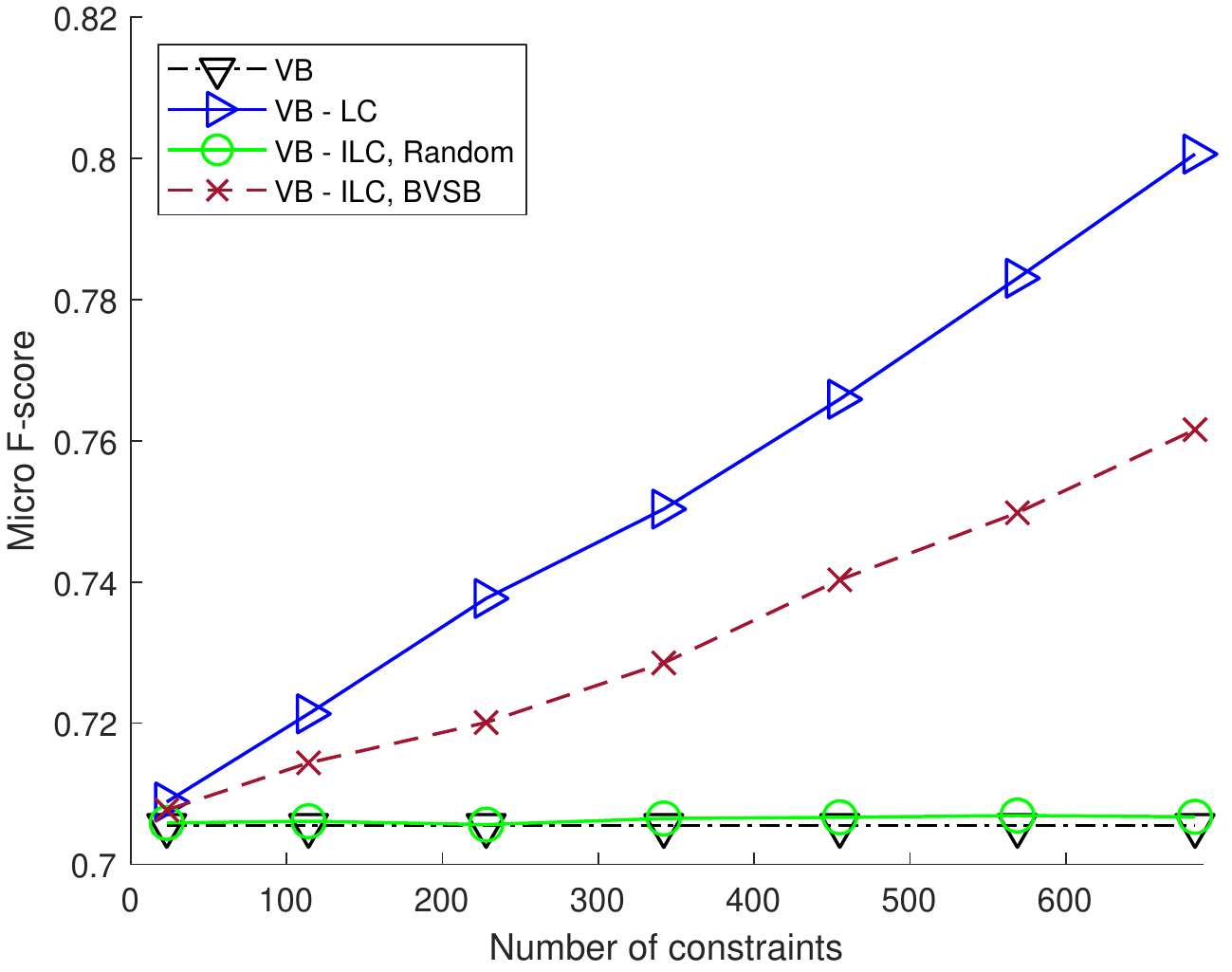}
         \caption{Micro F-score}
         \label{fig:trec_micro_exp2}
     \end{subfigure}
     \begin{subfigure}[b]{0.9\textwidth}
         \centering
         \includegraphics[width=\textwidth]{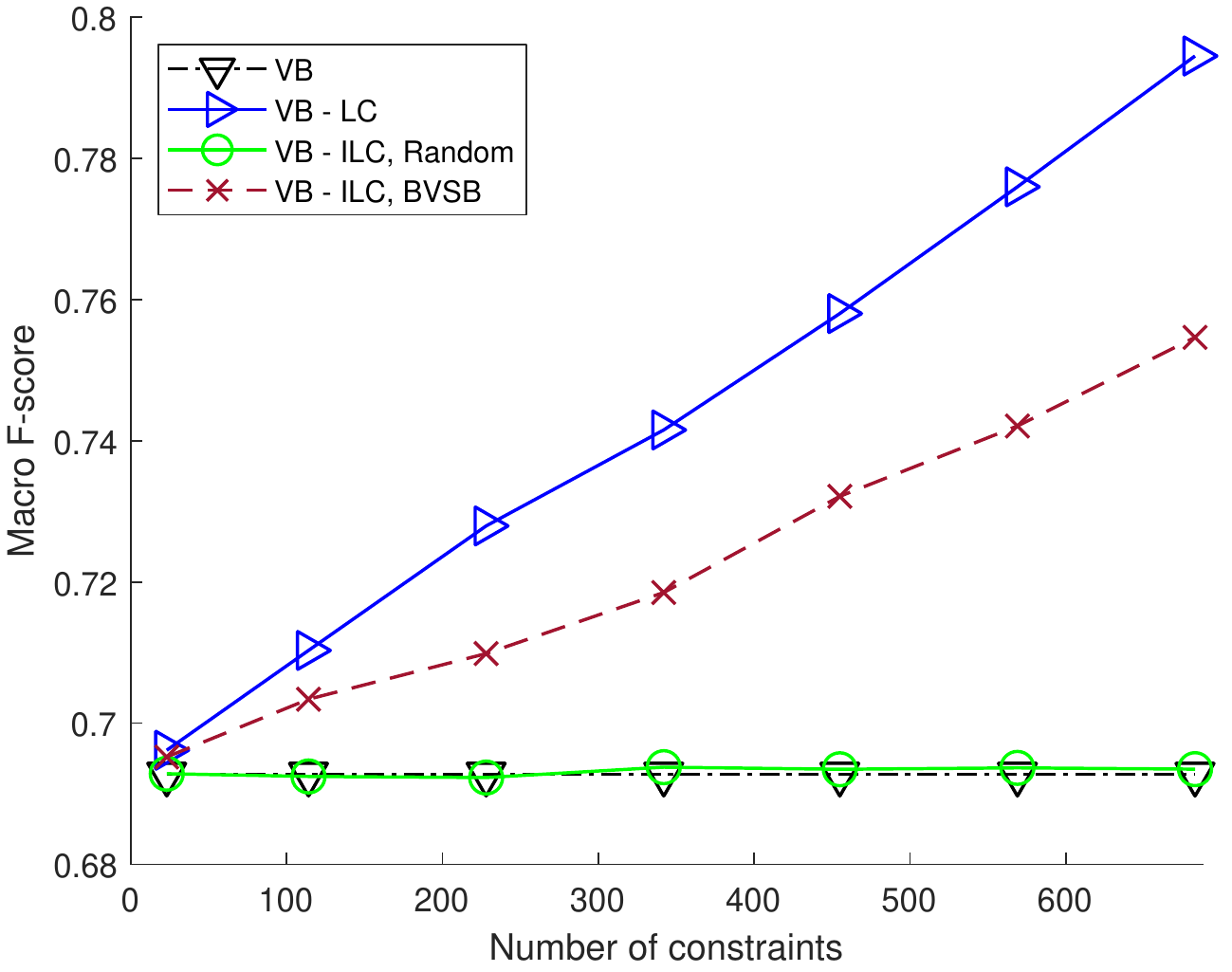}
         \caption{Macro F-score}
         \label{fig:trec_macro_exp2}
     \end{subfigure}
        \caption{Results for the TREC~\cite{Lease11-trec} dataset, with constraints selected via uncertainty sampling.}
        \label{fig:trec_res_exp2}
    \end{minipage}
\end{figure}

\begin{figure}[tb]
     \centering
     \begin{subfigure}[b]{0.3\textwidth}
         \centering
         \includegraphics[width=\textwidth]{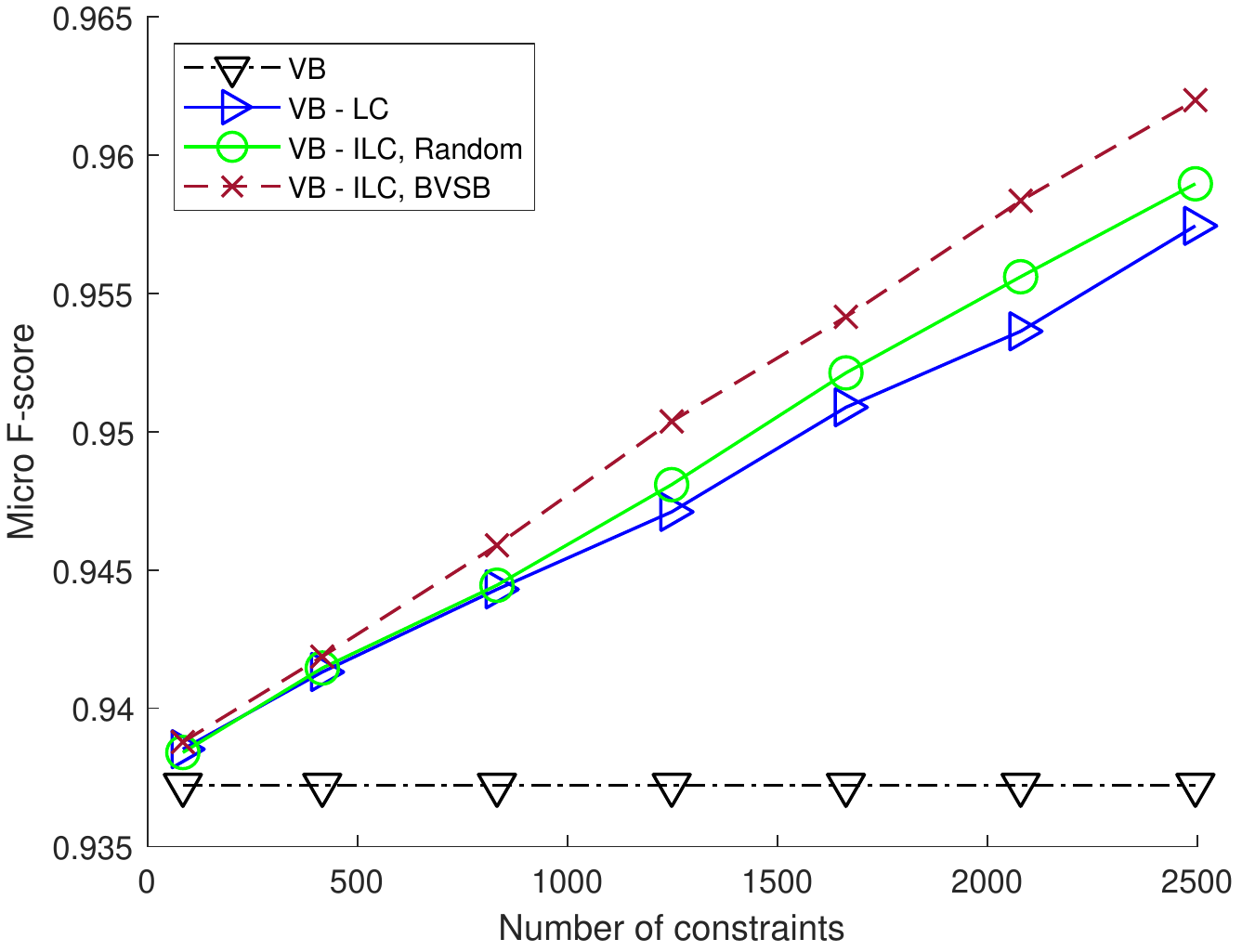}
         \caption{Micro F-score}
         \label{fig:prod_micro_exp2}
     \end{subfigure}\\
     \begin{subfigure}[b]{0.3\textwidth}
         \centering
         \includegraphics[width=\textwidth]{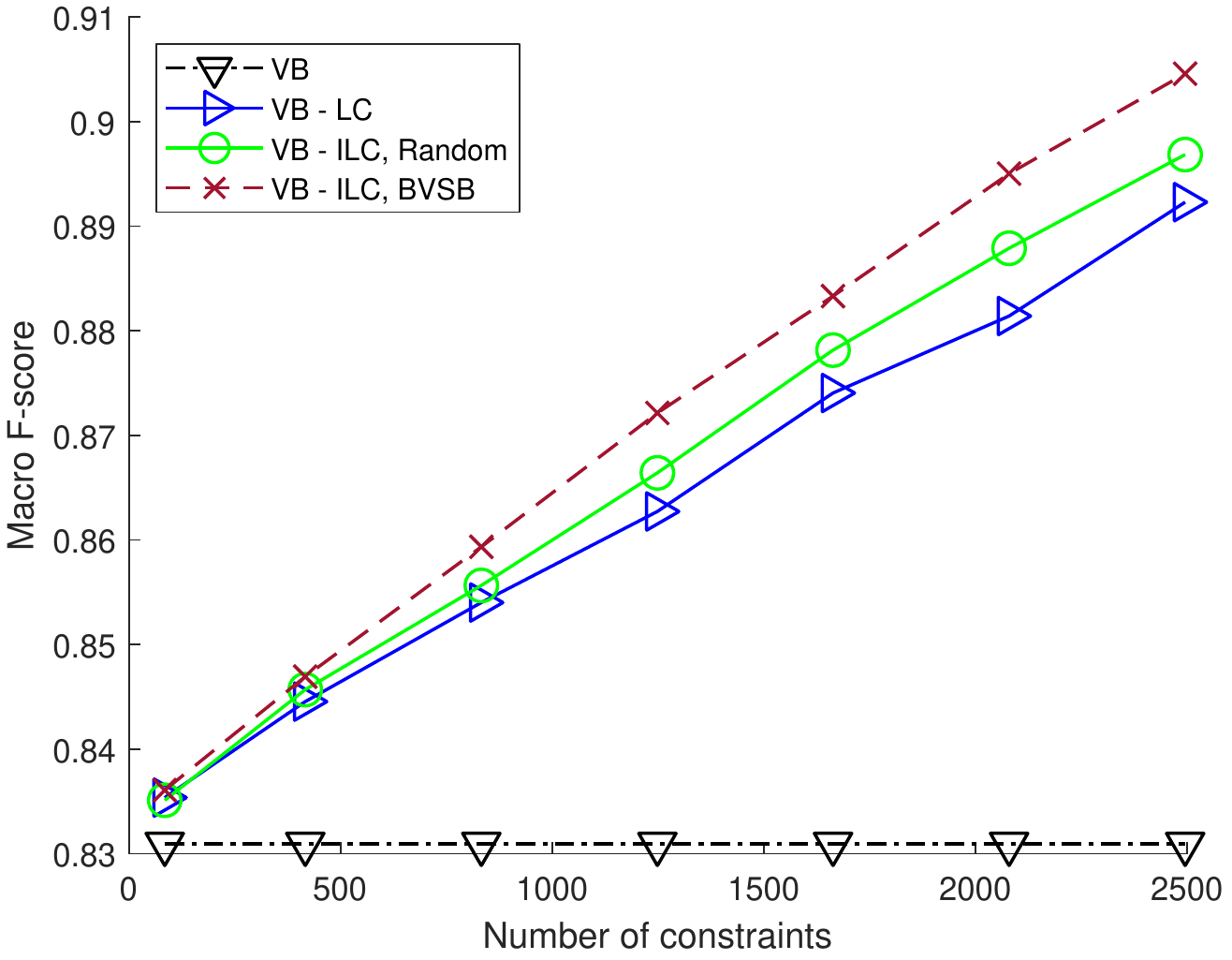}
         \caption{Macro F-score}
         \label{fig:prod_macro_exp2}
     \end{subfigure}
        \caption{Results for the Product~\cite{crowder} dataset, with constraints selected via uncertainty sampling.}
        \label{fig:prod_res_exp2}
\end{figure}

\begin{figure}[tb]
\centering
\begin{subfigure}[b]{0.24\textwidth}
\centering
\includegraphics[width=\textwidth]{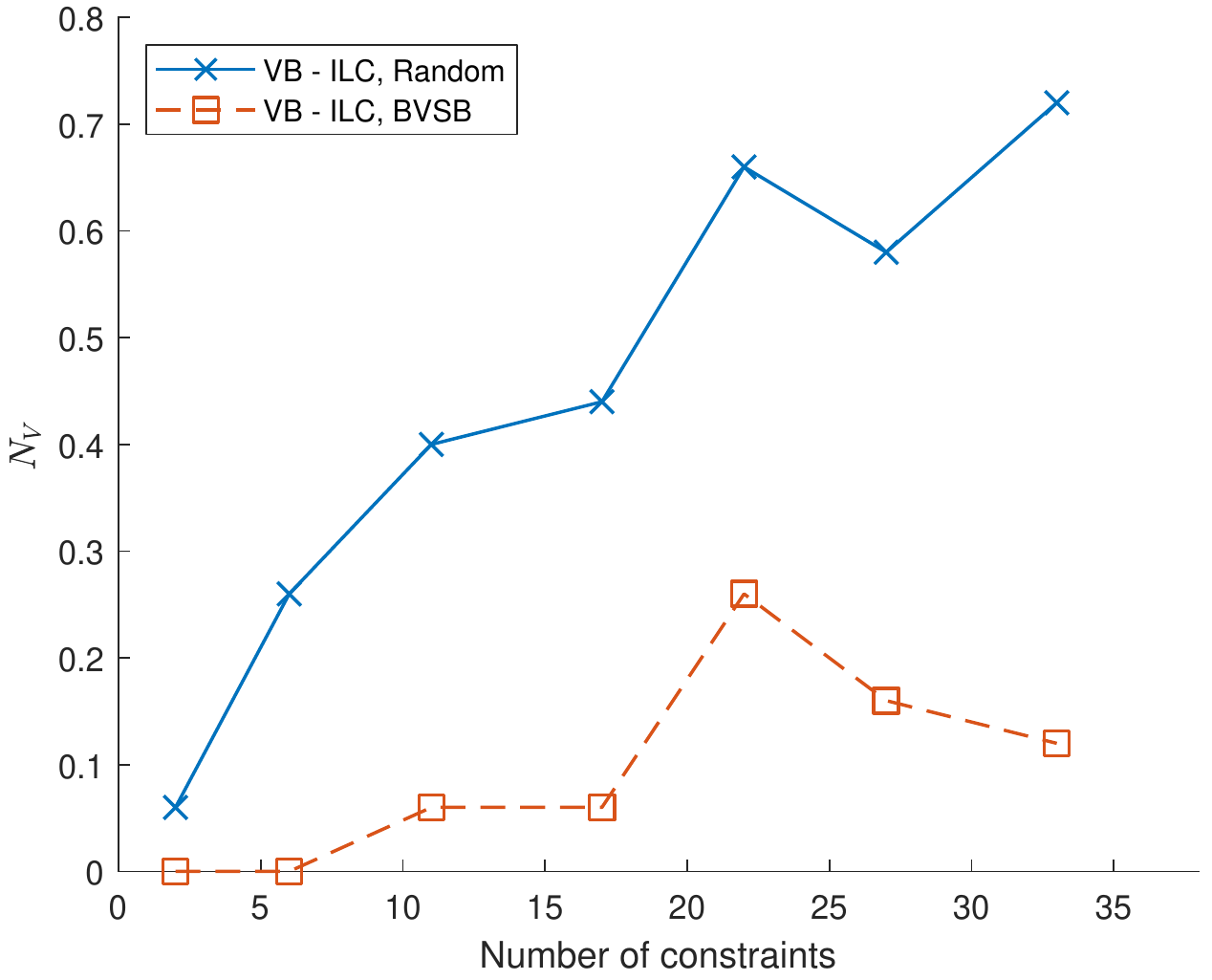}
\caption{Bluebird}
\label{fig:exp2:num_violated:bluebird}
\end{subfigure}
\begin{subfigure}[b]{0.24\textwidth}
\centering
\includegraphics[width=\textwidth]{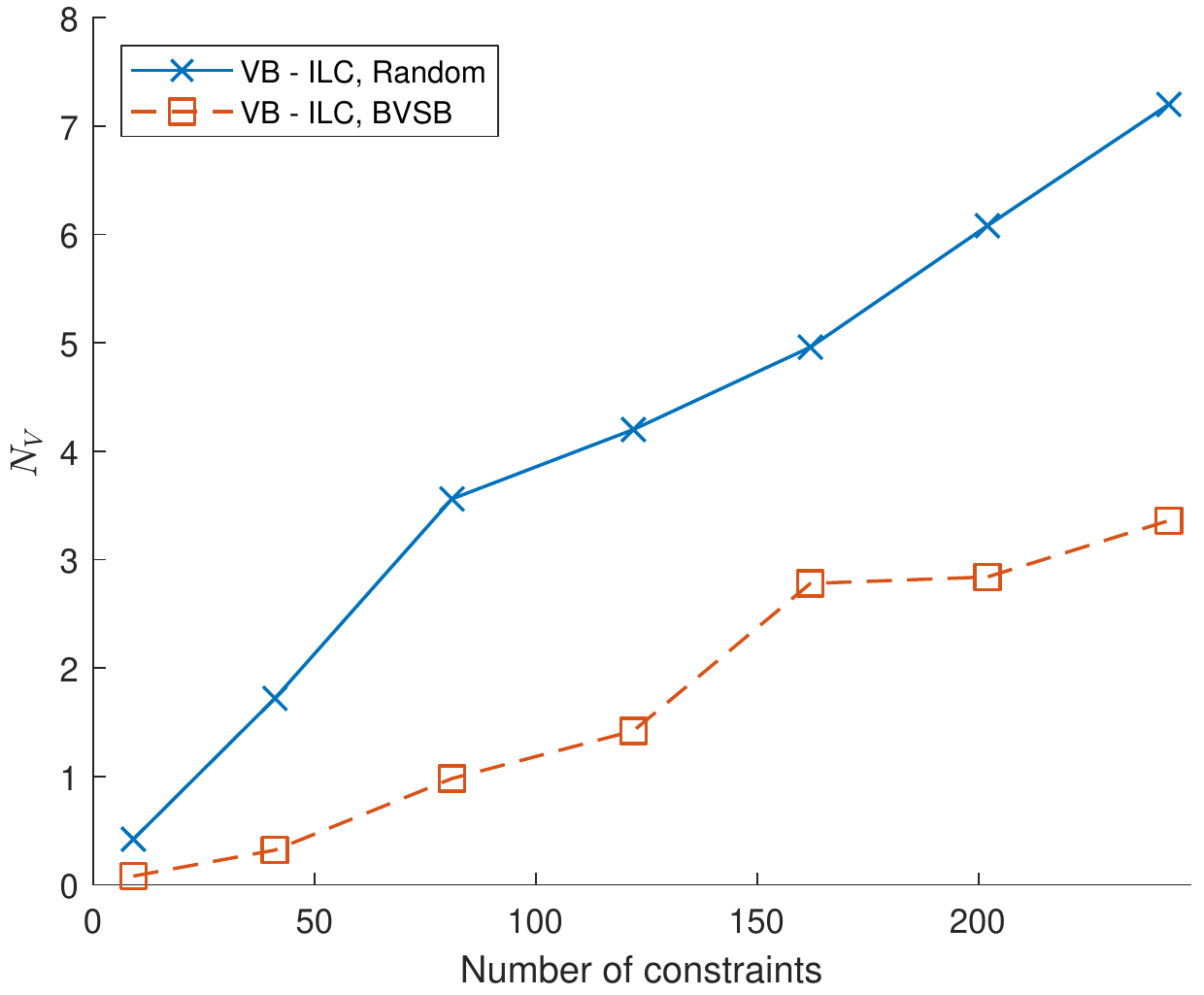}
\caption{Dog}
\label{fig:exp2:num_violated:dog}
\end{subfigure}
\begin{subfigure}[b]{0.24\textwidth}
\centering
\includegraphics[width=\textwidth]{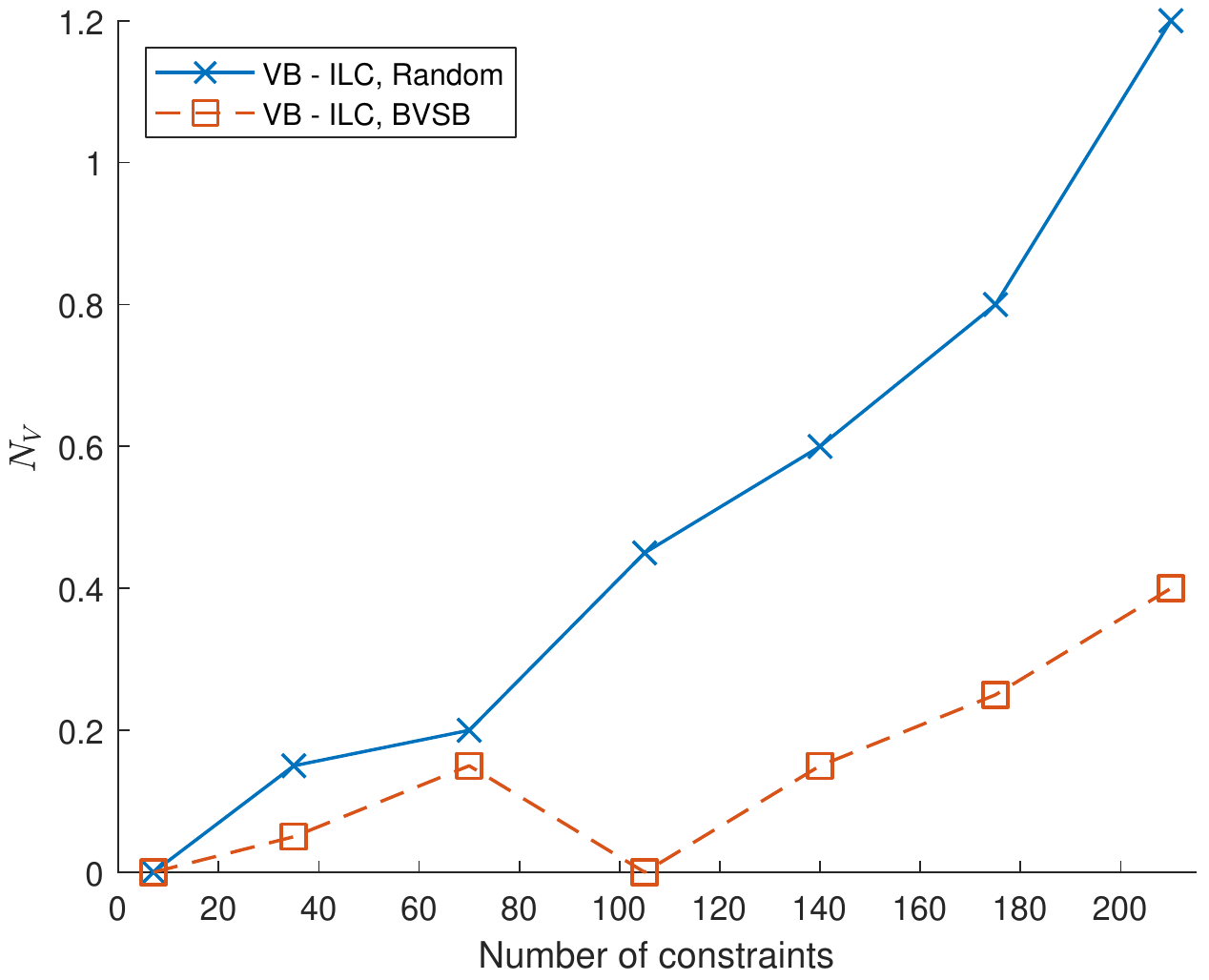}
\caption{Music Genre}
\label{fig:exp2:num_violated:music}
\end{subfigure}
\begin{subfigure}[b]{0.24\textwidth}
\centering
\includegraphics[width=\textwidth]{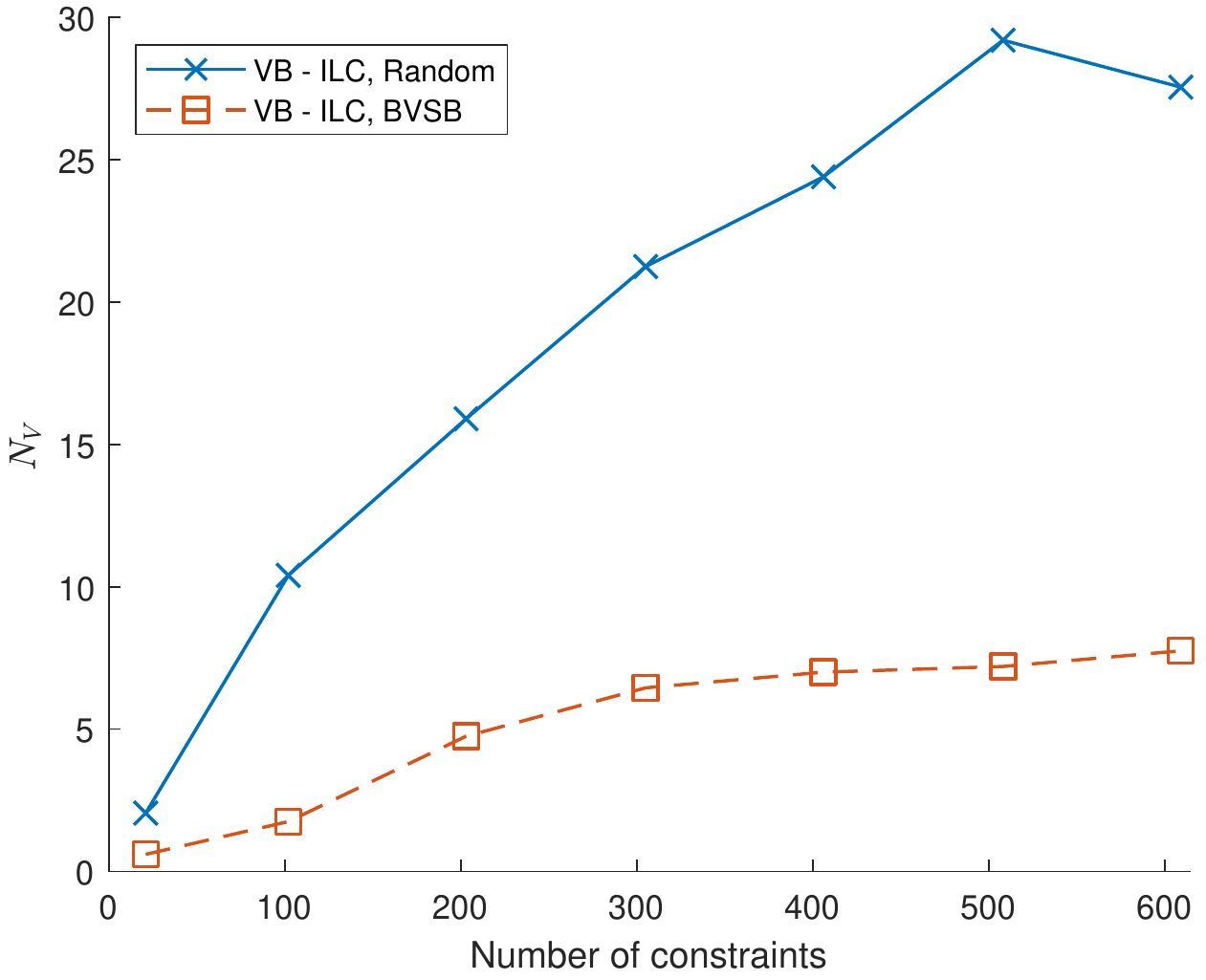}
\caption{ZenCrowd India}
\label{fig:exp2:num_violated:ZenCrowd_in}
\end{subfigure}
\\
\begin{subfigure}[b]{0.24\textwidth}
\centering
\includegraphics[width=\textwidth]{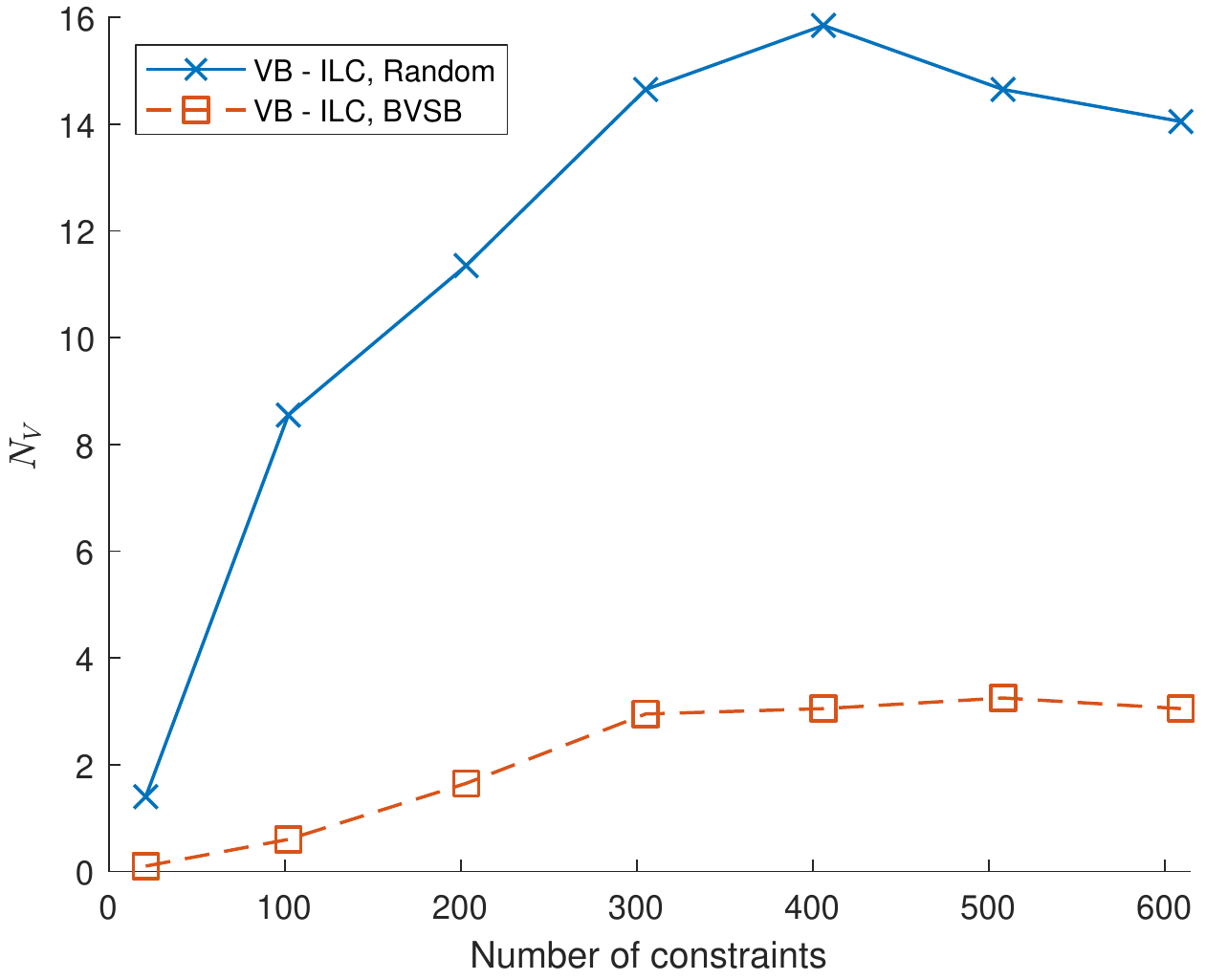}
\caption{ZenCrowd US}
\label{fig:exp2:num_violated:ZenCrowd_US}
\end{subfigure}
\begin{subfigure}[b]{0.24\textwidth}
\centering
\includegraphics[width=\textwidth]{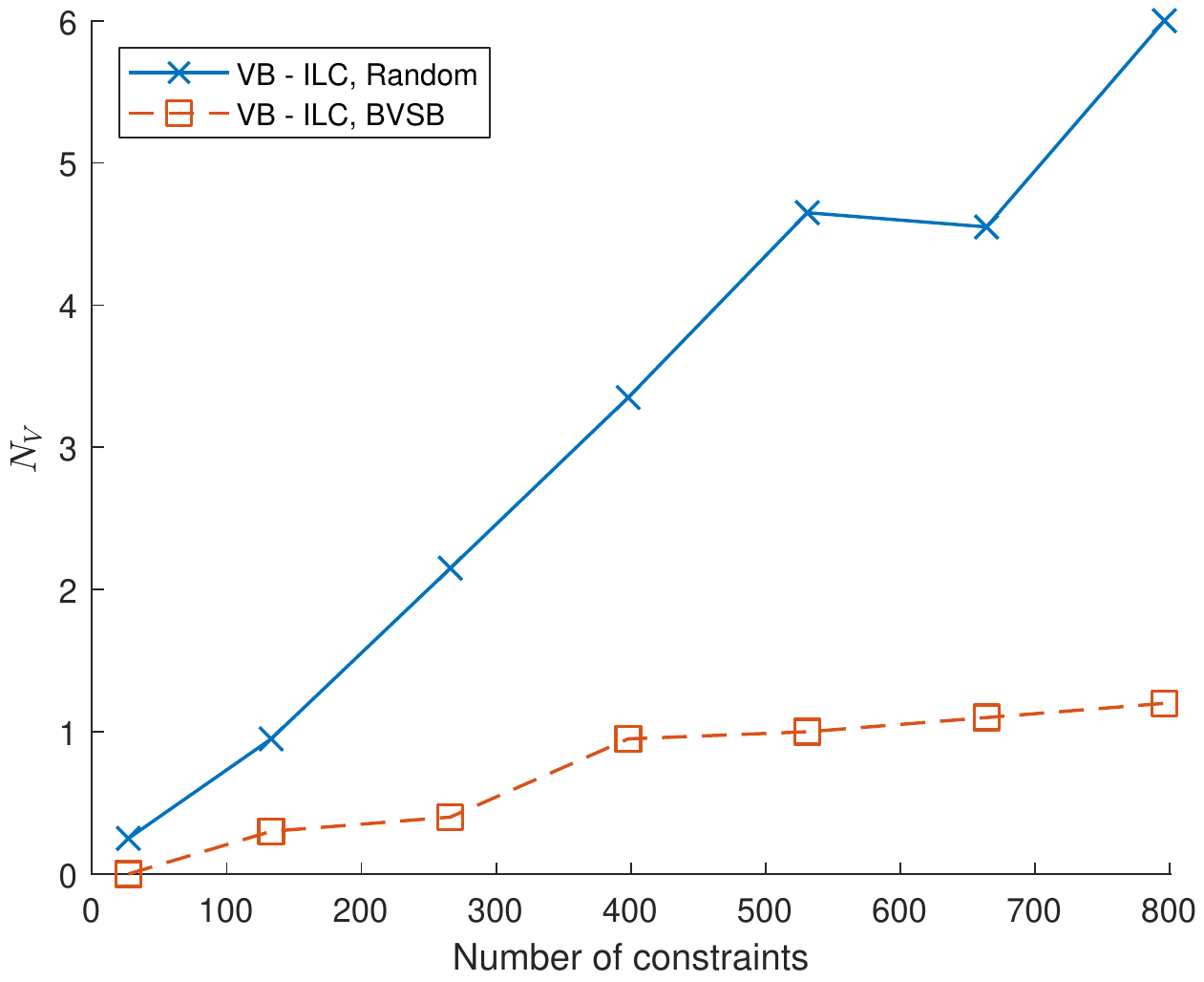}
\caption{Web}
\label{fig:exp2:num_violated:web}
\end{subfigure}
\begin{subfigure}[b]{0.24\textwidth}
\centering
\includegraphics[width=\textwidth]{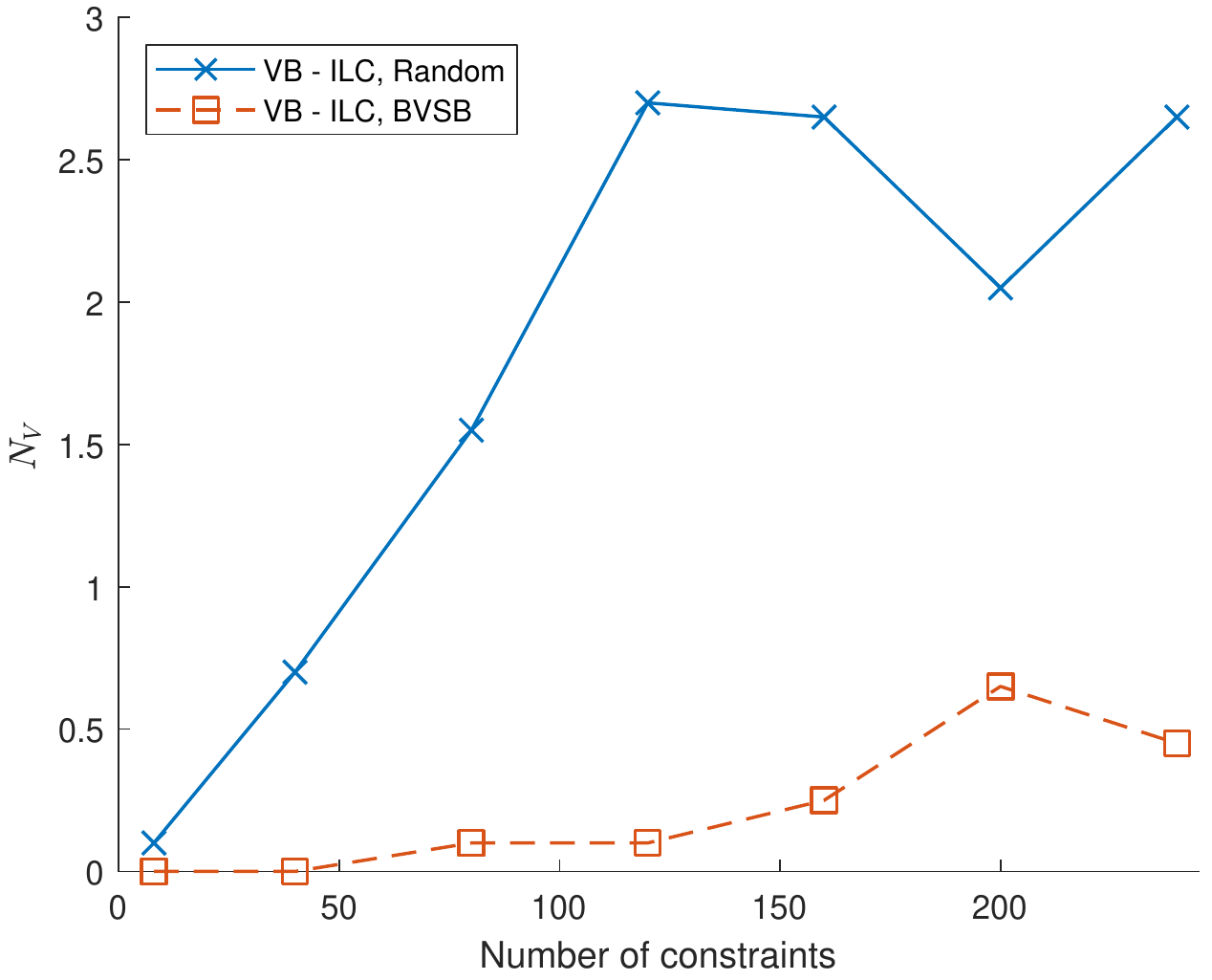}
\caption{RTE}
\label{fig:exp2:num_violated:rte}
\end{subfigure}
\begin{subfigure}[b]{0.24\textwidth}
\centering
\includegraphics[width=\textwidth]{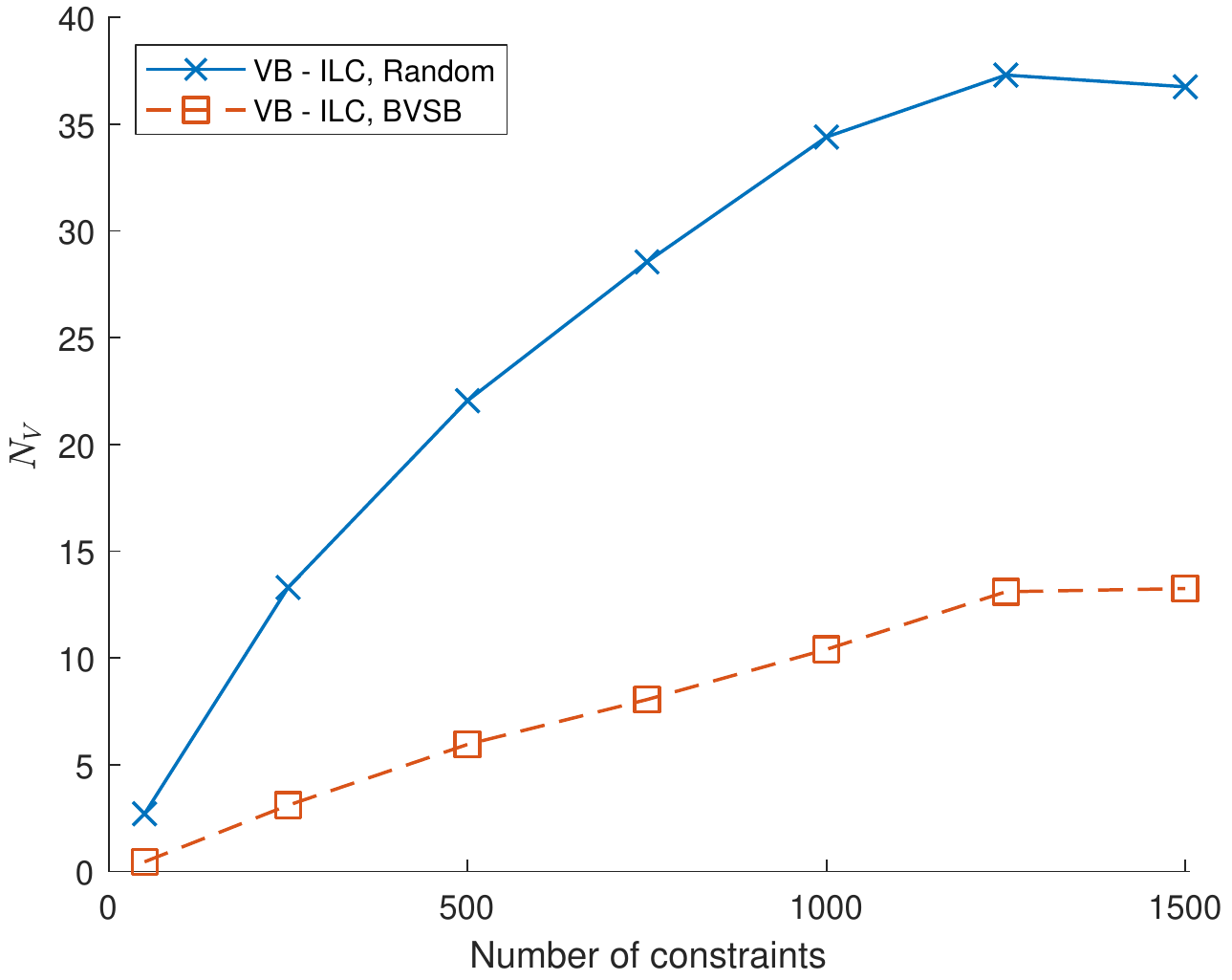}
\caption{Sentence Polarity}
\label{fig:exp2:num_violated:sen_polarity}
\end{subfigure}
\\
\begin{subfigure}[b]{0.24\textwidth}
\centering
\includegraphics[width=\textwidth]{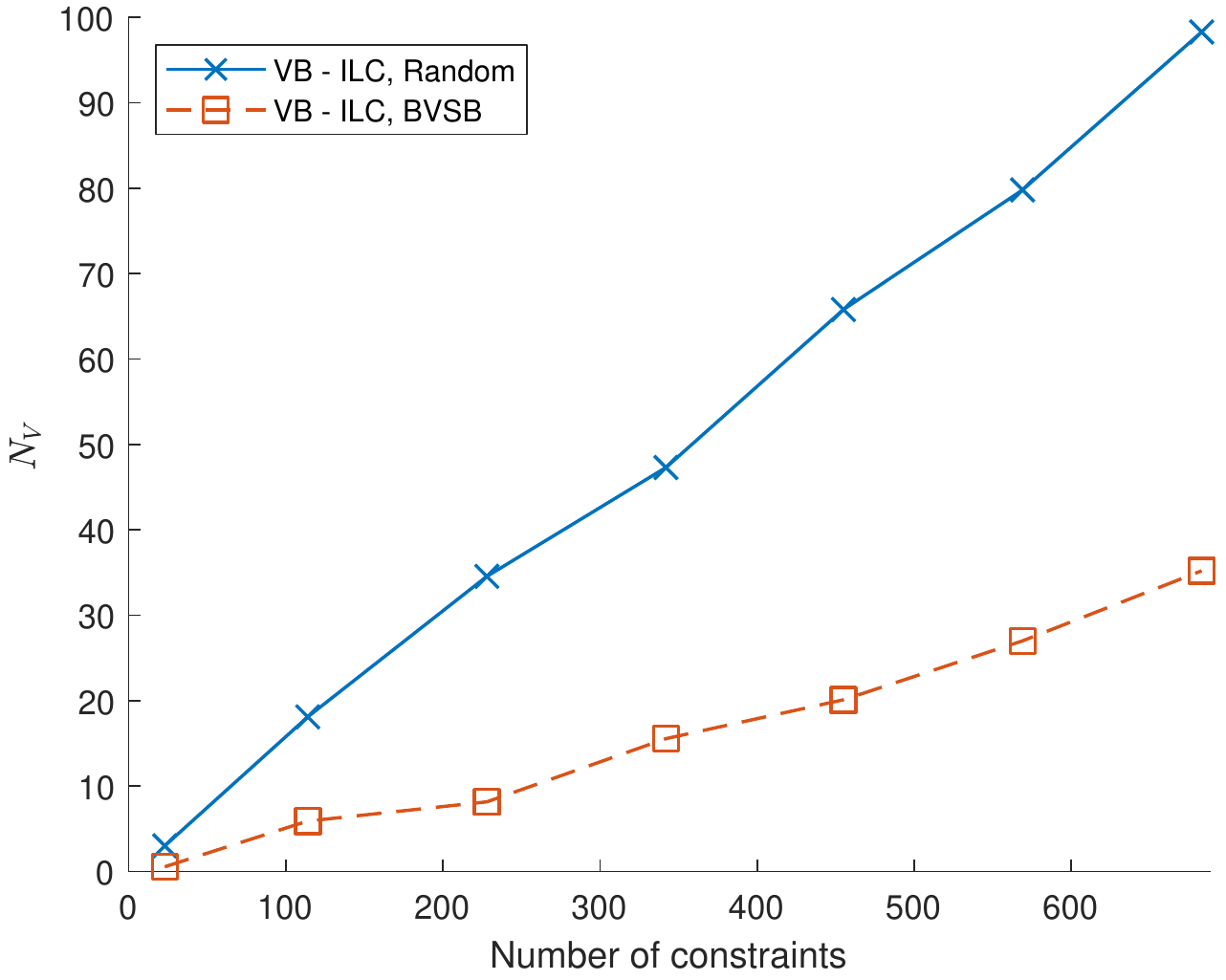}
\caption{TREC}
\label{fig:exp2:num_violated:trec}
\end{subfigure}
\begin{subfigure}[b]{0.24\textwidth}
\centering
\includegraphics[width=\textwidth]{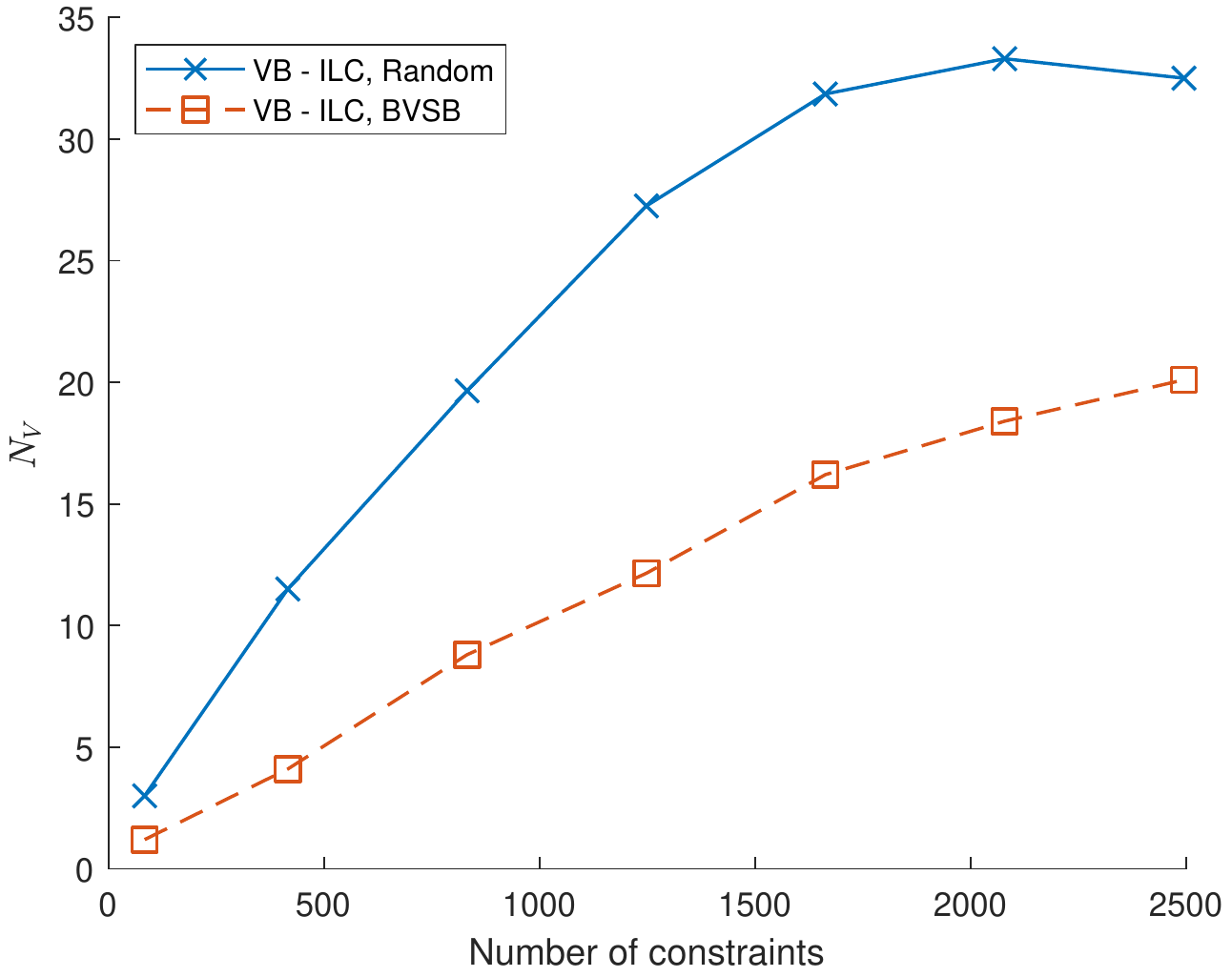}
\caption{Product}
\label{fig:exp2:num_violated:prod}
\end{subfigure}
\caption{Number of violated constraints for each dataset, with constraints selected via uncertainty sampling.}
\label{fig:exp2:num_violated}
\end{figure}

For this set of numerical tests, random selection of constraints is compared to the uncertainty sampling methods introduced in Sec. 4.2. of the original text for \emph{VB - ILC}. Figs. \ref{fig:bluebird_res_exp2}-\ref{fig:prod_res_exp2} show the results w.r.t. Micro and Macro-averaged F-scored for the datasets considered, as the number of constraints increases.  In all datasets, uncertainty sampling  provides a  modest performance improvement for\emph{VB - ILC}, with the exception of the Dog and TREC datasets where the performance improvement was more significant. Fig.~\ref{fig:exp2:num_violated} shows the number of violated constraints $N_V$ for \emph{VB - ILC, Random} and \emph{VB - ILC, BVSB} for each dataset. In all cases, \emph{VB - ILC, BVSB} enjoys a significantly lower number of violated constraints, further explaining the improved performance that is exhibited in most datasets.

\subsection{Label derived constraints} 
\begin{figure}
    \centering
    \begin{minipage}{0.32\textwidth}
        \centering
        \begin{subfigure}[b]{0.9\textwidth}
         \centering
         \includegraphics[width=\textwidth]{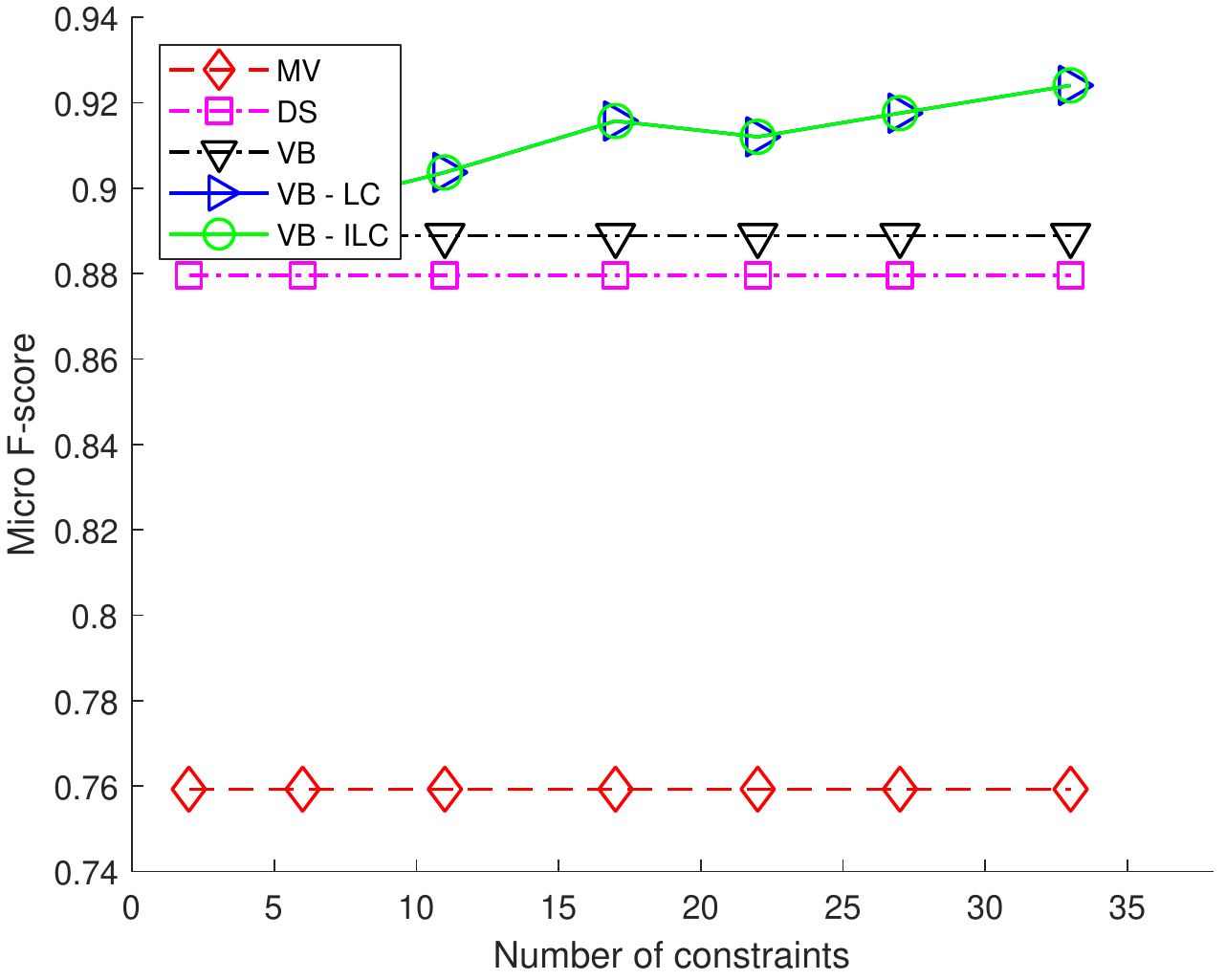}
         \caption{Micro F-score}
         \label{fig:bluebird_micro_exp3}
     \end{subfigure}\\
     \begin{subfigure}[b]{0.9\textwidth}
         \centering
         \includegraphics[width=\textwidth]{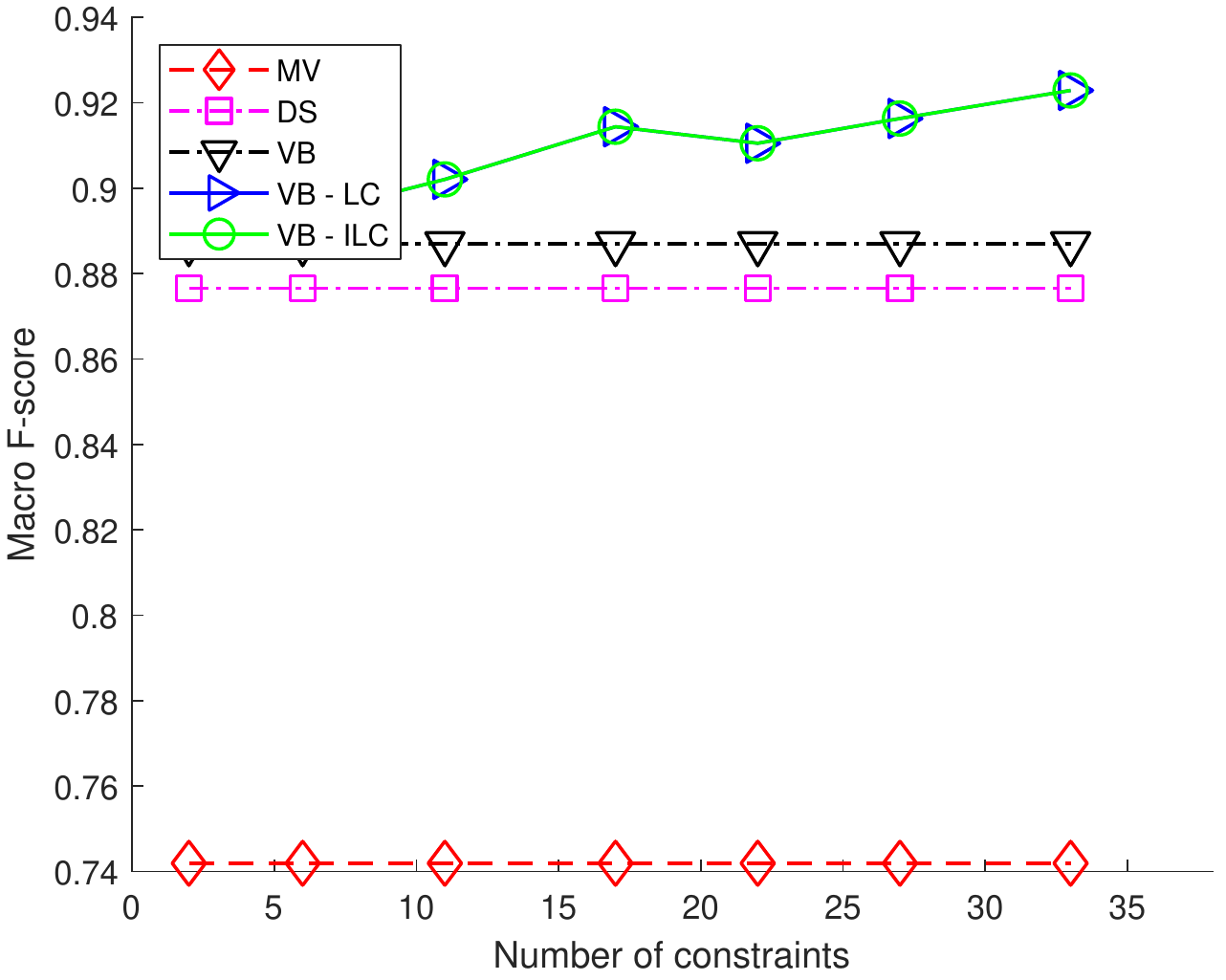}
         \caption{Macro F-score}
         \label{fig:bluebird_macro_exp3}
     \end{subfigure}
        \caption{Results for the Bluebird~\cite{multidim_wisdom} dataset.}
        \label{fig:bluebird_res_exp3}
    \end{minipage}\hfill
    \begin{minipage}{0.32\textwidth}
        \centering
     \begin{subfigure}[b]{0.9\textwidth}
         \centering
         \includegraphics[width=\textwidth]{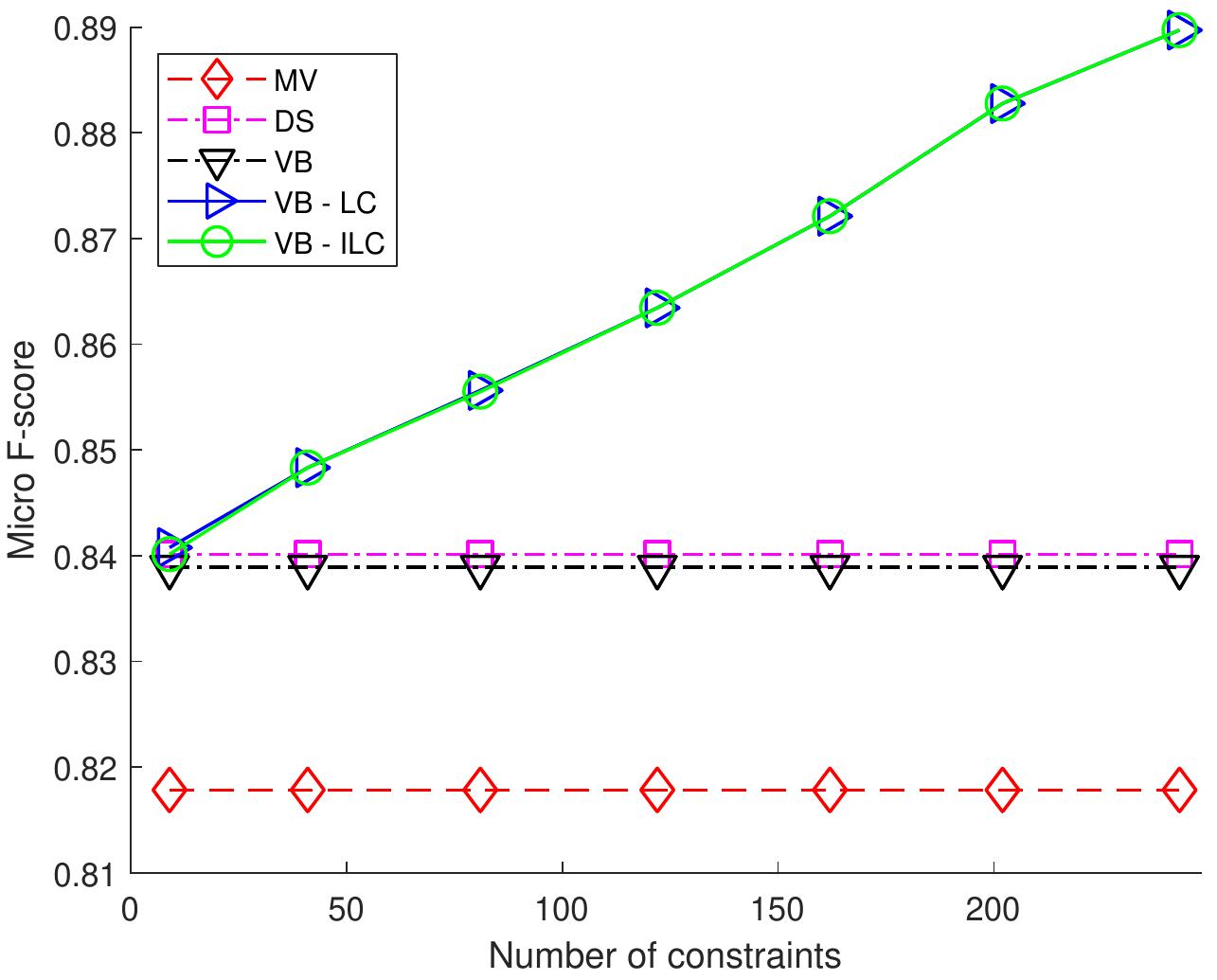}
         \caption{Micro F-score}
         \label{fig:dog_micro_exp3}
     \end{subfigure}
     \begin{subfigure}[b]{0.9\textwidth}
         \centering
         \includegraphics[width=\textwidth]{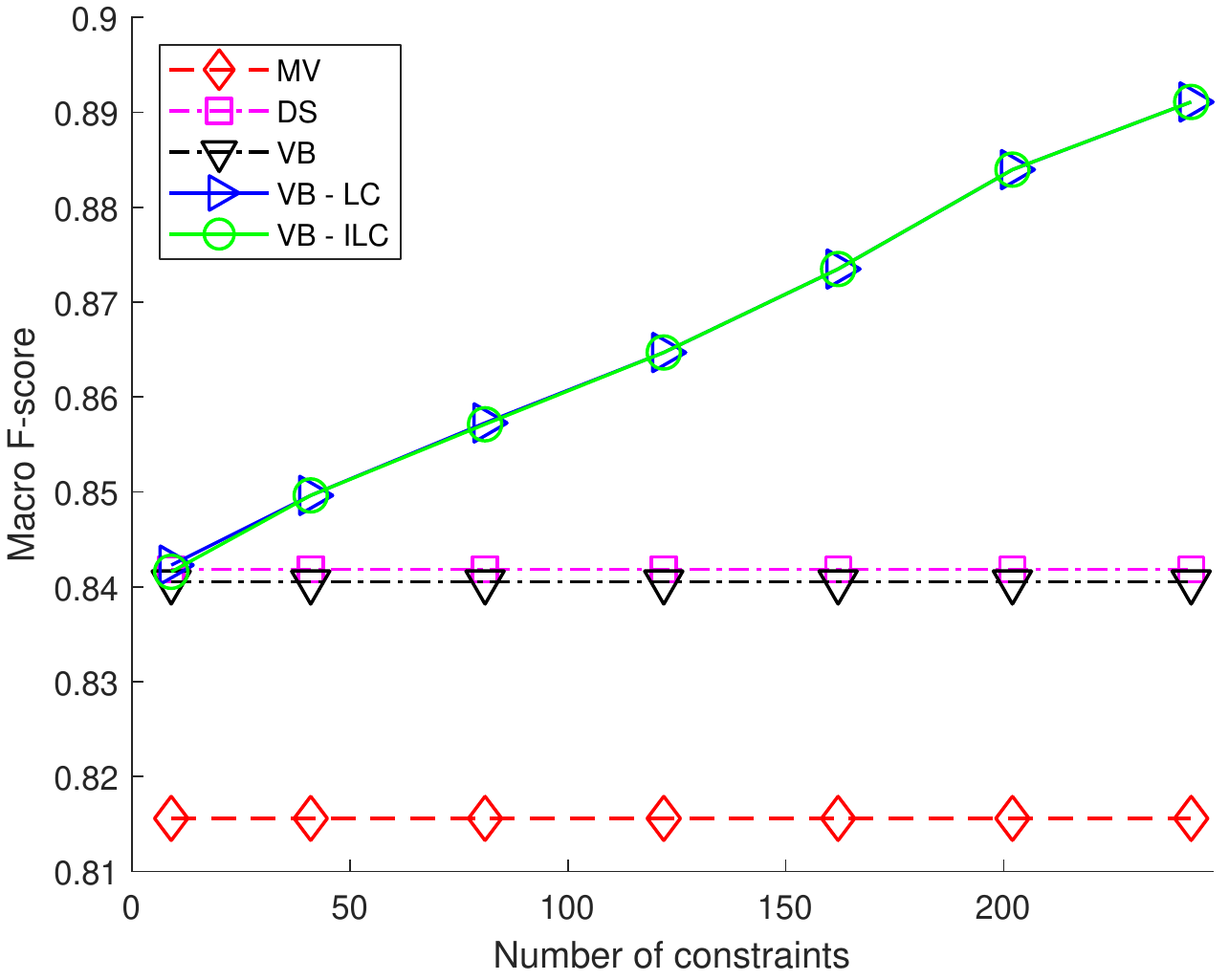}
         \caption{Macro F-score}
         \label{fig:dog_macro_exp3}
     \end{subfigure}
        \caption{Results for the Dog~\cite{imagenet} dataset.}
        \label{fig:dog_res_exp3}
    \end{minipage}
    \hfill
    \begin{minipage}{0.32\textwidth}
        \centering
        \begin{subfigure}[b]{0.9\textwidth}
         \centering
         \includegraphics[width=\textwidth]{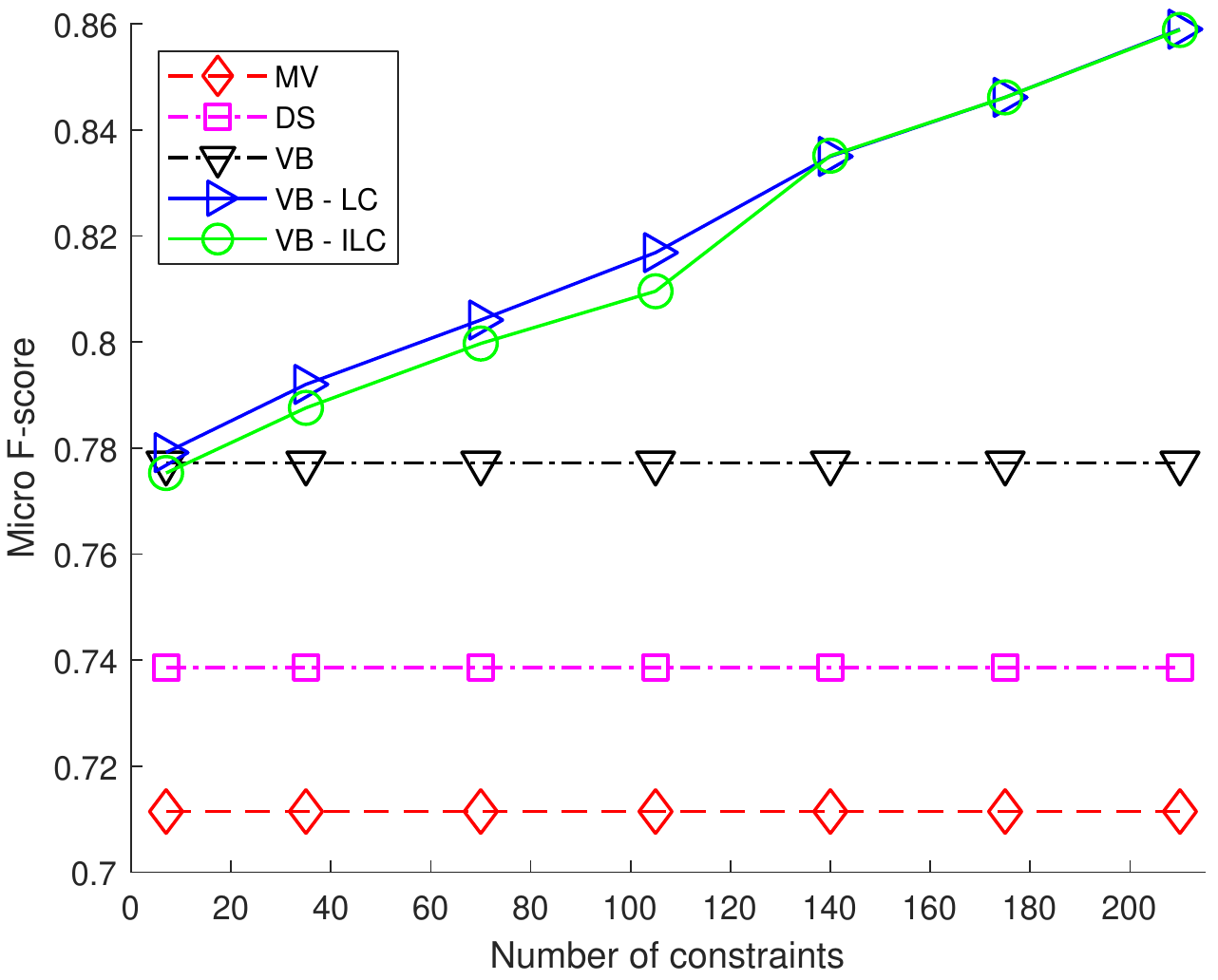}
         \caption{Micro F-score}
         \label{fig:music_micro_exp3}
     \end{subfigure}
     \begin{subfigure}[b]{0.9\textwidth}
         \centering
         \includegraphics[width=\textwidth]{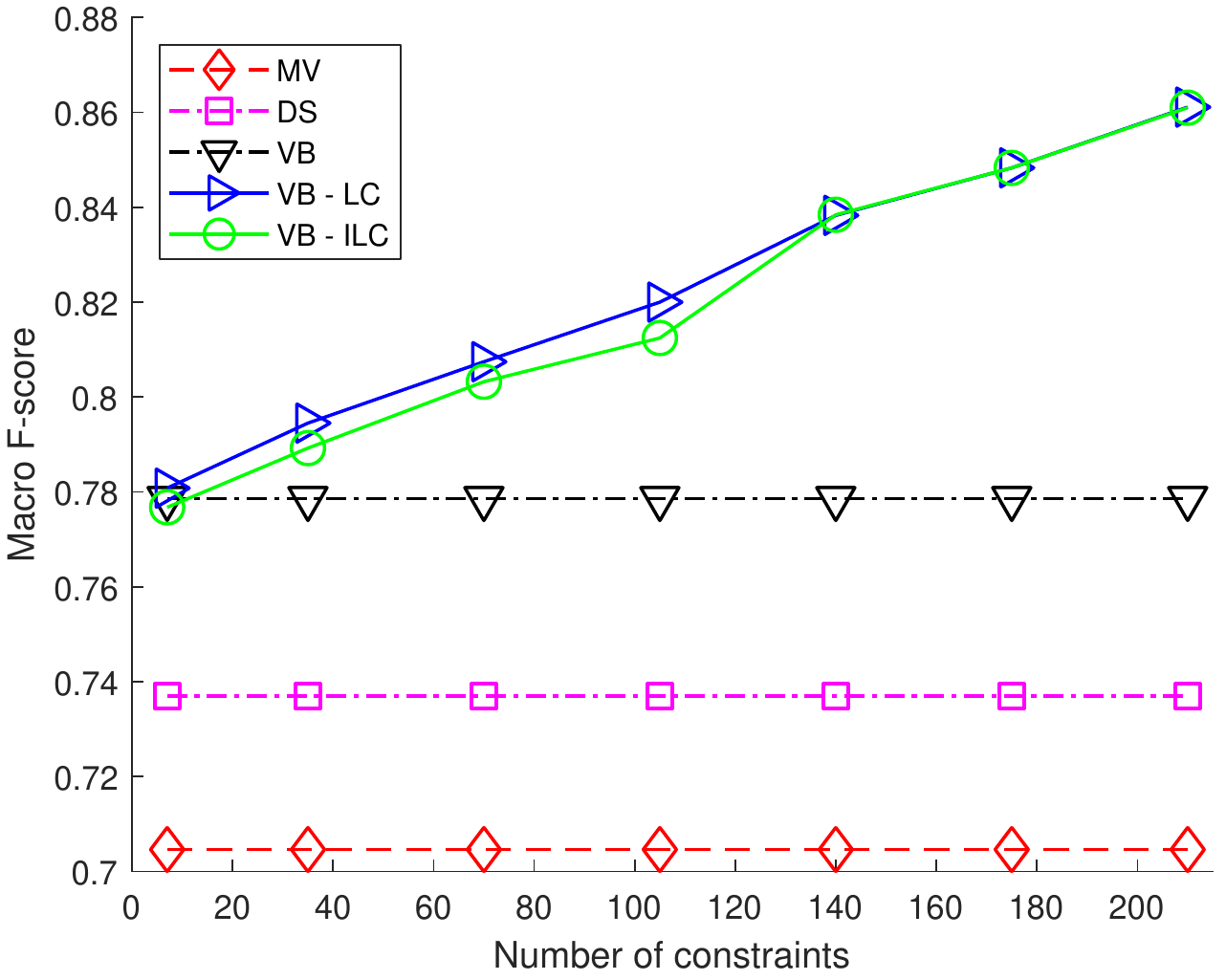}
         \caption{Macro F-score}
         \label{fig:music_macro_exp3}
     \end{subfigure}
        \caption{Results for the Music Genre~\cite{musicgenre_senpoldata} dataset.}
        \label{fig:music_res_exp3}
    \end{minipage}
\end{figure}

\begin{figure}
    \centering
    \begin{minipage}{0.32\textwidth}
        \centering
        \begin{subfigure}[b]{0.9\textwidth}
         \centering
         \includegraphics[width=\textwidth]{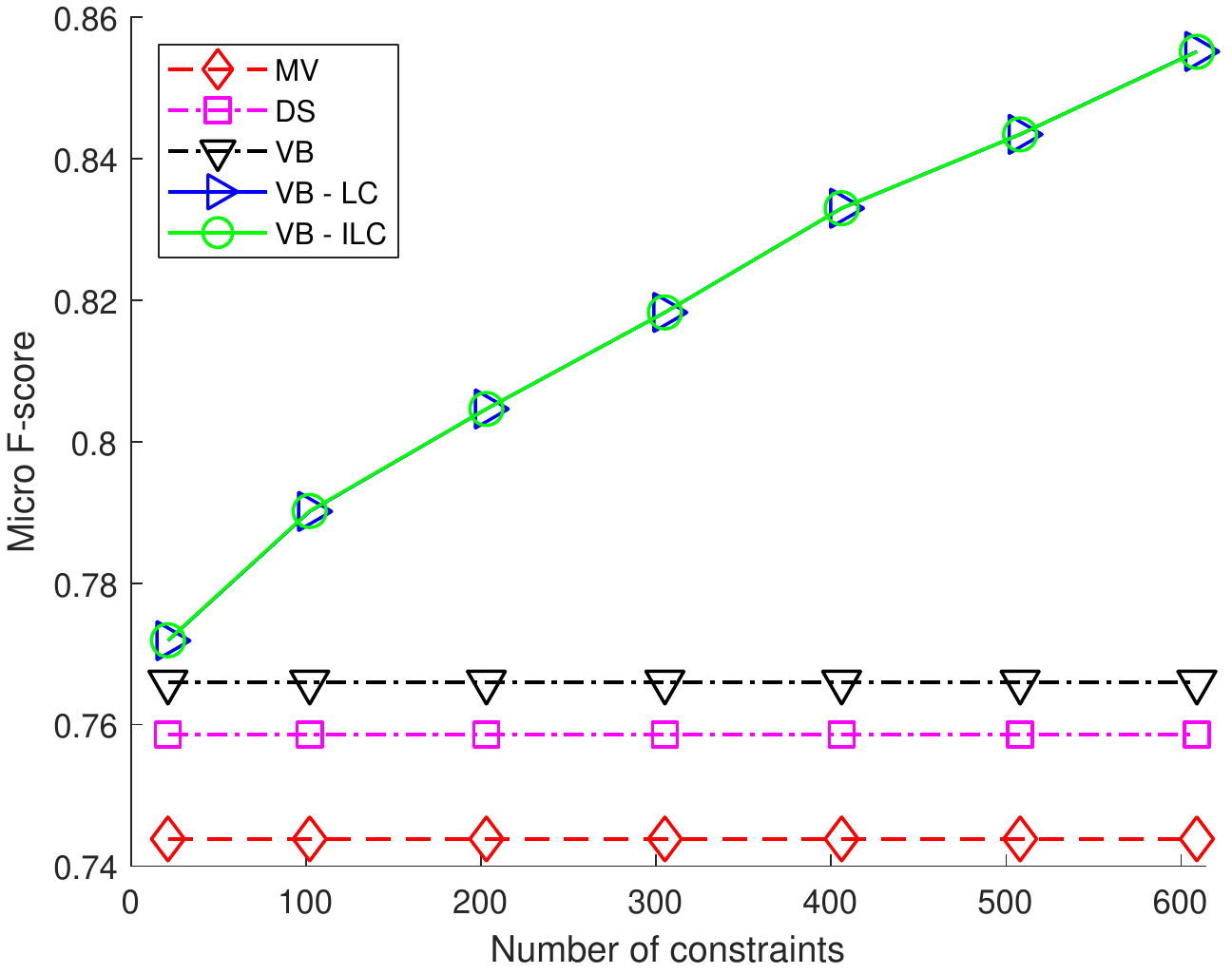}
         \caption{Micro F-score}
         \label{fig:ZenCrowd_in_micro_exp3}
     \end{subfigure}
     \begin{subfigure}[b]{0.9\textwidth}
         \centering
         \includegraphics[width=\textwidth]{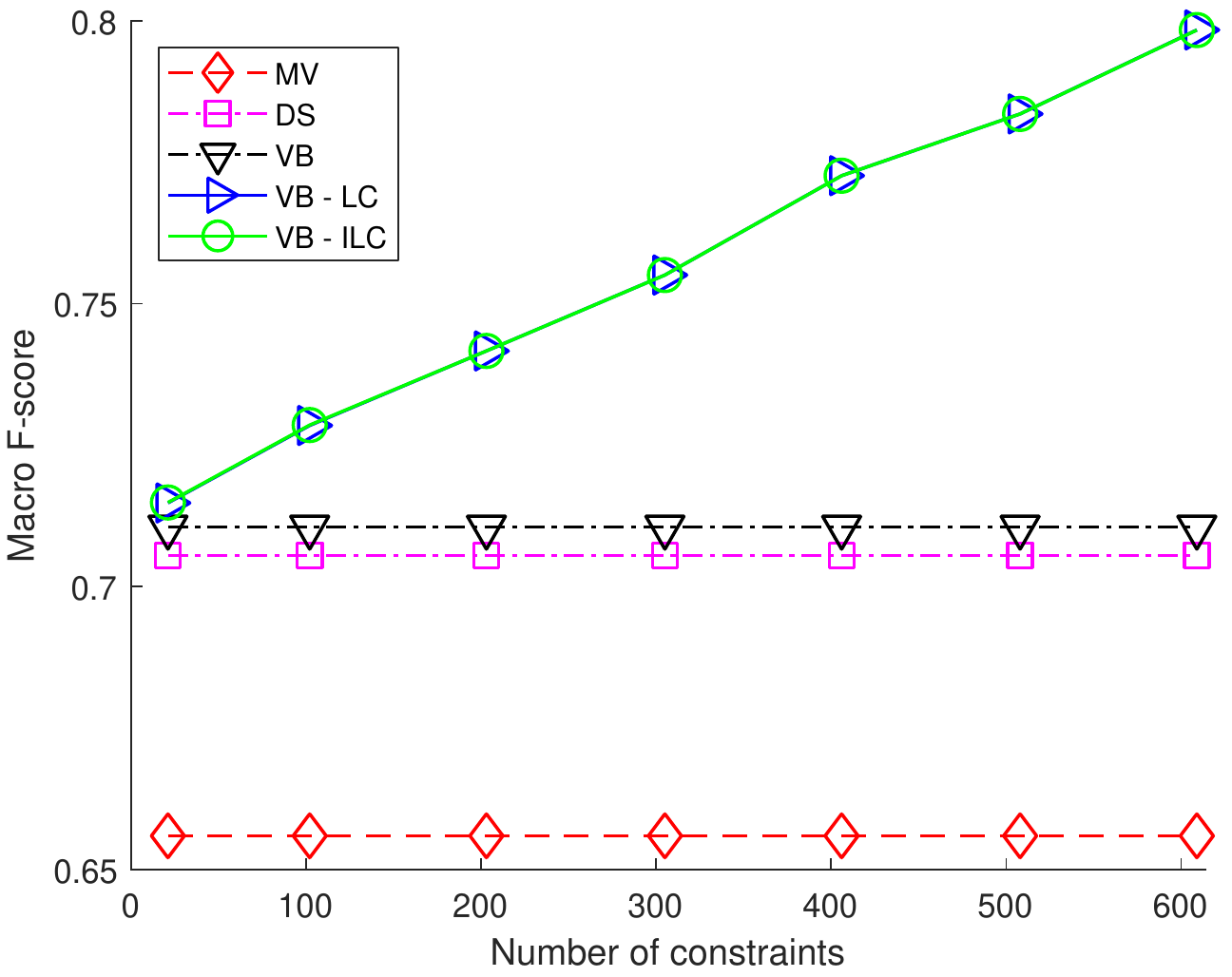}
         \caption{Macro F-score}
         \label{fig:ZenCrowd_in_macro_exp3}
     \end{subfigure}
        \caption{Results for the ZenCrowd India~\cite{ZenCrowd} dataset.}
        \label{fig:ZenCrowd_in_res_exp3}
    \end{minipage}\hfill
    \begin{minipage}{0.32\textwidth}
        \centering
        \begin{subfigure}[b]{0.9\textwidth}
         \centering
         \includegraphics[width=\textwidth]{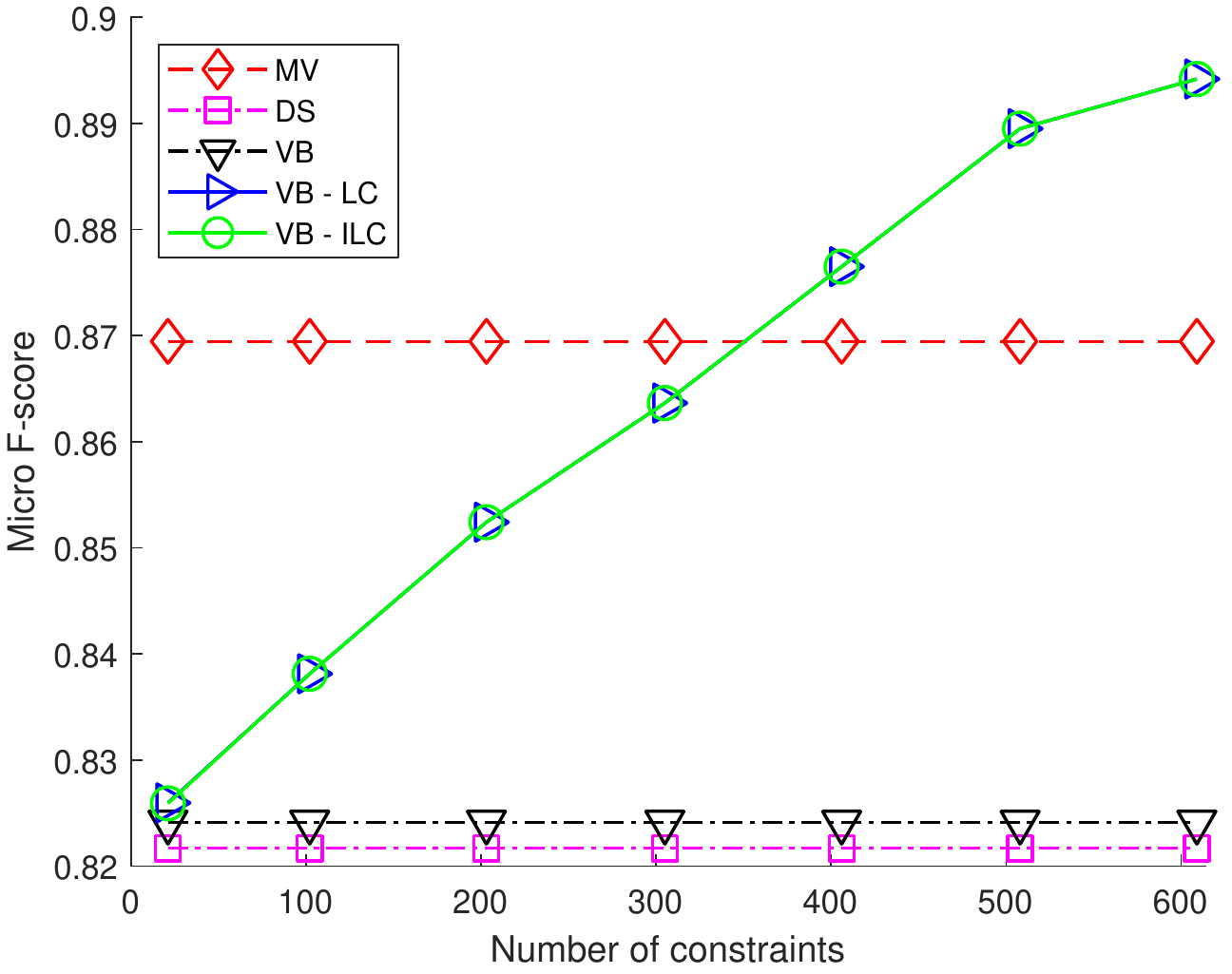}
         \caption{Micro F-score}
         \label{fig:ZenCrowd_us_micro_exp3}
     \end{subfigure}
     \begin{subfigure}[b]{0.9\textwidth}
         \centering
         \includegraphics[width=\textwidth]{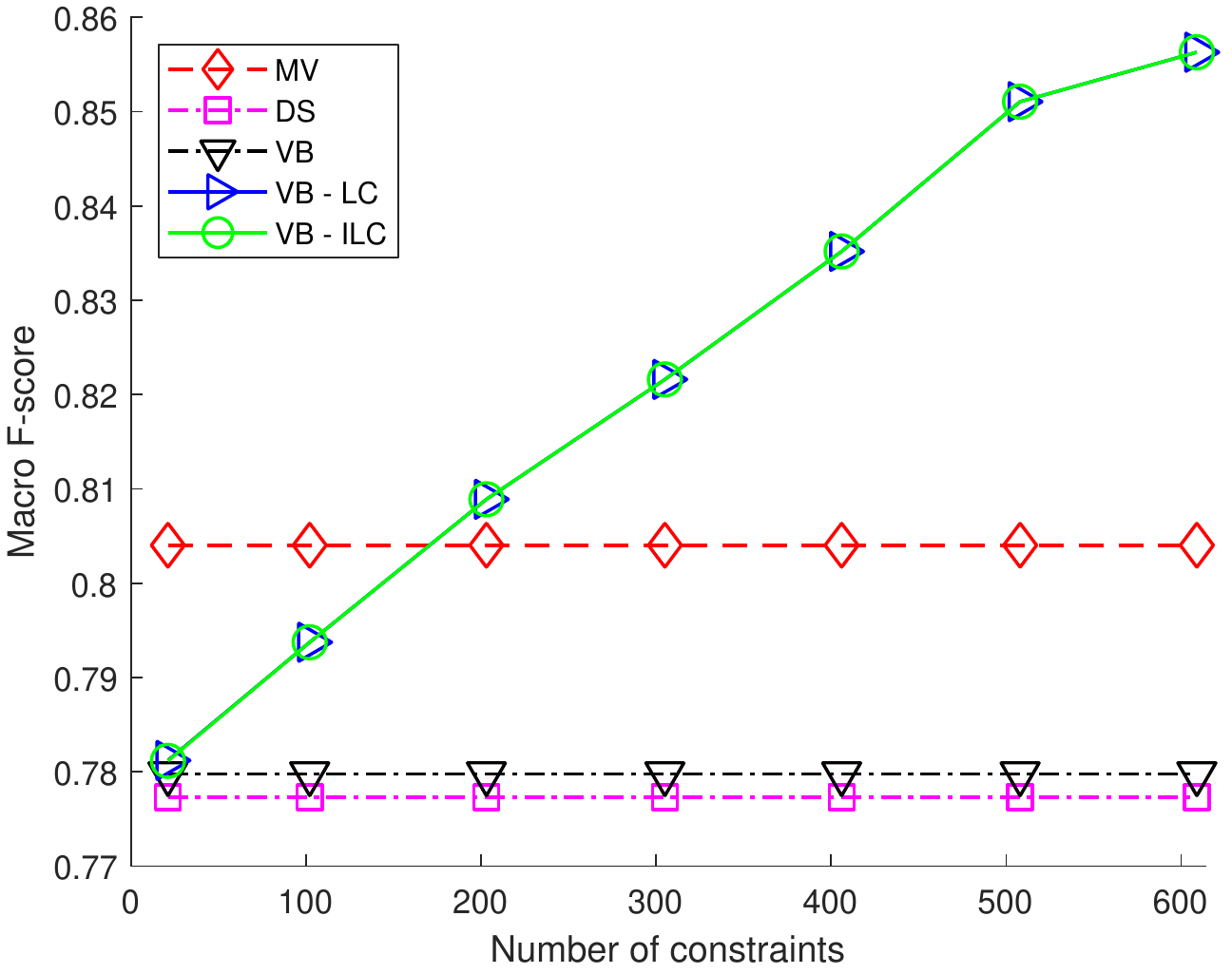}
         \caption{Macro F-score}
         \label{fig:ZenCrowd_us_macro_exp3}
     \end{subfigure}
        \caption{Results for the ZenCrowd US~\cite{ZenCrowd} dataset.}
        \label{fig:ZenCrowd_us_res_exp3}
    \end{minipage}
    \hfill
    \begin{minipage}{0.32\textwidth}
        \centering
        \begin{subfigure}[b]{0.9\textwidth}
         \centering
         \includegraphics[width=\textwidth]{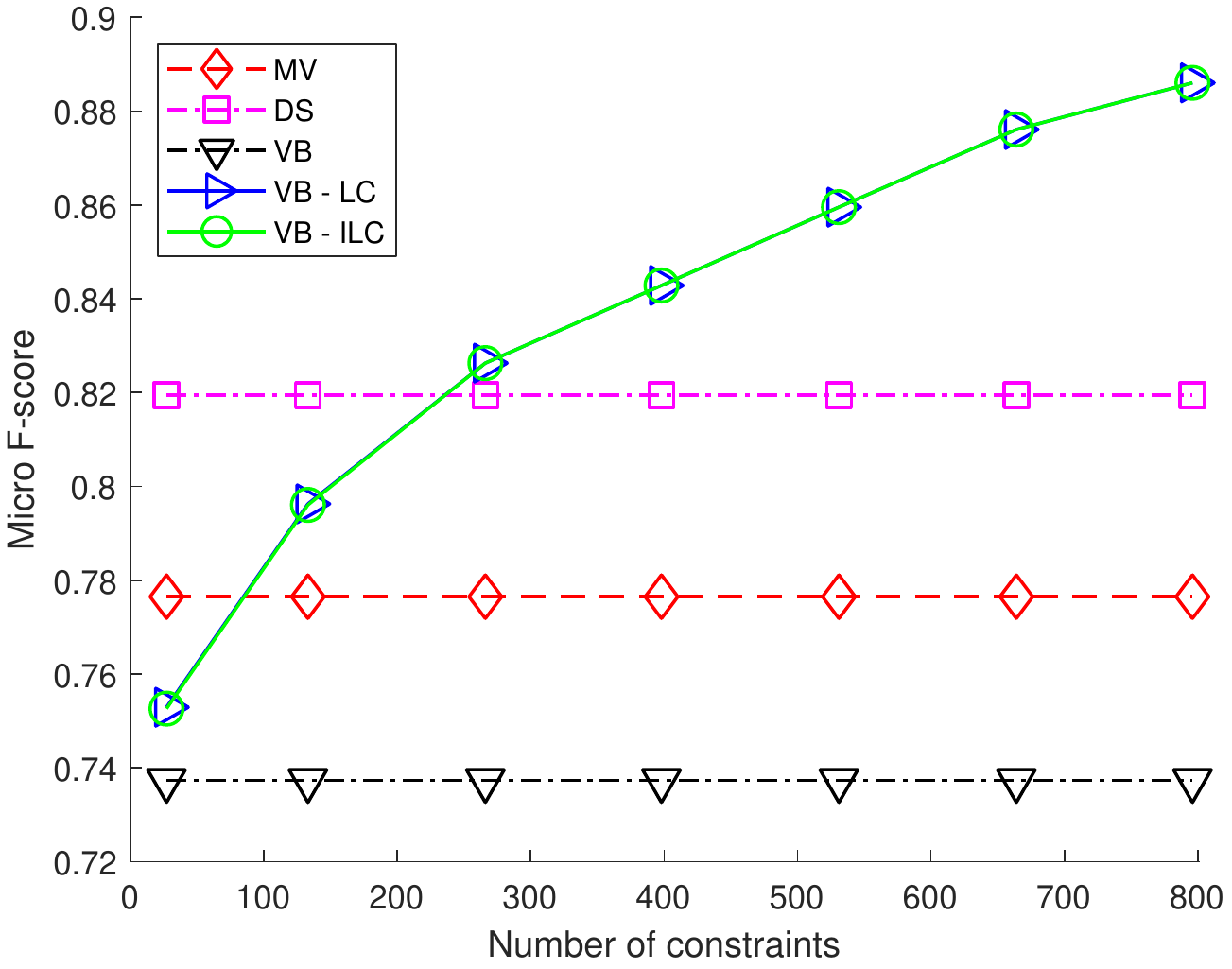}
         \caption{Micro F-score}
         \label{fig:web_micro_exp3}
     \end{subfigure}
     \begin{subfigure}[b]{0.9\textwidth}
         \centering
         \includegraphics[width=\textwidth]{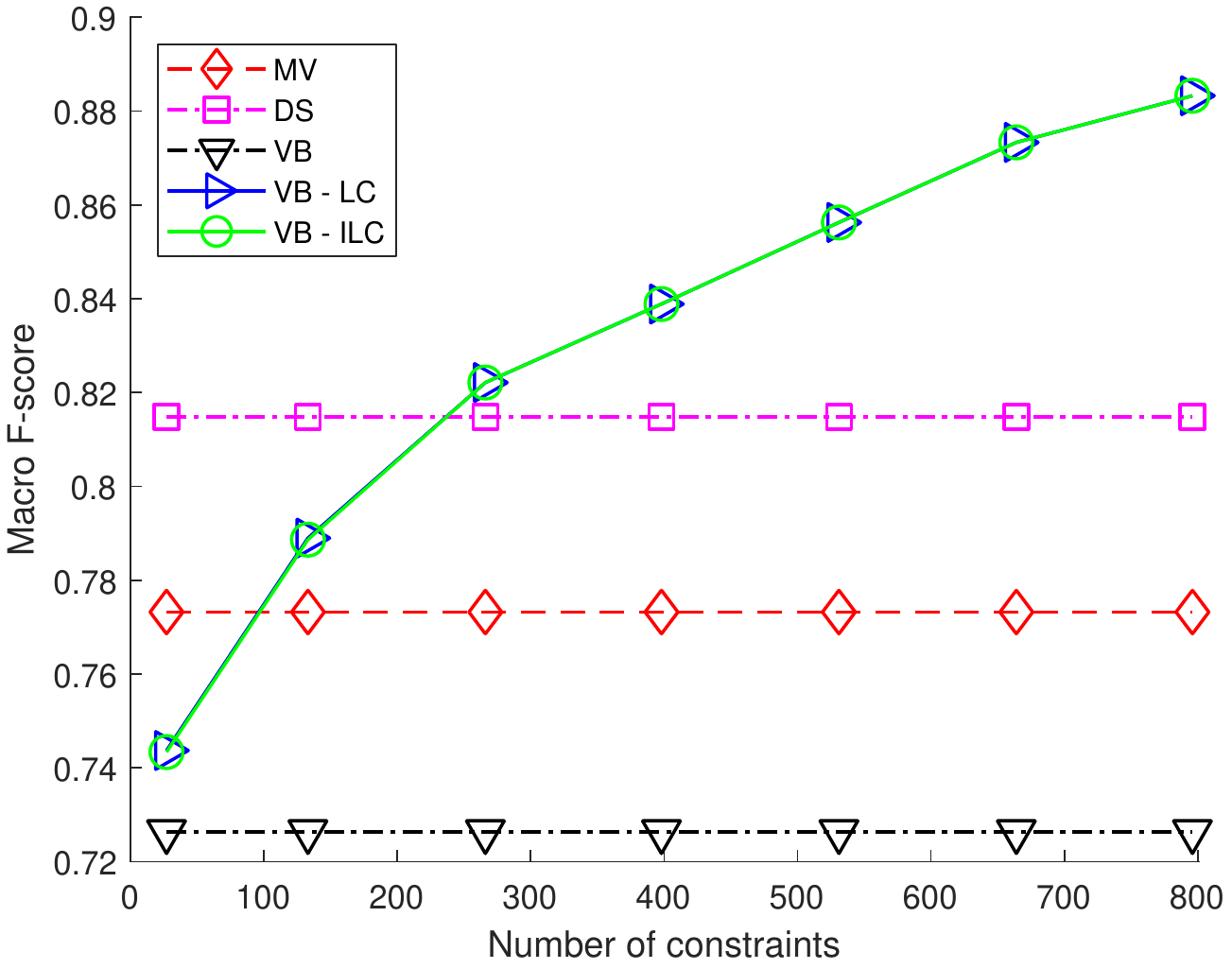}
         \caption{Macro F-score}
         \label{fig:web_macro_exp3}
     \end{subfigure}
        \caption{Results for the Web~\cite{minimax_crowd} dataset.}
        \label{fig:web_res_exp3}
    \end{minipage}
\end{figure}

\begin{figure}
    \centering
    \begin{minipage}{0.32\textwidth}
        \centering
        \begin{subfigure}[b]{0.9\textwidth}
         \centering
         \includegraphics[width=\textwidth]{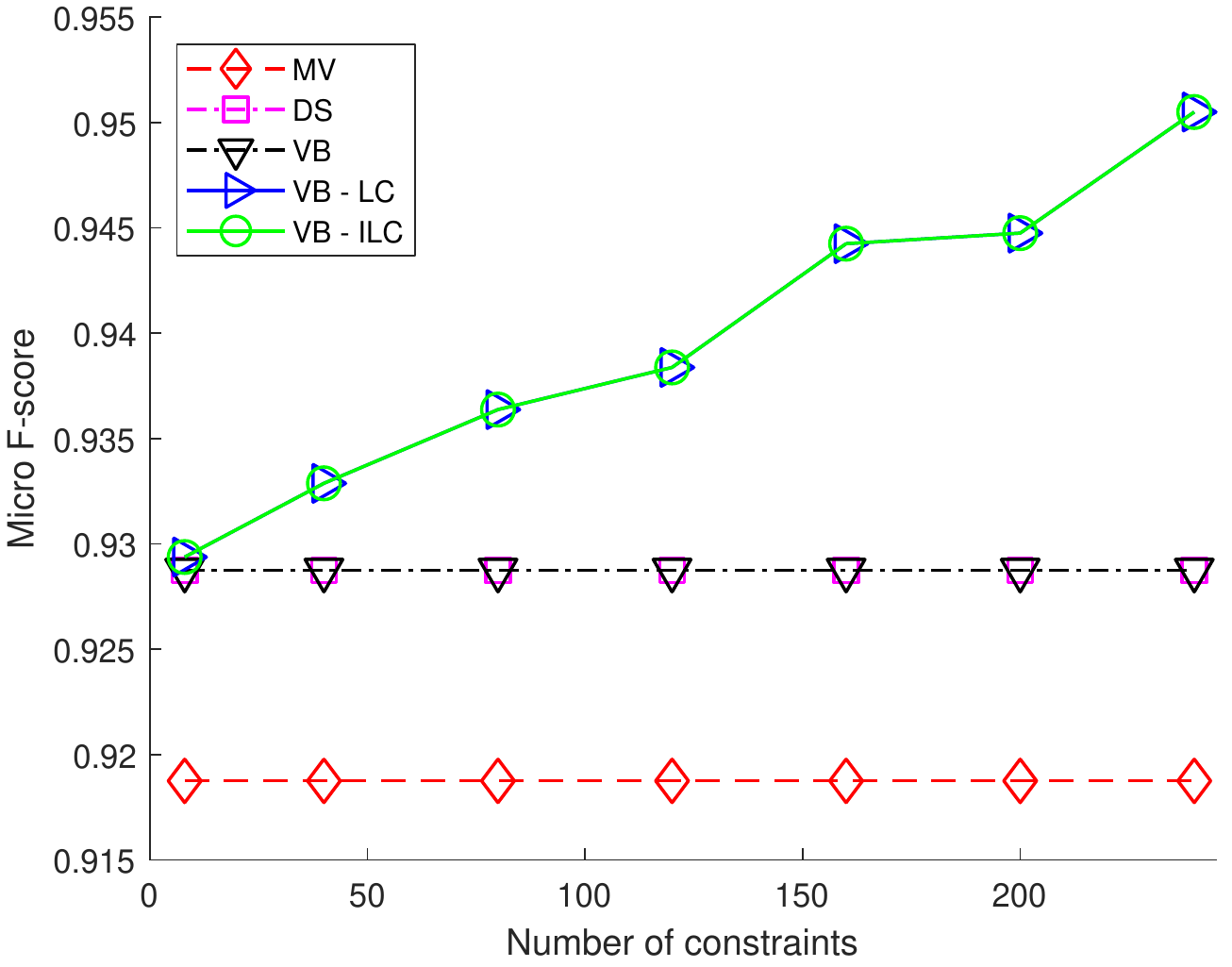}
         \caption{Micro F-score}
         \label{fig:rte_micro_exp3}
     \end{subfigure}
     \begin{subfigure}[b]{0.9\textwidth}
         \centering
         \includegraphics[width=\textwidth]{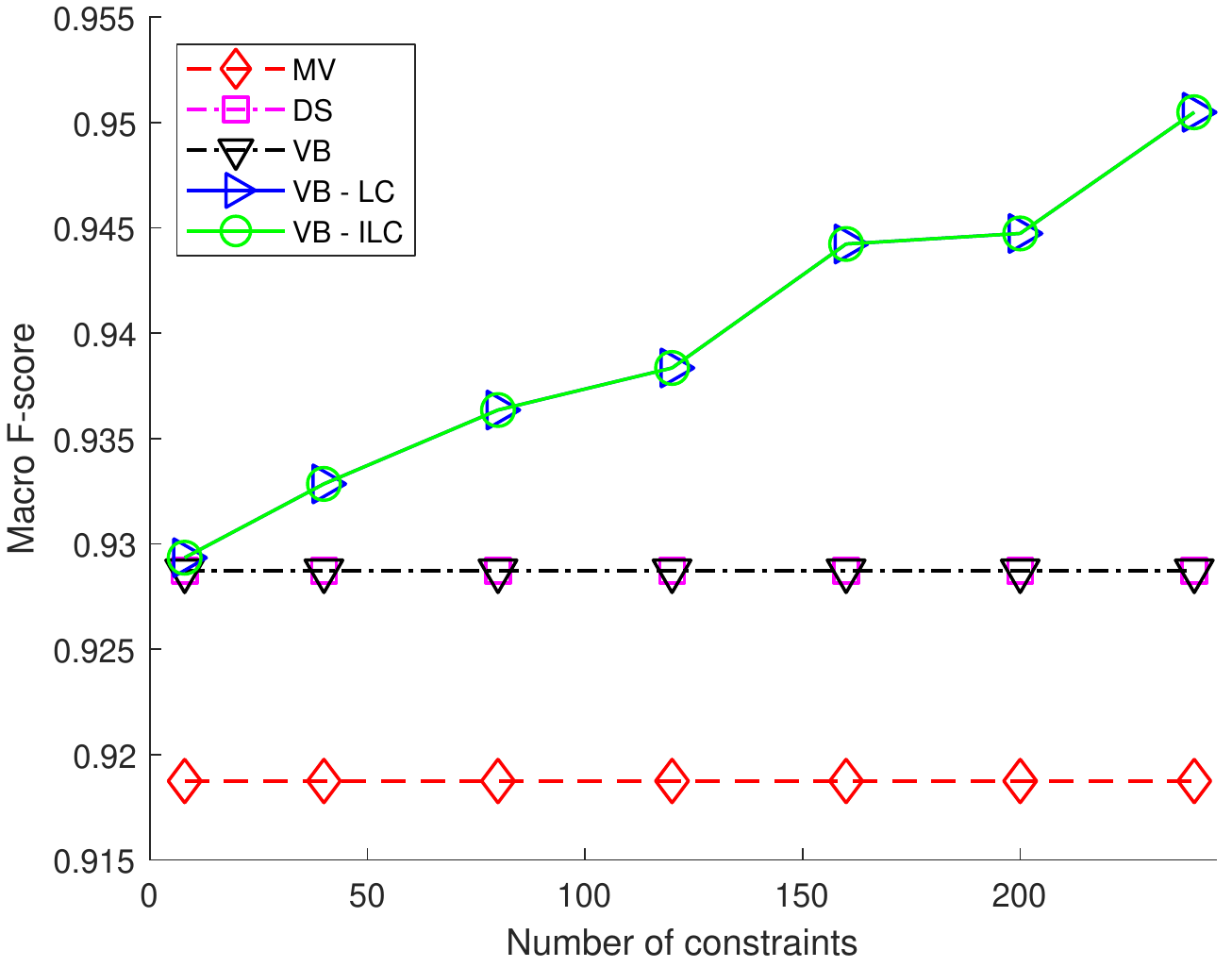}
         \caption{Macro F-score}
         \label{fig:rte_macro_exp3}
     \end{subfigure}
        \caption{Results for the RTE~\cite{cheapnfast} dataset.}
        \label{fig:rte_res_exp3}
    \end{minipage}\hfill
    \begin{minipage}{0.32\textwidth}
        \centering
        \begin{subfigure}[b]{0.9\textwidth}
         \centering
         \includegraphics[width=\textwidth]{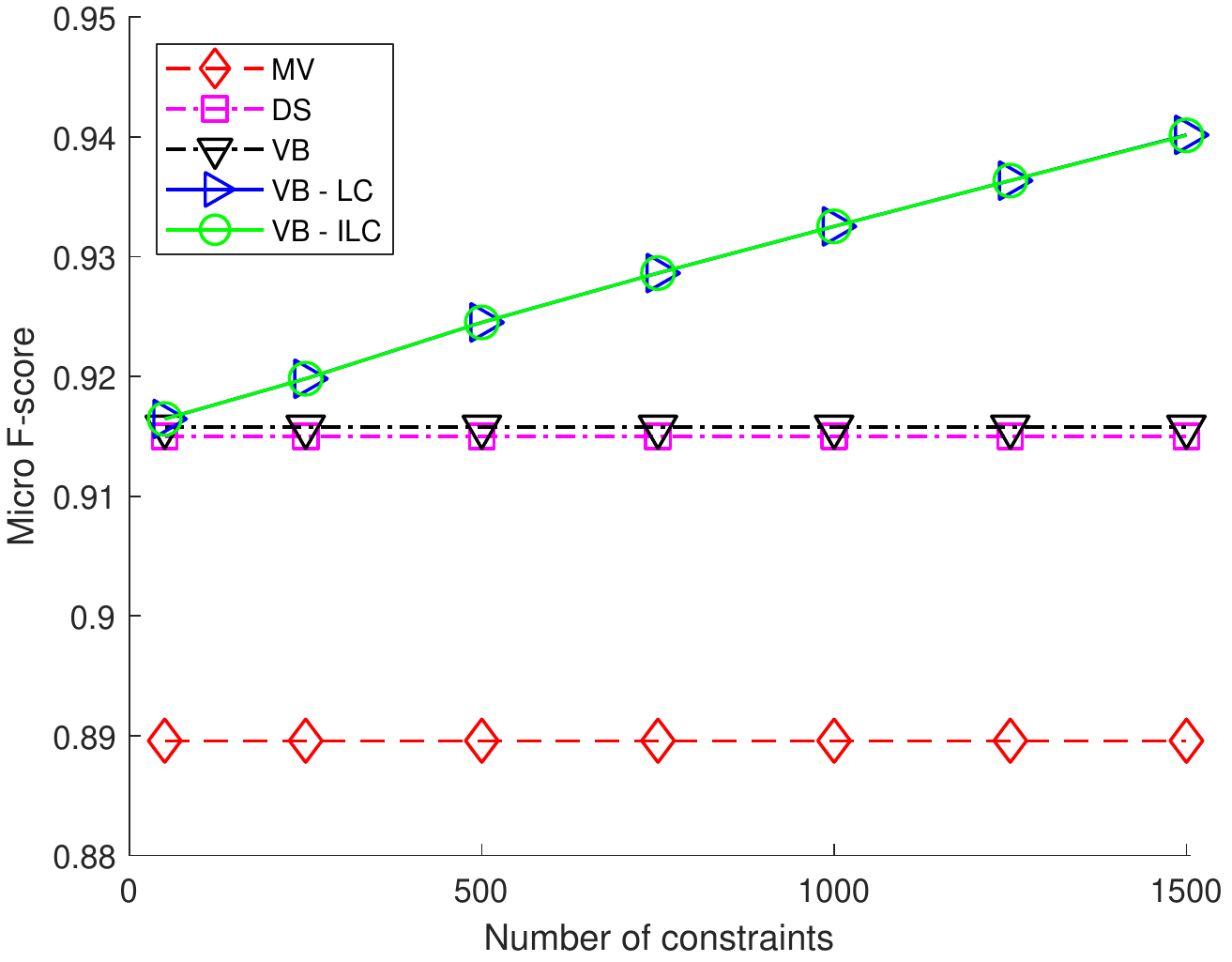}
         \caption{Micro F-score}
         \label{fig:sentencepolarity_micro_exp3}
     \end{subfigure}
     \begin{subfigure}[b]{0.9\textwidth}
         \centering
         \includegraphics[width=\textwidth]{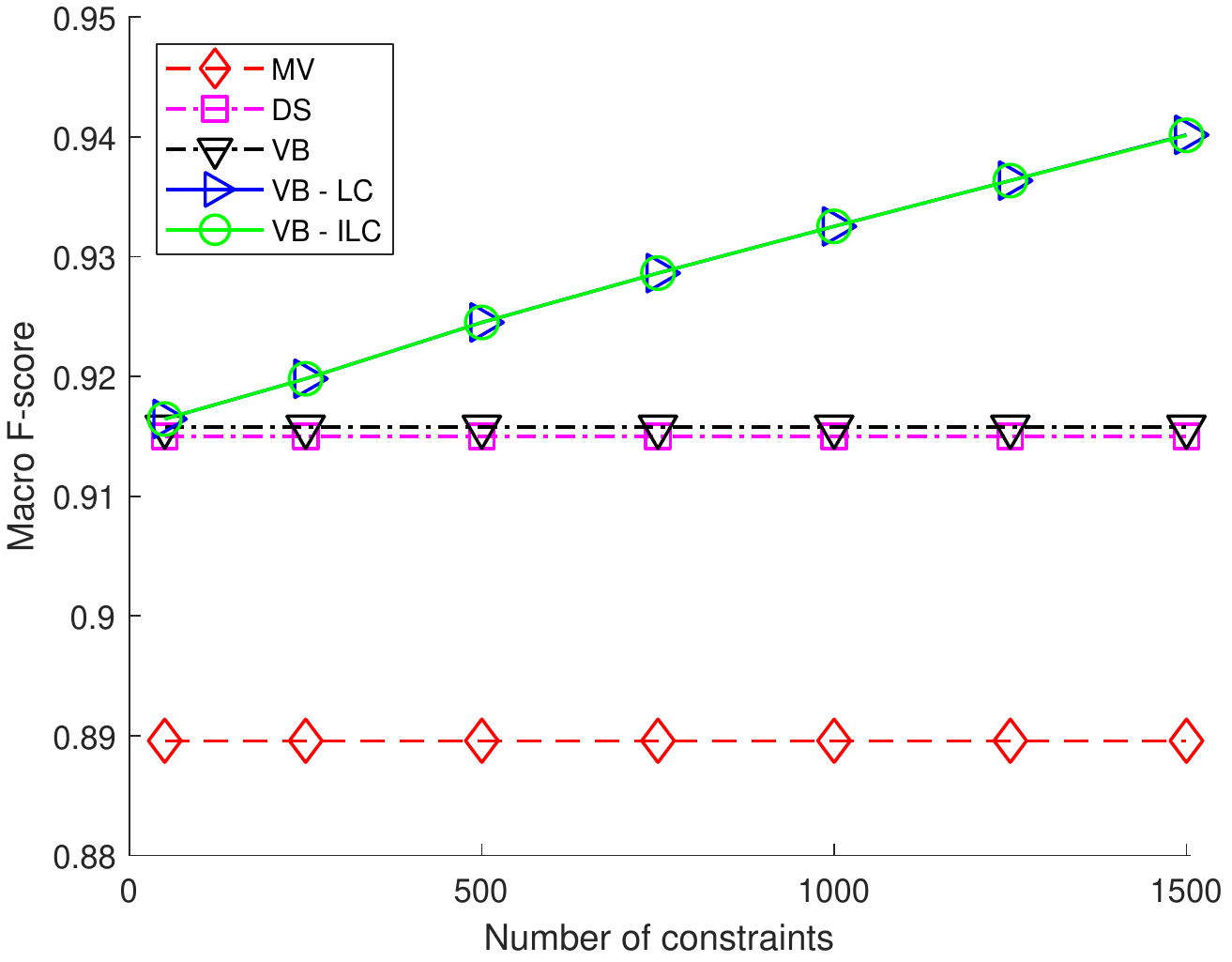}
         \caption{Macro F-score}
         \label{fig:sentencepolarity_macro_exp3}
     \end{subfigure}
        \caption{Results for the Sentence Polarity~\cite{musicgenre_senpoldata} dataset.}
        \label{fig:sentencepolarity_res_exp3}
    \end{minipage}
    \hfill
    \begin{minipage}{0.32\textwidth}
        \centering
        \begin{subfigure}[b]{0.9\textwidth}
         \centering
         \includegraphics[width=\textwidth]{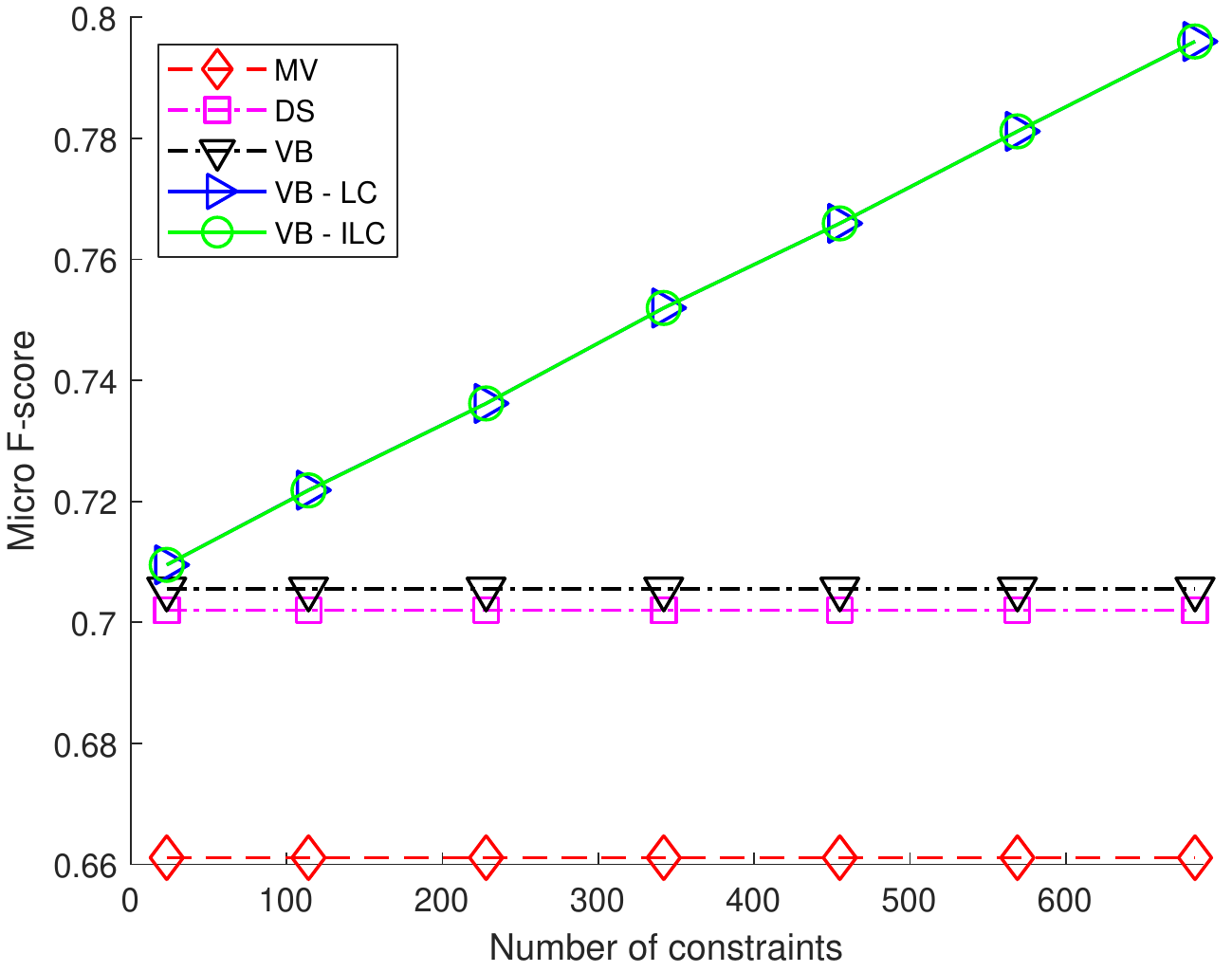}
         \caption{Micro F-score}
         \label{fig:trec_micro_exp3}
     \end{subfigure}
     \begin{subfigure}[b]{0.9\textwidth}
         \centering
         \includegraphics[width=\textwidth]{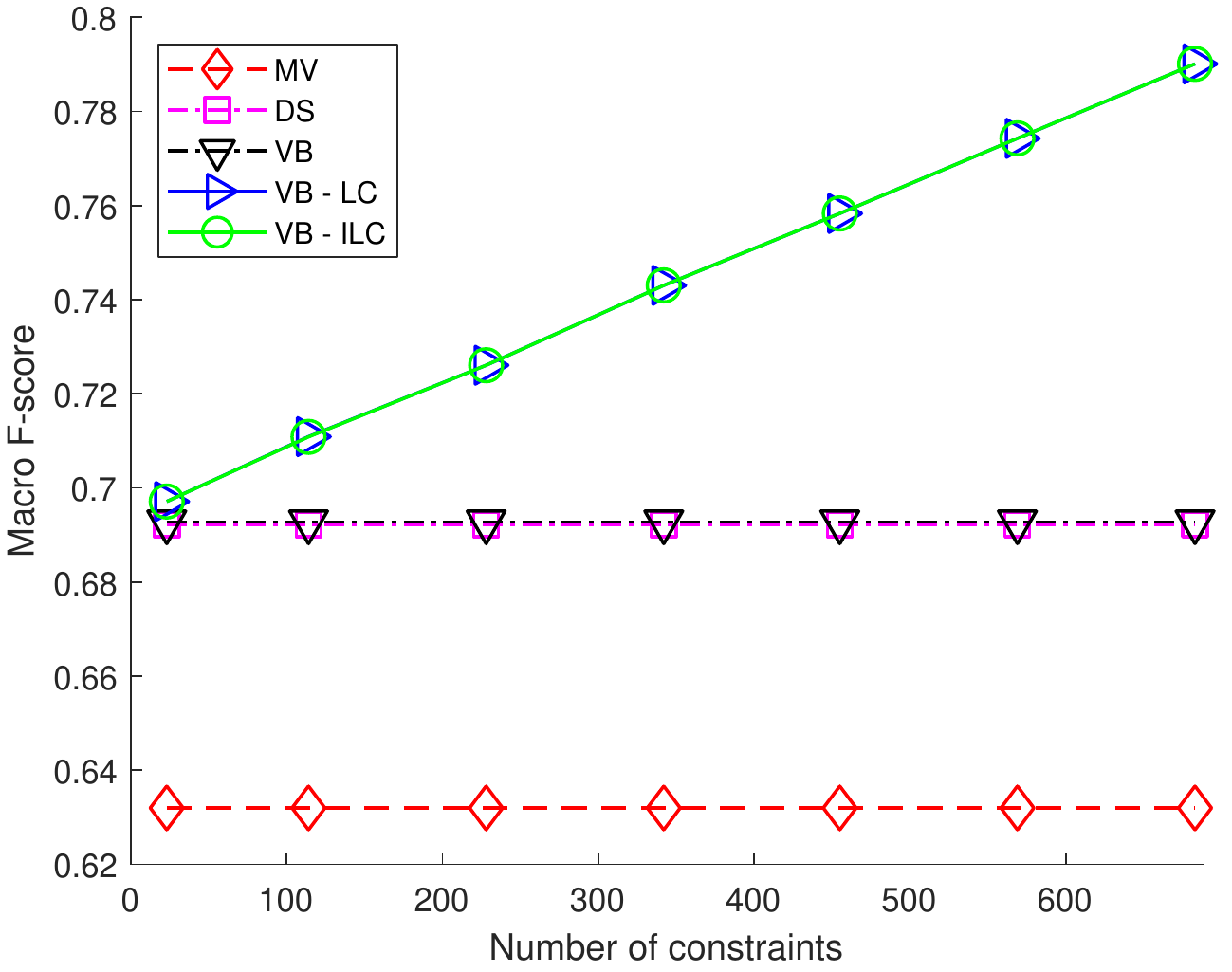}
         \caption{Macro F-score}
         \label{fig:trec_macro_exp3}
     \end{subfigure}
        \caption{Results for the TREC~\cite{Lease11-trec} dataset.}
        \label{fig:trec_res_exp3}
    \end{minipage}
\end{figure}

\begin{figure}[tb]
     \centering
     \begin{subfigure}[b]{0.3\textwidth}
         \centering
         \includegraphics[width=\textwidth]{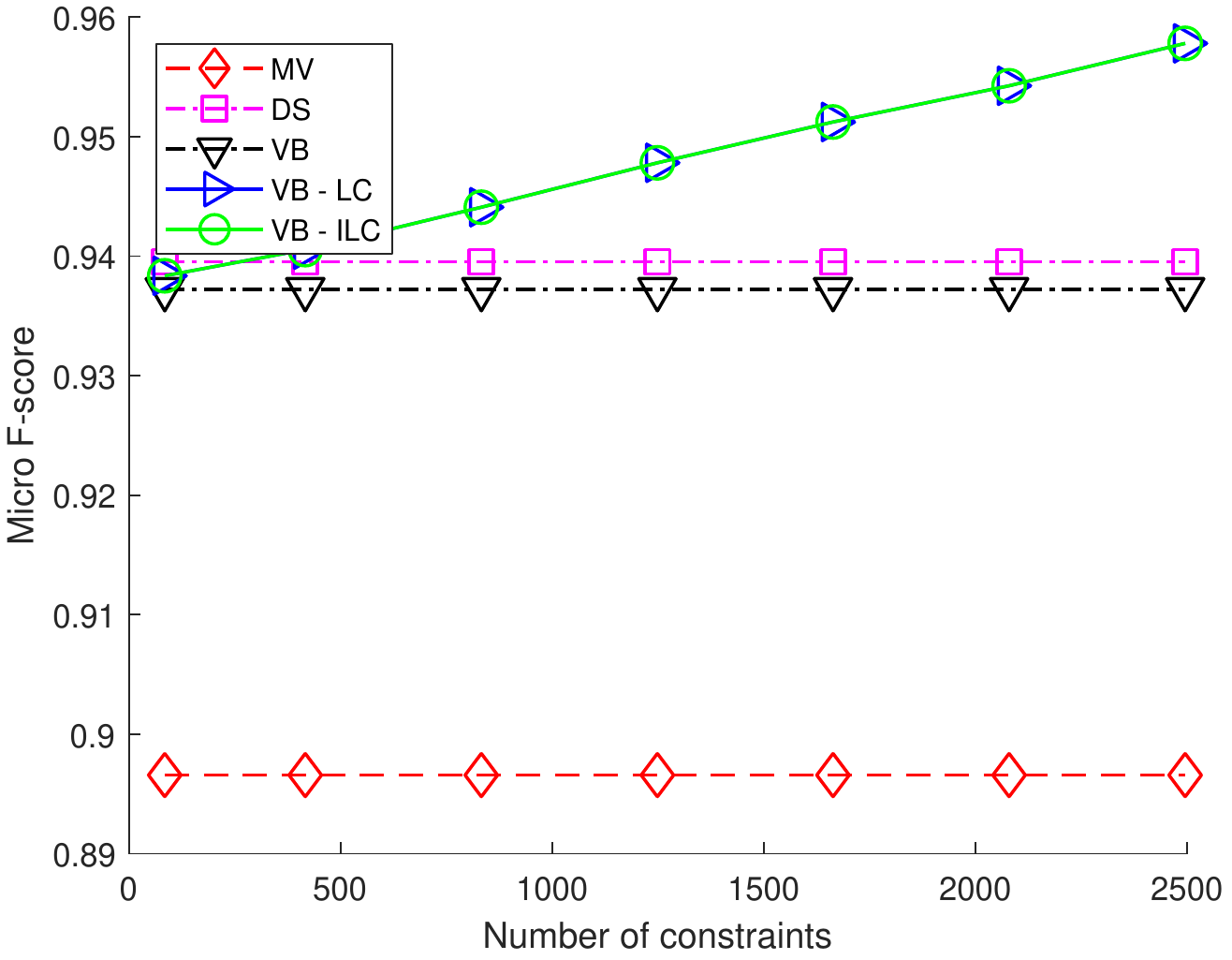}
         \caption{Micro F-score}
         \label{fig:prod_micro_exp3}
     \end{subfigure}\\
     \begin{subfigure}[b]{0.3\textwidth}
         \centering
         \includegraphics[width=\textwidth]{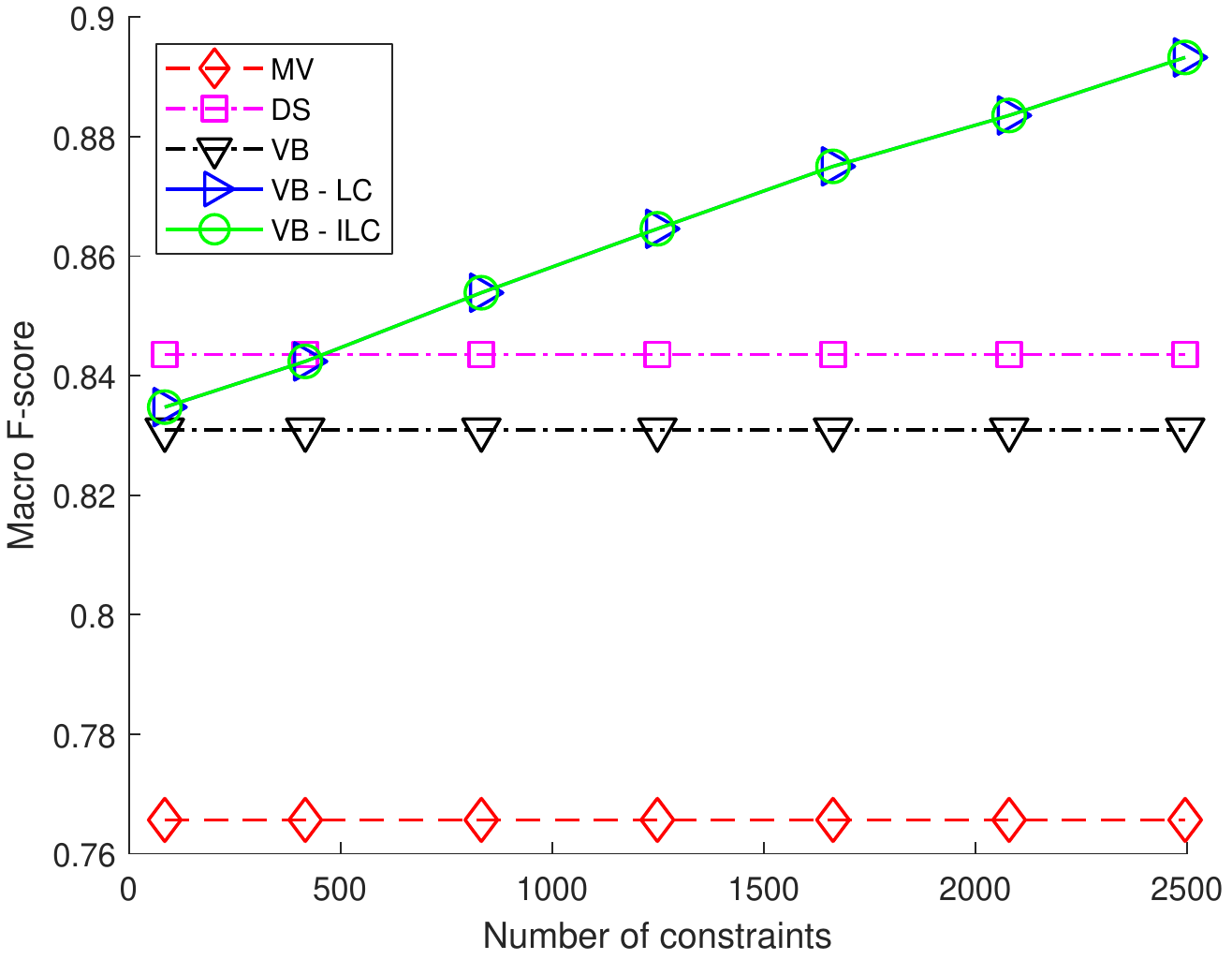}
         \caption{Macro F-score}
         \label{fig:prod_macro_exp3}
     \end{subfigure}
        \caption{Results for the Product~\cite{crowder} dataset.}
        \label{fig:prod_res_exp3}
\end{figure}

The final set of numerical tests involves a similar setup to the previous subsection. Here, $N_C$ label constraints are randomly selected from each dataset and provided to \emph{VB - LC}. From these label constraints, instance level constraints are derived, and then provided to \emph{VB - ILC}. The number of instance level constraints generated using this method will be much greater than $N_C$. Figs. \ref{fig:bluebird_res_exp3}-\ref{fig:prod_res_exp3} show the results under this setup as $N_C$ increases. For all datasets the performance of \emph{VB - LC} and \emph{VB - ILC} is nearly identical, indicating that \emph{VB - ILC} yields high quality results when provided with an adequate number of constraints. Note that, in this scenario and for all datasets, no constraints were violated by \emph{VB - ILC}.

\end{appendices}
